\documentclass{article}

\usepackage{microtype}
\usepackage{graphicx}
\usepackage{subcaption}
\usepackage{booktabs} 
\usepackage{multirow}   
\usepackage{amsmath}    
\usepackage{graphicx}   
\usepackage{pifont}    
\usepackage{amssymb}   
\usepackage[table]{xcolor} 
\usepackage[inline]{enumitem}

\usepackage[colorlinks=true,
            linkcolor=ucladblue,
            citecolor=ucladblue,
            filecolor=ucladblue,
            urlcolor=ucladblue,
            bookmarks=false]{hyperref}

\usepackage[accepted]{icml2026}

\usepackage{amsmath}
\usepackage{amssymb}
\usepackage{mathtools}
\usepackage{amsthm}

\usepackage[capitalize,noabbrev]{cleveref}

\theoremstyle{plain}

\theoremstyle{definition}

\theoremstyle{remark}

\usepackage[textsize=tiny]{todonotes}

\definecolor{Gray}{gray}{0.9}
\definecolor{mygreen}{rgb}{0.0, 0.5, 0.0}
\definecolor{myred}{rgb}{0.8, 0.25, 0.33}
\definecolor{myblue}{rgb}{0.19, 0.55, 0.91}
\definecolor{uclablue}{rgb}{0.15, 0.45, 0.68}
\definecolor{ucladblue}{rgb}{0.0, 0.33, 0.53}
\definecolor{ucladdblue}{rgb}{0.0, 0.23, 0.36}
\definecolor{uclagold}{rgb}{1.0, 0.82, 0.0}
\definecolor{ucladgold}{rgb}{1.0, 0.78, 0.17}
\definecolor{ucladdgold}{rgb}{1.0, 0.72, 0.11}
\definecolor{boxgreen}{rgb}{0.02, 0.66, 0.02}
\definecolor{boxred}{rgb}{0.66, 0.1, 0.1}
\definecolor{boxblue}{rgb}{0.01, 0.01, 0.73}

\usepackage{mdframed}
\usepackage{tikz}
\usepackage{nicefrac}
\usepackage{enumitem}
\usepackage{balance}
\usepackage{xspace}
\usepackage{setspace}
\usepackage{textcase}
\usepackage{mathabx,amsbsy,etoolbox}
\usepackage{acronym}
\usepackage{listings}
\usepackage{tabularx,colortbl,makecell}
\usepackage{overpic}
\usepackage[misc]{ifsym}
\usepackage{siunitx}
\usepackage{soul}
\usepackage[export]{adjustbox}
\usepackage{tcolorbox}
\tcbuselibrary{skins, breakable}
\usepackage[accsupp]{axessibility}

\usepackage[linesnumbered,ruled,vlined,algo2e]{algorithm2e}
\usepackage{dsfont}
\usepackage[dvipsnames]{xcolor}
\usepackage{wrapfig}

\makeatletter
\DeclareRobustCommand\onedot{\futurelet\@let@token\@onedot}
\def\@onedot{\ifx\@let@token.\else.\null\fi\xspace}
\def\eg{\emph{e.g}\onedot} 

\def\ie{\emph{i.e}\onedot}

\def\vs{\emph{vs}\onedot}

\makeatother

\makeatletter
\renewcommand{\paragraph}{%
  \@startsection{paragraph}{4}%
  {\z@}{0ex \@plus 0ex \@minus 0ex}{-1em}%
  {\hskip\parindent\normalfont\normalsize\bfseries}%
}
\makeatother

\crefname{algorithm}{Alg.}{Algs.}
\Crefname{algocf}{Algorithm}{Algorithms}
\crefname{section}{Sec.}{Secs.}
\Crefname{section}{Section}{Sections}
\crefname{table}{Tab.}{Tabs.}
\Crefname{table}{Table}{Tables}
\crefname{figure}{Fig.}{Fig.}
\Crefname{figure}{Figure}{Figure}

\definecolor{gblue}{HTML}{4285F4}
\definecolor{gred}{HTML}{DB4437}
\definecolor{ggreen}{HTML}{0F9D58}

\definecolor{mygray}{gray}{.92}

\acrodef{3dvl}[3D-VL]{3D Vision-Language}
\acrodef{cot}[CoT]{Chain-of-Thought}
\acrodef{sft}[SFT]{Supervised Fine-Tuning}
\acrodef{llm}[LLM]{Large Language Model}
\acrodef{3dqa}[3D-QA]{3D Question-Answering}
\acrodef{mllm}[MLLM]{Multi-modal Large Language Model}
\acrodef{vqa}[VQA]{Visual Question-Answering}
\acrodef{rlvr}[RLVR]{Reinforcement Learning with Verifiable Rewards}
\acrodef{rft}[RFT]{Reinforcement Fine-Tuning}
\acrodef{rl}[RL]{Reinforcement Learning}
\acrodef{vla}[VLA]{Vision-Language-Action}
\acrodef{vlm}[VLM]{Vision-Language Model}
\acrodef{vl}[VL]{Vision-Language}
\acrodef{msqa}[MSQA]{Multi-modal Situated Question Answering}
\acrodef{mson}[MSNN]{Multi-modal Next-step Navigation}
\acrodef{grpo}[GRPO]{Group Relative Policy Optimization}
\acrodef{sft}[SFT]{Supervised Fine-Tuning}
\acrodef{f1score}[F1 Score]{F1 Score}
\acrodef{iou}[IoU]{Intersection over Union}
\acrodef{ce}[CE]{Cross-Entropy}
\acrodef{da}[DA]{Direct-Answer}
\acrodef{ta}[TA]{Think-Answer}
\acrodef{id}[ID]{In-Domain}
\acrodef{ood}[OOD]{Out-Of-Domain}

\newcommand{\sota}{state-of-the-art\xspace}

\newcommand{\model}{3D-RFT-4B\xspace}
\newcommand{\vgllml}{VG LLM-8B\xspace}

\newcommand{\vg}{3D Visual Grounding\xspace}

\newcommand{\spr}{3D spatial reasoning\xspace}
\newcommand{\videt}{3D video detection\xspace}
\newcommand{\vigrd}{3D visual grounding\xspace}
\newcommand{\vsu}{video-based 3D scene understanding\xspace}
\newcommand{\su}{3D scene understanding\xspace}
\newcommand{\vsi}{VSI-Bench\xspace}

\newcommand{\vgllmpaper}{VG LLM\xspace}
\newcommand{\vlmr}{VLM-3R\xspace}
\newcommand{\vggt}{VGGT\xspace}

\newcommand{\vr}{Verifiable Rewards\xspace}
\newcommand{\percep}{3D perception\xspace}

\newcommand{\scanrefer}{ScanRefer\xspace}
\newcommand{\scancap}{Scan2Cap\xspace}
\newcommand{\scannetdet}{ScanNetDetection\xspace}
\newcommand{\srdatanum}{298K\xspace}

\newcommand{\qwenvl}{Qwen2.5-VL\xspace}

\lstdefinestyle{pythonstyle}{
    language=Python,
    basicstyle=\ttfamily\small,
    keywordstyle=\color{blue},
    stringstyle=\color{green!50!black},
    commentstyle=\color{gray},
    numbers=left,
    numberstyle=\tiny\color{gray},
    stepnumber=1,
    numbersep=10pt,
    backgroundcolor=\color{white},
    showspaces=false,
    showstringspaces=false,
    showtabs=false,
    frame=single,
    tabsize=4,
    captionpos=b,
    breaklines=true,
    breakatwhitespace=false,
    escapeinside={\%*}{*)}
}

\definecolor{lightgraybg}{RGB}{235,235,235}
\definecolor{orangehl}{RGB}{220,120,20}
\definecolor{bluehl}{RGB}{40,90,180}
\definecolor{greenhl}{RGB}{20,140,90}

\icmltitlerunning{RFT for 3D Scene Understanding}

\begin{document}

\twocolumn[
  \icmltitle{3D-RFT: Reinforcement Fine-Tuning for Video-based 3D Scene Understanding}



  \icmlsetsymbol{equal}{*}
  \icmlsetsymbol{lead}{†}

  \begin{icmlauthorlist}
    \icmlauthor{Xiongkun Linghu}{bigai,equal,lead}
    \icmlauthor{Jiangyong Huang}{bigai,pku,equal}
    \icmlauthor{Baoxiong Jia}{bigai}
    \icmlauthor{Siyuan Huang}{bigai}
  \end{icmlauthorlist}


  \icmlaffiliation{bigai}{State Key Laboratory of General Artificial Intelligence, BIGAI}
  \icmlaffiliation{pku}{Peking University}

  \icmlcorrespondingauthor{Siyuan Huang}{huangsiyuan@ucla.edu}
  \icmlcorrespondingauthor{Baoxiong Jia}{baoxiongjia@g.ucla.edu}

  \icmlkeywords{3D Scene Understanding, Reinforcement Learning, ICML}

  \vskip 0.3in
]



\printAffiliationsAndNotice{\icmlEqualContribution; †Project lead.}


\begin{abstract}
\ac{rlvr} has emerged as a transformative paradigm for enhancing the reasoning capabilities of \acp{llm}, yet its potential in 3D scene understanding remains under-explored. Existing approaches largely rely on \ac{sft}, where the token-level cross-entropy loss acts as an indirect proxy for optimization, leading to a misalignment between training objectives and task performances. To bridge this gap, we present 3D-RFT, a systematic framework to extend \acs{rlvr} to video-based 3D perception and reasoning. 3D-RFT shifts the paradigm by directly optimizing the model towards evaluation metrics, performing reinforcement fine-tuning using \ac{grpo} with strictly verifiable reward functions, which directly stem from metrics like 3D IoU and F1-Score to provide more effective learning signals. Extensive experiments demonstrate that \model achieves \sota performance on various \vsu tasks. Notably, \model significantly outperforms larger models (\eg, \vgllml) on 3D video detection, 3D visual grounding, and spatial reasoning benchmarks. We further showcase advantages of 3D-RFT, such as robust efficacy, and provide valuable insights into training strategies and data impact. We hope 3D-RFT can serve as a robust and promising paradigm for future development of 3D scene understanding. Code is available on \href{https://3d-rft.github.io/}{project page}.
\end{abstract}

\section{Introduction}
\label{sec:intro}


The remarkable success of \acp{mllm} \citep{Qwen2.5-VL,Qwen3-VL,hurst2024gpt4o,liu2023visual,team2023gemini,comanici2025gemini2.5} has driven advancements in diverse fields, including \su \citep{huang2023embodied,huang2024chatscene,zhu2025llava3d,linghu2024multi,zheng2025video,zheng2025vgllm,chen2026lifting}, robotic manipulation \citep{driess2023palm,brohan2023rt2,team2025geminirobotics}, and embodied navigation \citep{zheng2024navillm,zhang2024navid,cheng2024navila}.

Treating 3D scenes as video streams offers a scalable alternative to traditional point-cloud and depth-based methods, bypassing the need for specialized sensors. By leveraging widely available RGB cameras and the temporal capabilities of \acp{mllm}, this paradigm has become a central focus of recent research \cite{zheng2025video,zheng2025vgllm,zhu2025llava3d,huang2025leovl,fan2025vlm}. While some works \citep{zheng2025video,zhu2025llava3d} still inject explicit 3D features, others \citep{zheng2025vgllm,fan2025vlm} extract 3D priors directly from video frames. Crucially, \vgllmpaper \citep{zheng2025vgllm} validates this approach on tasks like cross-frame detection and 3D visual grounding.

\begin{figure*}[t]
\centering
\includegraphics[width=\textwidth]{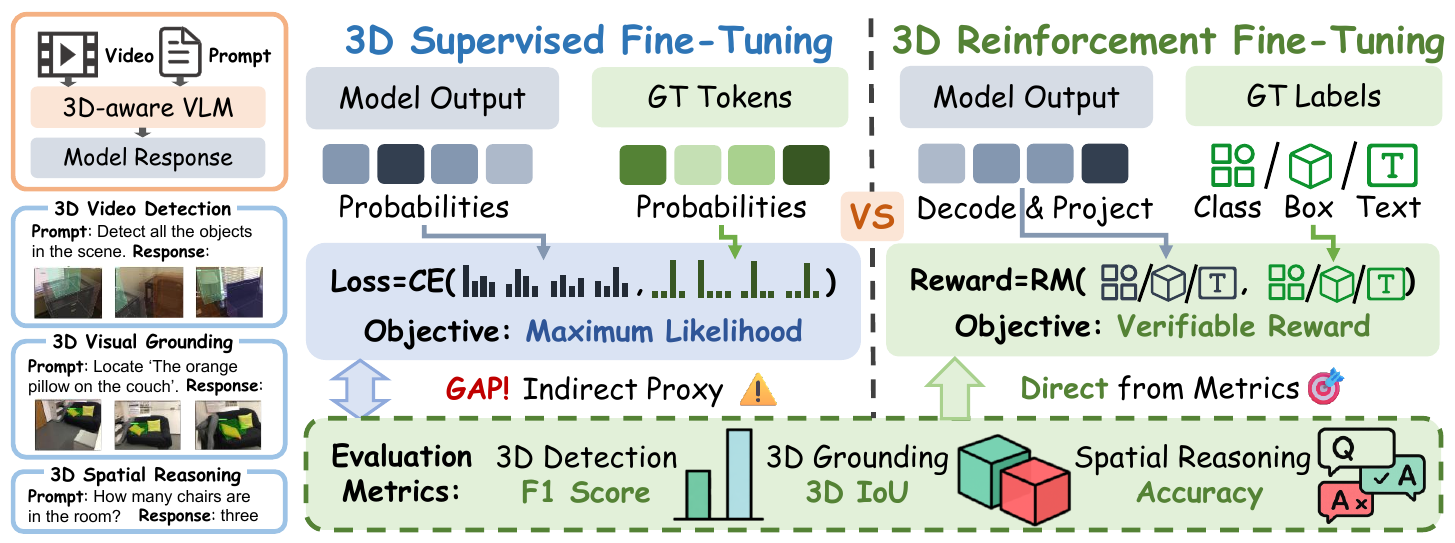}
\caption{\textbf{Comparison between SFT and 3D-RFT training paradigms.} 
Left: Standard Supervised Fine-Tuning (SFT) relies on a per-token Cross-Entropy loss which acts as an indirect proxy, leading to a gap between training objectives and final evaluation metrics. 
Middle: Identical output formats for 3D Video Detection, Grounding, and Spatial Reasoning tasks.
Right: 3D-RFT utilizes a Scalar Reward (Policy Gradient) derived directly from evaluation metrics (\eg, F1-Score, 3D IoU, and Accuracy) through a decoding and parsing module, ensuring the model directly optimizes for the final task performance.}
\label{fig:teaser}
\vspace{-1em}
\end{figure*}


However, these approaches primarily rely on \ac{sft} as the learning paradigm, which imposes an inherent performance ceiling due to the limitations of imitating labels. Specifically, in \percep tasks, model responses consist of 3D bounding boxes represented as sequences of textual floating-point numbers. \ac{sft} optimizes these sequences by minimizing a per-token \ac{ce} loss, which creates a critical misalignment: optimization occurs in the discrete token space, whereas evaluation is conducted in the continuous 3D coordinate system. Since the output tokens must be decoded and parsed into geometric structures to compute metrics like 3D \ac{iou}, the standard \ac{sft} objective acts only as an indirect proxy, failing to capture the underlying landscape of the predictions. 

Incorporating metric-driven loss functions directly into \ac{sft} remains a key challenge due to the non-differentiable nature of evaluation pipelines. Because standard \ac{sft} relies on backpropagation, it requires a fully differentiable path from the loss to the model's logits. Using target 3D metrics as an \ac{sft} loss is mathematically intractable; converting text to a 3D box requires discrete string parsing, and calculating the final metric involves non-continuous step functions (e.g., $IoU > 0.25$). If integrated directly, these operations produce zero or undefined gradients, halting the learning process.



How can we overcome the limits of \textbf{answer-only supervision} and the \textbf{non-differentiable barrier} of SFT when applying metric-driven supervised signals? \ac{rlvr} offers a principled solution by optimizing models directly against \textbf{\vr}, which are derived directly from distinct evaluation metrics or pre-defined rules. Unlike \ac{sft}, which constrains the model to mimic ground-truth sequences via a per-token \ac{ce} loss, \ac{rl} optimizes the policy based on scalar reward scores. This provides a superior optimization target that is strictly aligned with the evaluation process, encouraging sequence exploration without the limitations of token-level penalties. Recent breakthroughs, such as GPT-o1 \citep{jaech2024openaio1} and DeepSeek-R1 \citep{guo2025deepseekr1}, have widely demonstrated the success of \ac{rlvr} in advancing mathematical reasoning and code generation.


A natural question arises: \textit{Can this metrics-driven \ac{rl} paradigm generalize to \vsu?} To address this, we introduce 3D-RFT, a unified framework that successfully extends \ac{rl} to diverse \vsu tasks, encompassing both \textit{3D perception} and \textit{3D spatial reasoning}.


As illustrated in \cref{fig:teaser}, 3D-RFT fundamentally shifts the training objective. While \ac{sft} relies on an indirect proxy—minimizing \ac{ce} loss between predicted and ground-truth probabilities—this leads to a misalignment between training objectives and evaluation metrics. In contrast, 3D-RFT leverages policy gradients derived directly from verifiable reward signals (\eg, 3D IoU, Accuracy), ensuring the model is optimized explicitly for final task performance.


To overcome the lack of native 3D perception capabilities in existing \acp{mllm}, we design a robust two-stage training pipeline:
\begin{enumerate*}[label=\textbf{\arabic*)}]
    \item \textbf{SFT Warm-Up:} We first inject fundamental 3D awareness and scene understanding capabilities into the \ac{mllm} via \ac{sft}, establishing a stable policy initialization.
    \item \textbf{RL Training:} The model is then fine-tuned using the \ac{grpo} algorithm with verifiable reward functions strictly following evaluation protocols. For instance, in \vg, we parse the predicted 9-DoF box, project it into the global coordinate system, and compute the global 3D \ac{iou} as the reward signal.
\end{enumerate*}
This approach effectively transitions the learning paradigm from task-agnostic sequence imitation to metrics-driven policy optimization.


We validate the efficacy of 3D-RFT across typical \vsu tasks, including \percep (\eg, \videt \citep{wang2024embodiedscan}, \vigrd \citep{chen2020scanrefer}) and \spr \citep{yang2025thinking}. Experimental results demonstrate that 3D-RFT consistently enhances model performance, and our model \model outperforms larger baseline models. Our analysis yields several key findings:
1) 3D-RFT provides consistent gains across all three task domains.
2) In \percep tasks, 3D-RFT significantly boosts performance over the \ac{sft} baseline, surpassing even 8B-scale fine-tuned models on nearly all metrics, proving that metrics-driven optimization offers a more effective learning objective.
3) For \spr, 3D-RFT effectively enhances model performance to \sota on \vsi, surpassing previous models at larger scales; we also reveal the effects of data diversity on 3D-RFT. 

In summary, our key contributions are as follows:
\begin{itemize}[nolistsep,noitemsep,leftmargin=*]
    \item We propose 3D-RFT, a systematic reinforcement fine-tuning framework that extends \ac{rlvr} to 3D video perception and reasoning, shifting the learning paradigm from sequence imitation to metrics-driven policy optimization.
    \item We design task-specific verifiable reward functions derived directly from evaluation metrics (\eg, 3D IoU, F1-Score) to enable efficient and robust policy updates.
    \item We conduct extensive experiments on standard \vsu benchmarks, demonstrating that 3D-RFT achieves significant improvements over SFT baselines and surpasses larger-scale models in both 3D perception and reasoning tasks.
\end{itemize}

\section{Related Work}
\label{sec:related_work}

\paragraph{\acp{mllm} for 3D Scene Understanding.}
\nocite{huang2022perceive,gong2023arnold}
3D perception and spatial reasoning are fundamental pillars of 3D scene understanding. Early research focused on developing unified \acp{mllm} based on 3D sensory inputs (\eg, point clouds) \citep{zhu20233d,xu2024pointllm,huang2023embodied,yang2024llm,chen2024ll3da,qi2024gpt4point,zhu2024unifying,linghu2024multi,fu2025scene,chu2024unified,deng20253d,mao2025spatiallm}. Recently, advancements in video understanding establish video-based \acp{mllm} as a compelling approach due to the streamlined input and strong performance \citep{el2024probing,man2024lexicon3d,zhang2025video,zhu2025llava3d,zheng2025video,qi2025gpt4scene,huang2025leovl,zhu2025struct2d,li2025seeground,cheng20253d}. To bolster the 3D awareness of video-based representation, there are efforts in geometry enhancement \citep{wang2025ross3d,huang2025mllms,zheng2025reg3d,hu2025g} and data scaling \citep{chen2024spatialvlm,cheng2024spatialrgpt,cai2025spatialbot,ma2025spatialllm,xu2025multi,song2025robospatial,zhang2025flatland,zhou2025roborefer,brown2025sims,yang2025cambrian,cai2025scaling}. However, these efforts are insufficient to resolve the persistent bottlenecks in 3D perception and spatial reasoning \citep{fu2024blink,huang2025beacon3d,yu2025far}. In contrast, we contend that more critical issues reside in the \ac{sft} paradigm, \eg, suboptimal objectives for perception tasks and overfitting issues for reasoning tasks. We address these bottlenecks through our meticulously designed 3D-RFT framework.

\paragraph{Reinforcement Learning for \acp{mllm}.}
\ac{rlvr} has catalyzed significant breakthroughs in reasoning capabilities of \acp{llm} \citep{jaech2024openai-o1,shao2024deepseekmath,guo2025deepseekr1,yu2025dapo,zheng2025group,liu2025understanding,liu2025part,yue2025does,wen2025reinforcement} and \acp{mllm} \citep{huang2025vision,liu2025visual,sarch2025grounded,tan2025reason}. Recent efforts in video-based models have advanced \ac{cot} data generation \citep{fei2024video,wu2024mind,shao2024visual,linghu2025scenecot,chen2025reasoning,zheng2025spatialreasoner} and employed \ac{rlvr} for video understanding \citep{feng2025video,li2025videochat,wang2025videorft}, 3D perception \citep{yuan2025scener1,huang20253d,wang2025n3d,ma2025spatialreasoner}, spatial reasoning \citep{zhan2025actial,chen2025think,wu2025spatial,wu2025reinforcing}, and embodied intelligence \citep{qi2025vln,zhao2025embodied,yuan2025embodied}. However, the efficacy of \ac{rlvr} for 3D scene understanding tasks is under-explored \citep{liao2025improved,zhong2025rethinking}. A concurrent work, VST~\citep{yang2025visual}, explores \ac{rlvr} for training a generalist video model on various spatial tasks. In contrast, we present a systematic study of \ac{rlvr} spanning 3D perception, spatial-temporal grounding, and spatial reasoning, offering insights into learning objectives, model components, data configurations, and training dynamics.

\section{Methodology}
\label{sec:method}

\subsection{Preliminary}
\label{sec:method:pre}

\textbf{Reinforcement Learning with Verifiable Rewards (RLVR).} \ac{rlvr} has emerged as a powerful paradigm for enhancing reasoning in \acp{llm} \cite{guo2025deepseekr1,team2025kimik1.5}. Unlike \ac{sft} that maximizes the likelihood of expert data, \ac{rlvr} maximizes the expected value of a verifiable outcome. Given an input prompt $\mathbf{x}$, the policy $\pi_{\theta}$ generates a response $\mathbf{y}$, which is evaluated by a deterministic verifier $\mathcal{V}$ to yield a reward $R=\mathcal{V}(\mathbf{x},\mathbf{y})$. The objective is to maximize the expected reward over the data distribution:
\begin{equation}
    J(\theta) = \mathbb{E}_{\mathbf{x}\sim\mathcal{D},\mathbf{y}\sim \pi_{\theta}(\cdot |\mathbf{x})}[R].
\end{equation}
\textbf{Group Relative Policy Optimization (GRPO).} \ac{grpo} \cite{shao2024deepseekmath} is a memory-efficient variant of PPO \citep{schulman2017proximal} that eliminates the need for a separate critic network. Instead of learning a value function to estimate the baseline, \ac{grpo} samples a group of outputs $\{ \mathbf{y_1}, \dots, \mathbf{y_G} \}$ for each prompt $\mathbf{x}$ from the old policy $\pi_{\theta_{old}}$. The advantage $A_i$ for the $i$-th sample is then computed by normalizing its reward against the group statistics:
\begin{equation}
\label{equation:advantage}
    A_i = \frac{R_i - \text{mean}(\{R_1, \dots, R_G\})}{\text{std}(\{R_1, \dots, R_G\})},
\end{equation}
where $G$ is the group size. This approach significantly reduces memory overhead by removing the value model.

\begin{figure*}[t]
\centering
\includegraphics[width=\textwidth]{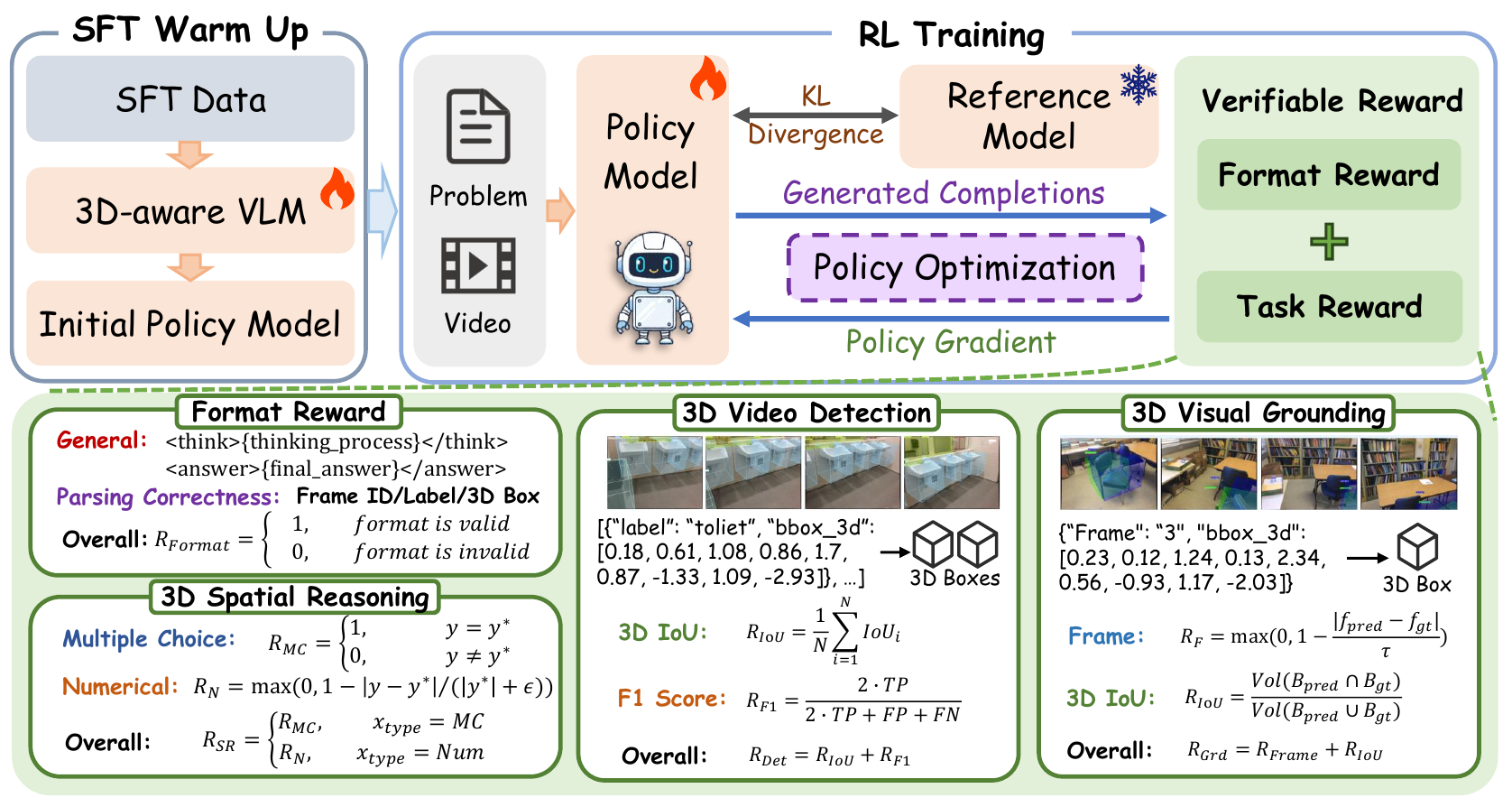}
\caption{\textbf{Overview of the 3D-RFT training framework.} 
The process consists of two main stages: 
(1) \textbf{SFT Warm Up}: Initial training of the 3D-aware VLM using SFT data to establish a baseline policy. 
(2) \textbf{RL Training}: The Policy Model generates completions for video-based problems. A Verifiable Reward is calculated based on \textbf{Format Reward} (adherence to structured output) and \textbf{Task Reward} (performance in 3D Video Detection, 3D Visual Grounding, and Spatial Reasoning using metrics like 3D IoU and F1 Score). The model is optimized via Policy Gradient while maintaining a KL Divergence constraint relative to the frozen Reference Model.}
\label{fig:framework}
\vspace{-1em}
\end{figure*}

\subsection{Task Formulation} 
\label{sec:task_formulation}


We formulate \vsu as a conditional text generation problem. Given the visual input $\mathbf{I}$ (composed of RGB video frames) and a textual query $\mathbf{x}$, the model generates a response sequence $\mathbf{y}$. To facilitate verifiable reward computation, we enforce a structured output format: the model must first generate a reasoning chain enclosed in \texttt{<think>} tags, followed by the final prediction within \texttt{<answer>} tags. We consider two perception tasks (3D video detection and 3D visual grounding) and a reasoning task (3D spatial reasoning). For \percep tasks, the final answer is formatted as a JSON object where 3D bounding boxes are represented as 9-DoF tuples: $\mathbf{b} = (x, y, z, w, h, d, \psi, \theta, \phi)$.

\paragraph{3D Video Detection.}
The goal is to detect all objects $\mathcal{O} = \{(\mathbf{b_i}, c_i)\}$ appearing throughout the image frames, where $\mathbf{b_i}$ refers to the 9-DoF bounding box and $c_i$ refers to the object category. All the bounding boxes are unified into the \textbf{coordinate system of the first frame}. 

\paragraph{3D Visual Grounding.}
Given a grounding text, the model predicts a tuple $(f_{\text{pred}}, \mathbf{b_{\text{pred}}})$, identifying both the frame index and the object's 9-DoF bounding box. Unlike detection, $\mathbf{b_{\text{pred}}}$ is defined in the \textbf{local coordinate system of the predicted frame $f_{\text{pred}}$}.

\paragraph{3D Spatial Reasoning.}
Given a question, the model must generate accurate text answers. This task mainly focuses on spatial attributes and relations within 3D scenes.

\subsection{Training Pipeline}
\label{sec:training_pipeline}

As illustrated in Figure \ref{fig:framework}, our training pipeline consists of two stages: (1) \textbf{SFT Warm Up}, which activates the model's capabilities of 3D scene understanding and establishes an initial policy model; and (2) \textbf{RL Training}, where the model is optimized using strictly verifiable rewards. This second stage leverages policy gradients derived from both format constraints and task-specific metrics (\eg, 3D IoU) to directly refine the model's performance on \vsu tasks.

\subsubsection{Stage 1: SFT Warm Up}
\label{sec:training_pipeline:sft}
Before \ac{rl} training, the model must learn to follow the required output format (\ie, enclosing reasoning in \texttt{<think>} tags and presenting the final answer in proper format) and acquire basic capabilities to provide a good starting point for the subsequent \ac{rl} stage. In the \ac{sft} stage, the model is trained to maximize the log-likelihood of the ground-truth response $\mathbf{y^*}$ given the visual input $\mathbf{I}$ and textual query $\mathbf{x}$:
\begin{equation}
    \mathcal{L}_{\text{SFT}}(\theta) = -\sum_{t=1}^{T} \log \pi_\theta(\mathbf{y^*_t} \mid \mathbf{x}, \mathbf{I}, \mathbf{y^*_{<t}}).
\end{equation}

\subsubsection{Stage 2: \ac{rl} Training}
\label{sec:training_pipeline:rft}

In the second stage, we proceed to refine the model by employing \ac{grpo} with verifiable rewards. For each visual-textual input $(\mathbf{I}, \mathbf{x})$, we sample a group of $G$ outputs $\{\mathbf{y_1}, \dots, \mathbf{y_G}\}$ from the current policy $\pi_{\theta_{\text{old}}}$. The advantage $A_i$ for each sample is computed by normalizing the rewards within the group. The objective maximizes the expected advantage while constraining the policy shift via a KL-divergence penalty with the reference model $\pi_{\text{ref}}$:
{\small
\begin{equation}
    \mathcal{L}_{\text{GRPO}}(\theta) = - \frac{1}{\sum_{i=1}^G T_i} \sum_{i=1}^{G} \sum_{t=1}^{T_i} \mathcal{L}_{i,t}(\theta) + \beta \mathds{D}_\text{KL}[\pi_\theta || \pi_{\text{ref}}],
\end{equation}
\vspace{-5pt}
\begin{equation}
    \mathcal{L}_{i,t}(\theta) = \min\bigg( r_{i,t} A_{i}, \text{clip}\left( r_{i,t}, 1-\epsilon, 1+\epsilon \right) A_{i} \bigg),
\end{equation}
\vspace{-5pt}
\begin{equation}
    r_{i,t} = \frac{\pi_\theta(\mathbf{y_{i,t}} | \mathbf{I_i}, \mathbf{x_i}, \mathbf{y_{i,<t}})}{\pi_{\theta_{\text{old}}}(\mathbf{y_{i,t}} | \mathbf{I_i}, \mathbf{x_i}, \mathbf{y_{i,<t}})}.
\end{equation}
}

The advantage $A_i$ is defined in \cref{equation:advantage}. To mitigate the high memory cost of long video contexts during training, we adopt a loss chunking technique (see details in \cref{appendix:efficient_training}).

\subsection{Verifiable Reward Design}
\label{sec:rlvr}


As shown in \Cref{fig:framework}, our rewards include two components: a \textbf{Format Reward} to enforce structural validity (\eg, correct JSON syntax and bounding box tuples), and \textbf{Task Rewards} that directly optimize geometric and semantic metrics. We will elaborate on \textbf{Task Rewards} as follows.


\subsubsection{3D Video Detection: 3D IoU and F1-Score}
\label{sec:rlvr:3d_detection}

In the 3D video detection task, the model outputs a set of predicted bounding boxes $\mathcal{B}_{\text{pred}} = \{\mathbf{b_1}, \dots, \mathbf{b_N}\}$. Each predicted bounding box $\mathbf{b_i}$ is directly compared against the ground-truth labels $\mathcal{B}_{\text{gt}}$ to derive the maximum 3D \ac{iou} $\mathcal{I}_i$.
We adopt \textbf{Average IoU Reward} to provide a dense learning signal by averaging the maximum \acp{iou} across all predictions:
\begin{equation}
    R_{\text{IoU}}^{(\text{Det})} = \frac{1}{N} \sum_{i=1}^{N} \mathcal{I}_i.
\end{equation}




To further enhance detection capability, we introduce the \textbf{F1-Score Reward}. A prediction is counted as a True Positive (TP) if it matches an unmatched ground-truth bounding box with a 3D \ac{iou} exceeding $\tau_{\text{F1}}$ (set to $0.25$ by default); otherwise, it is treated as a False Positive (FP). Ground-truth bounding boxes that remain unmatched are regarded as False Negatives (FN). The F1-Score Reward is then computed as:
\begin{equation}
    R_{\text{F1}} = \frac{2 \cdot \text{TP}}{2 \cdot \text{TP} + \text{FP} + \text{FN}}.
\end{equation}
The \textbf{Average \ac{iou} Reward} ensures valid geometric precision, while the \textbf{F1-Score Reward} directly optimizes the final evaluation metric. The combined verifiable reward for 3D video detection is: 
\begin{equation}
    R_{\text{Det}} = R_{\text{IoU}}^{(\text{Det})} + R_{\text{F1}}.
\end{equation}

\subsubsection{3D Visual Grounding: Frame and 3D IoU}
\label{sec:rlvr:3d_visual_grounding}
Video-based \vigrd requires spatio-temporal localization of a specific object according to the grounding text. We consider both temporal and spatial precision.

\paragraph{Temporal Reward.}
Locating the target frame index $f_{\text{gt}}$ is formulated as a regression task. To provide a dense optimization signal, we employ a smoothed linear decay function based on the absolute temporal distance between the predicted frame index $f_{\text{pred}}$ and the ground truth, governed by a tolerance threshold $\tau_{\text{frame}}$ (set to $5$ by default):
\begin{equation}
    R_{\text{frame}} = \max \left( 0, 1 - \frac{|f_{\text{pred}} - f_{\text{gt}}|}{\tau_{\text{frame}}} \right).
\end{equation}
\paragraph{\ac{iou} Reward.}
To align with the evaluation metric, we compute the 3D \ac{iou} in the global coordinate system. Given the predicted bounding box $\mathbf{b_{\text{pred}}}$ in the local camera coordinates of frame $f_{\text{pred}}$, we first transform it to the global scene coordinates using the frame's extrinsic matrix $M_{c \to g}^{(f_{\text{pred}})}$ and the scene's axis-alignment matrix $M_{\text{align}}$: $\mathbf{b_{\text{pred}}'} = M_{\text{align}} M_{c \to g}^{(f_{\text{pred}})} \mathbf{b_{\text{pred}}}$. The \ac{iou} reward is then calculated by standard 3D IoU between the aligned prediction and the ground-truth box $\mathbf{b_{\text{gt}}}$:
\begin{equation}
    R_{\text{IoU}}^{(\text{Grd})} = \frac{|\mathbf{b_{\text{pred}}'} \cap \mathbf{b_{\text{gt}}}|}{|\mathbf{b_{\text{pred}}'} \cup \mathbf{b_{\text{gt}}}|}.
\end{equation}
The final reward for 3D visual grounding is defined as: 
\begin{equation}
    R_{\text{Grd}} = R_{\text{frame}} + R_{\text{IoU}}^{(\text{Grd})}.
\end{equation}
\subsubsection{3D Spatial Reasoning: Accuracy}
\label{sec:rlvr:reasoning}

For the \spr task, we design the accuracy reward $R_{\text{acc}}$ by dispatching task-specific verifiers based on the question type:
\begin{itemize}[leftmargin=*,nolistsep]
    \item \textbf{Multiple Choice.} For multi-choice scenarios, we use an exact-match indicator function: 
\begin{equation}
    R_{\text{MC}} = \mathds{1}(\mathbf{y} = \mathbf{y^*})
\end{equation}
    \item \textbf{Numerical Reasoning.} For numerical answers (\eg, counting), we employ the Mean Relative Accuracy (MRA) with $\mathcal{C} = \{0.50,0.55,\dots,0.95\}$ \citep{yang2025thinking}:
\begin{equation}
    R_{\text{num}} = \frac{1}{10} \sum\limits_{\tau_{\text{num}} \in \mathcal{C}} \mathds{1}\left(\frac{|\mathbf{y} - \mathbf{y^*}|}{|\mathbf{y^*}|} < 1 - \tau_{\text{num}} \right).
\end{equation}
\end{itemize}

\section{Experiments}

\label{sec:experiments}

\begin{table*}[ht]
    \centering
    \caption{\textbf{Quantitative results on \scannetdet.} In this table, we present the comparison with baseline models and report the performance improvement between the \ac{sft} baseline (VG LLM-4B) and \model (ours). }
    \label{tab:scannet_det_results}
    \small 
    \resizebox{0.97\textwidth}{!}{%
    \begin{tabular}{l|cccccccc|ccc}
        \toprule
        \multicolumn{1}{c|}{\multirow{2}{*}{\textbf{Model}}} & \multirow{2}{*}{\textbf{chair}} & \multirow{2}{*}{\textbf{cabinet}} & \multirow{2}{*}{\textbf{table}} & \multirow{2}{*}{\textbf{bin}} & \multirow{2}{*}{\textbf{couch}} & \multirow{2}{*}{\textbf{bed}} & \multirow{2}{*}{\textbf{bathtub}} & \multirow{2}{*}{\textbf{toilet}} & \multicolumn{3}{c}{\textbf{20 Common Classes}} \\
        \cmidrule{10-12}
        & & & & & & & & & $P_{25}$ & $R_{25}$ & $F1_{25}$ \\
        \midrule
        \multicolumn{12}{c}{\textit{4-Frame Setting}} \\
        \midrule
        Qwen2.5-VL-3B \citep{zheng2025vgllm} & 37.7 & 10.2 & 35.0 & 23.1 & 39.0 & 64.8 & 32.4 & 68.8 & 32.6 & 27.9 & 30.0 \\
        Qwen2.5-VL-7B \citep{zheng2025vgllm} & 41.2 & 11.6 & 36.5 & 30.2 & 41.1 & 68.2 & 36.6 & 68.7 & 34.6 & 31.0 & 32.5 \\
        VG LLM-4B \citep{zheng2025vgllm} & 49.7 & 13.1 & 41.3 & 39.2 & 44.6 & 71.2 & 33.5 & 83.4 & 41.7 & 35.7 & 38.2 \\
        VG LLM-8B \citep{zheng2025vgllm} & 54.0 & 17.1 & 46.5 & \textbf{39.8} & 47.0 & 74.1 & 42.1 & 82.5 & 43.4 & \textbf{39.6} & 41.2 \\
        3D-RFT-4B (ours) & \textbf{55.3}  & \textbf{18.7} & \textbf{48.2} & 39.1 & \textbf{49.8}  &  \textbf{77.1} &  \textbf{50.0}  &  \textbf{86.1} & \textbf{54.2}  &  38.2  &  \textbf{43.7} \\
        \textit{Improvement} & \color{ForestGreen}{+5.6} & \color{ForestGreen}{+5.6} & \color{ForestGreen}{+6.9} & -0.1 & \color{ForestGreen}{+5.2} & \color{ForestGreen}{+5.9} & \color{ForestGreen}{+16.5} & \color{ForestGreen}{+2.7} & \color{ForestGreen}{+12.5} & \color{ForestGreen}{+2.5} & \color{ForestGreen}{+5.5} \\
        \midrule
        \multicolumn{12}{c}{\textit{6-Frame Setting}} \\
        \midrule
        Qwen2.5-VL-3B \citep{zheng2025vgllm} & 32.8 & 7.8 & 31.3 & 20.9 & 32.2 & 58.8 & 36.5 & 66.1 & 27.8 & 24.1 & 25.7 \\
        Qwen2.5-VL-7B \citep{zheng2025vgllm} & 36.1 & 10.6 & 32.7 & 25.0 & 40.7 & 64.6 & 38.4 & 68.6 & 31.8 & 28.0 & 29.6 \\
        VG LLM-4B \citep{zheng2025vgllm} & 41.6 & 12.4 & 39.8 & 33.1 & 45.0 & 70.2 & 33.8 & 80.6 & 39.7 & 34.0 & 36.4 \\
        VG LLM-8B \citep{zheng2025vgllm} & 48.7 & 17.9 & 44.8 & \textbf{38.5} & 46.4 & \textbf{75.8} & 40.4 & \textbf{83.2} & 43.5 & \textbf{38.7} & 40.8 \\
        3D-RFT-4B (ours) & \textbf{49.2}  & \textbf{18.5}  & \textbf{44.9} & 34.6 &  \textbf{48.3}  &  74.1 & \textbf{51.4} & 81.1 & \textbf{53.4} &  35.7 &  \textbf{41.7} \\
        \textit{Improvement} & \color{ForestGreen}{+7.6} & \color{ForestGreen}{+6.1} & \color{ForestGreen}{+5.1} & \color{ForestGreen}{+1.5} & \color{ForestGreen}{+3.3} & \color{ForestGreen}{+3.9} & \color{ForestGreen}{+17.6} & \color{ForestGreen}{+0.5} & \color{ForestGreen}{+13.7} & \color{ForestGreen}{+1.7} & \color{ForestGreen}{+5.3}   \\
        \bottomrule
    \end{tabular}
     }
     \vspace{-1em}
\end{table*}

\begin{table}[t!]
    \centering
    \caption{\textbf{Quantitative results on ScanRefer.} The content in ``()'' indicates results with proposal refinement \citep{zheng2025vgllm}.}
    \small
    \resizebox{\linewidth}{!}{%
        \begin{tabular}{l c | c c}
            \toprule
            \multicolumn{1}{c}{\multirow{2}{*}{\textbf{Model}}} & \textbf{3D Scene} & \multirow{2}{*}{\textbf{Acc@0.25}} & \multirow{2}{*}{\textbf{Acc@0.5}} \\
            & \textbf{Input} & & \\
            \midrule
            ScanRefer \citep{chen2020scanrefer} & \checkmark & 37.3 & 24.3 \\
            MVT \citep{huang2022multi} & \checkmark & 40.8 & 33.3 \\
            ViL3DRel \citep{chen2022language} & \checkmark & 47.9 & 37.7 \\
            3D-LLM \citep{chen2024grounded3dllm} & \checkmark & 30.3 & - \\
            Chat-3D v2 \citep{huang2024chatscene} & \checkmark & 35.9 & 30.4 \\
            Grounded 3D-LLM \citep{chen2024grounded3dllm} & \checkmark & 47.9 & 44.1 \\
            ChatScene \citep{huang2024chatscene} & \checkmark & 55.5 & 50.2 \\
            LLaVA-3D \citep{zhu2025llava3d} & \checkmark & 54.1 & 42.4 \\
            Video-3D LLM \citep{zheng2025video} & \checkmark & \textbf{58.1} & \textbf{51.7} \\
            \midrule
            SPAR \citep{zhang2025flatland} & \ding{55} & 31.9 (48.8) & 12.4 (43.1) \\
            VG LLM-4B \citep{zheng2025vgllm} & \ding{55} & 36.4 (53.5) & 11.8 (47.5) \\
            VG LLM-8B \citep{zheng2025vgllm} & \ding{55} & 41.6 (57.6) & 14.9 (50.9) \\
            3D-RFT-4B (ours) & \ding{55} & \textbf{42.9} (54.6)  & \textbf{15.9} (48.1)  \\
            \textit{Improvement} &\ding{55} & \color{ForestGreen}{+6.5} (\color{ForestGreen}{+1.1}) &  \color{ForestGreen}{+4.1} (\color{ForestGreen}{+0.6})   \\
            \bottomrule
        \end{tabular}
    }
    \label{tab:scanrefer_results}
    \vspace{-1.2em}
\end{table}


In this section, we experiment with applying 3D-RFT to 3D perception and spatial reasoning tasks. We present the results and analyses of \percep tasks in \cref{sec:experiments:perception}, and \spr task in \cref{sec:experiments:spatial_reasoning}. Additionally, we present the training dynamics of both tasks in \cref{sec:exp:dynamics} to visualize the model's behaviors during 3D-RFT.


\paragraph{Model.} We build our model \model based on VG LLM-4B \citep{zheng2025vgllm}, which consists of an \ac{mllm} backbone, Qwen2.5-VL-3B-Instruct \citep{Qwen2.5-VL}, and a visual geometry backbone VGGT-1B \citep{wang2025vggt}. Features extracted from the VGGT backbone are first processed to align with the Qwen visual feature structure, and then combined via element-wise addition, producing a hybrid visual representation that is fed into the Qwen LLM.

\subsection{3D Perception Tasks}
\label{sec:experiments:perception}


\paragraph{Training Datasets.} For \percep tasks, we follow the settings of \vgllmpaper \citep{zheng2025vgllm}, utilizing \scanrefer \citep{chen2020scanrefer}, \scancap \citep{chen2021scan2cap}, and \scannetdet \citep{wang2024embodiedscan} for the \ac{sft} stage. We have not included \ac{cot} data in this stage due to the difficulty of acquiring high-quality data. In the second stage, we conduct \ac{rft} using \scannetdet and \scanrefer for \videt and \vigrd, respectively. We provide a detailed description of the data in \cref{appendix:training_data}.

\paragraph{Comparison Baselines and Evaluation Metrics.} For 3D perception tasks, we mainly compare our model (\ie, 3D-RFT-4B) with \vgllmpaper \citep{zheng2025vgllm} to demonstrate the effectiveness of 3D-\ac{rft}. Our evaluation metrics follow \vgllmpaper: for \scanrefer, we report the accuracy at IoU thresholds of $0.25$ and $0.5$; for \scannetdet, we report precision, recall, and F1-score for 20 common object classes at an IoU threshold of $0.25$. We use \texttt{lmms\_eval} \citep{zhang2025lmms} for evaluation and keep the same evaluation hyper-parameters as \vgllmpaper.

\paragraph{Results.} We report the detailed evaluation results of 3D video detection and 3D visual grounding in \cref{tab:scannet_det_results,tab:scanrefer_results}, respectively. The results show that \model achieves \sota performances on 3D perception tasks. We provide further interpretations and analyses as follows.

\subsubsection{3D Video Detection}


\begin{table}[t!]
    \centering
    \caption{\textbf{Ablation study of training strategies and 3D priors.} We report accuracy at IoU thresholds 0.25 and 0.5.}
    \label{tab:ablation}
    \setlength{\tabcolsep}{8pt} 
    \resizebox{0.9\linewidth}{!}{
    \begin{tabular}{llcc}
        \toprule
        \multirow{2}{*}{\begin{tabular}{@{}l@{}}\textbf{Training}\\\textbf{Strategy}\end{tabular}} & \multirow{2}{*}{\begin{tabular}{@{}l@{}}\textbf{3D}\\\textbf{Prior}\end{tabular}} & \multicolumn{2}{c}{\textbf{ScanRefer}} \\
        \cmidrule(lr){3-4}
         & & \textbf{Acc@0.25} & \textbf{Acc@0.5} \\
        \midrule
        SFT & None & 31.9 (49.9) & 9.3 (43.8) \\
        SFT $\to$ SFT & None & 34.2 (50.6) & 10.4 (44.9) \\
        SFT $\to$ RL & None & 38.2 (52.7) & 12.1 (46.6) \\
        SFT & VGGT & 36.4 (53.5) & 11.8 (47.5) \\
        SFT $\to$ RL & VGGT & \textbf{42.9} (54.6) & \textbf{15.9} (48.1) \\
        \bottomrule
    \end{tabular}
    }
    \vspace{-1.2em}
\end{table}

\textbf{3D-RFT significantly enhances detection performance over the SFT baseline.} As detailed in \cref{tab:scannet_det_results}, \model achieves substantial gains across all metrics. Under the \textit{4-frame setting}, our method improves Precision by +12.5\%, Recall by +2.5\%, and F1-Score by +5.5\% compared to the SFT baseline (VG LLM-4B). The improvements are robust across settings, with the \textit{6-frame setting} yielding even higher gains in Precision (+13.7\%) and comparable boosts in F1 (+5.3\%). Notably, the improvements are most pronounced for larger objects, such as ``bathtub'' (+16.5\%) and ``table'' (+6.9\%), whereas smaller objects like ``bin'' show more limited gains, suggesting that further increases in visual resolution could potentially benefit small-object detection.

\begin{table*}[t!]
    \centering
    \caption{\textbf{Quantitative results on VSI-Bench.} We include zero-shot results of base models, and test results of models after \ac{sft} and \ac{rft}.}
    \small
    \newcommand{\rot}[1]{\rotatebox{0}{#1}} 
    \setlength{\tabcolsep}{4pt} 
    \renewcommand{\arraystretch}{1.0}
    \resizebox{\linewidth}{!}{%
    \begin{tabular}{l m{28pt} c c c c c c c c}
        \toprule
        \multirow{2}{*}{\textbf{Model}} & \multirow{2}{*}{\textbf{Avg.}} & \multicolumn{4}{c}{\cellcolor[HTML]{FFF2E6} Numerical Answer} & \multicolumn{4}{c}{\cellcolor[HTML]{FFFFE6} Multiple-Choice Answer} \\
         & & \rot{Obj. Count} & \rot{Abs. Dist.} & \rot{Obj. Size} & \rot{Room Size} & \rot{Rel. Dist.} & \rot{Rel. Dir.} & \rot{Route Plan} & \rot{Appr. Order} \\
        \midrule
        \rowcolor[HTML]{F0F7FF} \textit{Base Models} & & & & & & & & & \\
        \rule{0pt}{2.2ex}GPT-4o \citep{hurst2024gpt4o} & 34.0 & 46.2 & 5.3 & 43.8 & 38.2 & 37.0 & 41.3 & 31.5 & 28.5 \\
        Gemini-2.5 Pro \citep{comanici2025gemini2.5} & 51.5 & 43.8 & 34.9 & 64.3 & 42.8 & 61.1 & 47.8 & \textbf{45.9} & 71.3 \\
        LLaVA-Video-7B \citep{zhang2025video} & 35.6 & 48.5 & 14.0 & 47.8 & 24.2 & 43.5 & 42.4 & 34.0 & 30.6 \\
        Qwen2.5-VL-7B \citep{Qwen2.5-VL} & 32.7 & 34.5 & 19.4 & 47.6 & 40.8 & 32.8 & 24.5 & 32.5 & 29.4 \\
        InternVL3-8B \citep{zhu2025internvl3} & 42.1 & 68.1 & 39.0 & 48.4 & 33.6 & 48.3 & 36.4 & 27.3 & 35.4 \\
        \midrule
        \rowcolor[HTML]{F0F7FF} \textit{Supervised Fine-Tuned} & & & & & & & & & \\
        \rule{0pt}{2.2ex}VG LLM-4B \citep{zheng2025vgllm} & 47.3 & 66.0 & 37.8 & 55.2 & 59.2 & 44.6 & 45.6 & 33.5 & 36.4 \\
        VLM-3R-7B \citep{fan2025vlm} & 60.9 & 70.2 & 49.4 & 69.2 & \textbf{67.1} & \textbf{65.4} & \textbf{80.5} & \textbf{45.4} & 40.1 \\
        Cambrian-S-3B \citep{yang2025cambrian} & 57.3 & 70.7 & 40.6 & 68.0 & 46.3 & 64.8 & 61.9 & 27.3 & \textbf{78.8} \\
        VST-SFT-3B \citep{yang2025visual} & 57.9 & 69.3 & 45.4 & 71.8 & 62.4 & 59.0 & 46.0 & 38.7 & 70.2 \\
        \midrule
        \rowcolor[HTML]{F0F7FF} \textit{Reinforcement Fine-Tuned} & & & & & & & & & \\
        \rule{0pt}{2.2ex}vsGRPO-2B \citep{liao2025improved} & 35.4 & 53.6 & 29.0 & 52.7 & 43.4 & 28.1 & 30.9 & 26.8 & 18.9 \\
        vsGRPO-7B \citep{liao2025improved} & 40.7 & 59.9 & 29.6 & 50.8 & 48.3 & 35.4 & 35.6 & 34.0 & 31.5 \\
        SpaceR-7B \citep{ouyang2025spacer} & 43.5 & 61.9 & 28.6 & 60.9 & 35.2 & 38.2 & 46.0 & 31.4 & 45.6 \\
        Spatial-MLLM-4B \citep{wu2025spatial} & 48.4 & 65.3 & 34.8 & 63.1 & 45.1 & 41.3 & 46.2 & 33.5 & 46.3 \\
        ViLaSR-7B \citep{wu2025reinforcing} & 45.4 & 63.5 & 34.4 & 60.6 & 30.9 & 48.9 & 45.2 & 30.4 & 49.2 \\
        SpatialLadder-3B \citep{li2025spatialladder} & 45.7 & 63.5 & 34.3 & 61.7 & 43.9 & 45.4 & 44.8 & 35.6 & 36.4 \\
        VST-RL-3B \citep{yang2025visual} & 57.7 & 66.6 & 45.0 & \textbf{72.8} & 60.9 & 59.9 & 47.6 & 40.7 & 68.3 \\
        \textbf{3D-RFT-4B (ours)} & \textbf{62.8} & \textbf{71.2} & \textbf{53.5} & 70.3 & 63.2 & 60.8 & 77.9 & 37.6 & 67.8 \\
        \bottomrule
    \end{tabular}
    }
    \label{tab:spatial_comparison}
    \vspace{-1em}
\end{table*}

\textbf{\model surpasses the larger \vgllml model.} With only half the parameters, \model consistently outperforms \vgllml on F1-Score and Precision for both \textit{4-frame} and \textit{6-frame} settings. Our model also achieves superior per-class performance on the majority of object categories. This demonstrates that 3D-RFT's metrics-driven optimization exploits the model's potential far more effectively than the standard \ac{sft}, enabling a 4B model to exceed the capabilities of an 8B baseline.

\subsubsection{3D Visual Grounding}



\textbf{3D-RFT significantly enhances performance compared to the SFT baseline.} As shown in \cref{tab:scanrefer_results}, \model achieves substantial gains over the VG LLM-4B baseline, elevating the Acc@IoU0.25 and Acc@IoU0.5 by $+6.5\%$ and $+4.5\%$ respectively. This performance leap confirms that the \ac{rft} stage effectively refines the policy's spatio-temporal grounding capabilities.

\textbf{\model outperforms the larger \vgllml.} With only half the parameters, \model surpasses \vgllml ($42.9\%$ \vs $41.6\%$ on Acc@IoU0.25). We attribute the advantage to the learning objective: while \ac{sft} focuses on maximizing the likelihood of ground-truth labels, 3D-RFT directly optimizes towards more geometrically precise grounding results. This effectively drives the model to learn tighter and more accurate bounding boxes, which is unlikely to emerge from the imitation learning of \ac{sft}.


\subsubsection{Analysis of \ac{rft} and 3D prior}
\label{sec:experiments:ablation_study}



\textbf{The efficacy of \ac{rft} is robust across diverse visual inputs.} We further ablate the VGGT input to investigate whether the efficacy of 3D-RFT is consistent: (1) vanilla \qwenvl (No 3D Prior), and (2) \qwenvl augmented with \vggt. The results demonstrate consistent improvements under both settings, with the \vggt-equipped model improving from $36.4\%$ to $42.9\%$ on Acc@0.25 after \ac{rft}. This confirms that 3D-RFT consistently enhances the model's capability regardless of visual inputs, \eg, the presence of 3D priors.

\subsection{3D Spatial Reasoning}
\label{sec:experiments:spatial_reasoning}

\paragraph{Formulation.} We consider two settings for the spatial reasoning task: (1) \ac{da}, where the model is prompted to directly output the final answer; and (2) \ac{ta}, where the model is prompted to think before giving the final answer. In this task, we conduct \ac{rft} in the \ac{ta} setting following common practices, and experiment with two schemes for the \ac{sft} stage: (1) default SFT (DA+TA), where we use both \ac{da} and \ac{ta} data; and (2) SFT (DA), where we exclude \ac{ta} data (\ie, \ac{cot} data).

\paragraph{Training Data.} We curate a data mixture from \vlmr \citep{fan2025vlm} and SpaceR \citep{ouyang2025spacer}, including \srdatanum data instances in total, termed VSI-298K. Additionally, we generate 10K high-quality \ac{cot} data using Qwen3-VL-32B-Thinking \citep{Qwen3-VL}, termed \ac{cot}-10K. By default, we use the mixture of VSI-298K (\ac{da}) and \ac{cot}-10K (\ac{ta}) for \ac{sft}, and VSI-298K for \ac{rft} (TA).

\begin{figure}[t!]
    \centering
    \begin{minipage}[t]{0.48\linewidth}
        \vspace{0pt} 
        \centering
        \includegraphics[width=\linewidth]{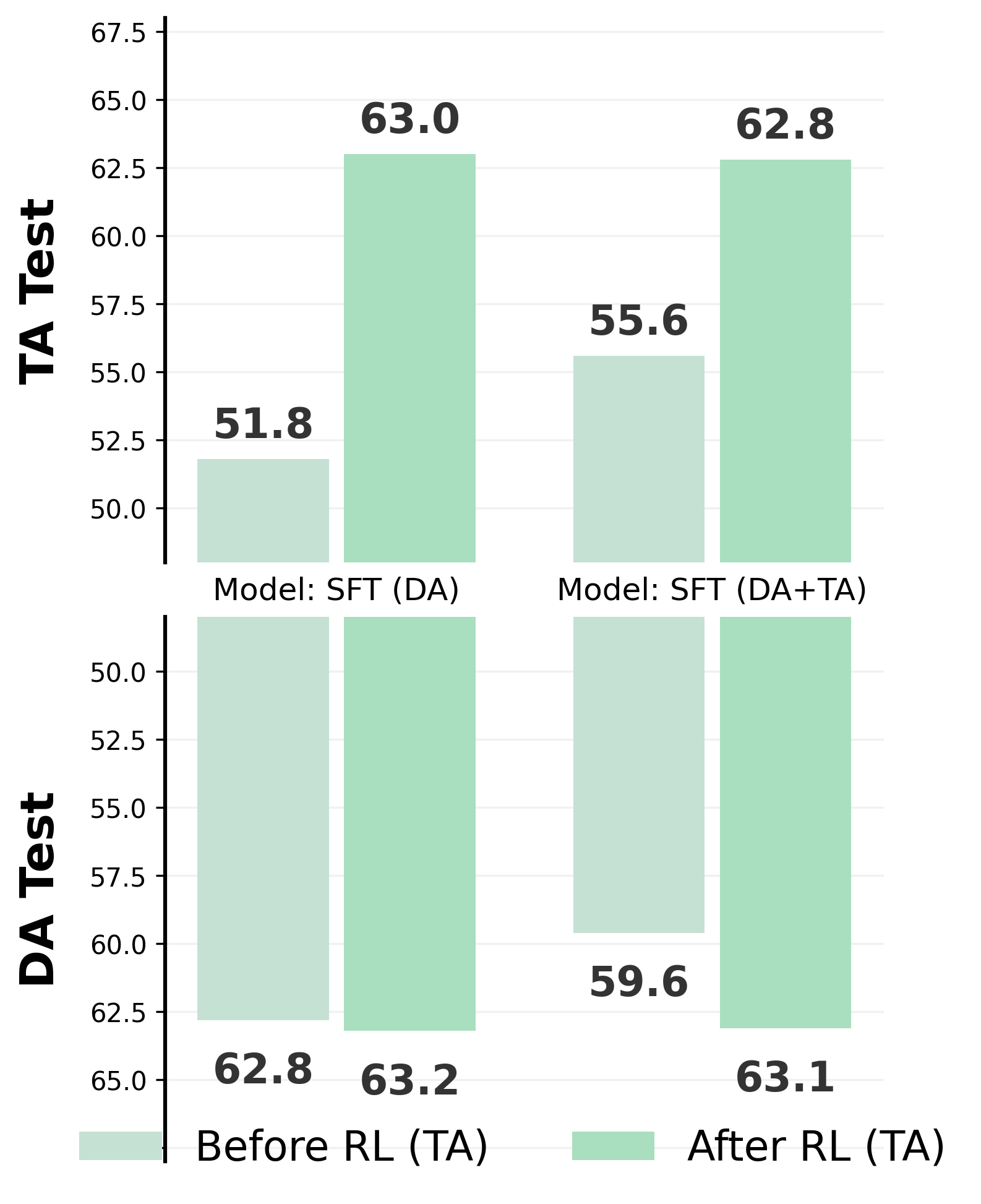}
        \captionof{figure}{\textbf{Evaluation results of SFT and RFT on VSI-Bench.} The results include comparison of SFT checkpoints, RFT efficacy, and TA/DA test settings.}
        \label{fig:spatial_reasoning_rl_ablation}
    \end{minipage}
    \hfill
    \begin{minipage}[t]{0.48\linewidth}
        \vspace{0pt} 
        \centering
        {
        \hspace{-14pt}
        \includegraphics[width=1.07\linewidth]{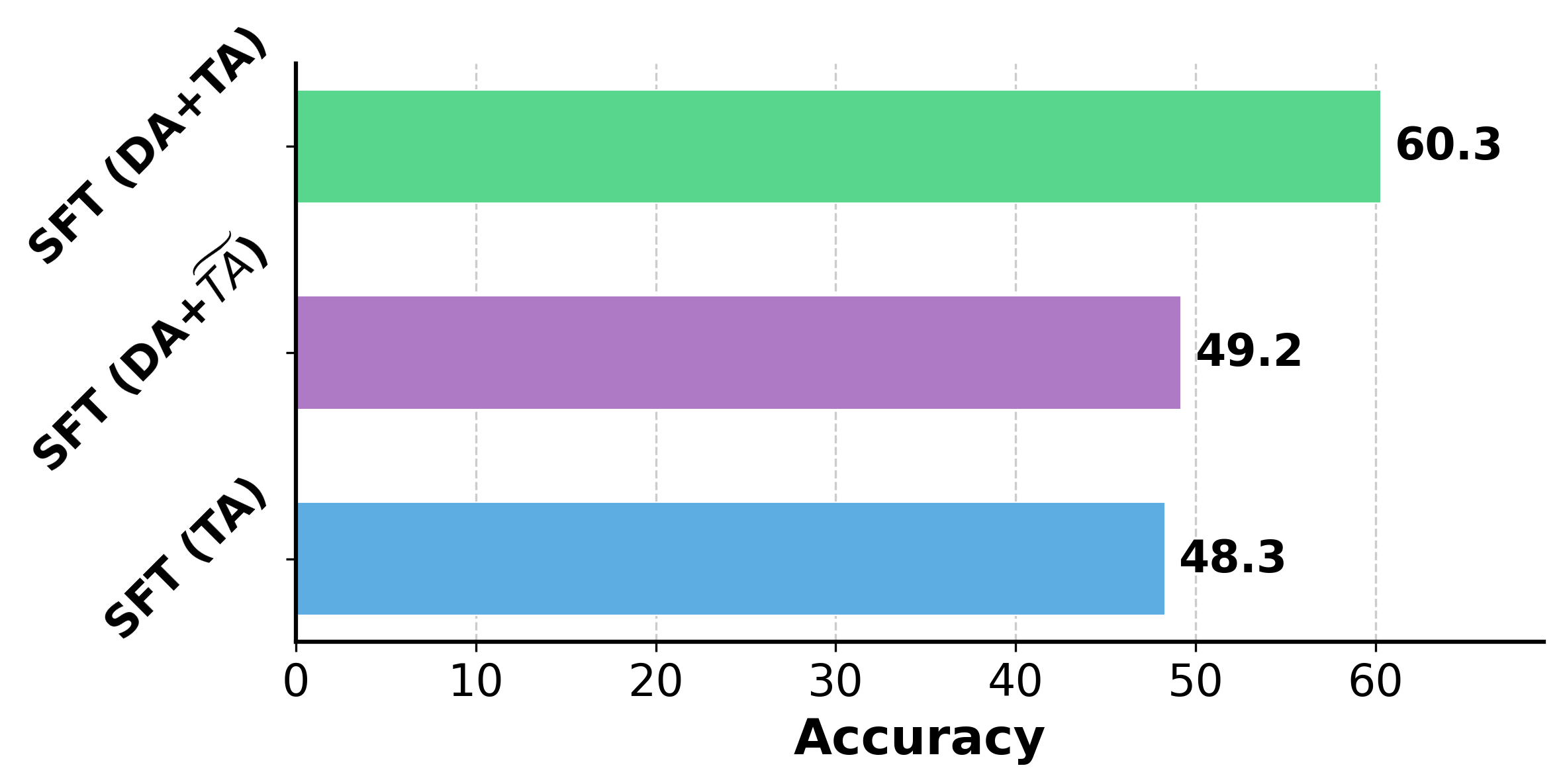}
        }
        \vspace{-1.5em}
        \captionof{figure}{\textbf{VSI-Bench accuracy after 800 steps of RFT.} $\widetilde{\text{TA}}$ denotes lower-quality \ac{ta} data.}
        \par\vspace{0.5em} 
        {
        \hspace{-6pt}
        \includegraphics[width=1.03\linewidth]{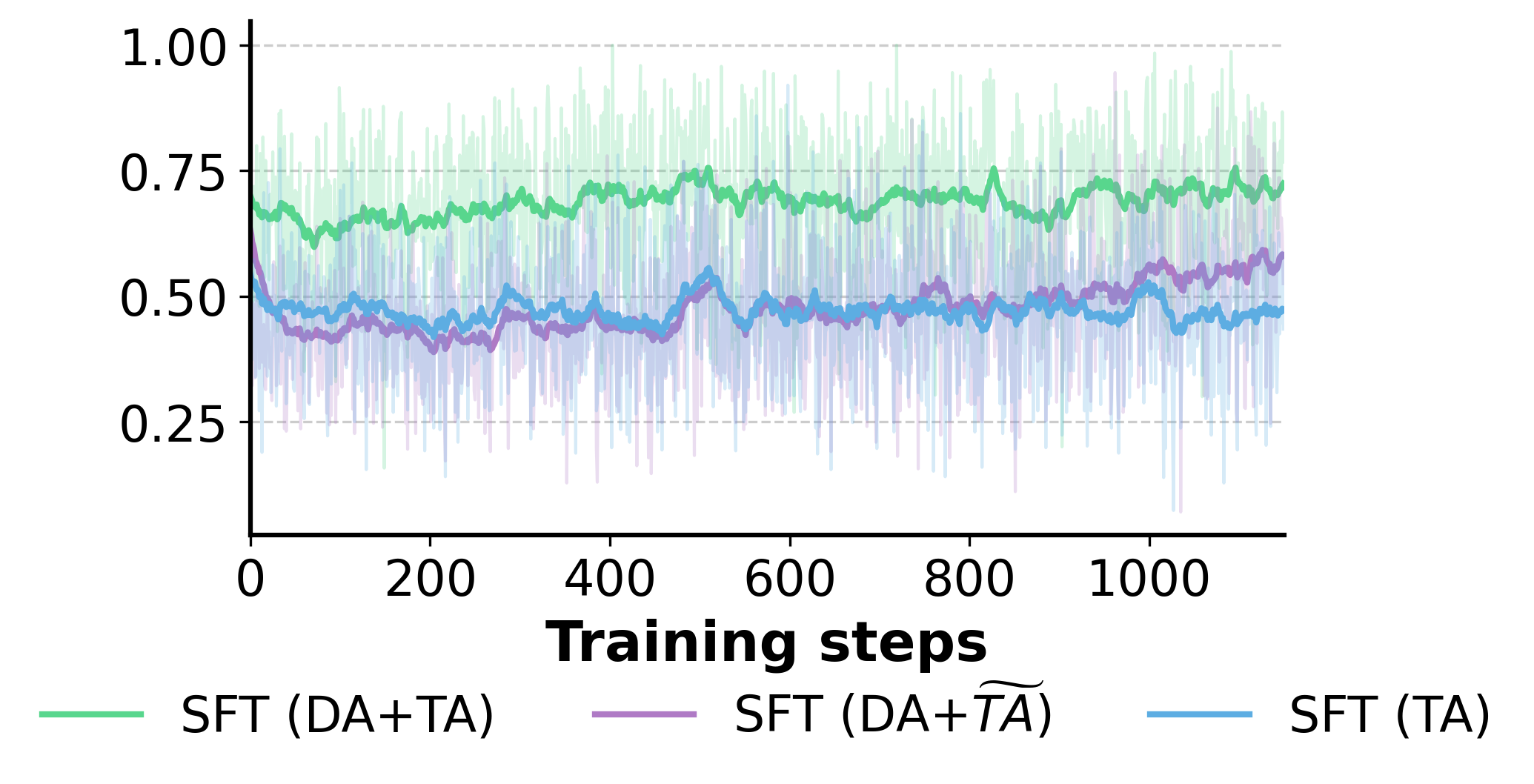}
        }
        \vspace{-1.8em}
        \captionof{figure}{\textbf{Accuracy rewards during RFT.} $\widetilde{\text{TA}}$ denotes lower-quality \ac{ta} data.}
        \label{fig:spatial_reasoning_cot_ablation}
    \end{minipage}
    \vspace{-1.5em}
\end{figure}

\paragraph{Results.} We present the evaluation results on VSI-Bench in \cref{tab:spatial_comparison}. The results show that \model outperforms prior methods by a large margin, especially on numerical reasoning categories. This suggests that \ac{rft} provides more effective learning signals compared to \ac{sft}. We provide additional results on more benchmarks in \cref{sec:additional_exp_vst}, and further analyses on \ac{rft} efficacy and data as follows.

\subsubsection{Analysis of RFT and Data}

\paragraph{3D-RFT consistently improves spatial reasoning performance.} As shown in \cref{fig:spatial_reasoning_rl_ablation}, we present evaluation results regarding two \ac{sft} checkpoints, and compare their performances between ``before'' and ``after'' \ac{rft}. The results demonstrate that \ac{rft} yields consistent improvements on VSI-Bench, especially on \ac{ta} test. Moreover, RFT on \ac{ta} task elicits improvements on \ac{da} task. In addition, we observe that continually training with \ac{sft} yields a moderate performance drop, which suggests the advantage of \ac{rft}.


\begin{figure}[t!]
    \centering
    \begin{subfigure}[b]{0.48\linewidth}
        \centering
        \includegraphics[width=\linewidth]{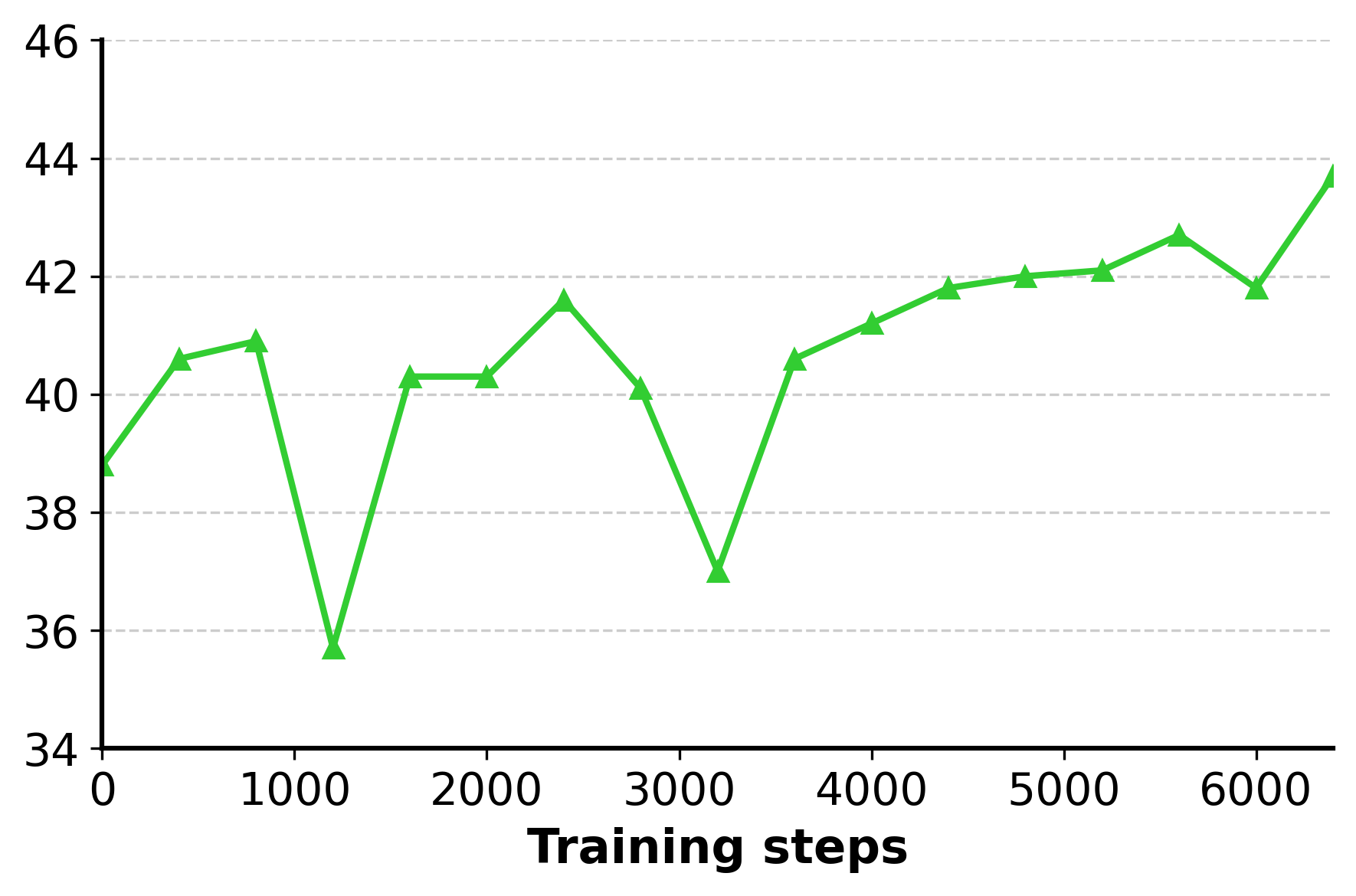}
        \caption{Evaluation F1 Score}
        \label{fig:dyn_eval}
    \end{subfigure}
    \hfill 
    \begin{subfigure}[b]{0.48\linewidth}
        \centering
        \includegraphics[width=\linewidth]{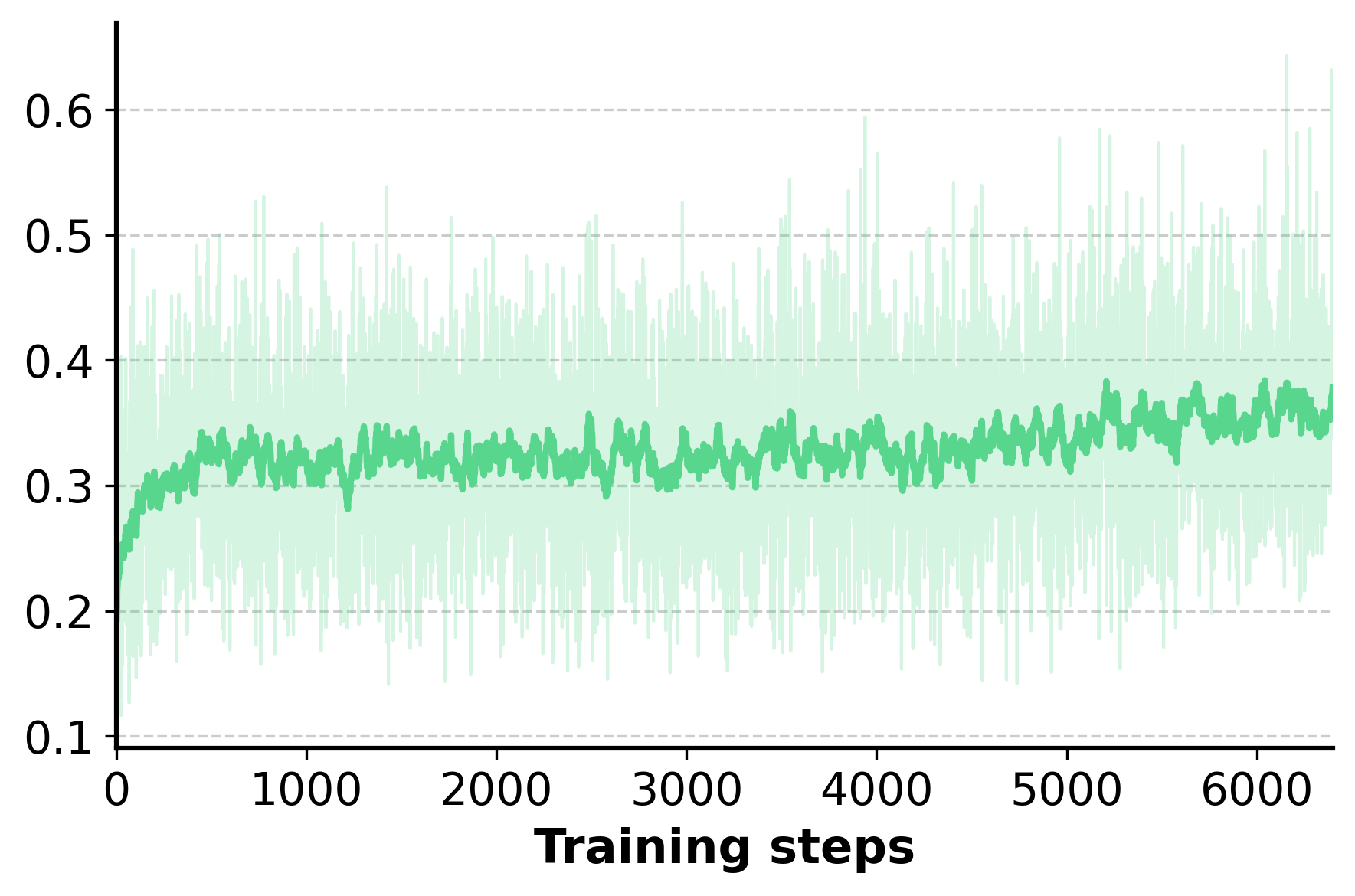}
        \caption{3D Detection F1 Reward}
        \label{fig:dyn_f1_reward}
    \end{subfigure}
    
    \vspace{0.5em} 
    
    \begin{subfigure}[b]{0.48\linewidth}
        \centering
        \includegraphics[width=\linewidth]{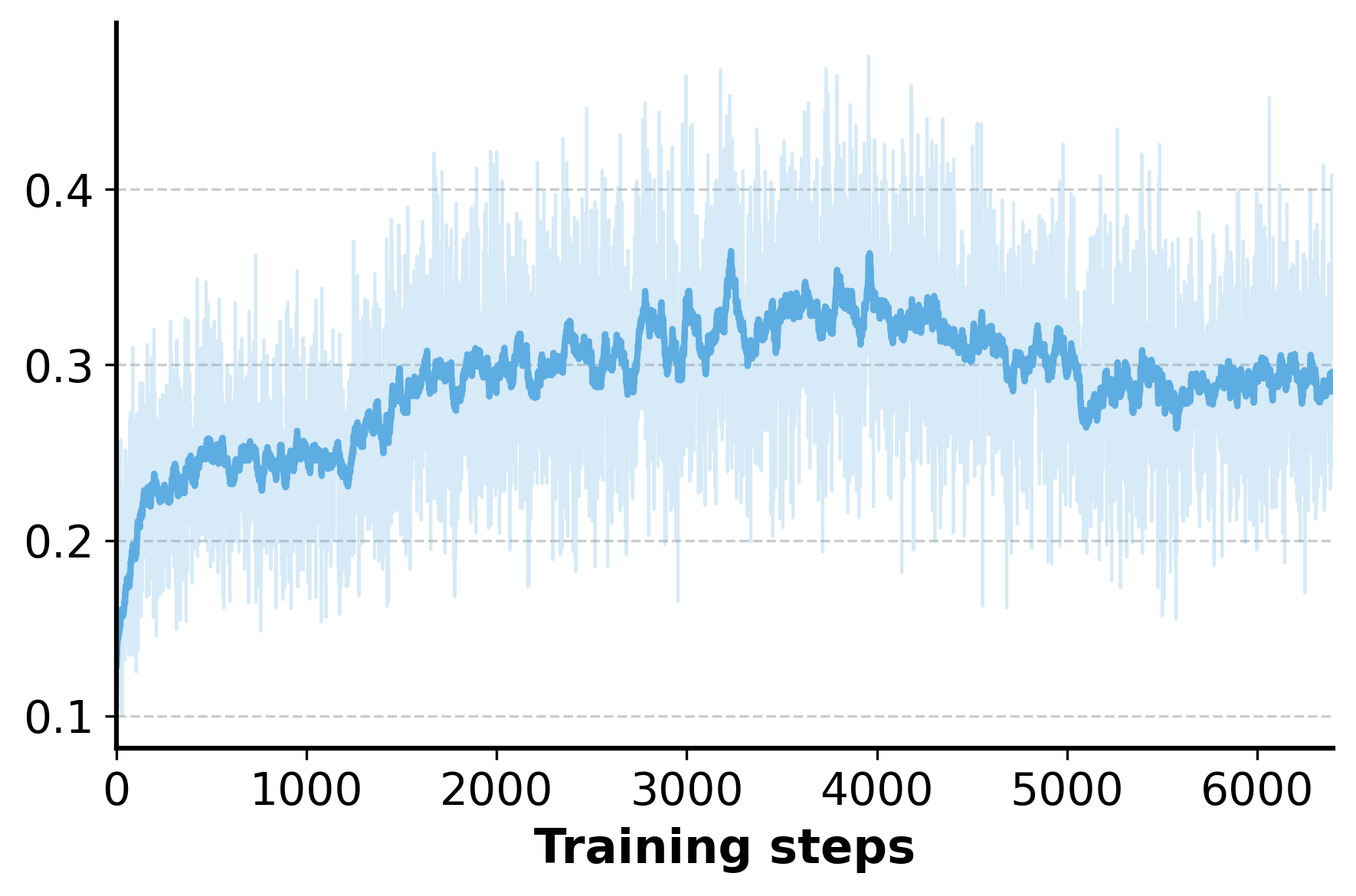}
        \caption{3D Detection IoU Reward}
        \label{fig:dyn_iou_reward}
    \end{subfigure}
    \hfill
    \begin{subfigure}[b]{0.48\linewidth}
        \centering
        \includegraphics[width=\linewidth]{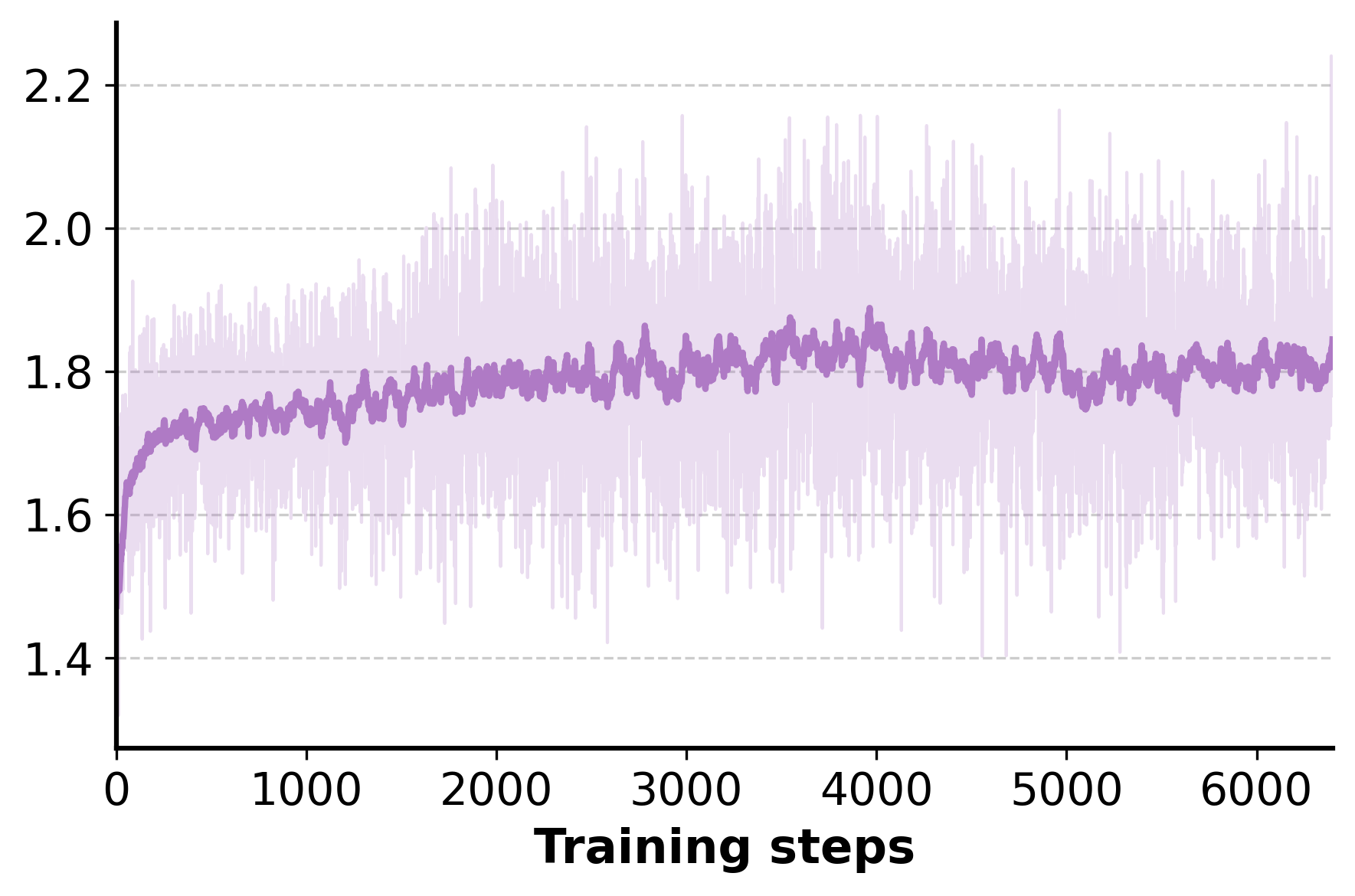}
        \caption{Total Reward}
        \label{fig:dyn_total_reward}
    \end{subfigure}
    
    \caption{\textbf{Training Dynamics Analysis (3D Video Detection).}}
    \label{fig:rl_training_dynamics_all}
    \vspace{-1em}
\end{figure}
\begin{figure}[t!]
    \centering
    \vspace{0.5em}
    \begin{subfigure}[b]{0.48\linewidth}
        \centering
        \includegraphics[width=\linewidth]{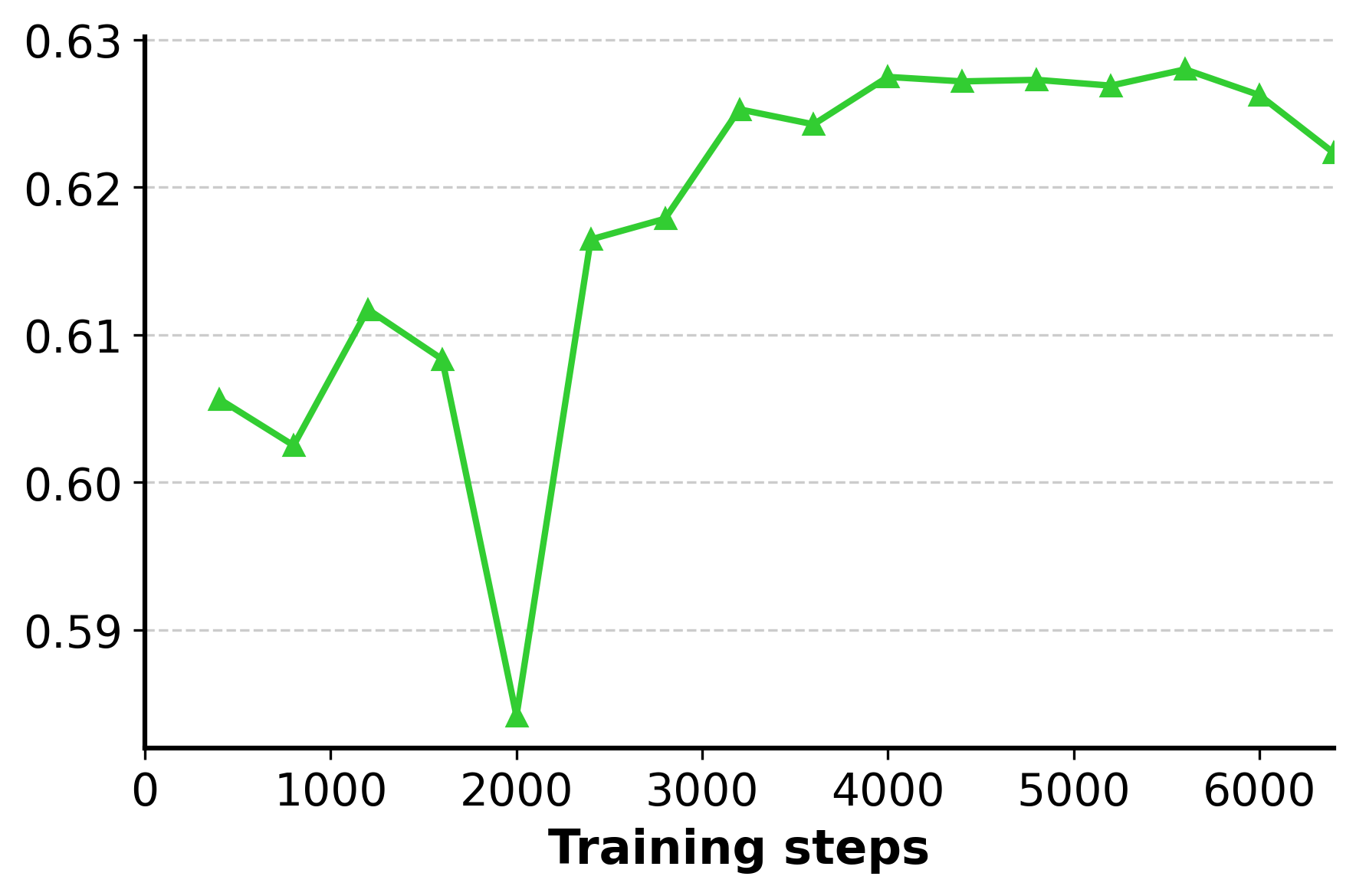}
        \caption{Evaluation Accuracy Score}
        \label{fig:dyn_eval_reasoning}
    \end{subfigure}
    \hfill 
    \begin{subfigure}[b]{0.48\linewidth}
        \centering
        \includegraphics[width=\linewidth]{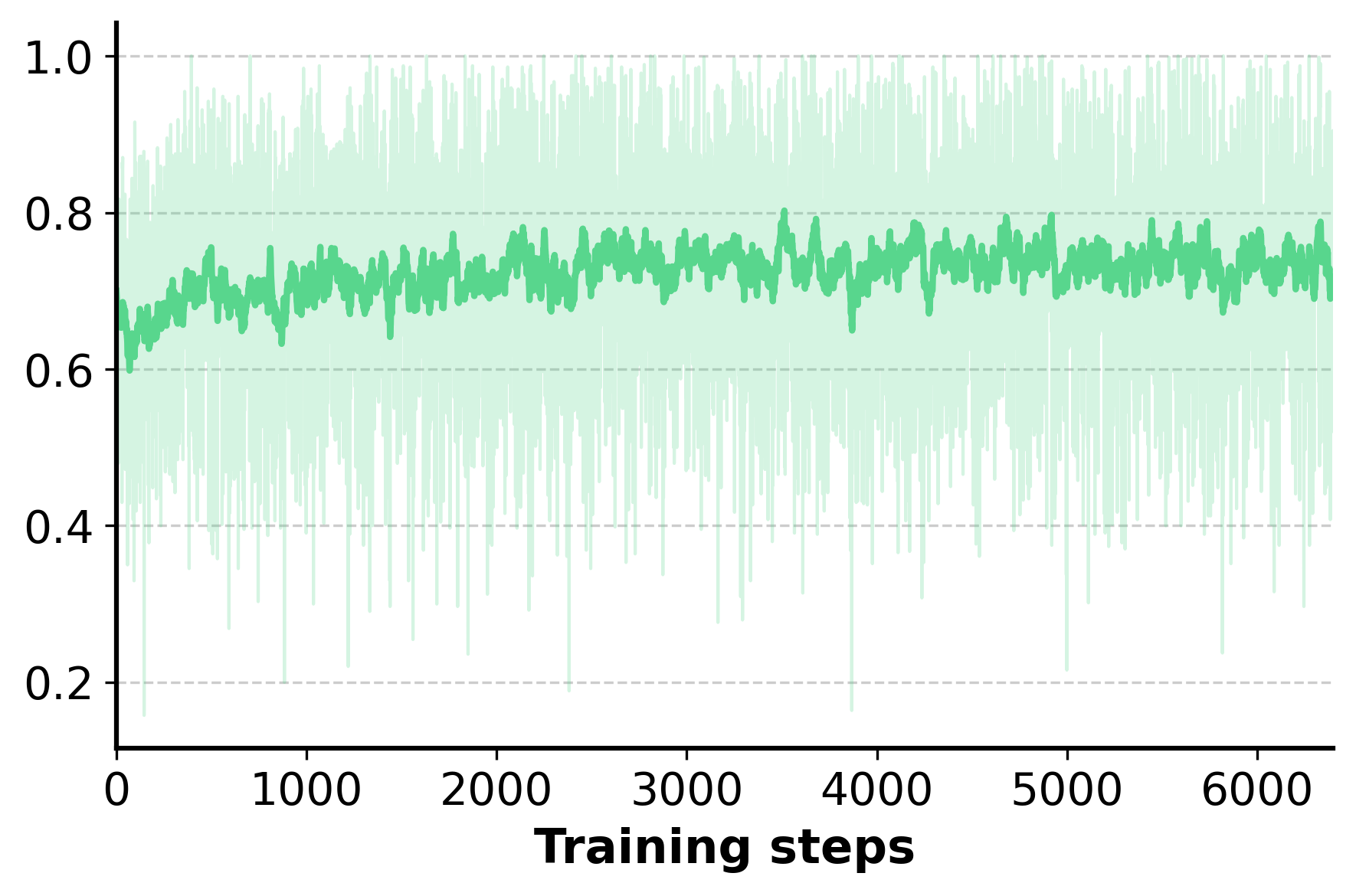}
        \caption{Accuracy Reward}
        \label{fig:dyn_acc_reward}
    \end{subfigure}
    
    \caption{\textbf{Training Dynamics Analysis (3D Spatial Reasoning).} }
    \label{fig:rl_training_dynamics_reasoning}
    \vspace{-1em}

\end{figure}

\paragraph{\ac{da} data and \ac{ta} data shape the foundation of \ac{rft}.} We compare the \ac{rft} results given different \ac{sft} data schemes: (1) default SFT (DA+TA); (2) SFT (DA+$\widetilde{\text{TA}}$), where we replace the \ac{cot}-10K data with 10K lower-quality \ac{cot} data generated by Qwen2.5-VL-72B-Instruct \citep{Qwen2.5-VL}; and (3) SFT (DA), which only uses \ac{cot}-10K during \ac{sft}. We compare their evaluation results and training rewards in \cref{fig:spatial_reasoning_cot_ablation}. The results show that the SFT (DA+TA) exhibits consistently higher accuracy and rewards than SFT (DA+$\widetilde{\text{TA}}$) and SFT (TA). This suggests that the foundation of \ac{rft} relies on both \ac{da} data that teaches \ac{id} knowledge and high-quality \ac{ta} data (\ie, \ac{cot} data) that teaches accurate reasoning behaviors.

\paragraph{\ac{ta} data prevents overfitting and ensures reliable reasoning behavior.} While SFT (DA) exhibits competitive \acf{id} performance on VSI-Bench (\cref{fig:spatial_reasoning_rl_ablation}), we observe it shows significantly poorer \ac{ood} performance than SFT (DA+TA) (\cref{appendix:sec:ood}). On the other hand, we observe that SFT (DA) can emerge reasoning output format, but the thoughts could be unreliable, \eg, tedious text like R1-Zero \citep{guo2025deepseekr1}. This suggests that \ac{ta} data is critical for generalization and reliable reasoning, and we should avoid training \acp{mllm} solely on \ac{da} data.



\subsection{Training Dynamics}

\label{sec:exp:dynamics}

\textbf{3D Video Detection.} \cref{fig:rl_training_dynamics_all} illustrates the training dynamics of 3D-RFT on \videt. The consistent rise in both Evaluation F1 Score (a) and F1 Reward (b) confirms that \ac{rft} effectively optimizes perception. Analyzing the reward components reveals a strategic shift: while the global F1 Reward (b) steadily increases, the IoU Reward (c) peaks early and then slightly declines. This indicates the policy transitions from initial \textit{geometric refinement} (tightening boxes) to \textit{recall maximization} (reducing false negatives). In the latter phase, the dense IoU Reward serves as a critical regulator, while the sparser F1 Reward optimizes the global balance between precision and recall. Moreover, the continuous improvement at the end implies that \ac{rlvr} is a stable and effective paradigm for \percep tasks.

\paragraph{3D Spatial Reasoning.} \cref{fig:rl_training_dynamics_reasoning} illustrates the training dynamics of 3D-RFT on the spatial reasoning task. Despite minor fluctuations, both evaluation accuracy and training rewards exhibit upward trends, confirming \ac{rft}'s efficacy. Notably, the curves show a sign of saturation after 4000 steps. We attribute this to the nature of the optimization landscape: unlike perception tasks where \ac{rft} continuously refines continuous geometric coordinates, the reasoning task features a discrete text space with coarser feedback. This leads to the earlier saturation than 3D perception tasks, and potentially limits the granularity of improvement over \ac{sft}.



\subsection{Analysis on Training Objectives of \ac{sft} and \ac{rft}}

While \cref{sec:intro} introduces how 3D-\ac{rft} provides a direct optimization signal compared to standard \ac{sft} in 3D scene understanding tasks, this section presents the theoretical analysis and empirical evidence to support this claim.

\subsubsection{Theoretical Analysis of Supervisory Signals}
\label{sec:theoretical_analysis}

\begin{figure}[t!]
    \centering

        \includegraphics[width=\linewidth]{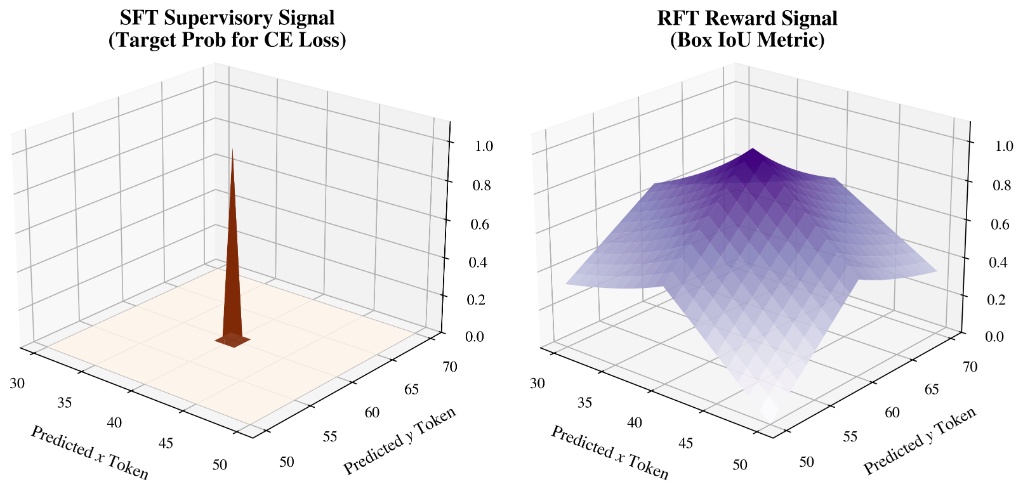}
        \caption{\textbf{Conceptual 1D Boundary Optimization Space.} We illustrate a simplified 1D proxy to illustrate the fundamental mathematical mismatch between token classification and geometric optimization. The axes represent the predicted left ($x$) and right ($y$) boundaries of a 1D interval.}
        \label{fig:theoretical_analysis}

    \vspace{-1em}
\end{figure}

We model the supervisory signals for a simplified bounding box prediction task to contrast the gradient landscapes of cross-entropy and metric-driven optimization (\cref{fig:theoretical_analysis}).

\textbf{The \ac{sft} supervisory signal}: Standard \ac{sft} optimizes next-token likelihood via CE loss against a one-hot target. As illustrated in \cref{fig:theoretical_analysis} (left), this formulation yields a sparse, binary target probability distribution—an isolated spike exactly at the ground truth. The objective lacks ordinal feedback, leaving the model blind to spatial proximity.

\textbf{The \ac{rft} reward signal}: In contrast, \ac{rft} bypasses this non-differentiable barrier by treating the final evaluation metric (e.g., IoU) as a black-box reward. This formulation constructs a smooth, continuous reward surface (\cref{fig:theoretical_analysis}, right). The model receives explicit feedback based on geometric overlap, which effectively guides the optimization trajectory toward high-precision regions even when initial predictions are imperfect.

\subsubsection{Empirical Evidence in 3D Visual Grounding}
\label{sec:empirical_evidence}

\begin{figure}[t!]
    \centering
    
        \includegraphics[width=0.75\linewidth]{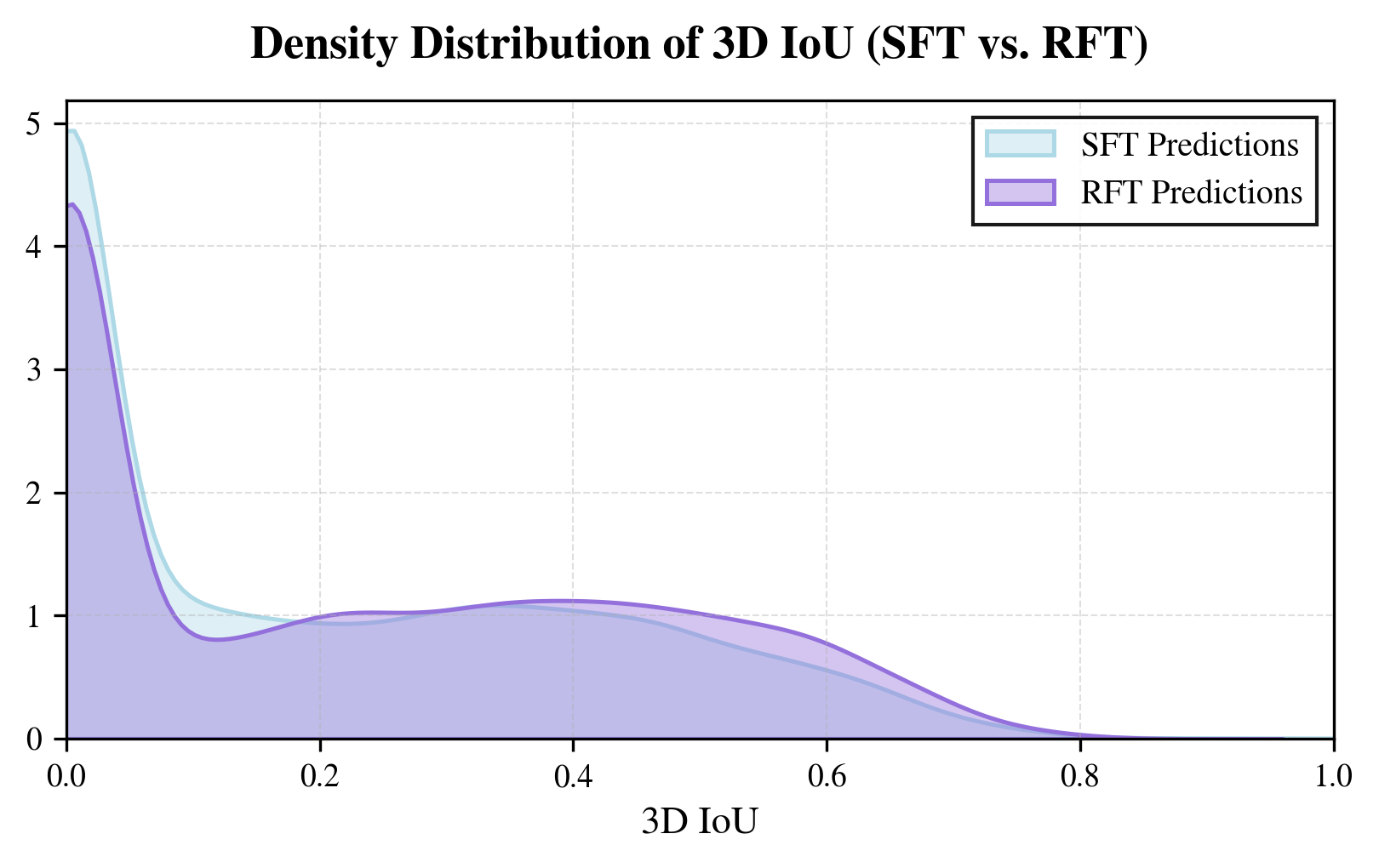}
        \caption{\textbf{Distribution of 3D IoU in 3D Visual Grounding.} A comparison of the raw 3D IoU distributions on the validation set between the baseline model trained with SFT and SFT.}
        \label{fig:prediction_distribution}

    \vspace{-1em}
\end{figure}

To evaluate how these theoretical differences dictate real-world performance, we analyze the probability density distributions of IoU predictions on the ScanRefer evaluation set for both the \ac{sft} baseline and the \ac{rft} model (\cref{fig:prediction_distribution}).

\textbf{SFT baseline (all-or-nothing phenomona)}: The \ac{sft} distribution (\cref{fig:prediction_distribution}, light blue) exhibits a severe density spike at exactly $\text{IoU} = 0.0$. Because token-level CE loss provides no spatial smoothing, a minor error in the token generation sequence often results in a completely dislocated 3D bounding box, leading to catastrophic failure.

\textbf{RFT advance (graceful degradation)}: By optimizing directly against the continuous evaluation metric, \ac{rft} fundamentally alters this error distribution. As shown by the purple curve in \cref{fig:prediction_distribution}, \ac{rft} substantially suppresses the zero-overlap spike and shifts the probability mass to the right. This forms a prominent "spatial hump"—particularly pronounced between IoU thresholds of 0.25 and 0.50—indicating that errors are bounded to localized regions.

The empirical evidence echoes our assumption in \cref{sec:theoretical_analysis} that \ac{rft} provides better optimization signals than \ac{sft}.

\section{Conclusion}

We introduce 3D-RFT, a framework that applies \ac{rlvr} to \vsu. By shifting the learning paradigm from token-level imitation to direct metrics-driven optimization, \ie, utilizing verifiable rewards like 3D IoU and F1-Score, 3D-RFT effectively aligns training objectives with task performance. We demonstrate that 3D-RFT effectively enhances model performance across perception tasks and reasoning tasks, including 3D video detection, 3D visual grounding, and spatial reasoning. Our model \model even outperforms larger baselines like VG LLM-8B in 3D video detection and VLM-3R-7B in spatial reasoning. Our analyses reveal good properties of 3D-RFT such as robust efficacy, and valuable insights like the impact of data diversity. We hope our work can unveil the potential of \ac{rlvr} for 3D scene understanding and offer valuable insights for future research.

\clearpage


\section*{Impact Statement}
This paper presents work whose goal is to advance the field of Machine Learning. There are many potential societal consequences of our work, none of which we feel must be specifically highlighted here.

\bibliography{ref}
\bibliographystyle{icml2026}

\newpage
\appendix
\onecolumn

\section{Implementation Details}
\label{appendix:implementation_details}

\subsection{Data}
\label{appendix:training_data}
\paragraph{3D Video Detection.} For 3D video detection, we re-use the data from VG LLM \citep{zheng2025vgllm}, which was initially curated from EmbodiedScan \citep{wang2024embodiedscan}. The data comprises consecutive frames and the corresponding visible object annotations in indoor scenes, covering 958 scenes for training and 243 scenes for evaluation. The object 3D bounding boxes are transformed to the coordinate system of the first frame.

\paragraph{3D Visual Grounding.} We use the ScanRefer \citep{chen2020scanrefer} dataset for 3D visual grounding, which is a common practice. ScanRefer includes 37K pairs of object grounding text and axis-aligned 3D bounding boxes across 562 indoor scans. We follow prior works \citep{zhang2025flatland,zheng2025vgllm} in terms of spatio-temporal grounding formulation and data processing procedures.

\paragraph{3D Spatial Reasoning.} Our \ac{sft} data for the spatial reasoning task includes VSI-207K from VLM-3R \citep{fan2025vlm} and Sr-91K from SpaceR \citep{ouyang2025spacer}, spanning ScanNet \citep{dai2017scannet}, ScanNet++ \citep{yeshwanth2023scannet++}, and ARKitScenes \citep{baruch2021arkitscenes}. Our \ac{cot} data is generated by Qwen3-VL-32B-Thinking \citep{Qwen3-VL} and curated by selecting those whose accuracy rewards exceed $0.85$. The curated \ac{cot}-10K comprises 9.2K from VLM-3R-VSI subset \citep{fan2025vlm} and 0.8K from SQA3D \citep{ma2023sqa3d}.


\subsection{Training}
\label{appendix:efficient_training}

\paragraph{Training Settings.} All experiments are conducted on 8 Nvidia A100 GPUs. For the \videt and \vigrd tasks, we adopt the full fine-tuning strategy following the settings in \vgllmpaper. In contrast, for \spr, we employ LoRA \citep{hu2022lora} and set the KL-Divergence penalty coefficient to zero during policy training.

\paragraph{GPU Memory Bottleneck in 3D-RFT.} 
A critical bottleneck in \vsu tasks is the prohibitive GPU memory consumption required by \ac{grpo}. The necessity of maintaining a large group size leads to substantial memory spikes. Specifically, processing long-context video inputs with high-resolution vision encoders (e.g., VGGT) generates extensive computation graphs. In standard implementations, storing activations for all $K$ samples simultaneously often exceeds VRAM capacity, resulting in Out-Of-Memory (OOM) errors.

\paragraph{Loss Backward Chunking.} 
To mitigate this, we introduce a gradient accumulation strategy that partitions the global batch indices $\{1, \dots, K\}$ into $M$ disjoint micro-chunks $\{C_1, \dots, C_M\}$. The size of each chunk is constrained such that $|C_m| \leq M_{\text{micro}}$ to fit within GPU memory limits. Accordingly, we reformulate the total loss as a sum of chunked losses:
\begin{align}
    \mathcal{L}_{\text{chunked}}(\theta) &= \sum_{m=1}^{M} \frac{|C_m|}{K} \mathcal{L}_{C_m}(\theta), \\
    \text{where} \quad \mathcal{L}_{C_m}(\theta) &= \frac{1}{|C_m|} \sum_{k \in C_m} \sum_{t} \mathcal{L}_{k,t}(\theta).
\end{align}
By iteratively computing and accumulating gradients for each $\mathcal{L}_{C_m}$, we reduce the peak memory complexity from $\mathcal{O}(B \times G)$ to $\mathcal{O}(M_{\text{micro}})$. This approach enables significant scaling of the group size $G$ without hurting the training performance. In our experiments, we set the micro-chunk size to 2 for \videt, 4 for \vigrd, and 1 for \spr.


\section{Additional Results}

\subsection{Quantitative Results}\label{sec:additional_exp_vst}

\begin{table}[t!]
\centering
\caption{\textbf{Additional results across more spatial reasoning benchmarks and different model scales.}}
\resizebox{0.75\linewidth}{!}{%
\begin{tabular}{lccccc}
\toprule
 & VSI-Bench & MindCube & MMSI-Bench & All-Angles & ERQA \\
\midrule
VST-3B~\citep{yang2025visual}   & 57.7 & 35.9 & 31.3 & -    & -    \\
3D-RFT-4B (ours) & 62.8 & 36.4 & 29.9 & 41.6 & 34.2 \\
\midrule
VST-7B~\citep{yang2025visual}   & 61.2 & 39.7 & 34.8 & -    & -    \\
3D-RFT-8B (ours) & 64.0 & 42.6 & 28.0 & 43.1 & 38.5 \\
\bottomrule
\end{tabular}
}
\label{tab:additional_exp_vst}
\end{table}

We report the evaluation results on more spatial reasoning benchmarks that adopt multi-view input, including VSI-Bench~\citep{yang2025thinking}, MindCube~\citep{yin2025spatial}, MMSI-Bench~\citep{yang2025mmsi}, All-Angles~\citep{yeh2026seeing}, and ERQA~\citep{team2025geminirobotics}. VSI-Bench adopts 32-frame inputs, while the other benchmarks adopt only several images as input. Specifically, we compare the performance of 3D-RFT to VST~\citep{yang2025visual}, a generalist model trained on over 6M samples spanning single-image, multi-image, and video settings. As shown in \cref{tab:additional_exp_vst}, we report the following findings: (1) 3D-RFT-8B performs better than 3D-RFT-4B in general, confirming the effect of model scaling under the 3D-RFT framework. (2) 3D-RFT achieves better in-distribution performance (VSI-Bench) compared to VST, including both 3B-4B and 7B-8B scales. (3) 3D-RFT slightly outperforms VST on MindCube and underperforms VST on MMSI-Bench. We attribute this to the substantial difference in training data: VST is trained on 6M samples spanning single-image, multi-image, and video settings, whereas our spatial reasoning model uses only 300K video-focused samples. The limited scale and diversity place our model at a disadvantage on this out-of-distribution benchmark, and also explain why performance does not scale positively from 4B to 8B on MMSI-Bench as it does on VSI-Bench. Nonetheless, our primary goal is to develop an RLVR framework for video-based 3D understanding rather than a generalist model, and we believe that the out-of-distribution performance can be aided by incorporating more diverse training data.

\subsection{Qualitative Results}

\subsubsection{3D Video Detection}

\begin{figure*}[t]
    \centering
    \setlength{\tabcolsep}{1.5pt} 
    \renewcommand{\arraystretch}{1.2} 
    
    \begin{tabular}{c c c c c c c}
        

        \rotatebox{90}{\small VG LLM-4B} & 
        \includegraphics[width=0.15\linewidth]{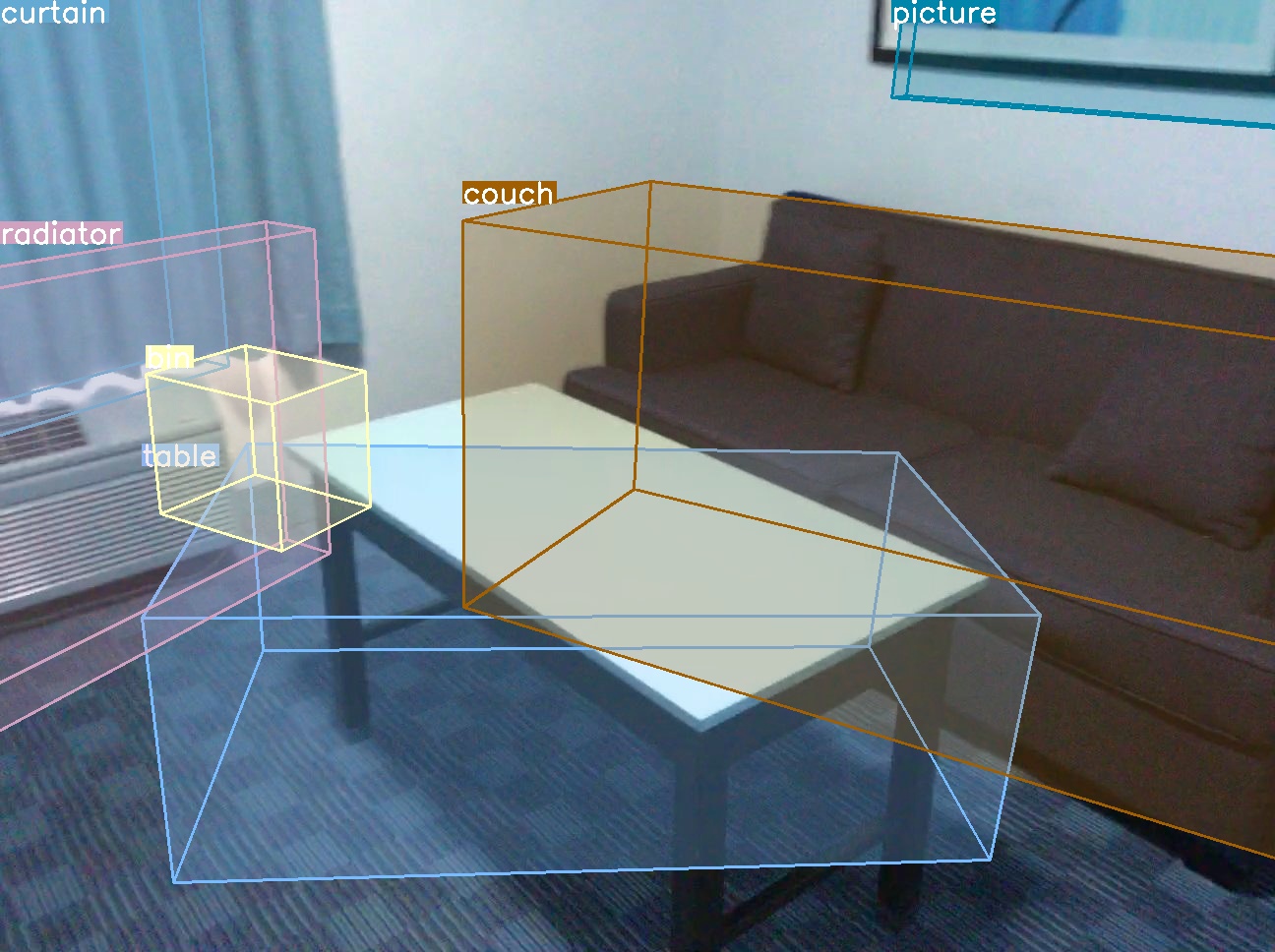} & 
        \includegraphics[width=0.15\linewidth]{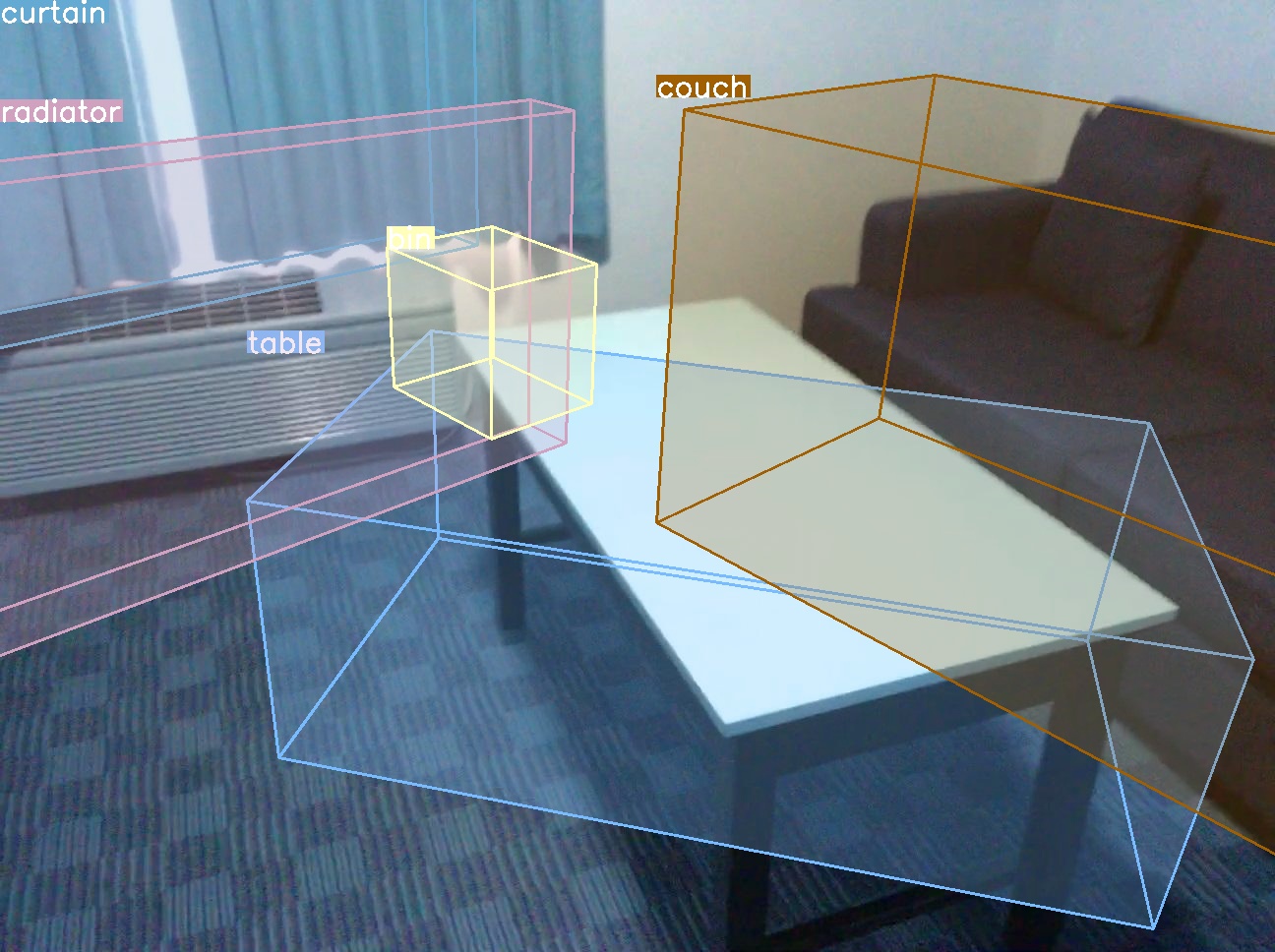} &
        \includegraphics[width=0.15\linewidth]{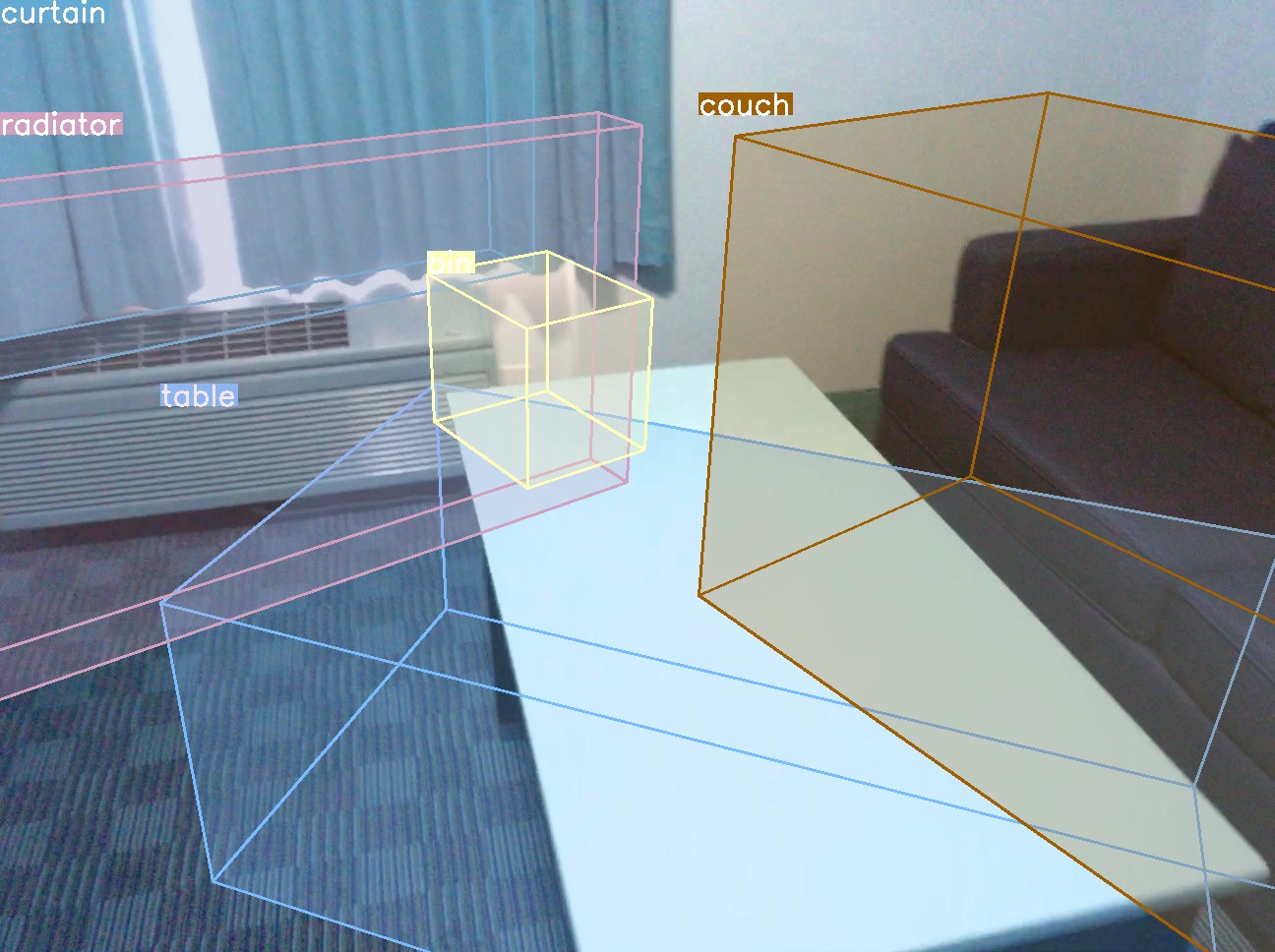} & 
        \includegraphics[width=0.15\linewidth]{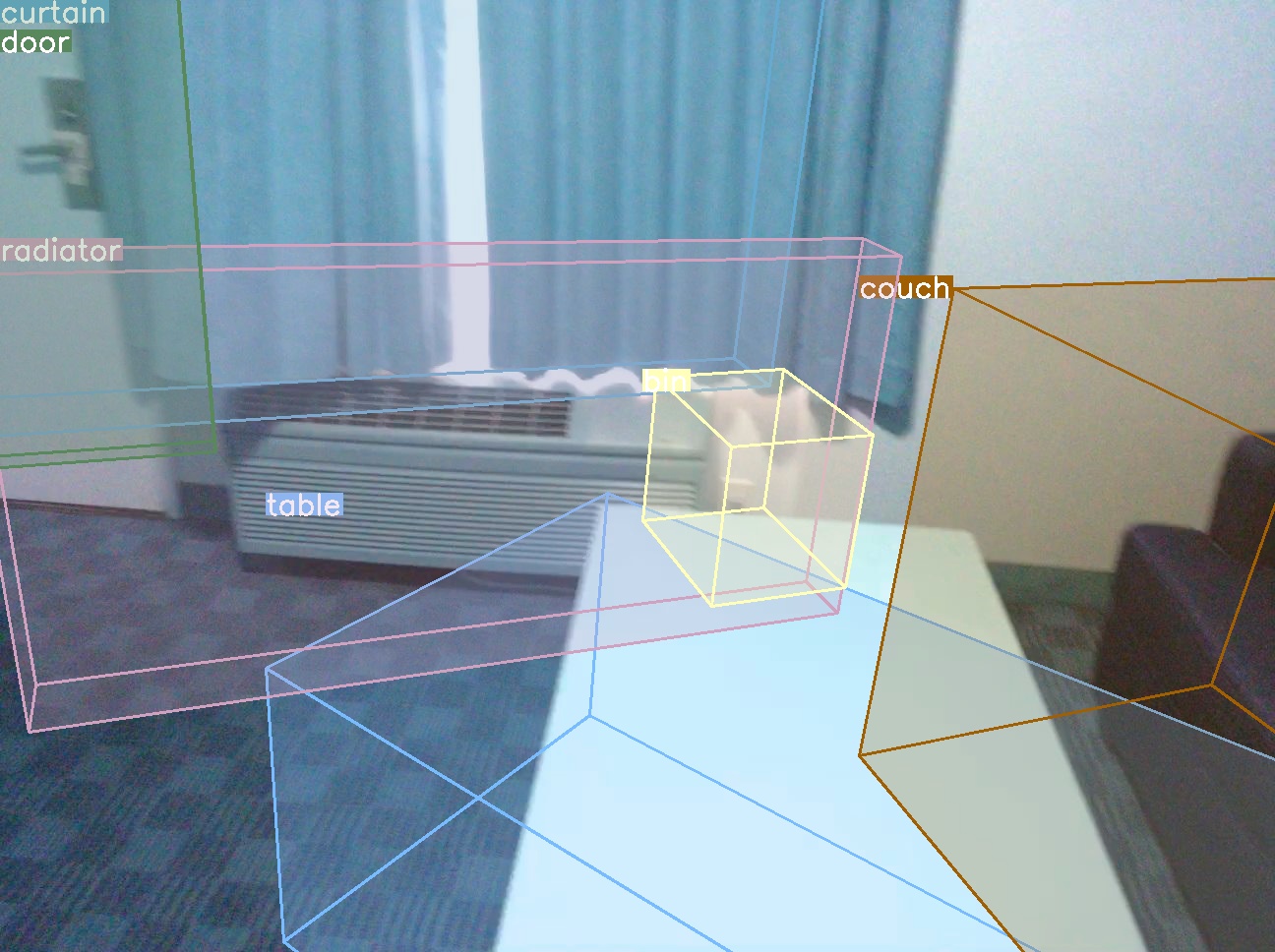} & 
        \includegraphics[width=0.15\linewidth]{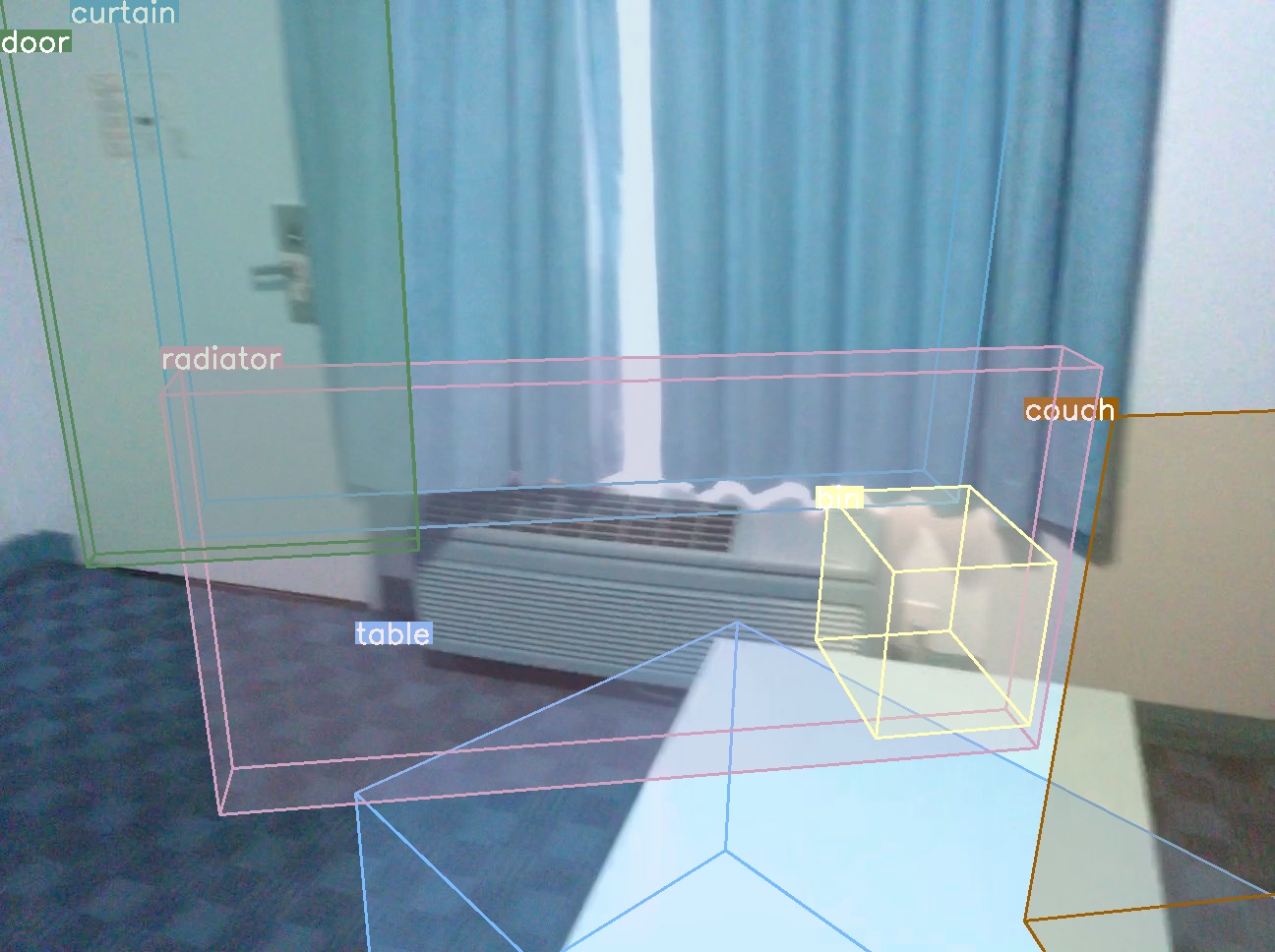} &
        \includegraphics[width=0.15\linewidth]{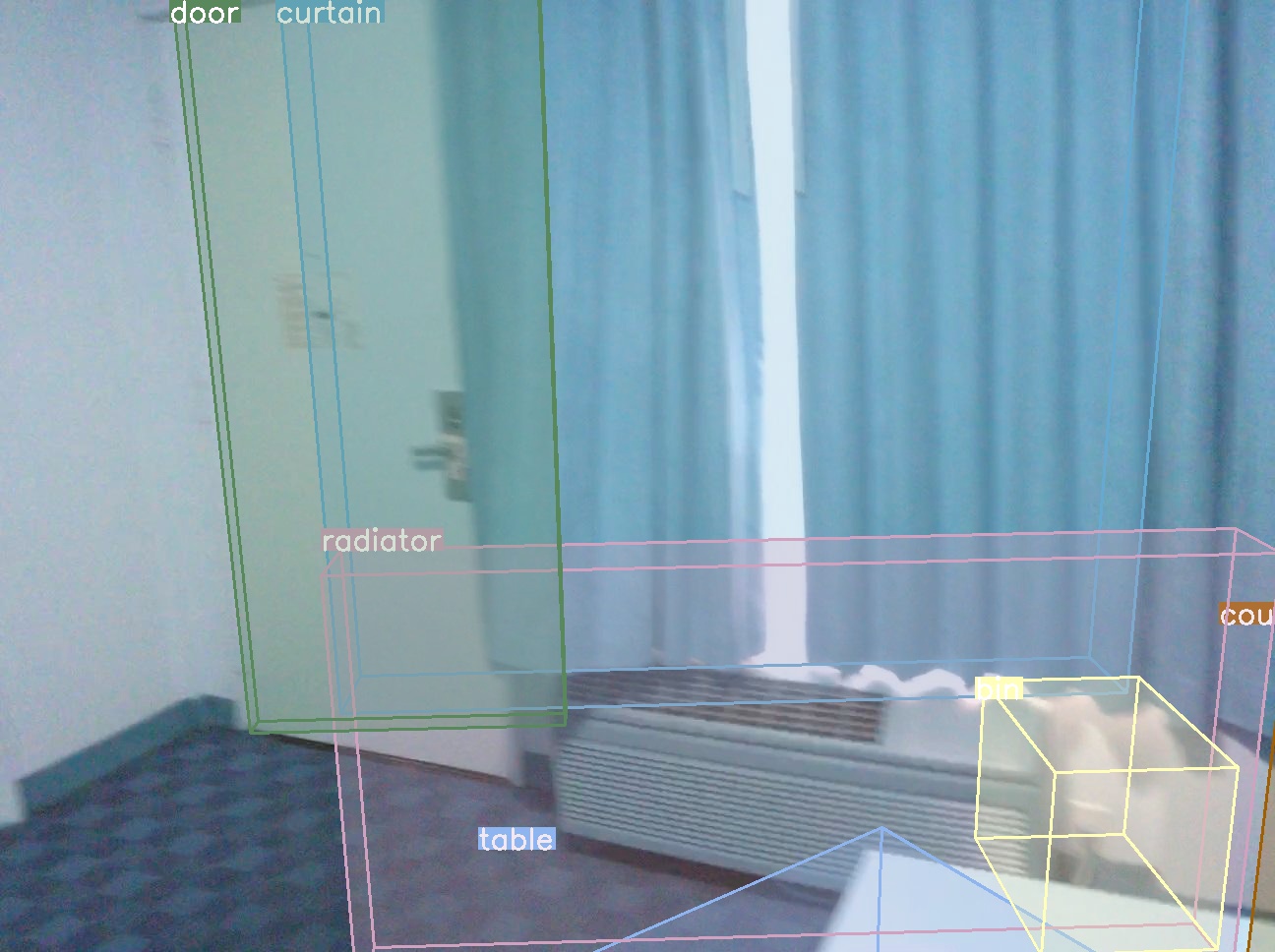} \\
        
        \rotatebox{90}{\small \model } & 
        \includegraphics[width=0.15\linewidth]{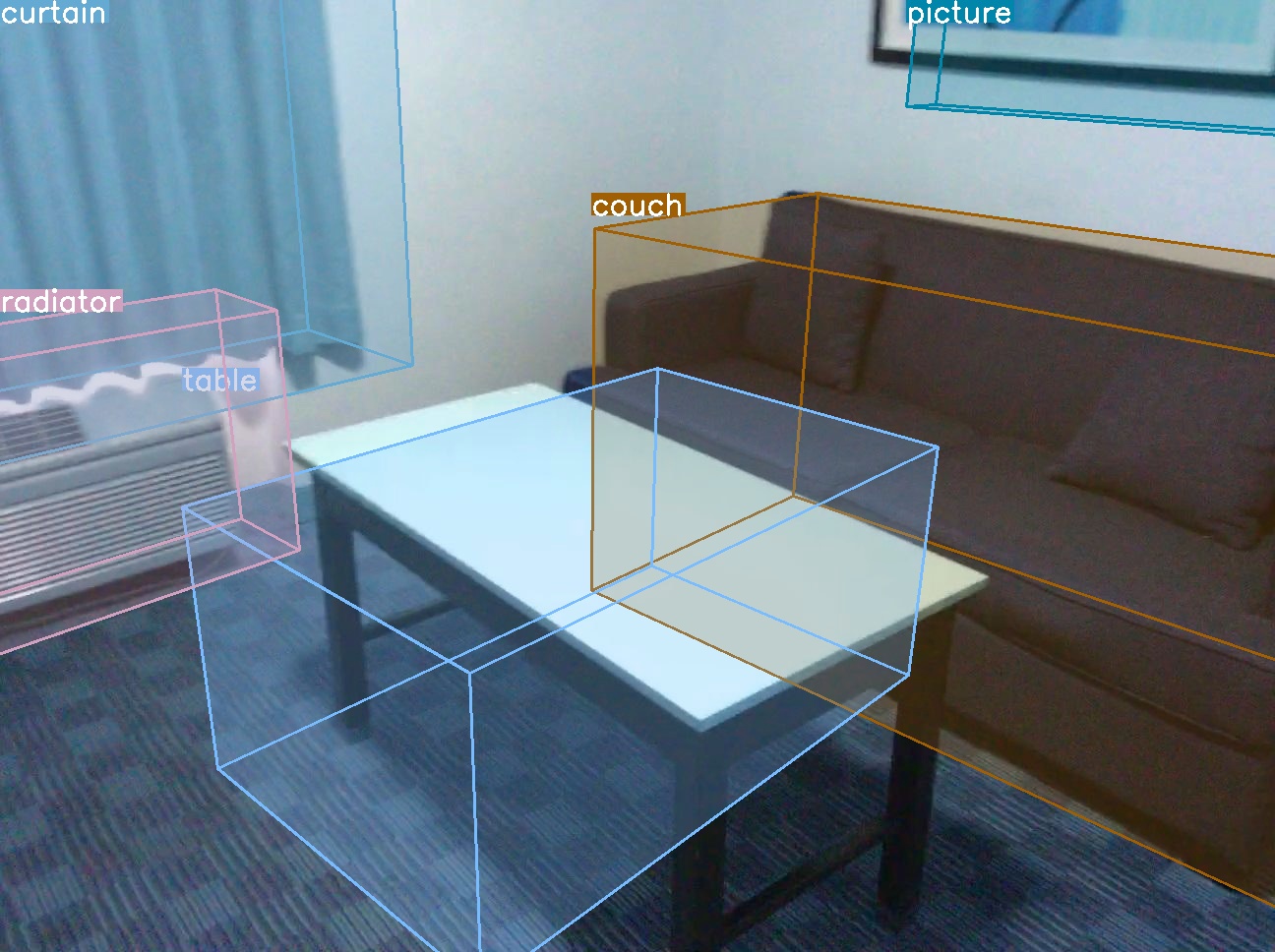} & 
        \includegraphics[width=0.15\linewidth]{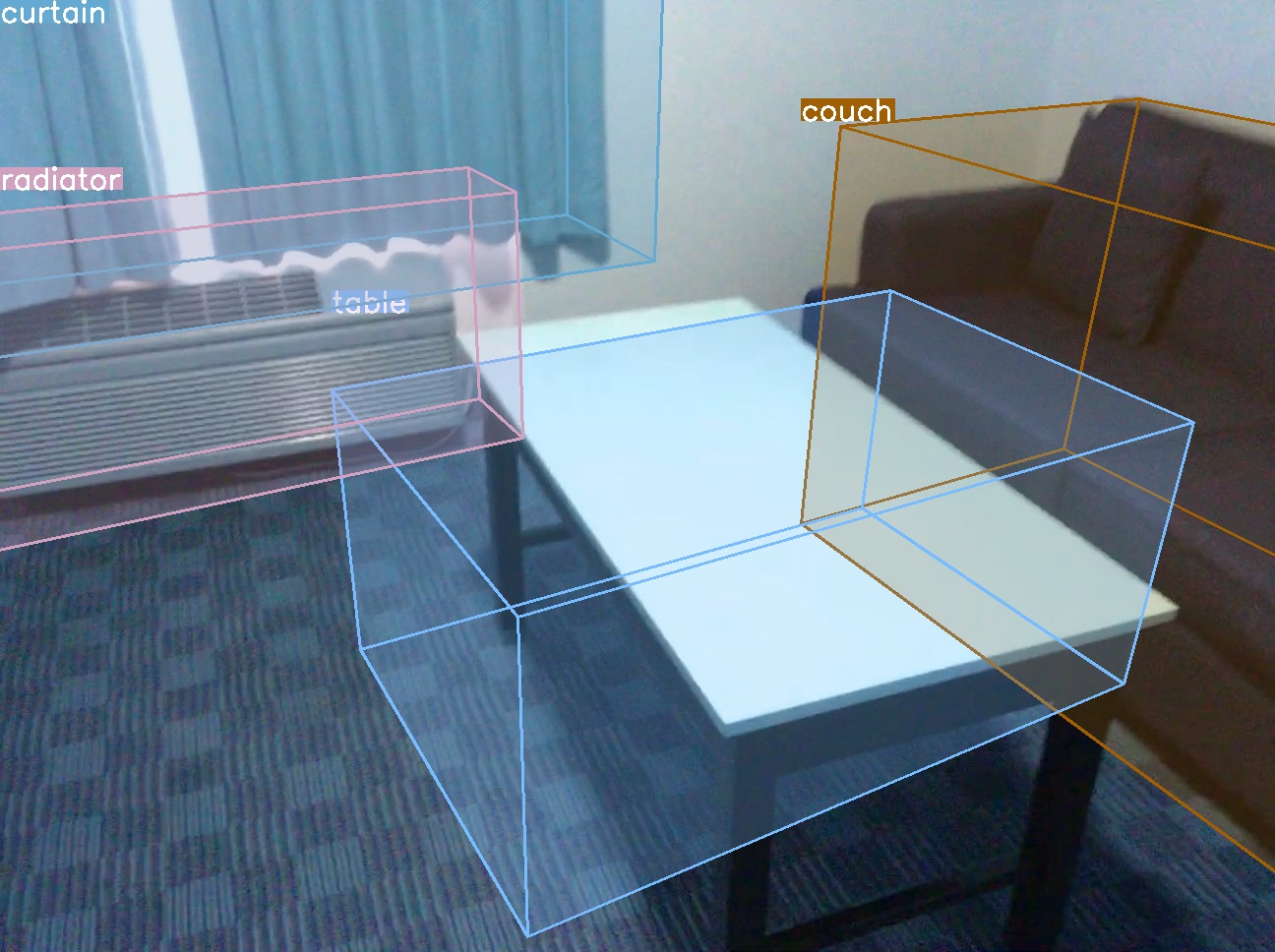} &
        \includegraphics[width=0.15\linewidth]{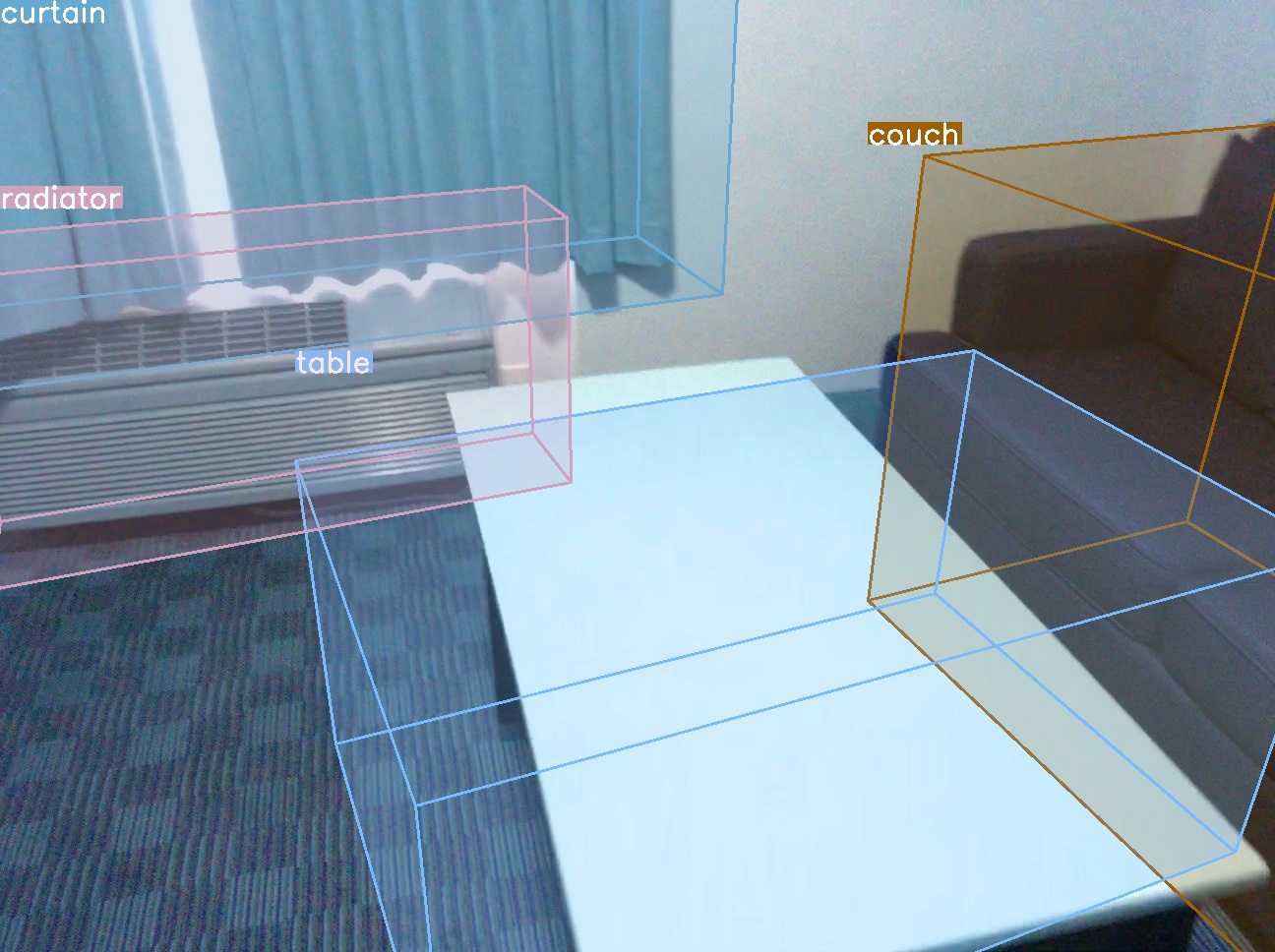} & 
        \includegraphics[width=0.15\linewidth]{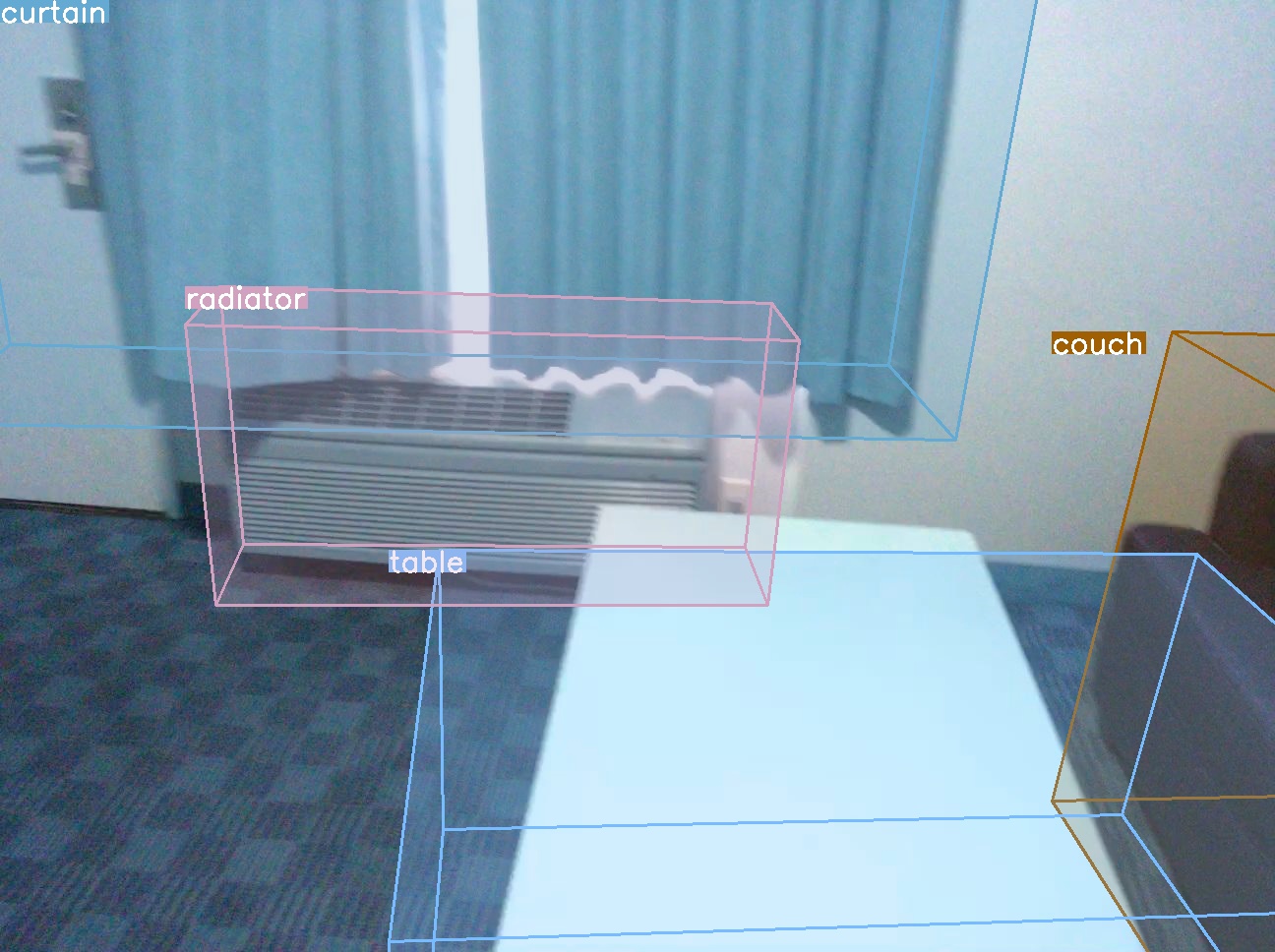} & 
        \includegraphics[width=0.15\linewidth]{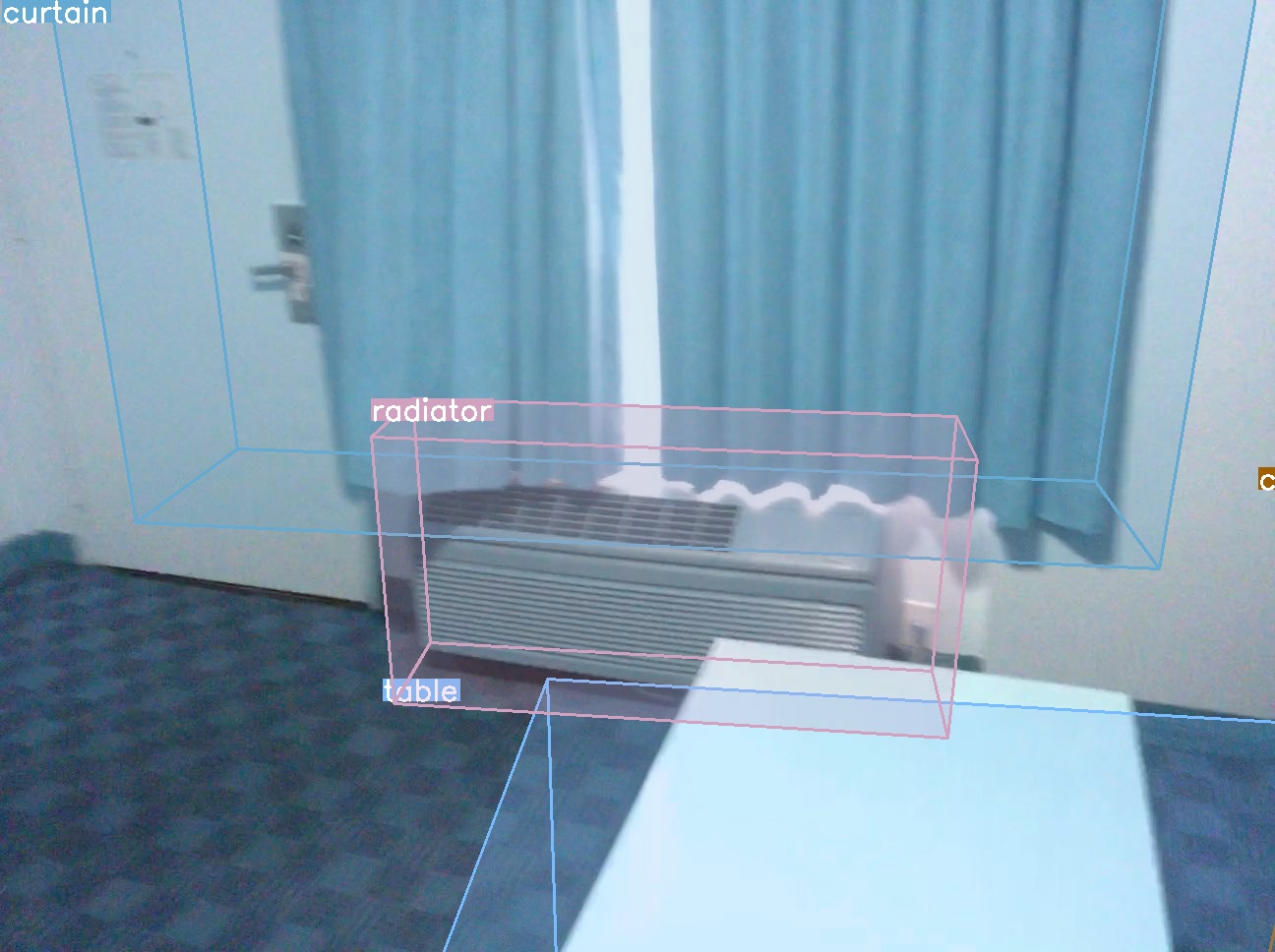} &
        \includegraphics[width=0.15\linewidth]{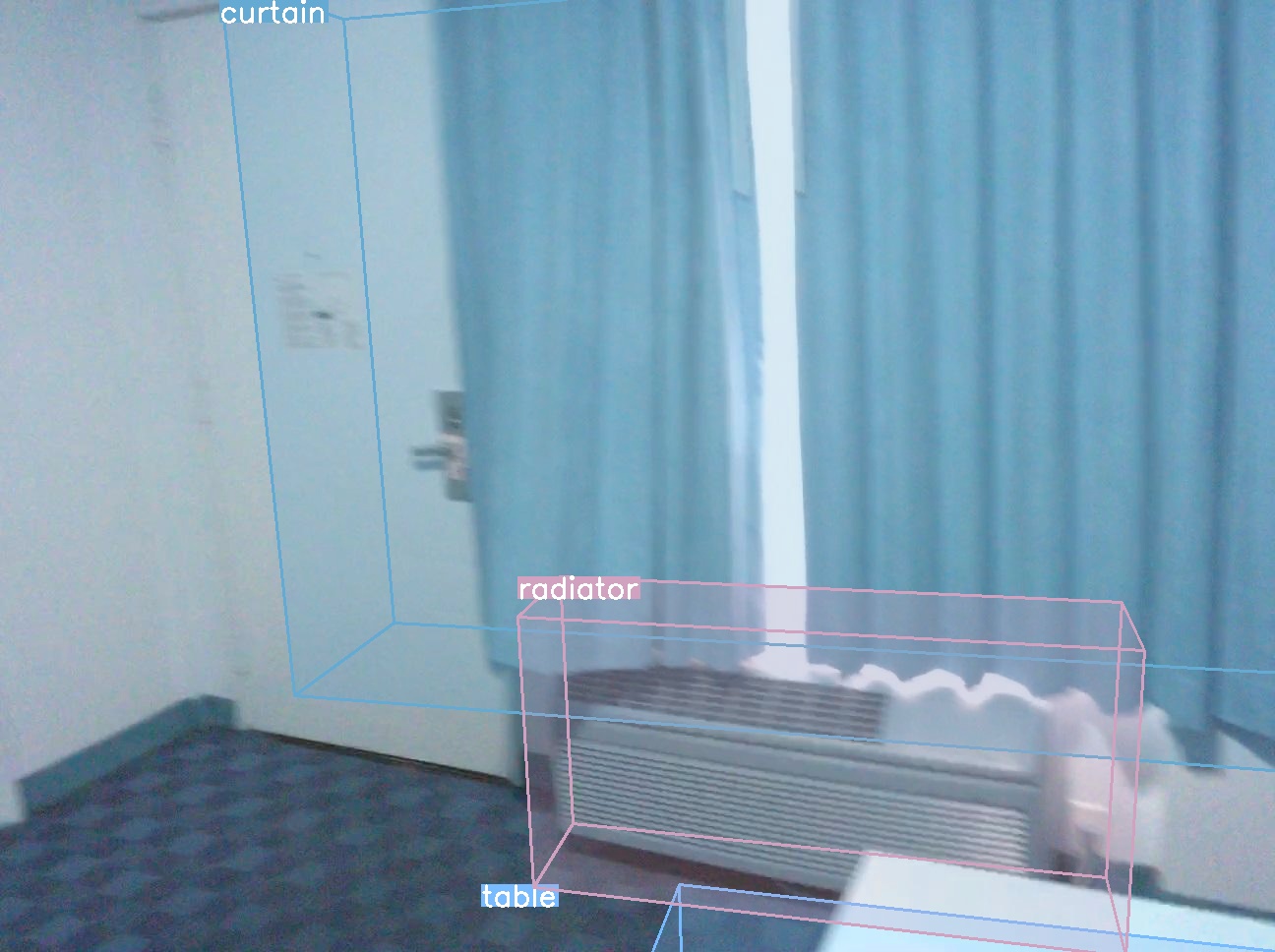} \\
        \\
        \\
        
        \rotatebox{90}{\small VG LLM-4B} & 
        \includegraphics[width=0.15\linewidth]{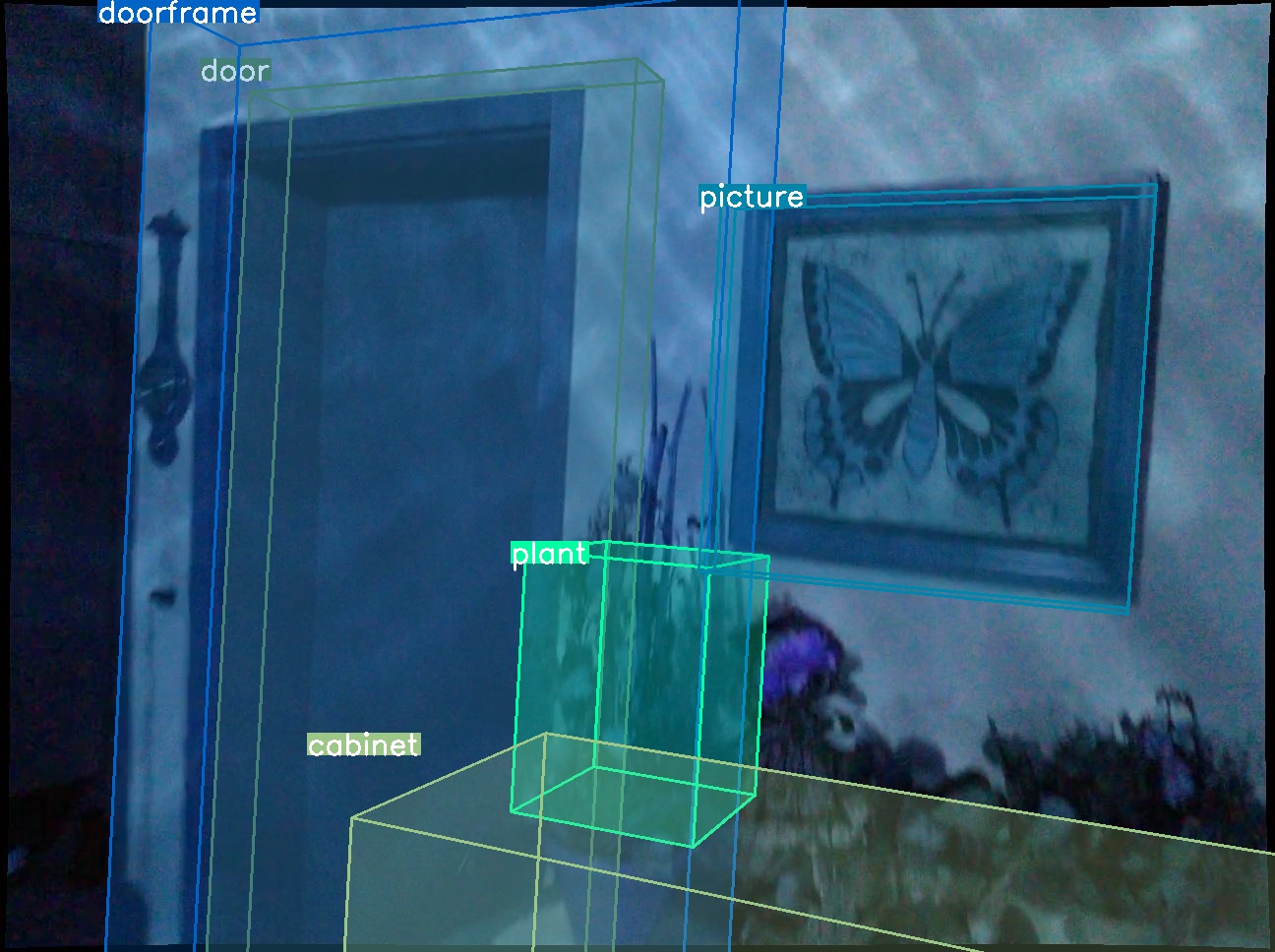} & 
        \includegraphics[width=0.15\linewidth]{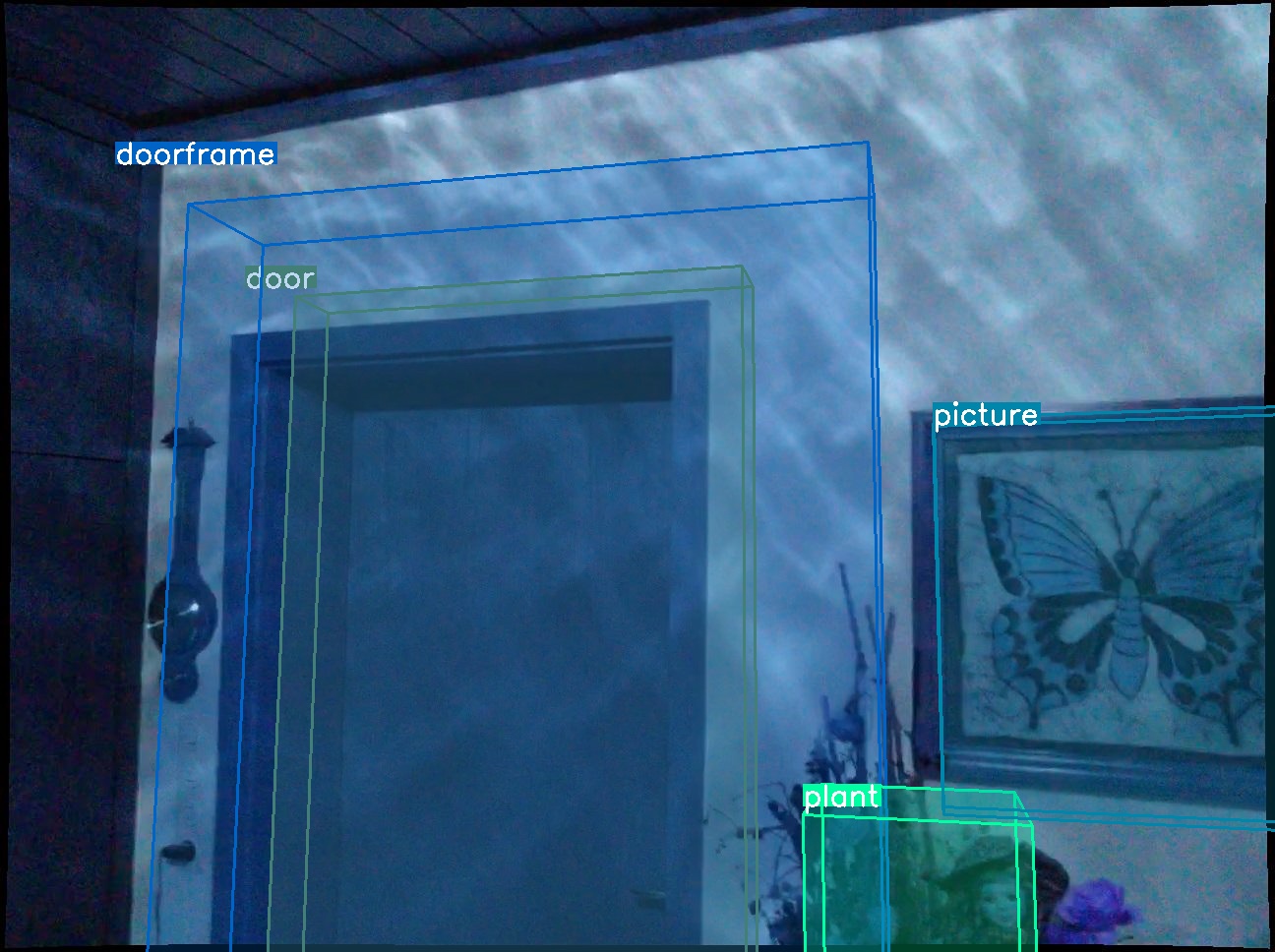} &
        \includegraphics[width=0.15\linewidth]{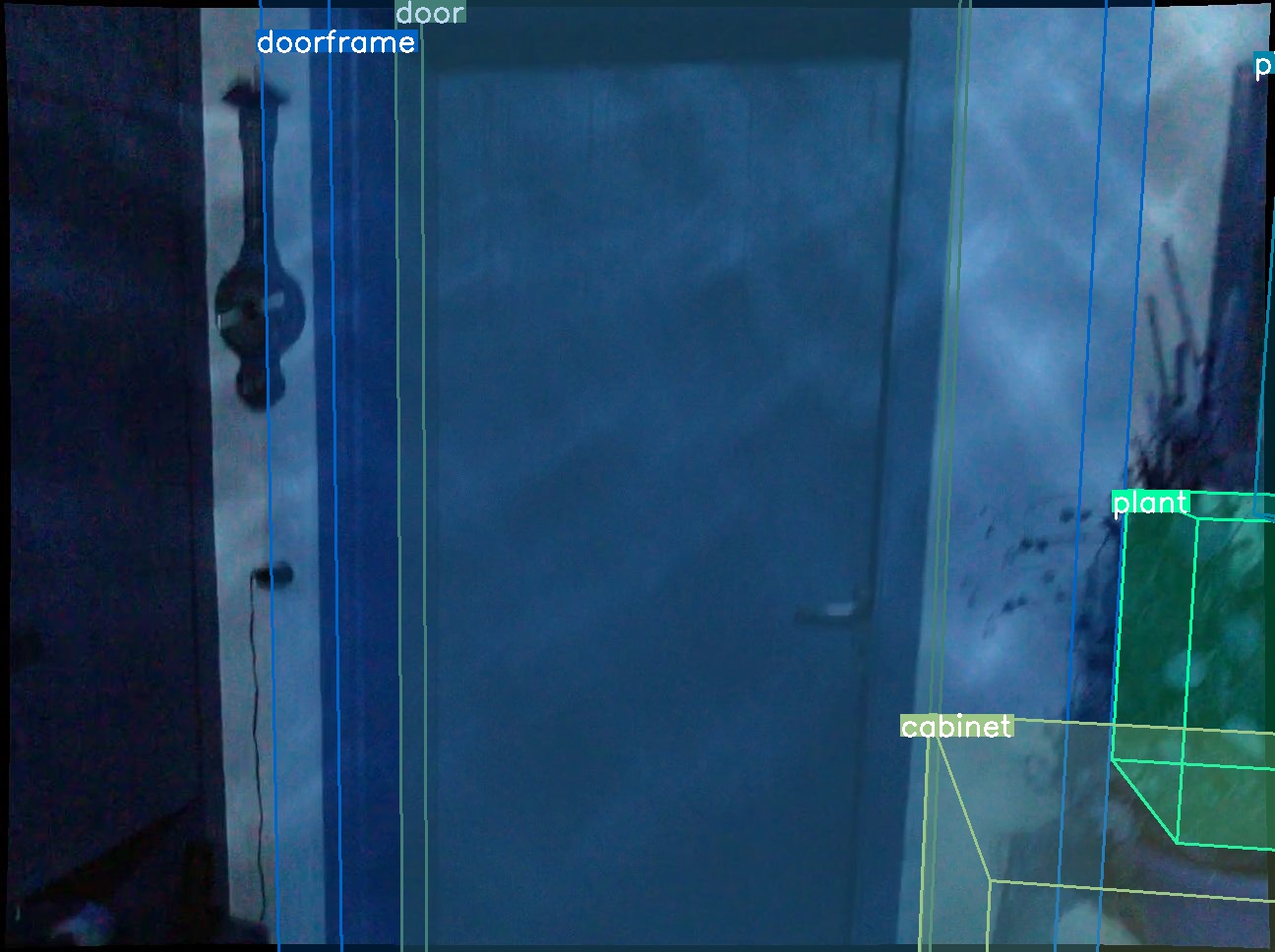} & 
        \includegraphics[width=0.15\linewidth]{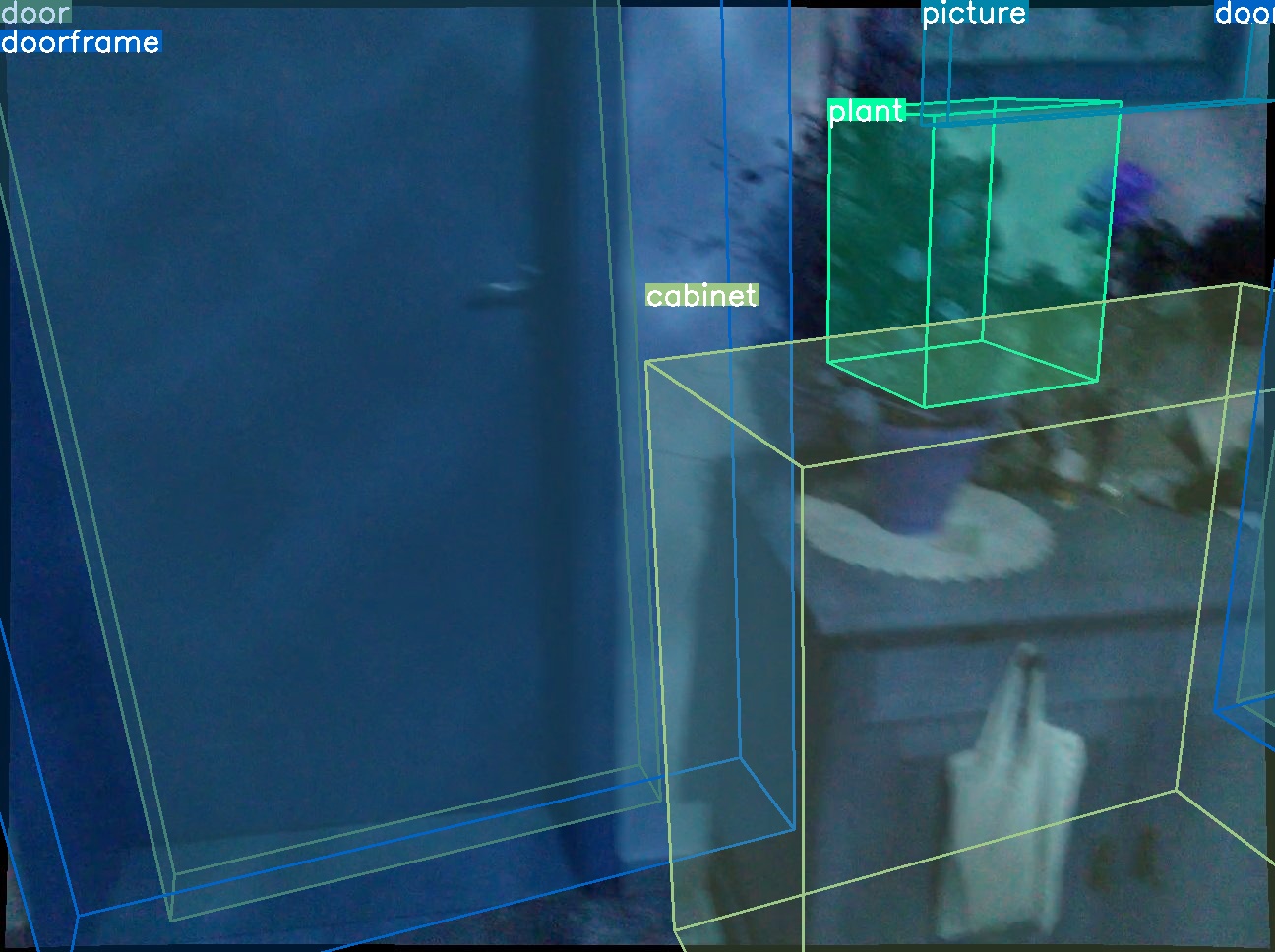} & 
        \includegraphics[width=0.15\linewidth]{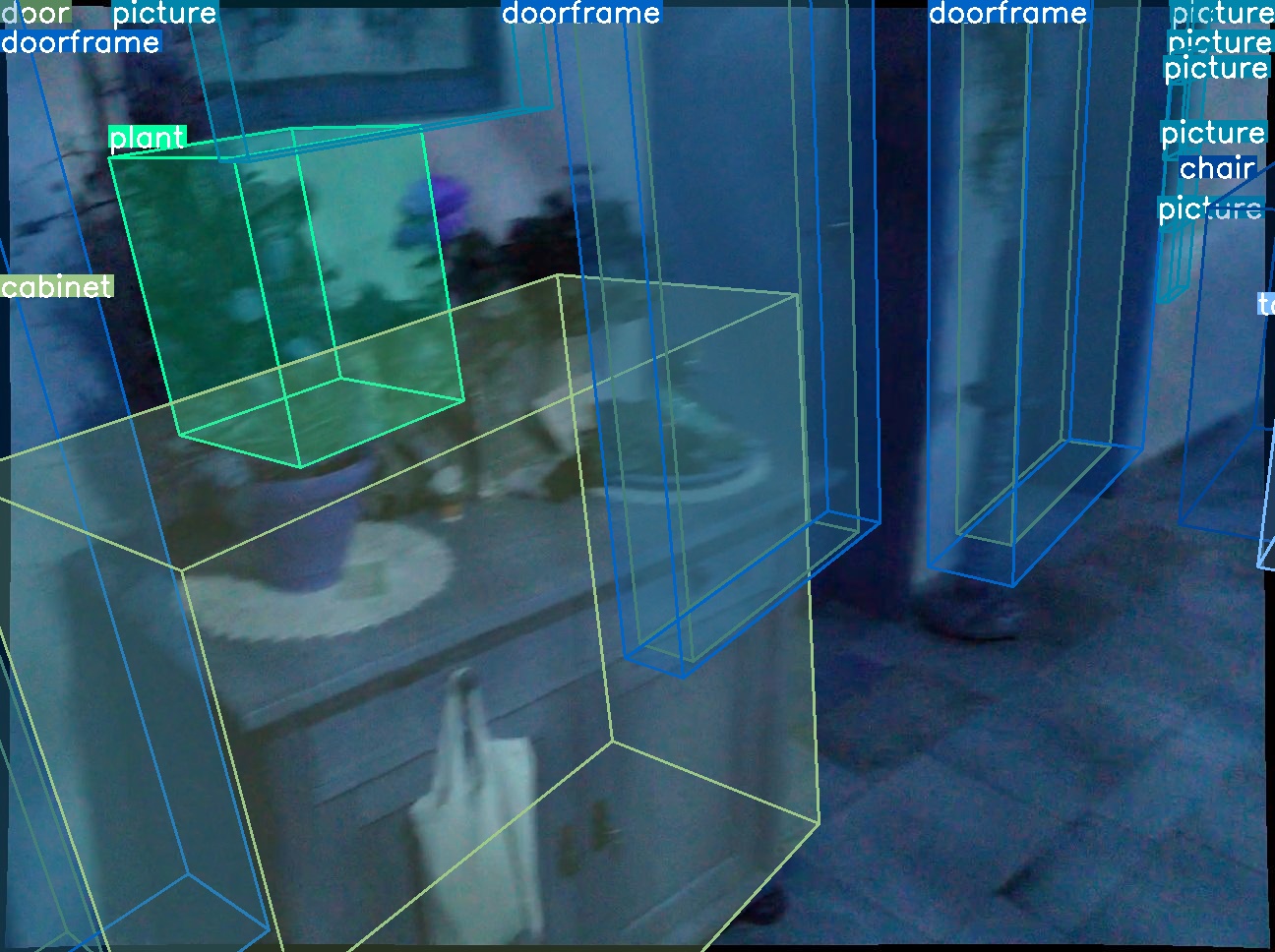} &
        \includegraphics[width=0.15\linewidth]{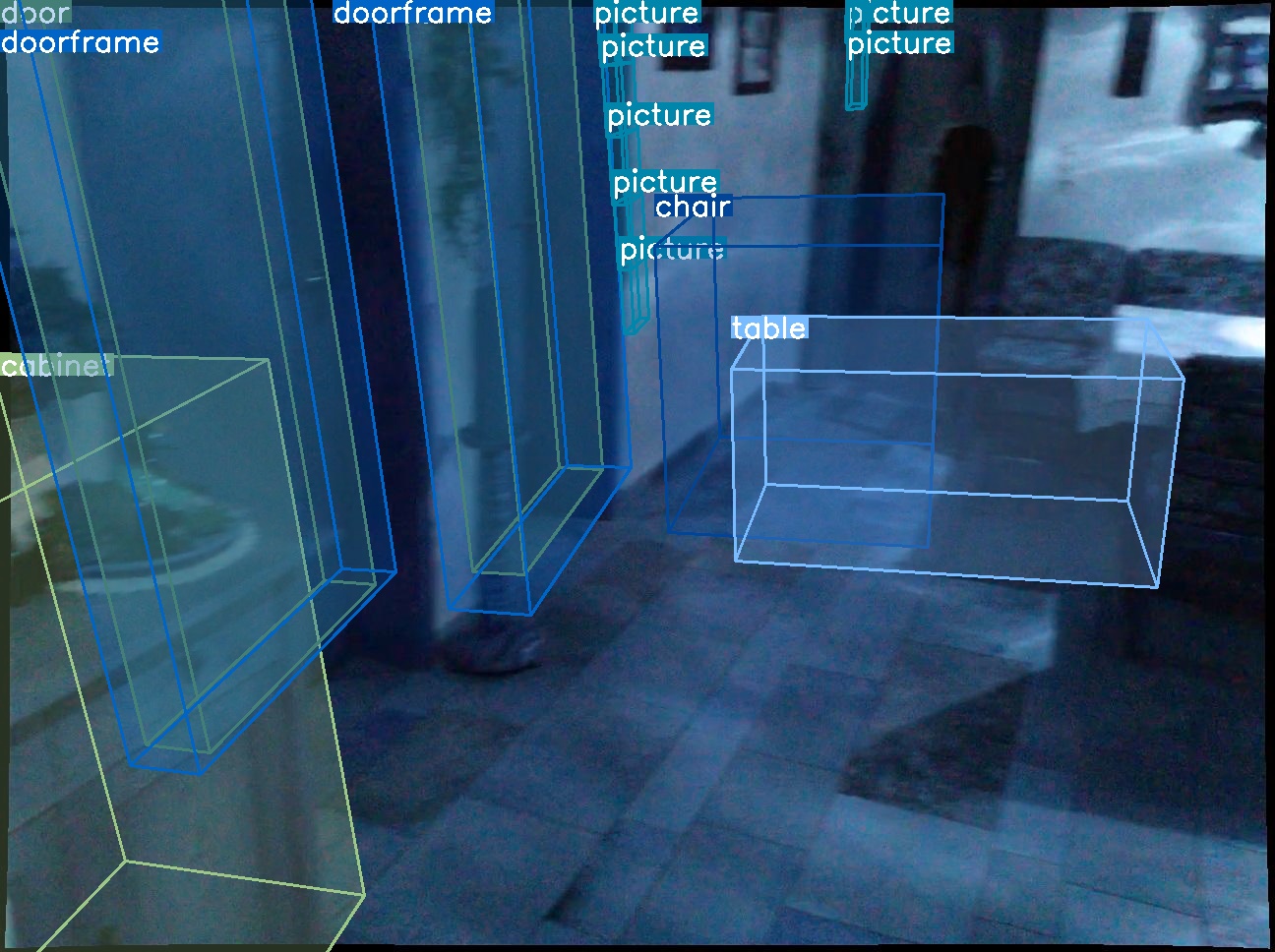} \\
        
        \rotatebox{90}{\small \model } & 
        \includegraphics[width=0.15\linewidth]{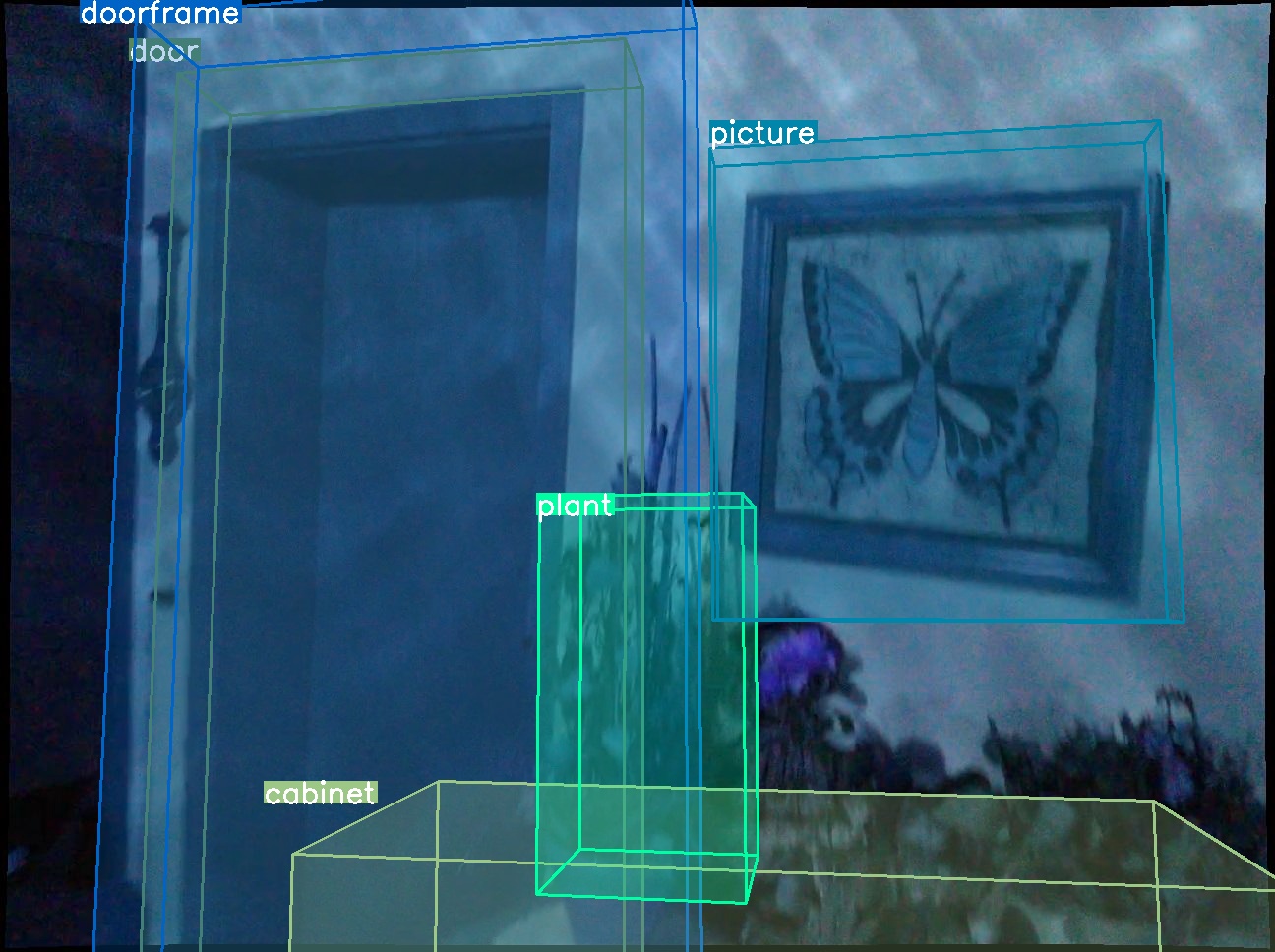} & 
        \includegraphics[width=0.15\linewidth]{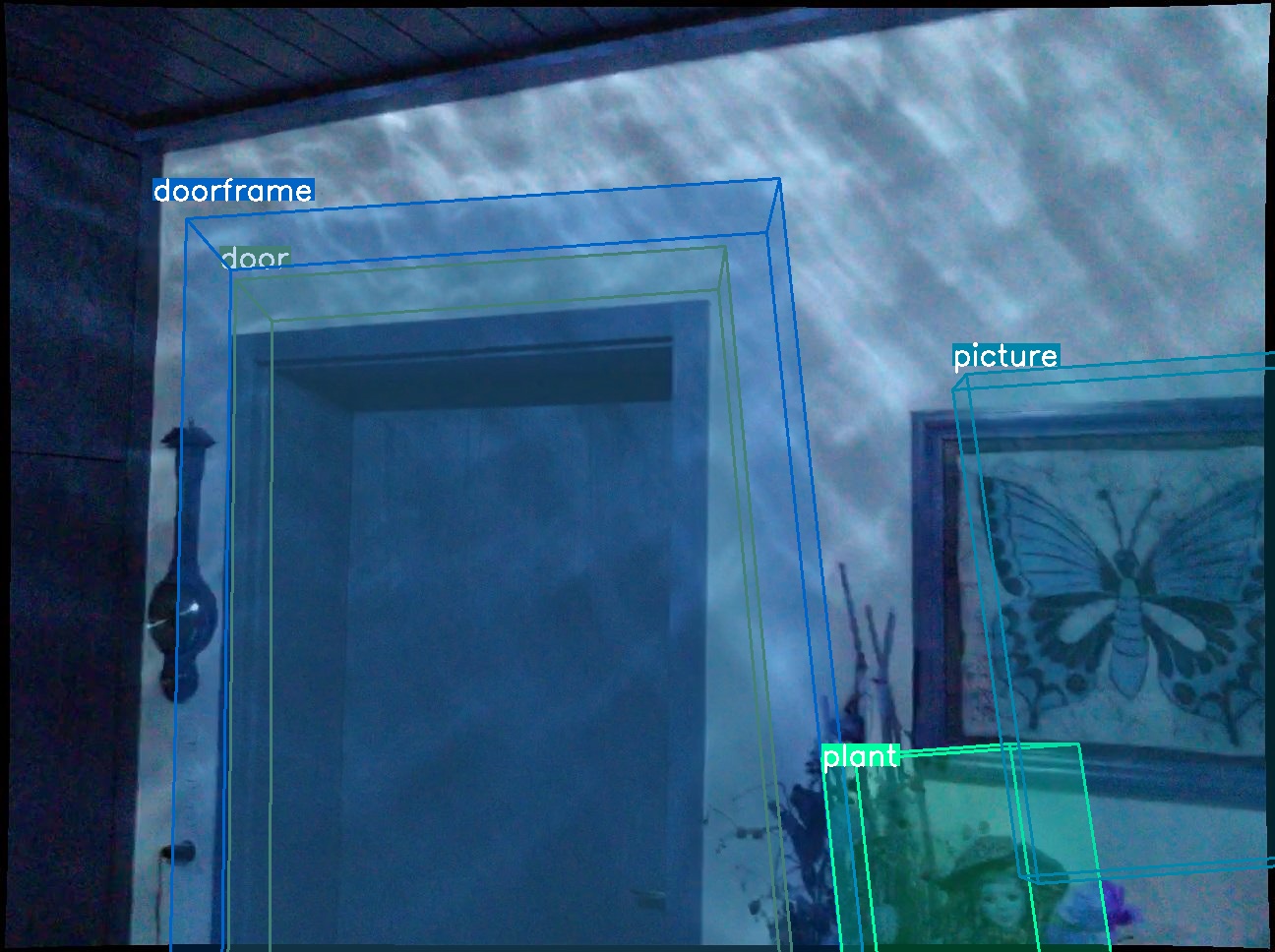} &
        \includegraphics[width=0.15\linewidth]{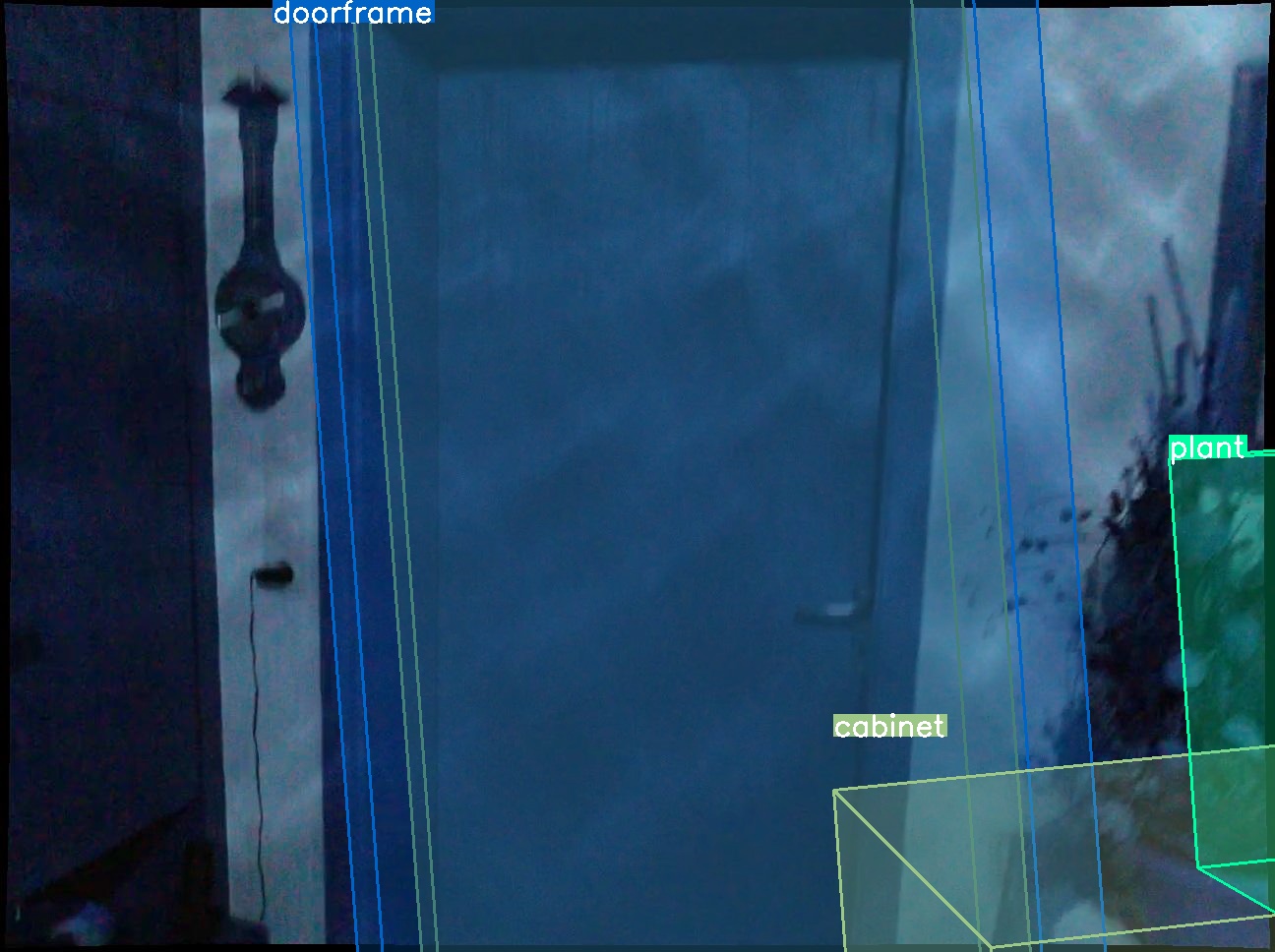} & 
        \includegraphics[width=0.15\linewidth]{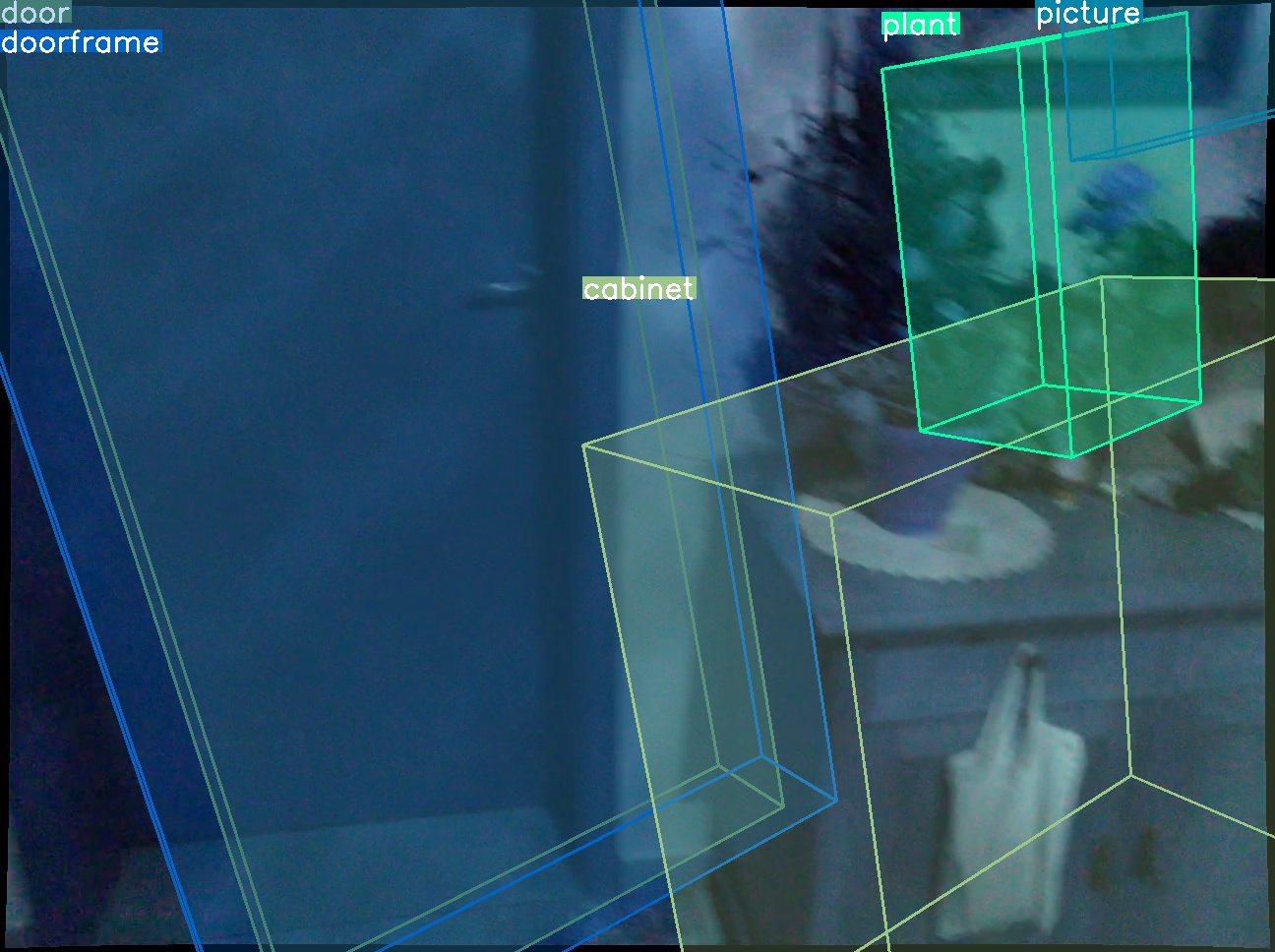} & 
        \includegraphics[width=0.15\linewidth]{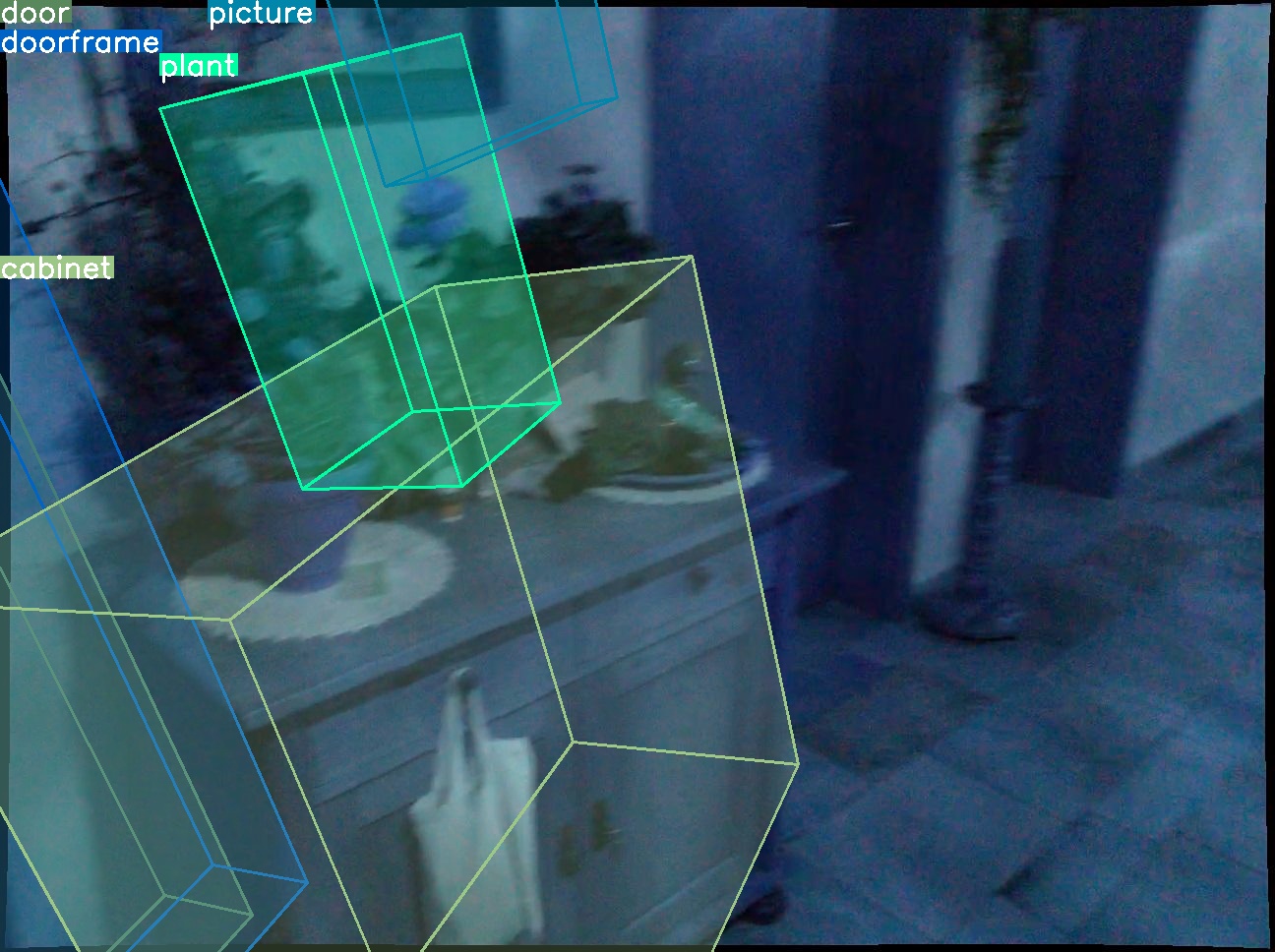} &
        \includegraphics[width=0.15\linewidth]{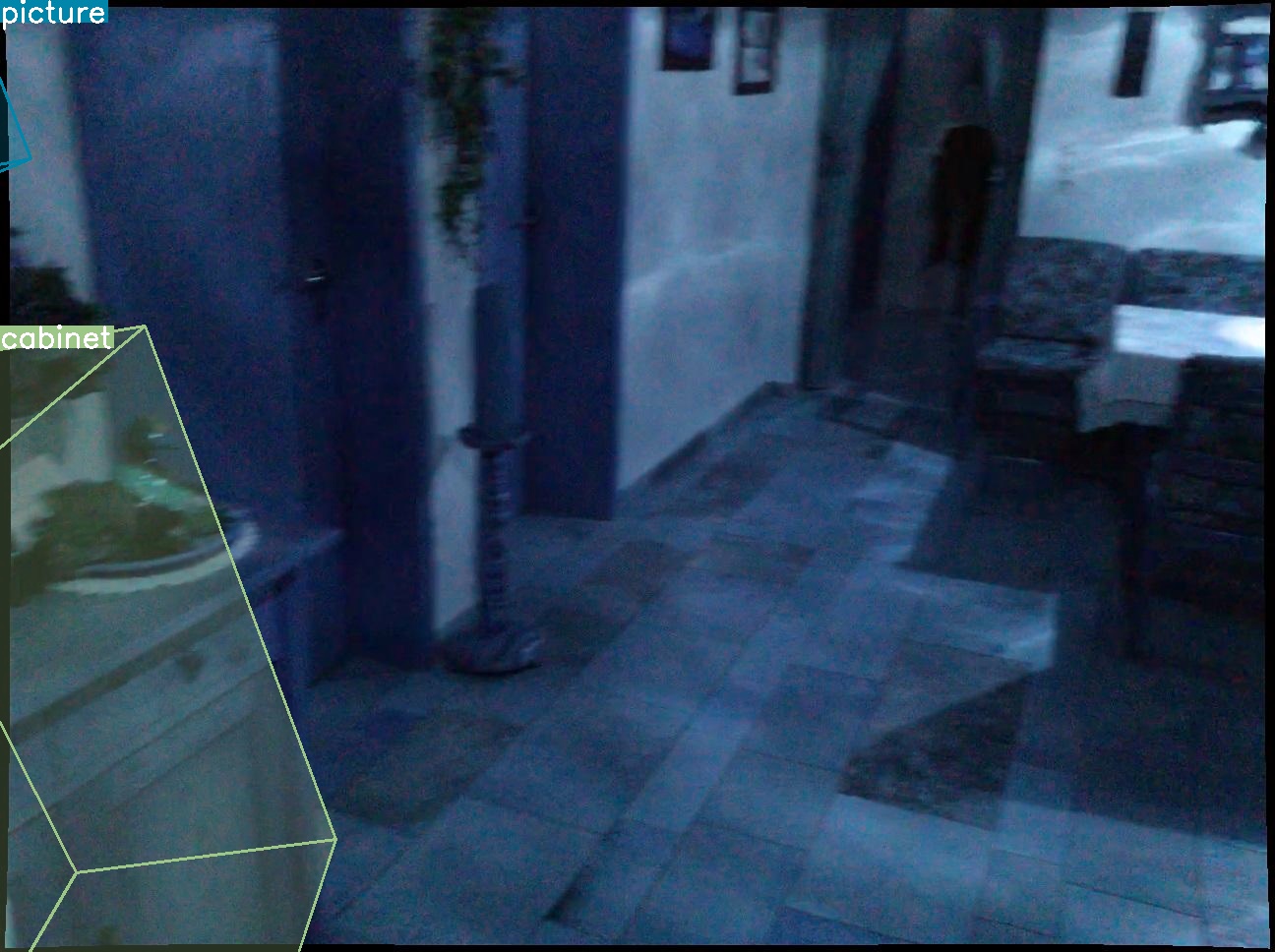} \\

        \\
        \\
        
        \rotatebox{90}{\small VG LLM-4B }& 
        \includegraphics[width=0.15\linewidth]{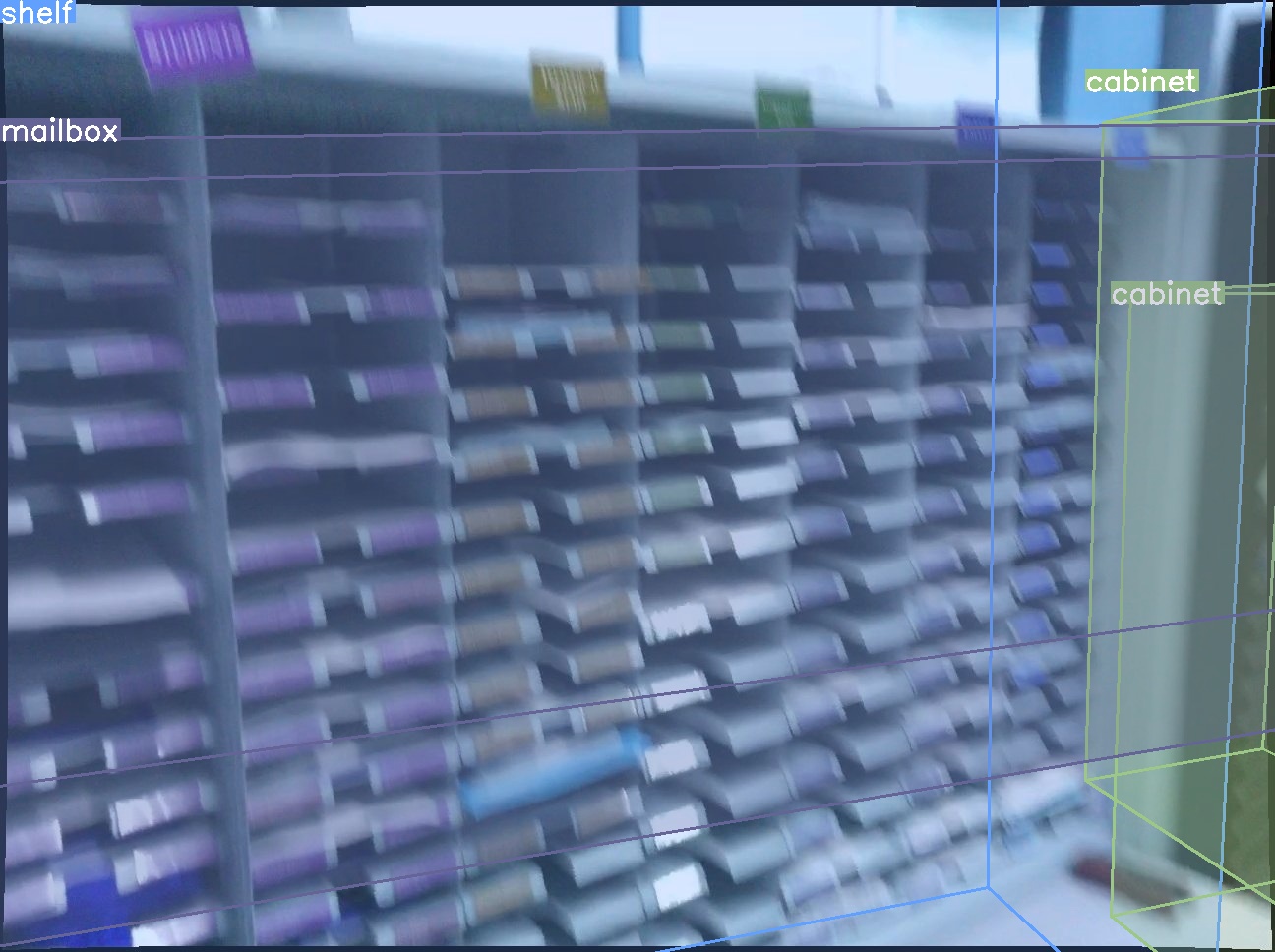} & 
        \includegraphics[width=0.15\linewidth]{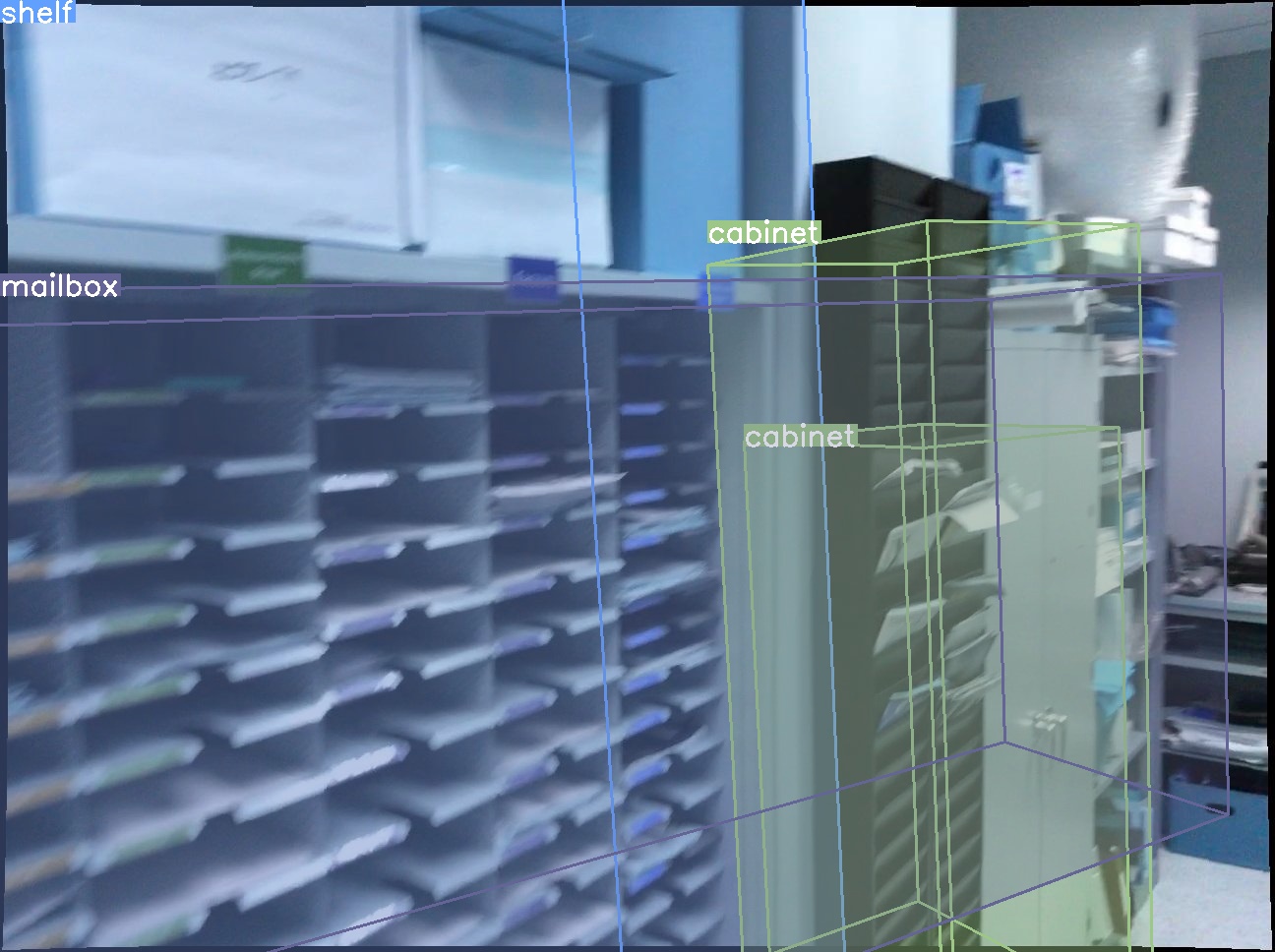} &
        \includegraphics[width=0.15\linewidth]{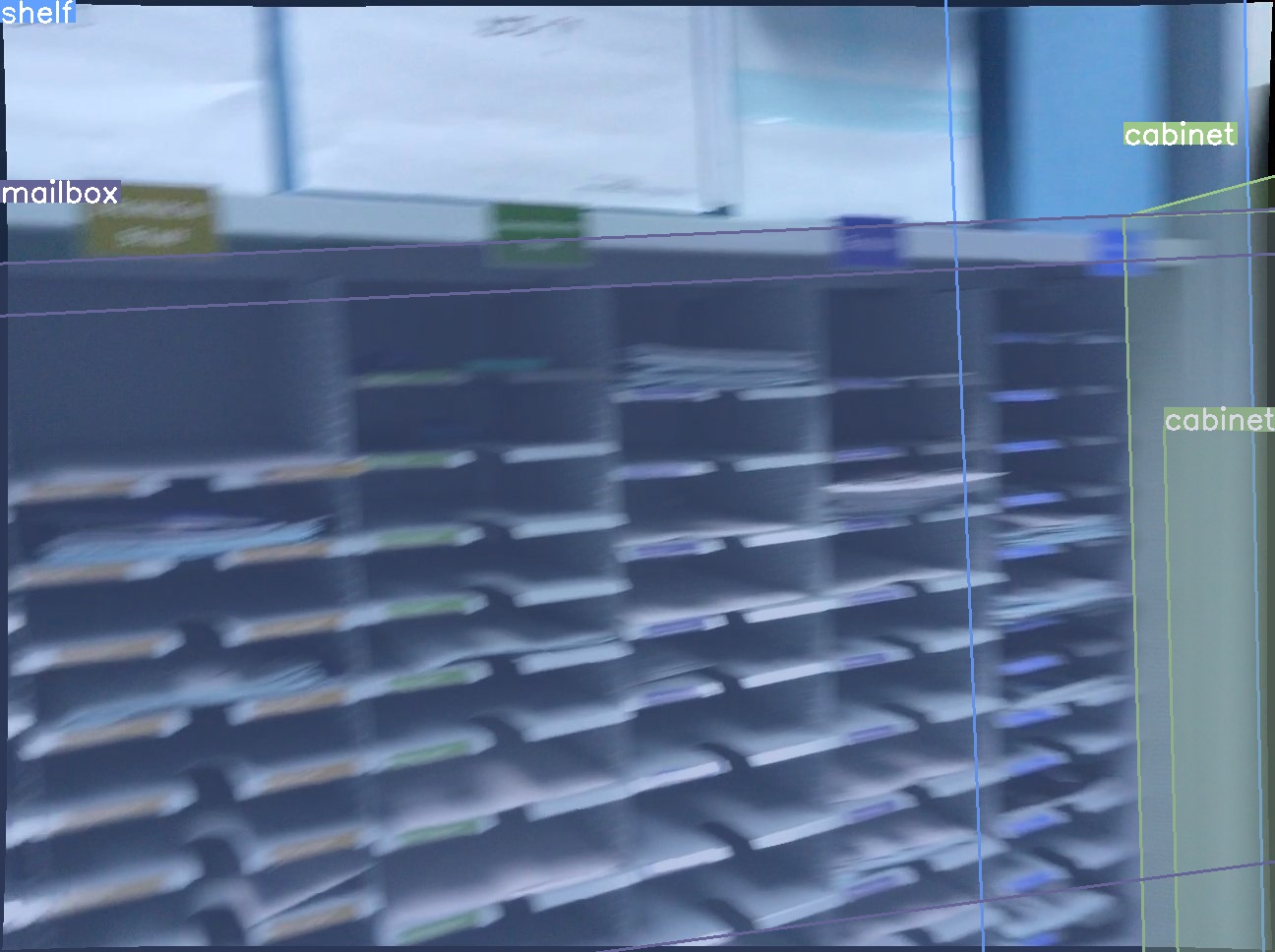} & 
        \includegraphics[width=0.15\linewidth]{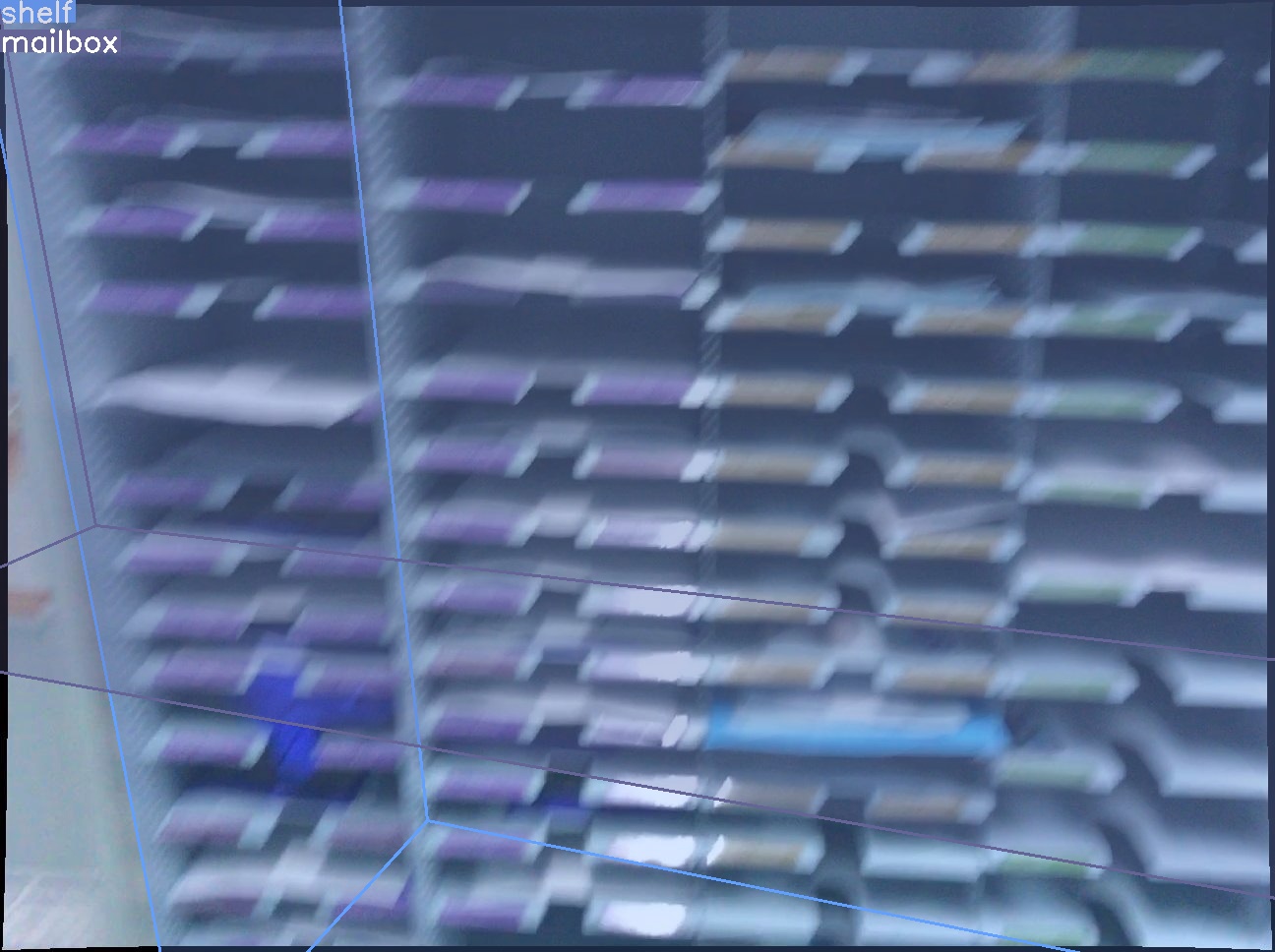} & 
        \includegraphics[width=0.15\linewidth]{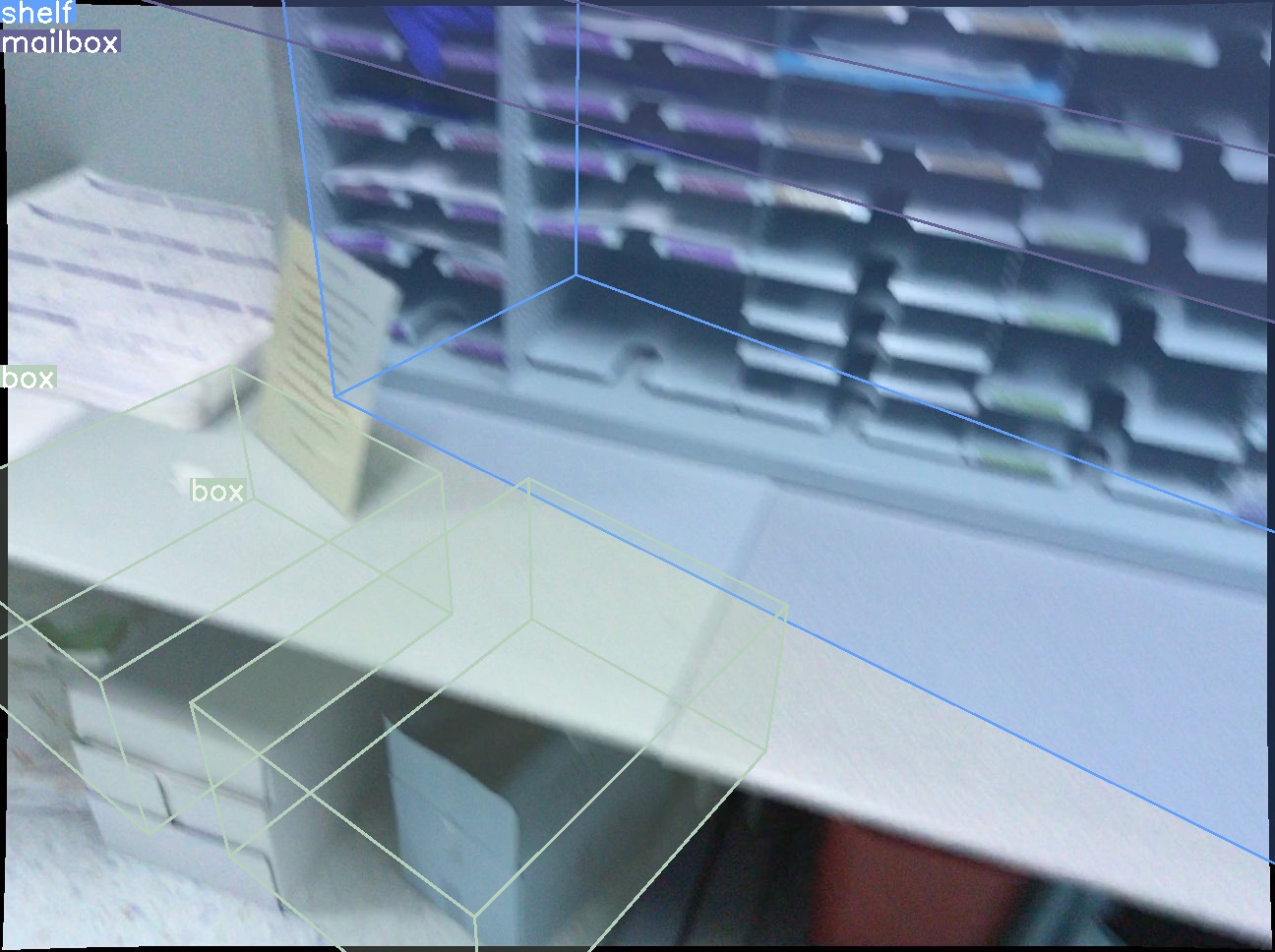} &
        \includegraphics[width=0.15\linewidth]{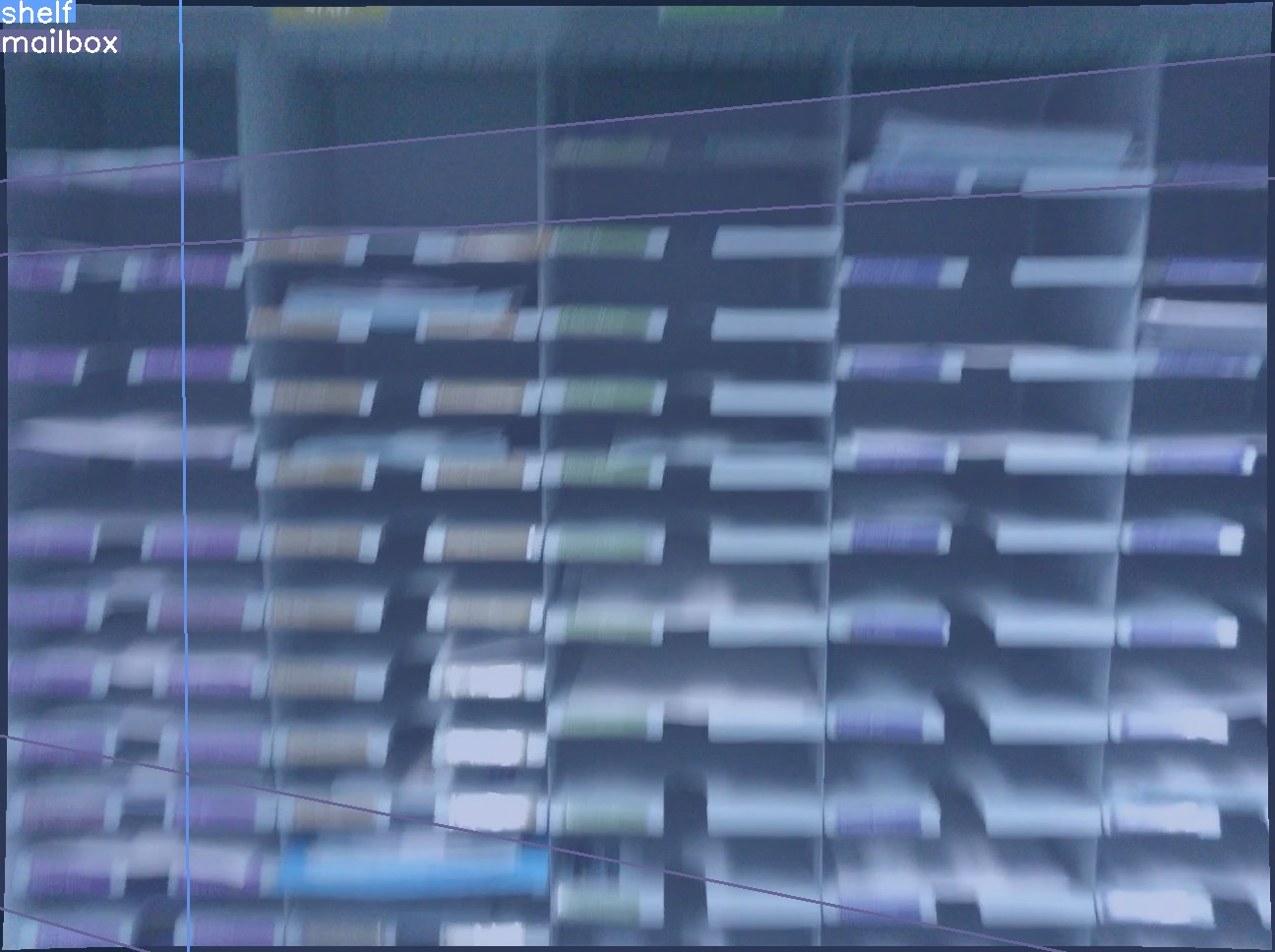} \\
        
        \rotatebox{90}{\small \model } & 
        \includegraphics[width=0.15\linewidth]{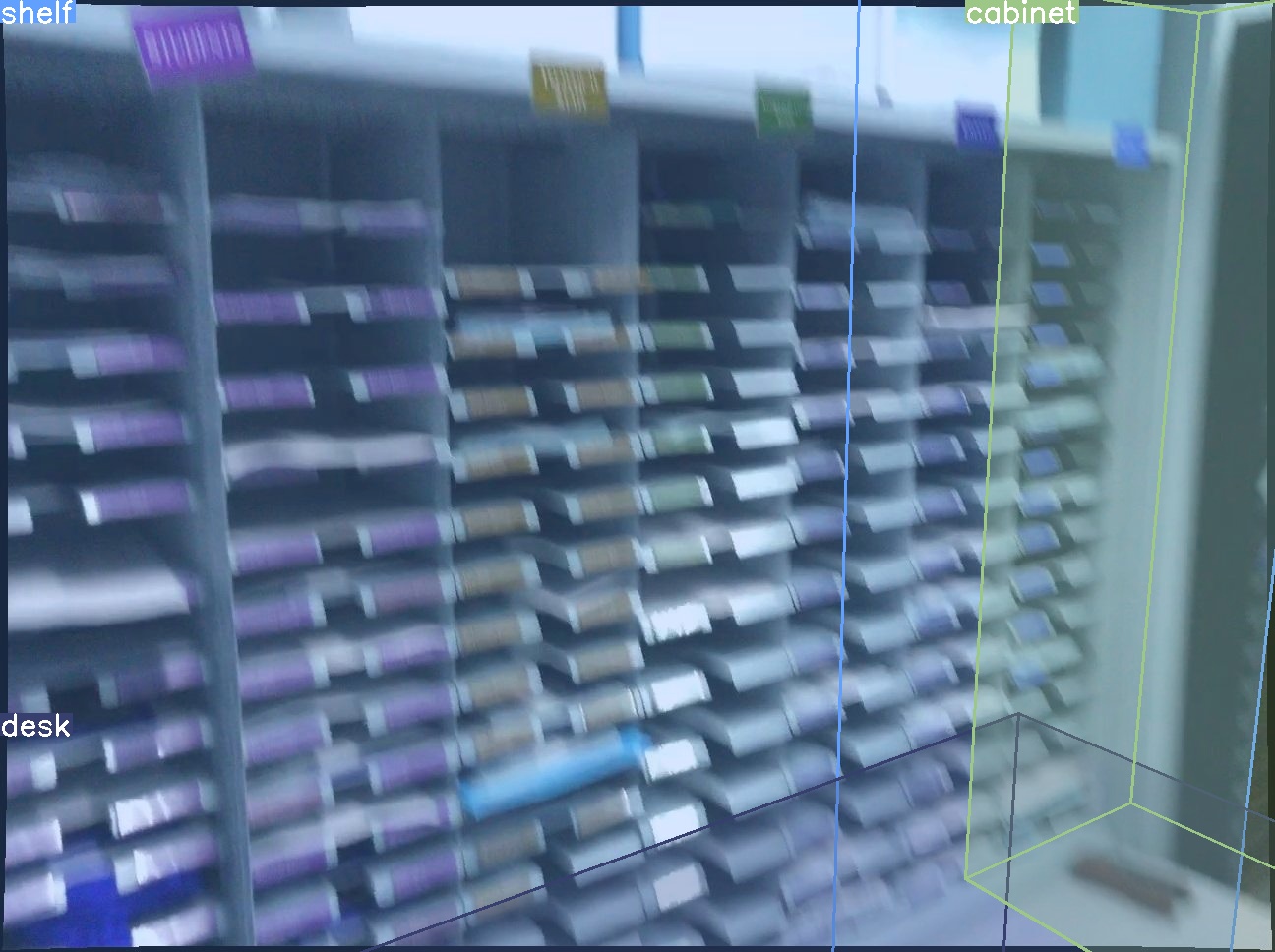} & 
        \includegraphics[width=0.15\linewidth]{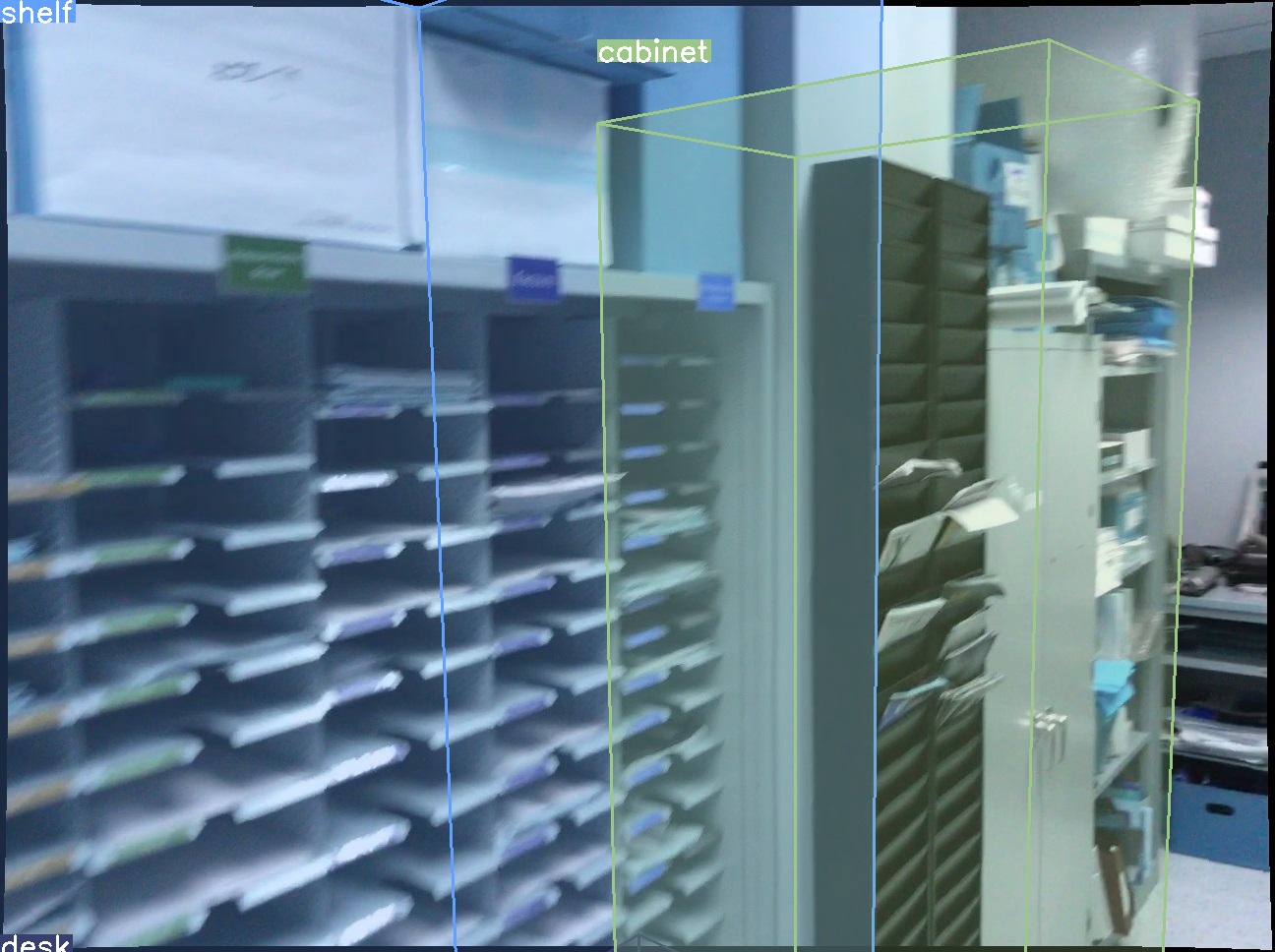} &
        \includegraphics[width=0.15\linewidth]{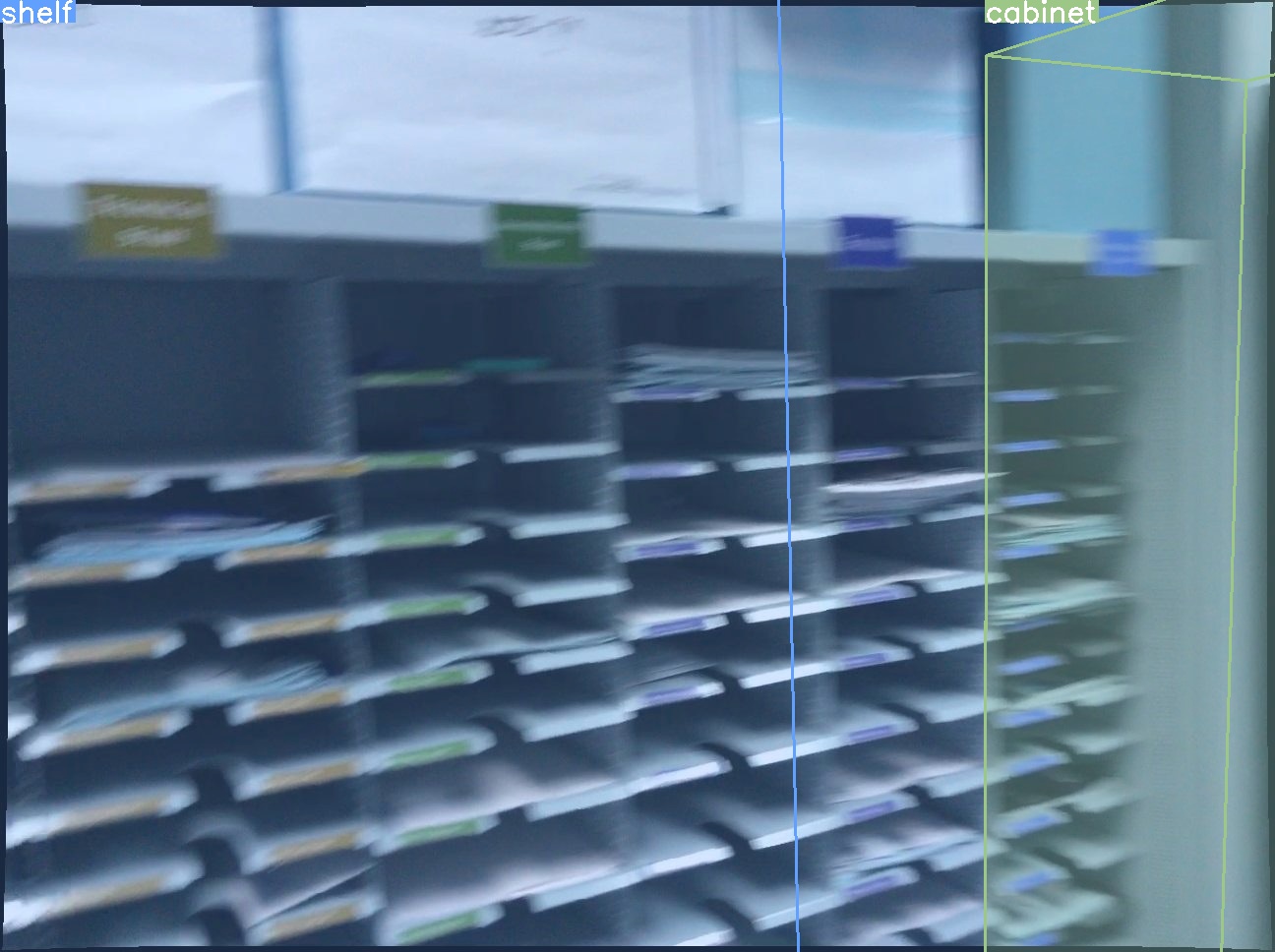} & 
        \includegraphics[width=0.15\linewidth]{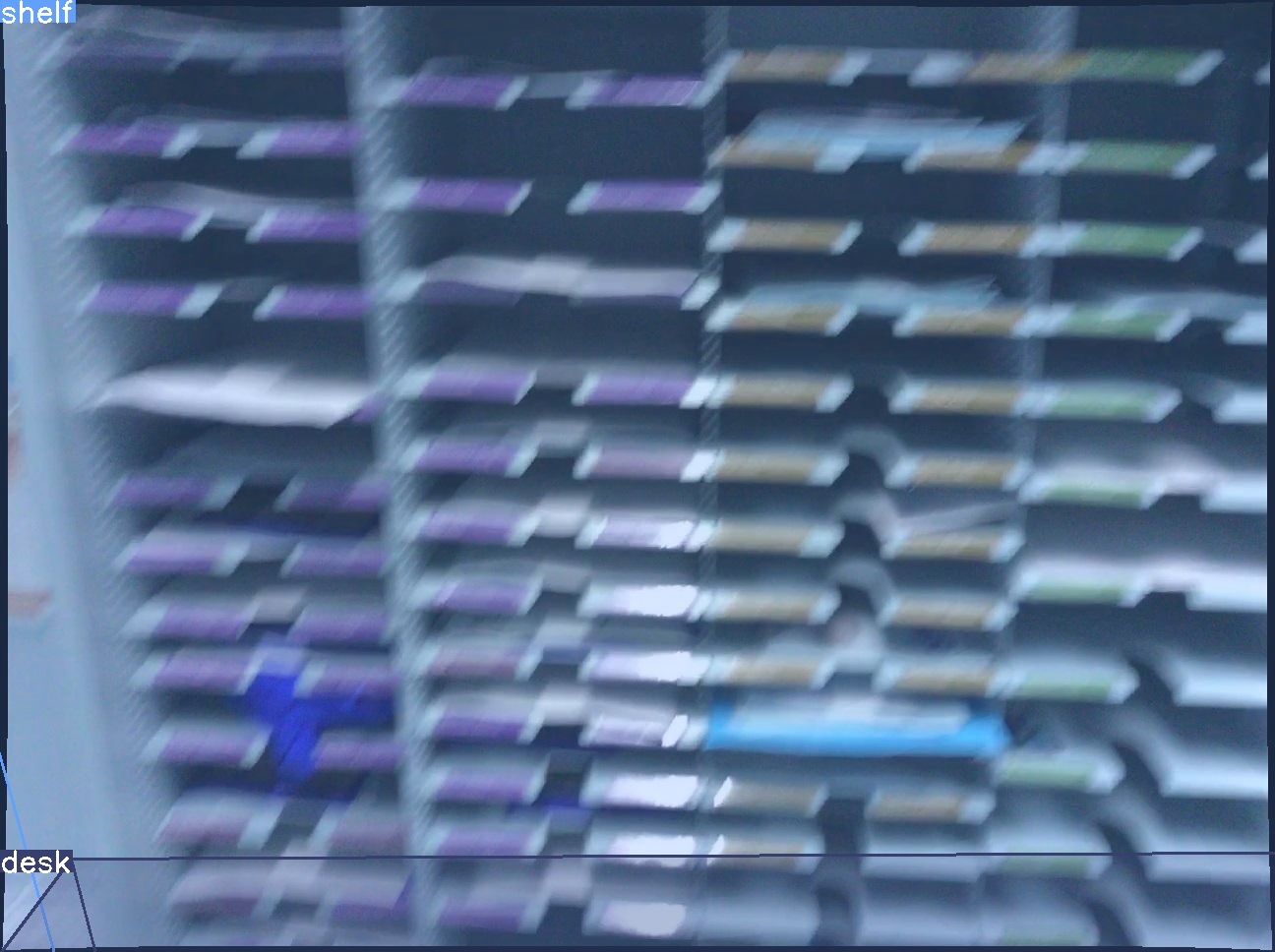} & 
        \includegraphics[width=0.15\linewidth]{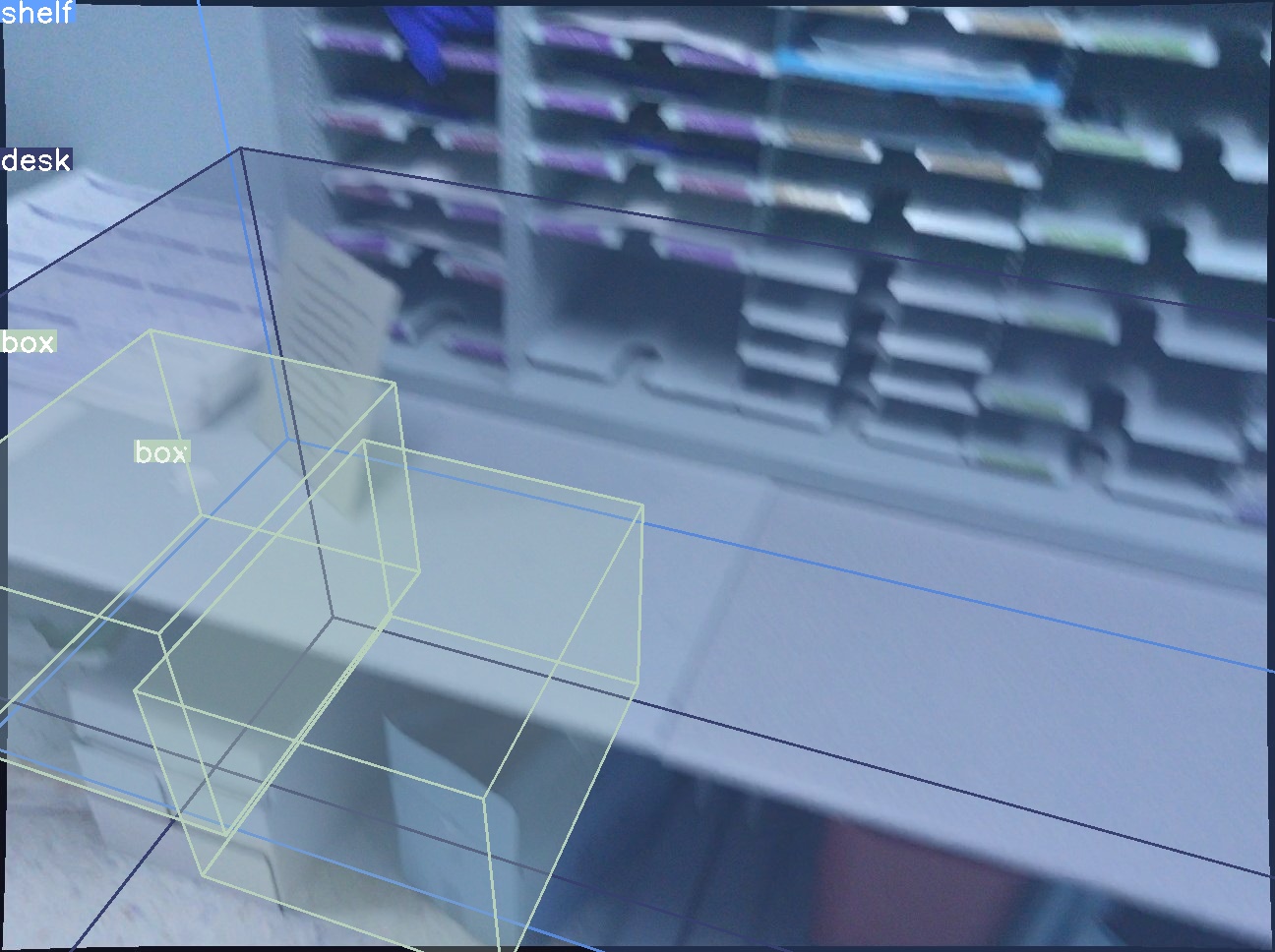} &
        \includegraphics[width=0.15\linewidth]{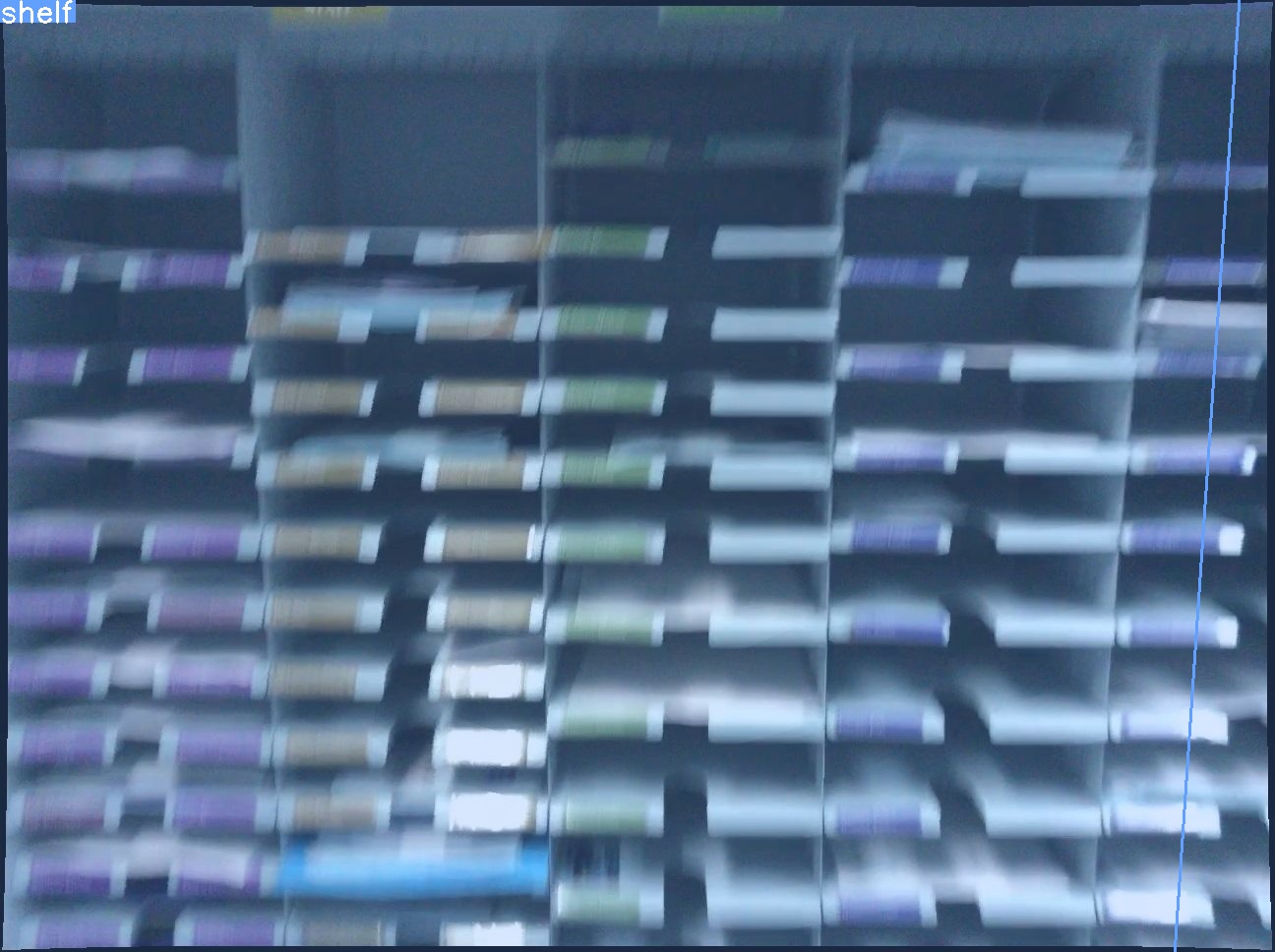} \\

        \\
        \\
        
        \rotatebox{90}{\small VG LLM-4B} & 
        \includegraphics[width=0.15\linewidth]{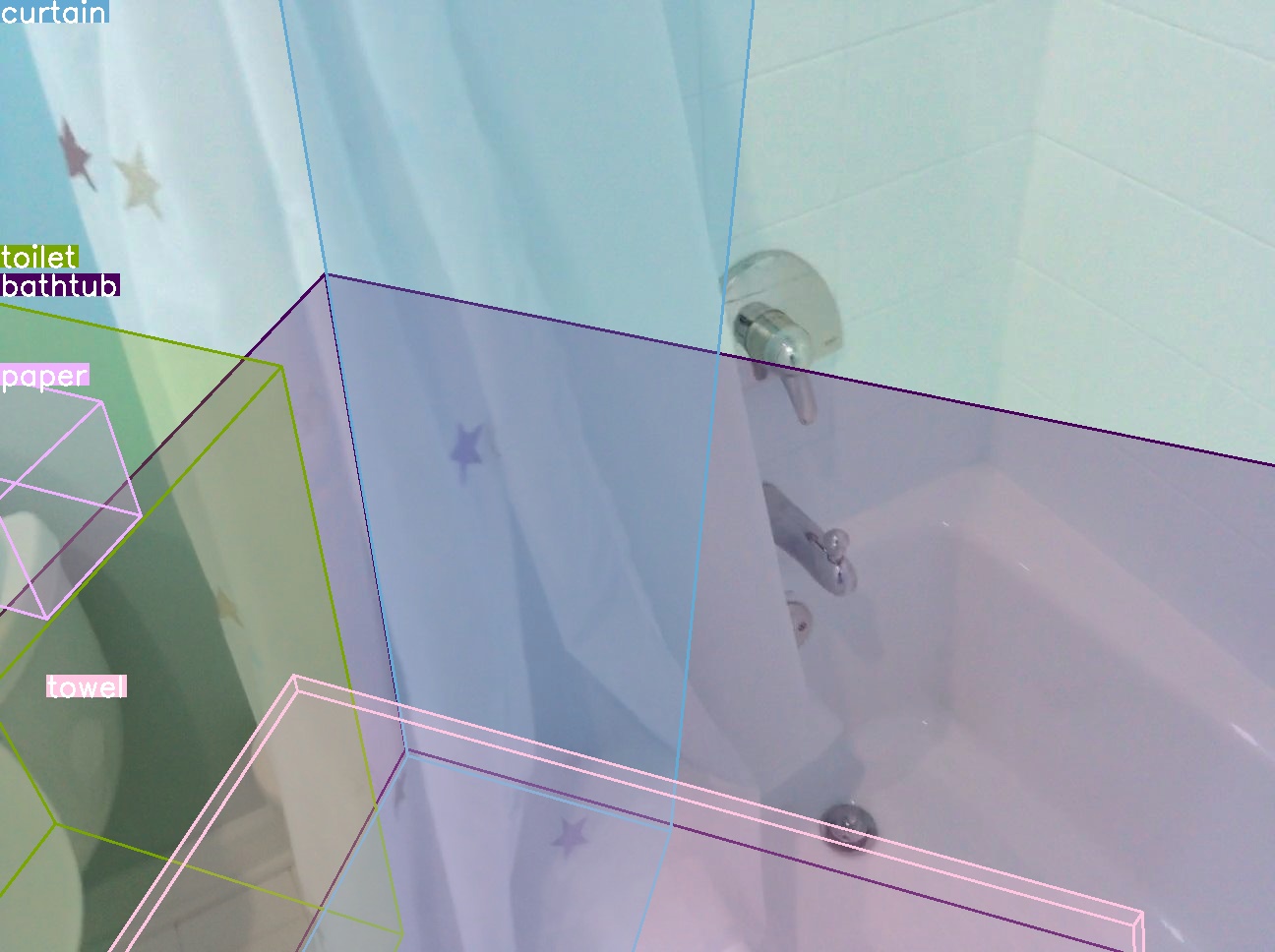} & 
        \includegraphics[width=0.15\linewidth]{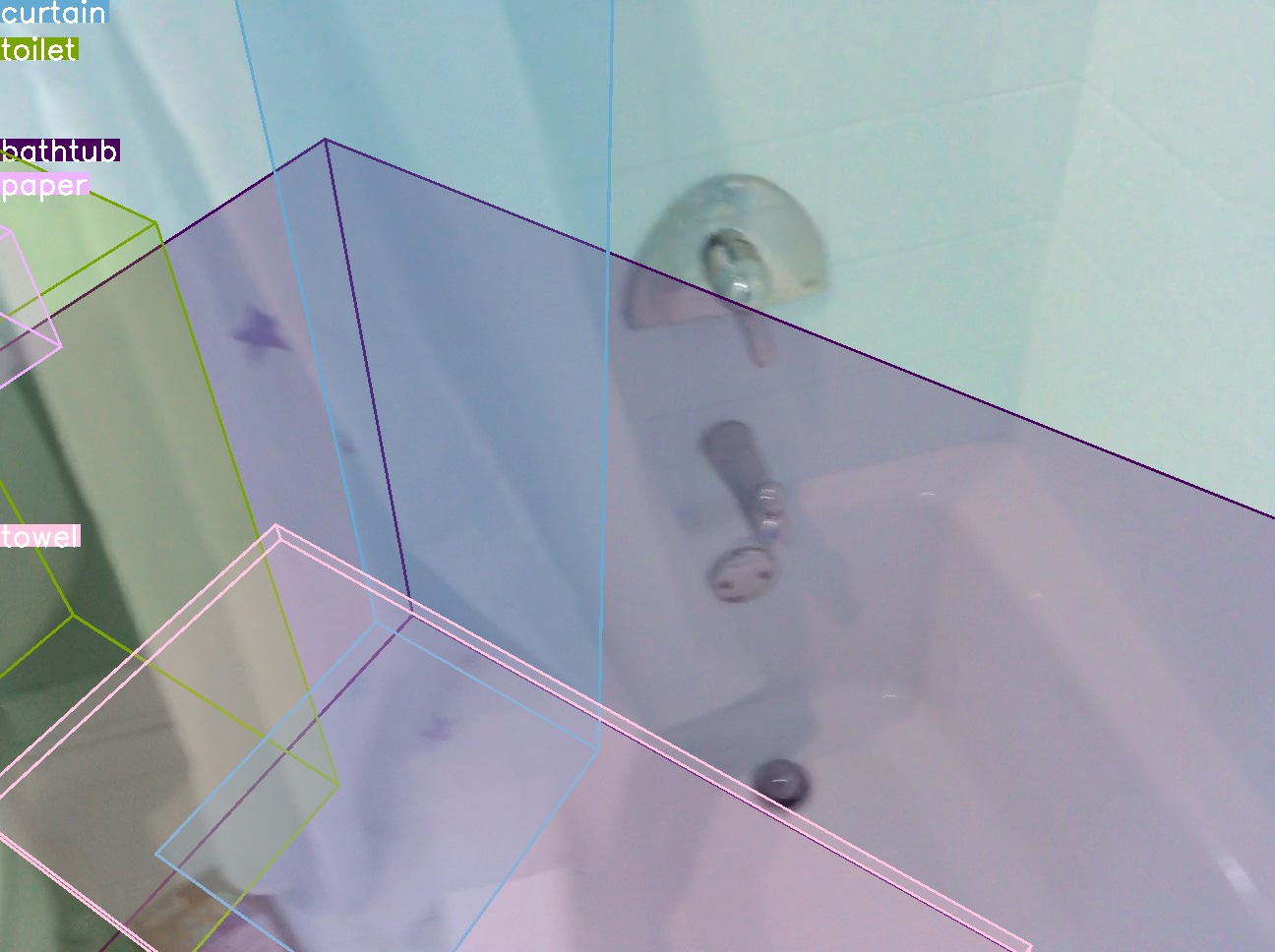} &
        \includegraphics[width=0.15\linewidth]{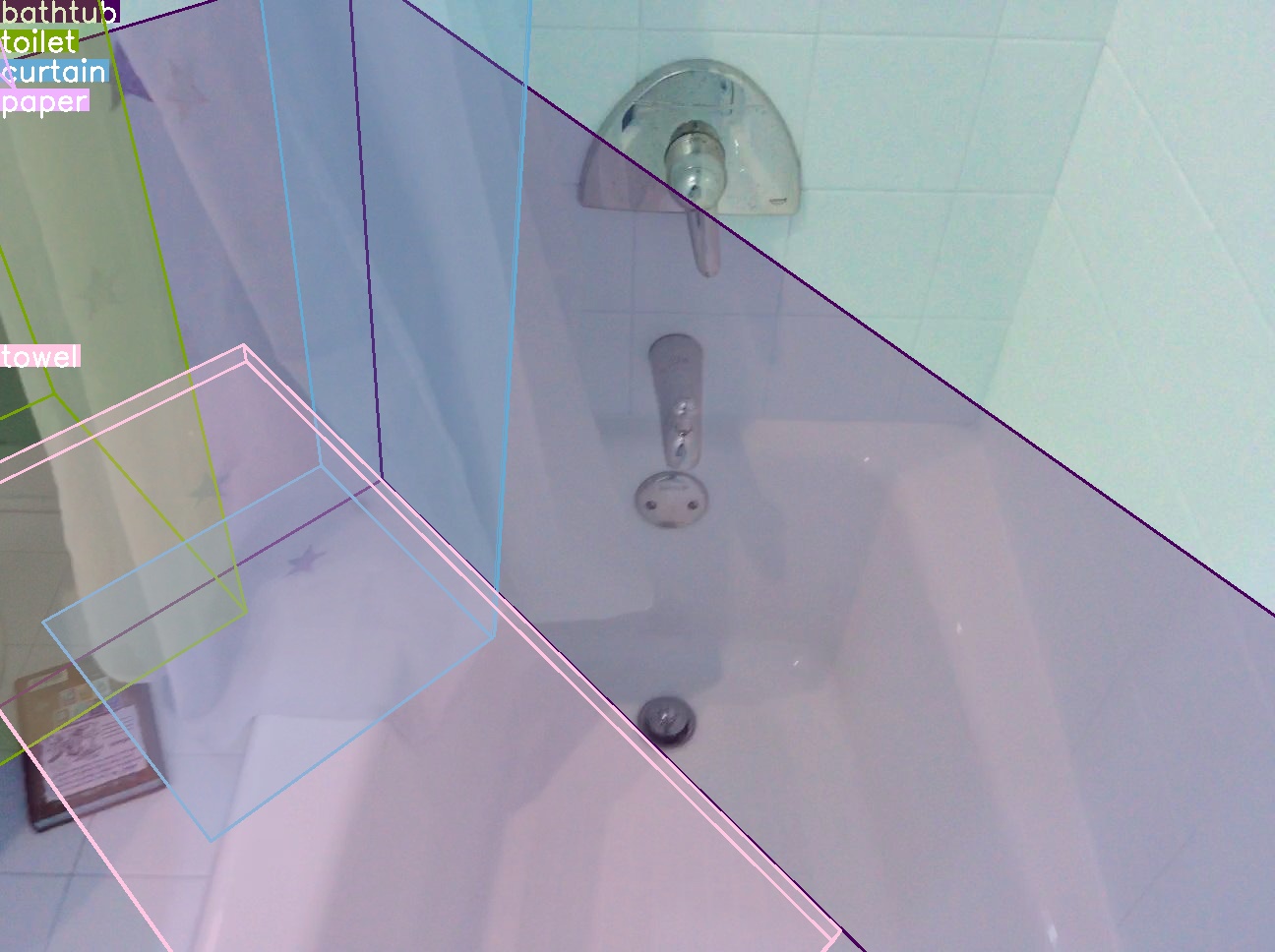} & 
        \includegraphics[width=0.15\linewidth]{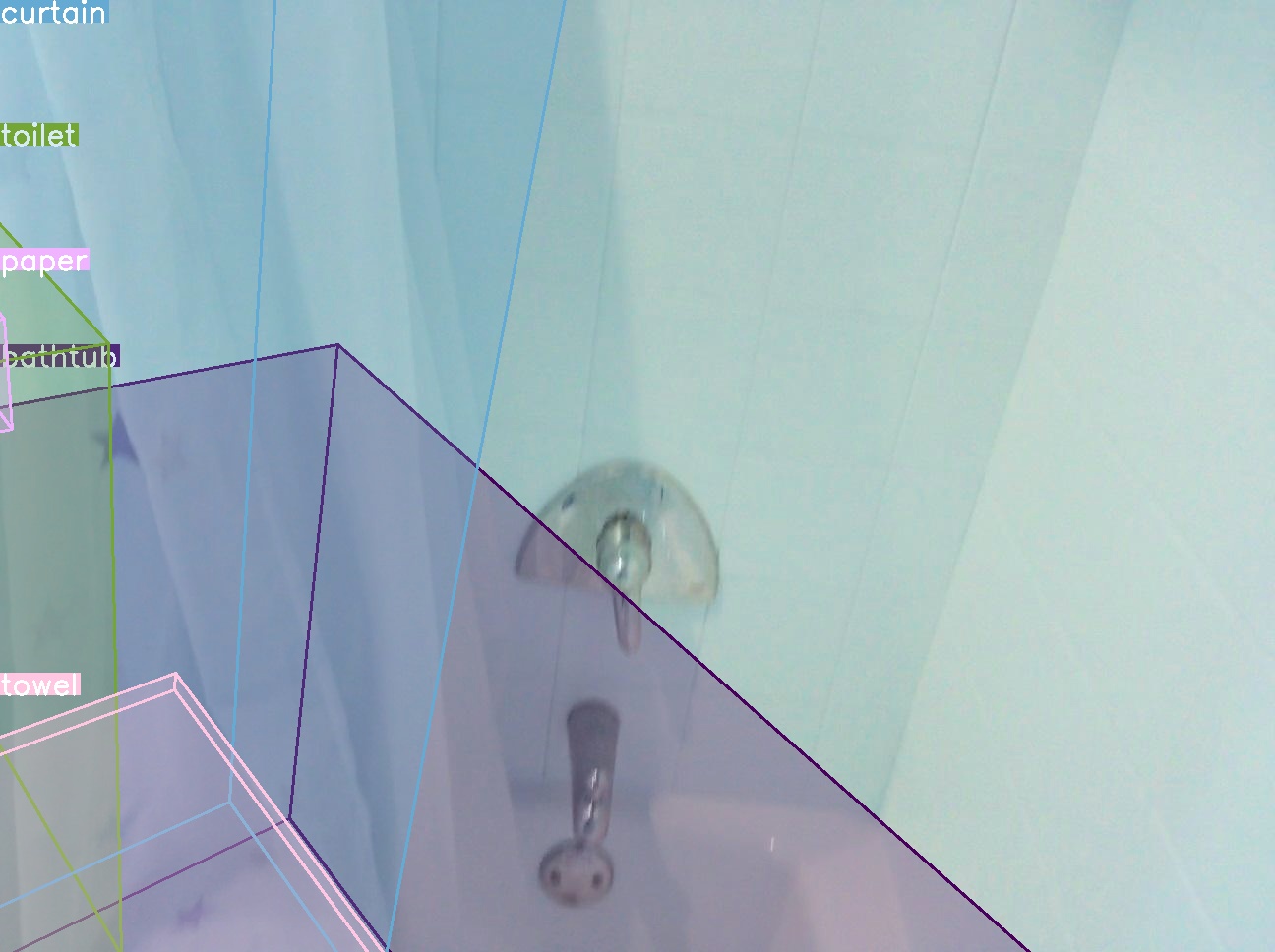} & 
        \includegraphics[width=0.15\linewidth]{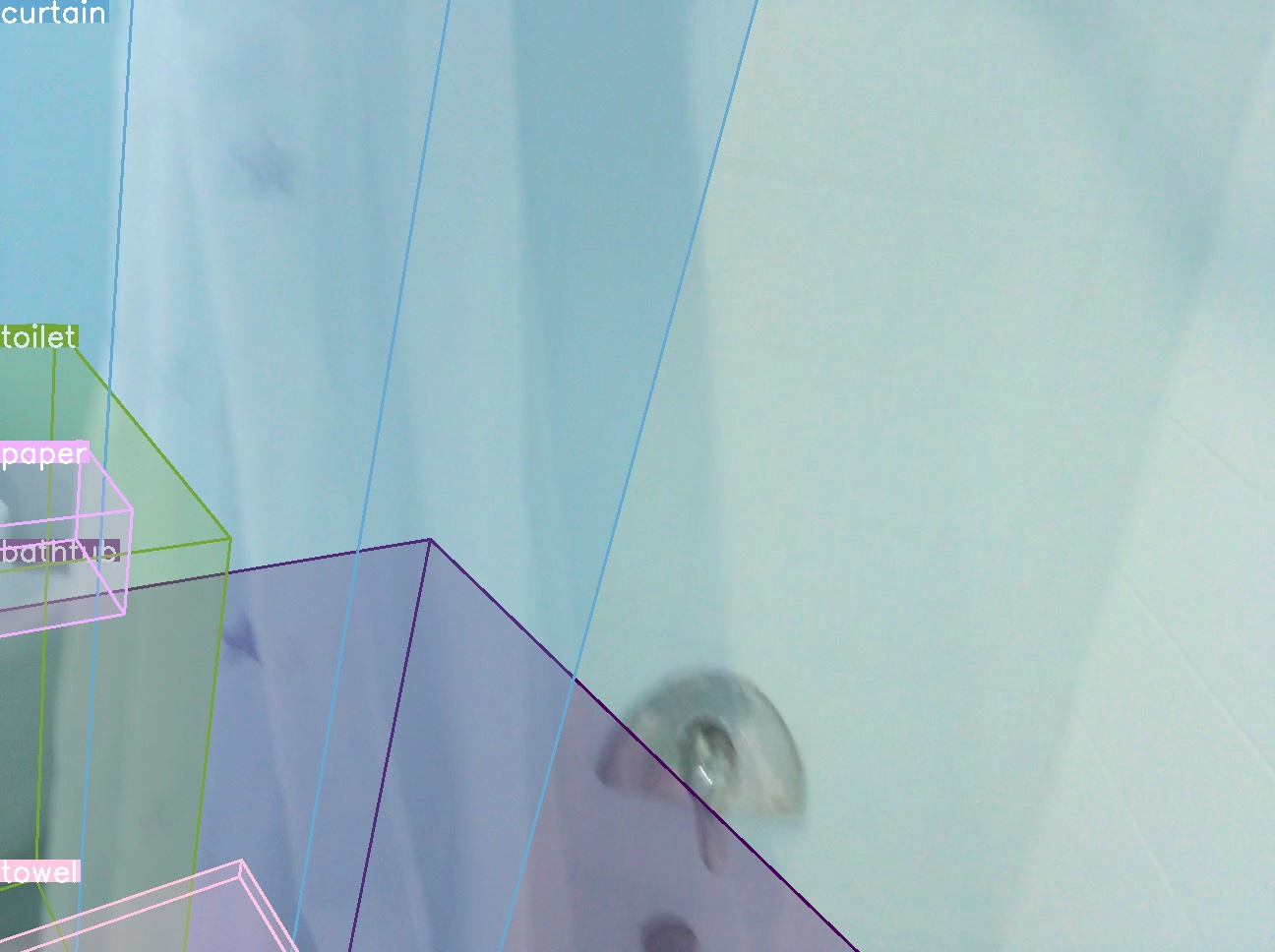} &
        \includegraphics[width=0.15\linewidth]{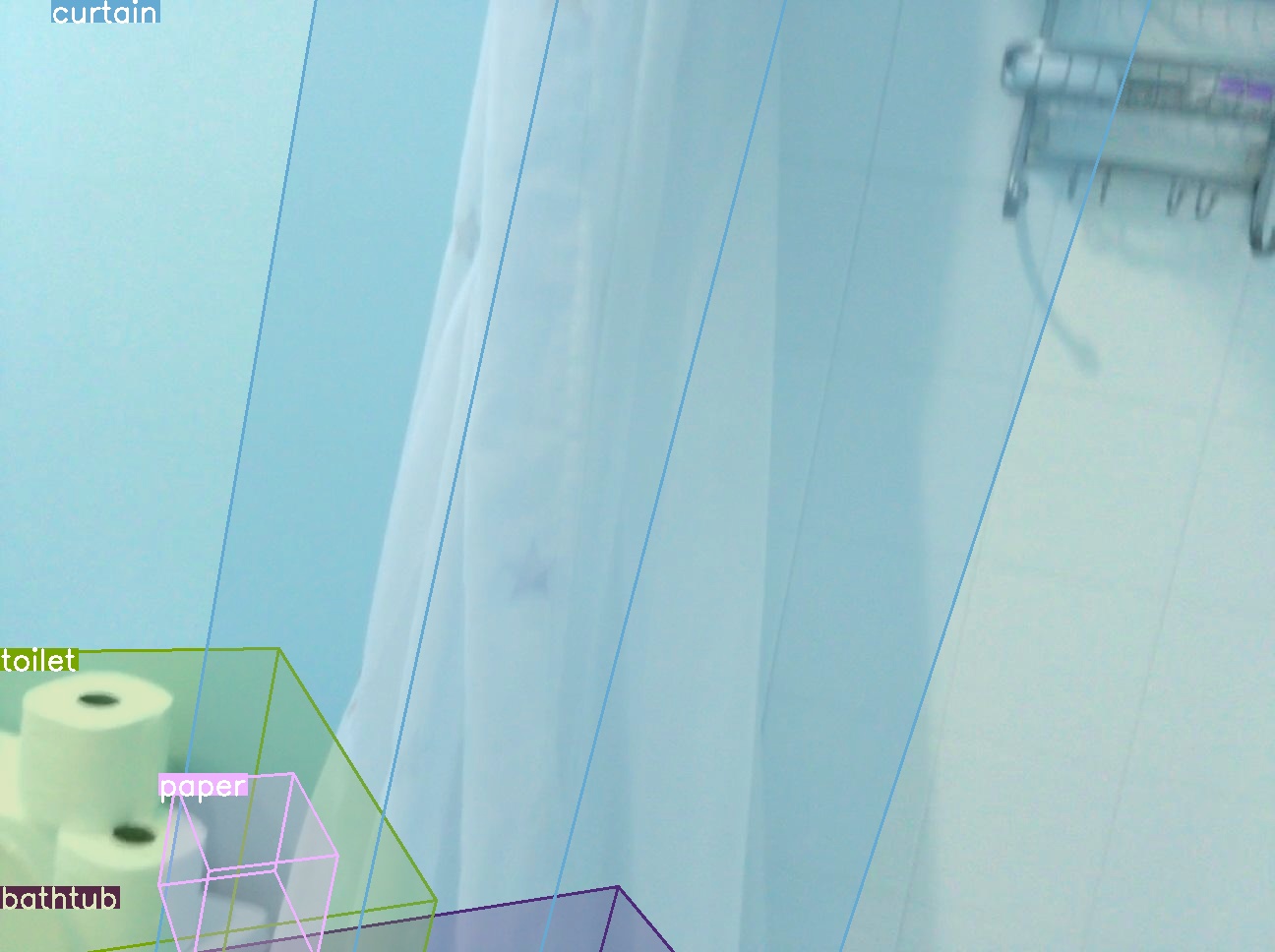} \\
        
        \rotatebox{90}{\small \model } & 
        \includegraphics[width=0.15\linewidth]{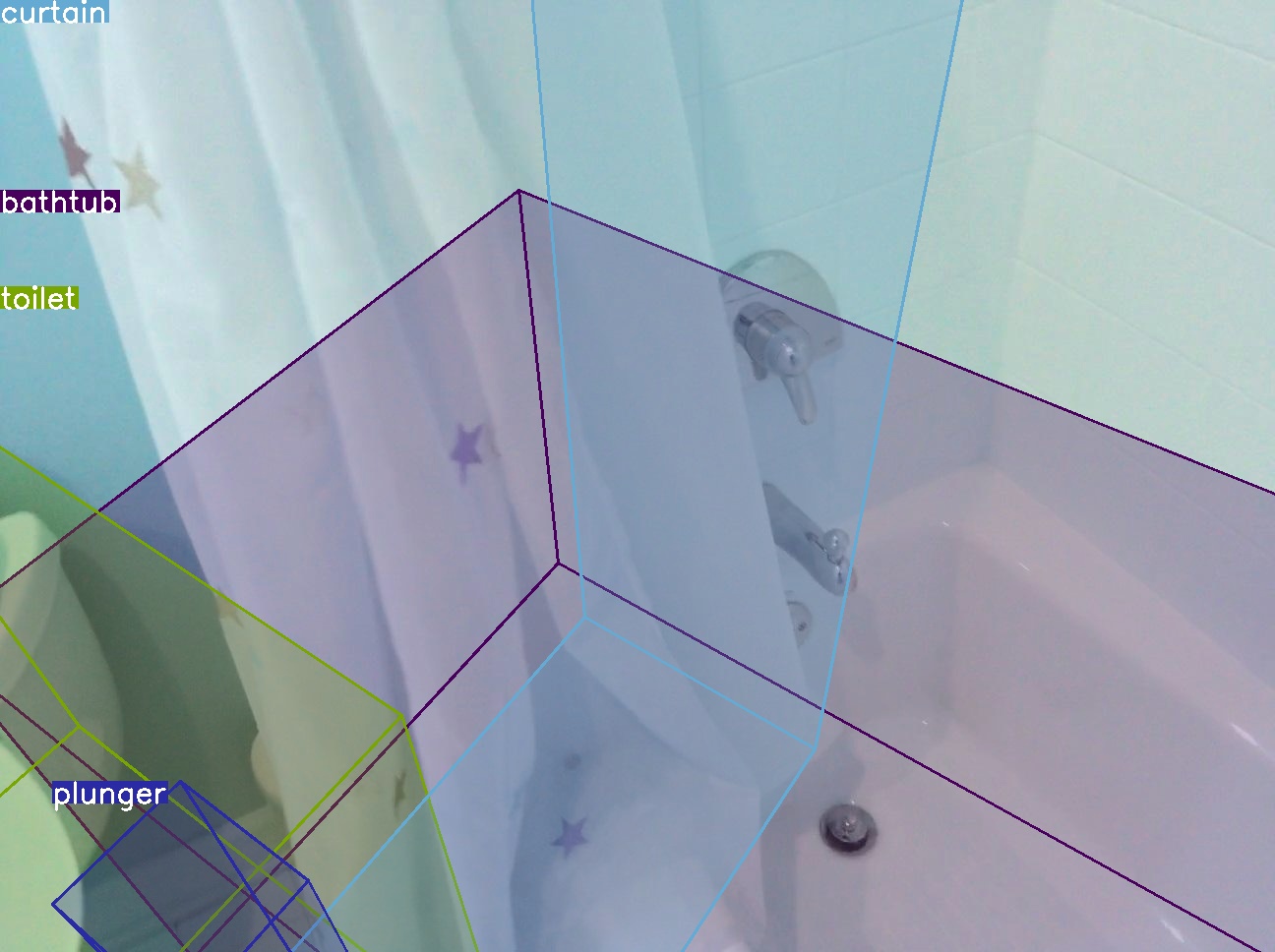} & 
        \includegraphics[width=0.15\linewidth]{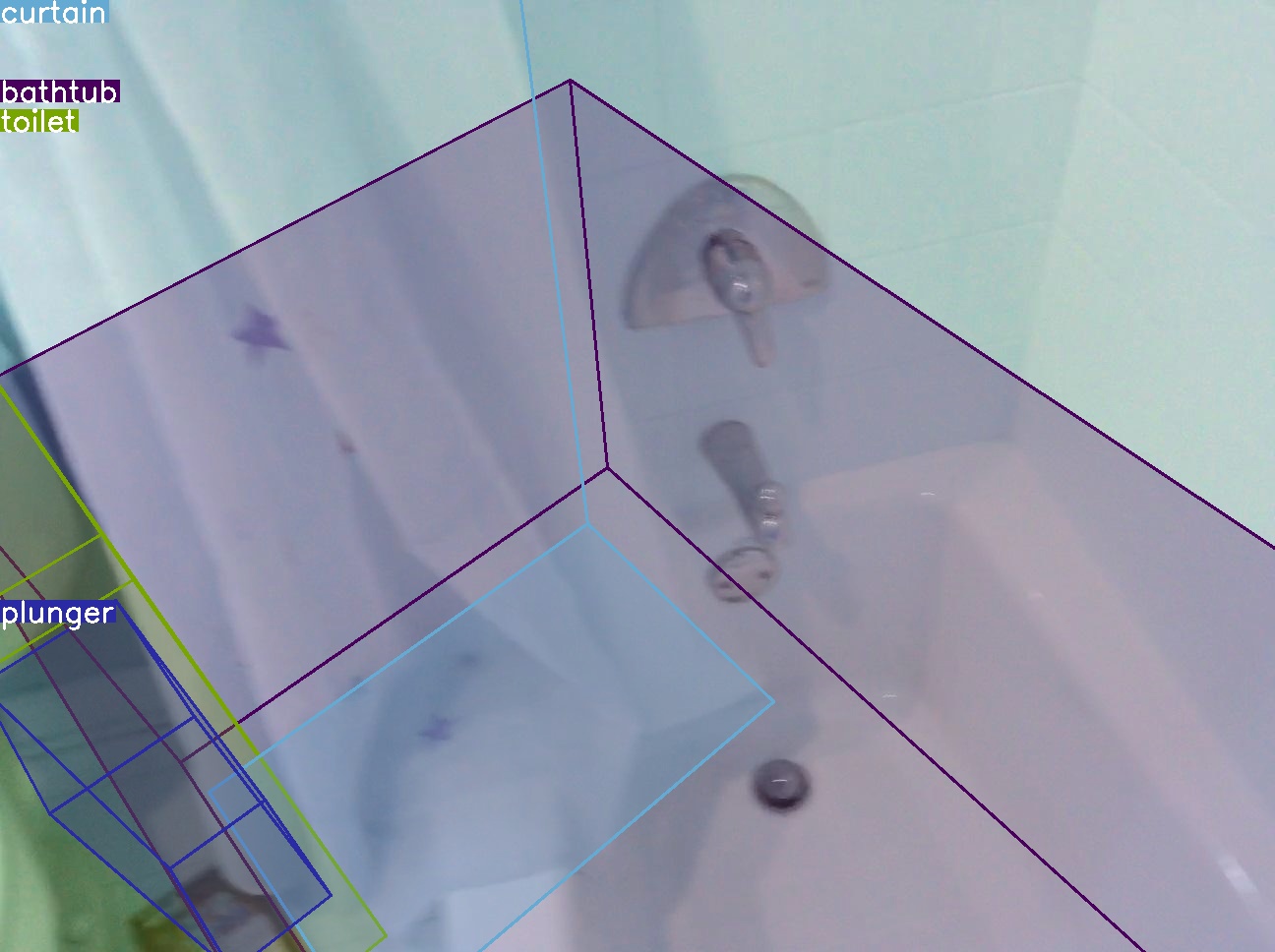} &
        \includegraphics[width=0.15\linewidth]{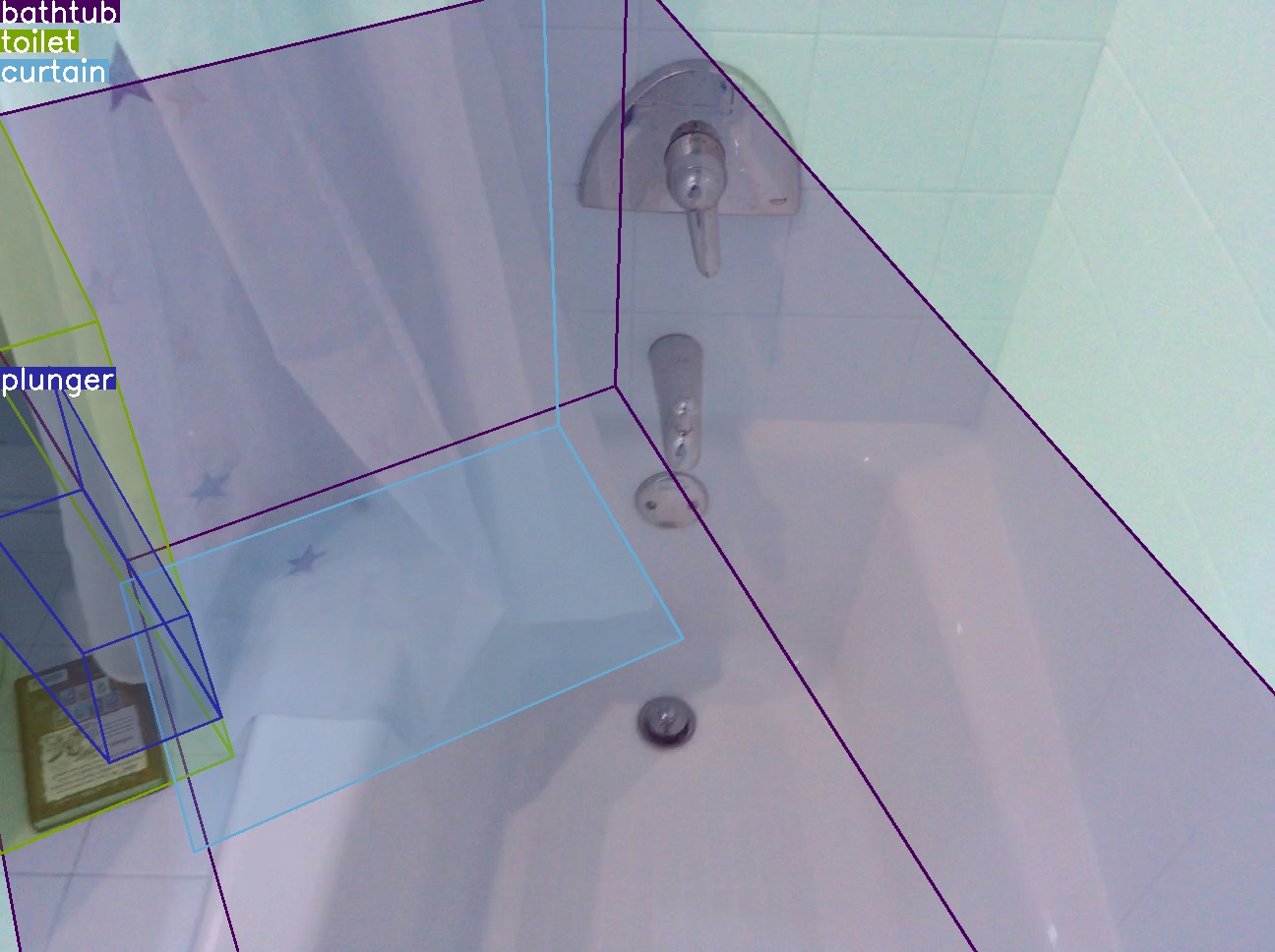} & 
        \includegraphics[width=0.15\linewidth]{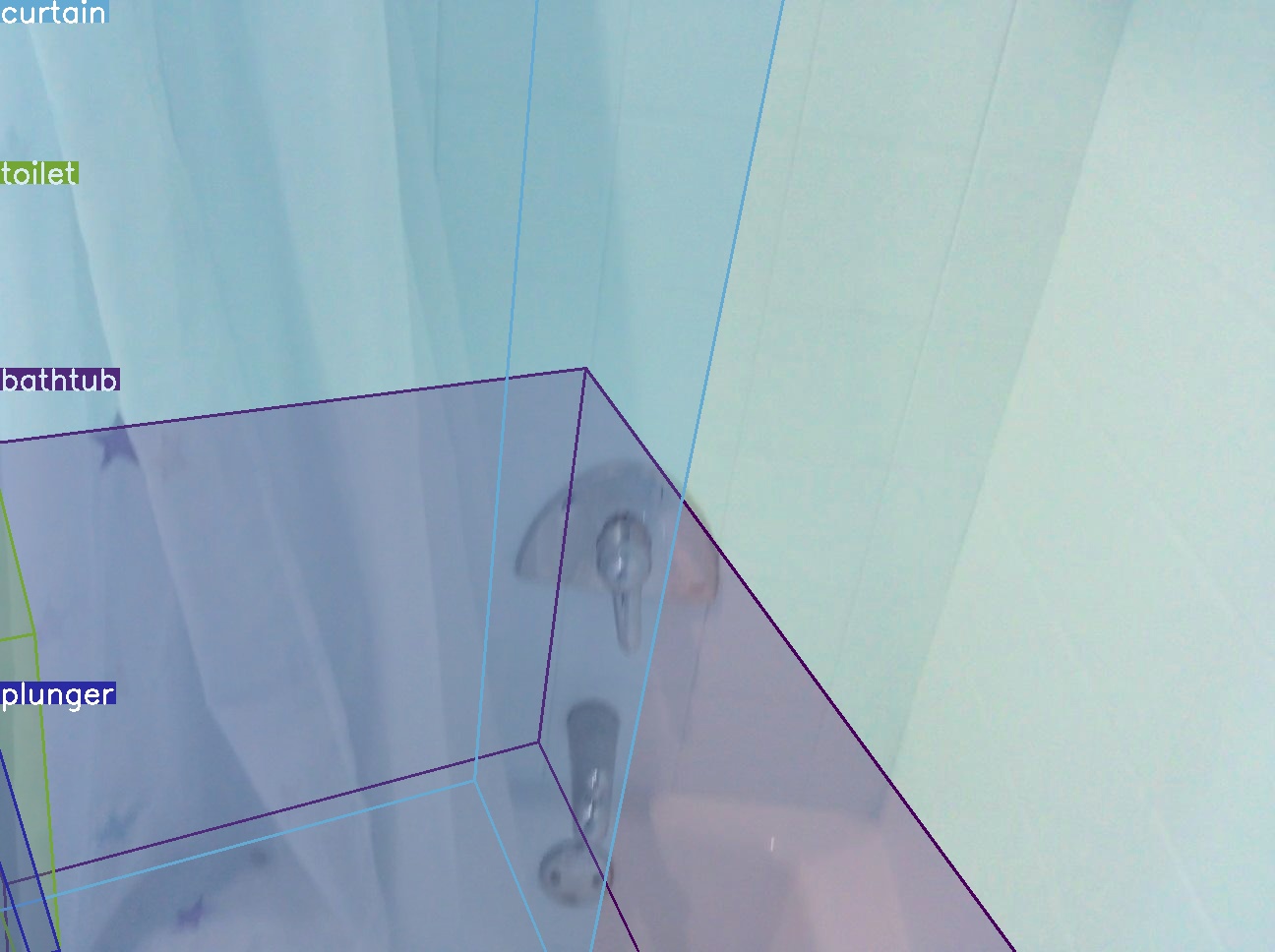} & 
        \includegraphics[width=0.15\linewidth]{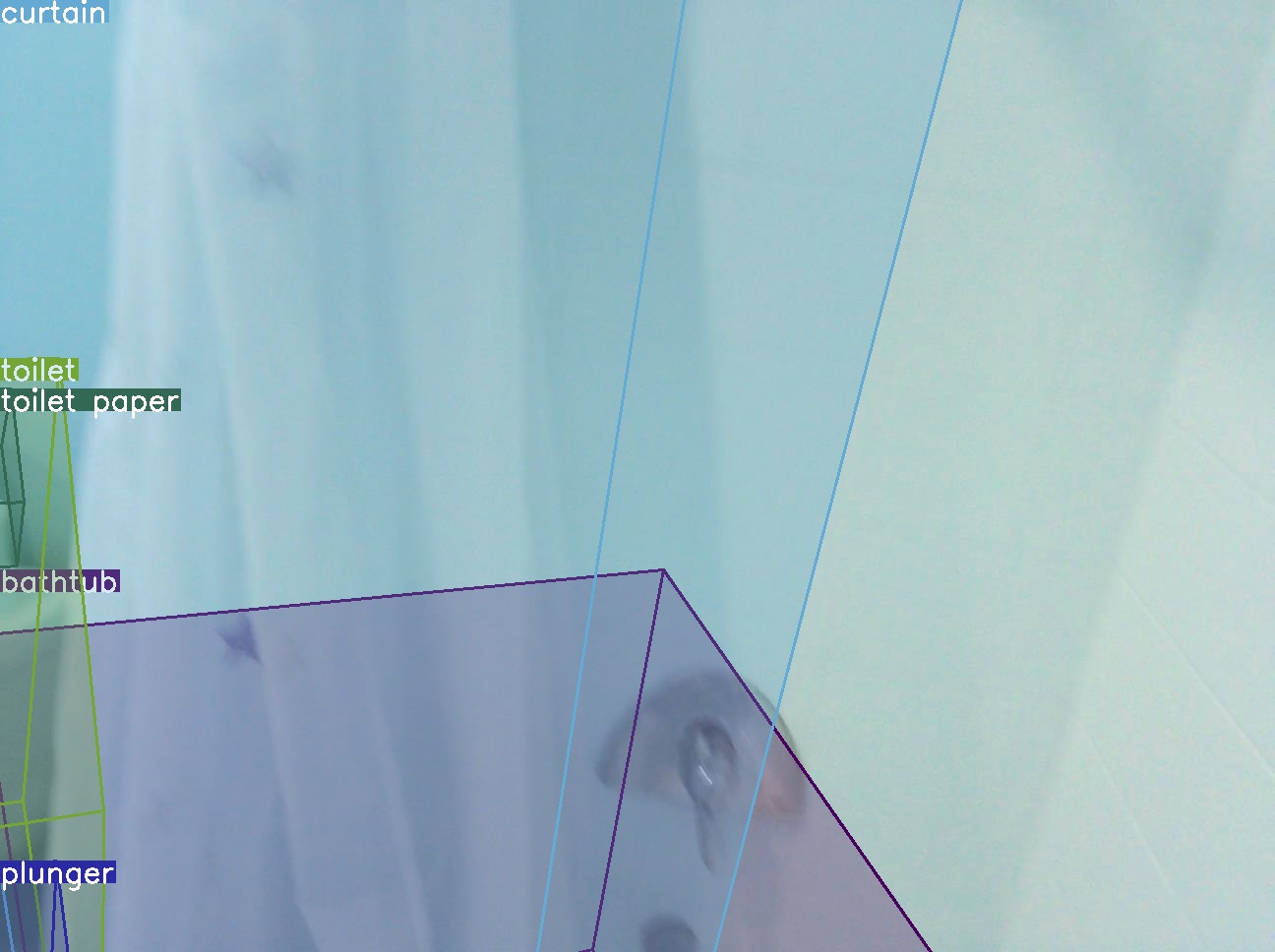} &
        \includegraphics[width=0.15\linewidth]{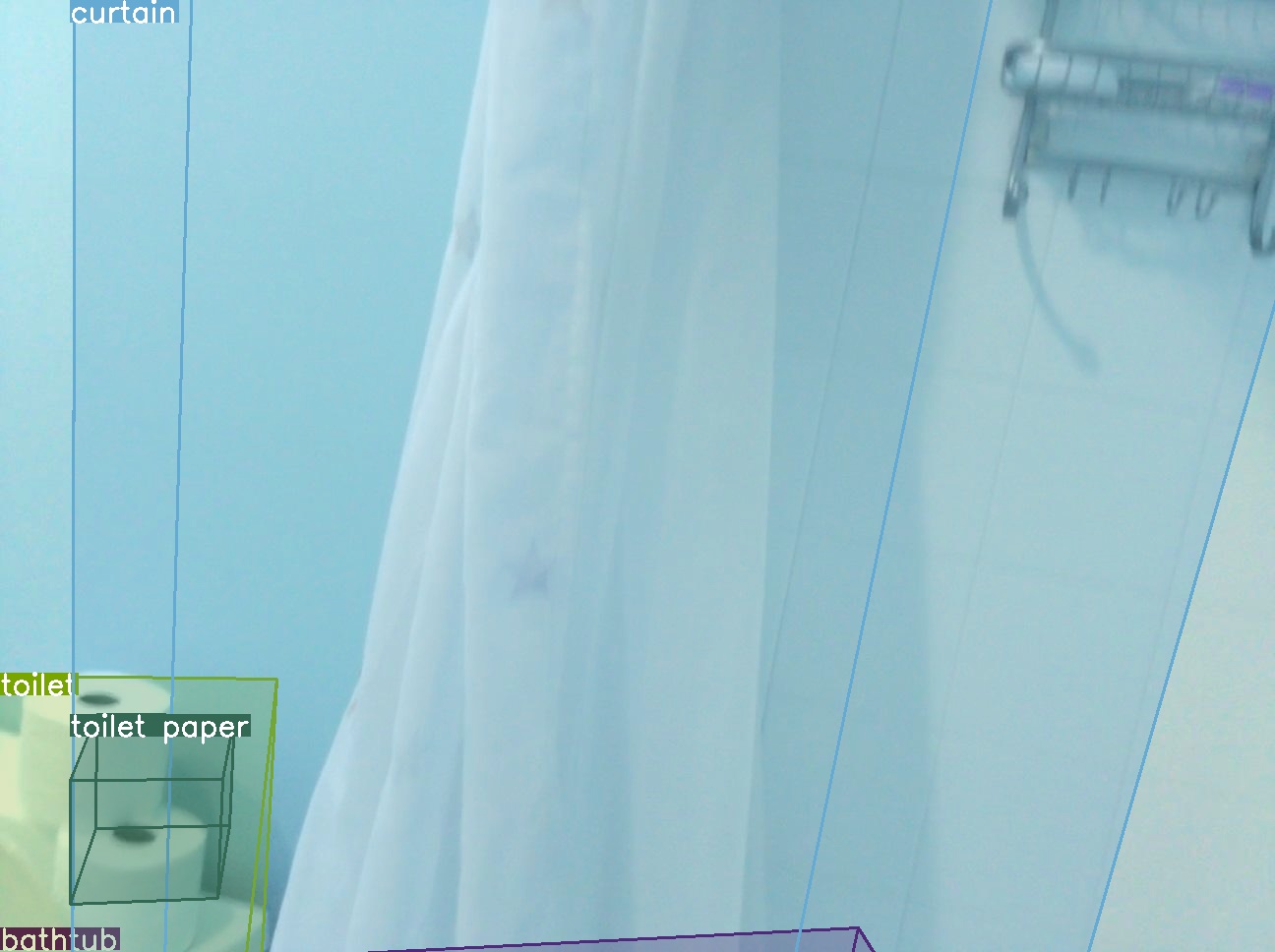} \\
        
    \end{tabular}
    
    \caption{\textbf{Qualitative comparison of 3D object detection (20 common classes) results.} We compare the predictions of the baseline \textbf{VG LLM-4B} against our \textbf{\model} model.}
    \label{fig:vis_3d_detection}
\end{figure*}

The qualitative results for \videt in \cref{fig:vis_3d_detection} corroborate our quantitative analysis, highlighting the enhanced precision of 3D-RFT. Consistent with the reduction in false positives reported in \cref{tab:scannet_det_results}, \model successfully avoids the hallucinations generated by the baseline. For instance, in the first row, VG LLM-4B misinterprets a curtain's shadow as a valid box, and in the second row, it hallucinates a table on the floor; our model correctly suppresses both errors. Furthermore, \model demonstrates superior semantic granularity and recall: in the last row, it precisely classifies ``toilet paper'' (improving upon the baseline's coarser label ``paper''), and in the third row, it successfully detects a ``desk'' that the baseline misses.

\subsubsection{3D Visual Grounding}

\begin{figure*}[t]
    \centering
    \setlength{\tabcolsep}{1pt}
    \renewcommand{\arraystretch}{0.5}

    \begin{minipage}{0.9\linewidth}
        \centering
        \textbf{Text:} \textit{``There are two black chairs situated between six brown chairs and a black couch. this black chair is next to the black couch. it appears to be leather. it is black. there is a snack machine on the opposite wall.''}\\
        \vspace{2pt}
        
        \begin{tabular}{ccccc}
            \includegraphics[width=0.19\linewidth]{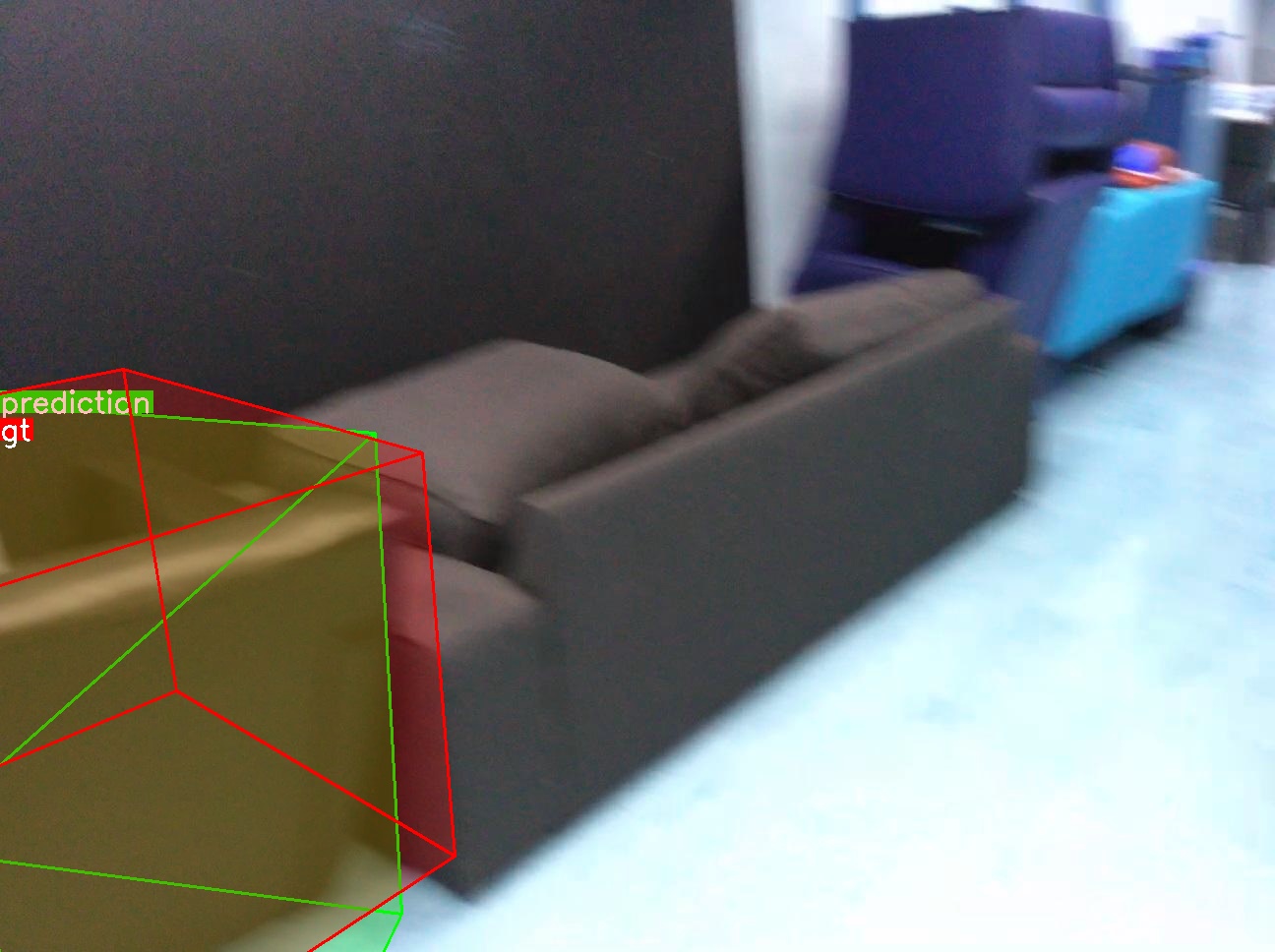} &
            \includegraphics[width=0.19\linewidth]{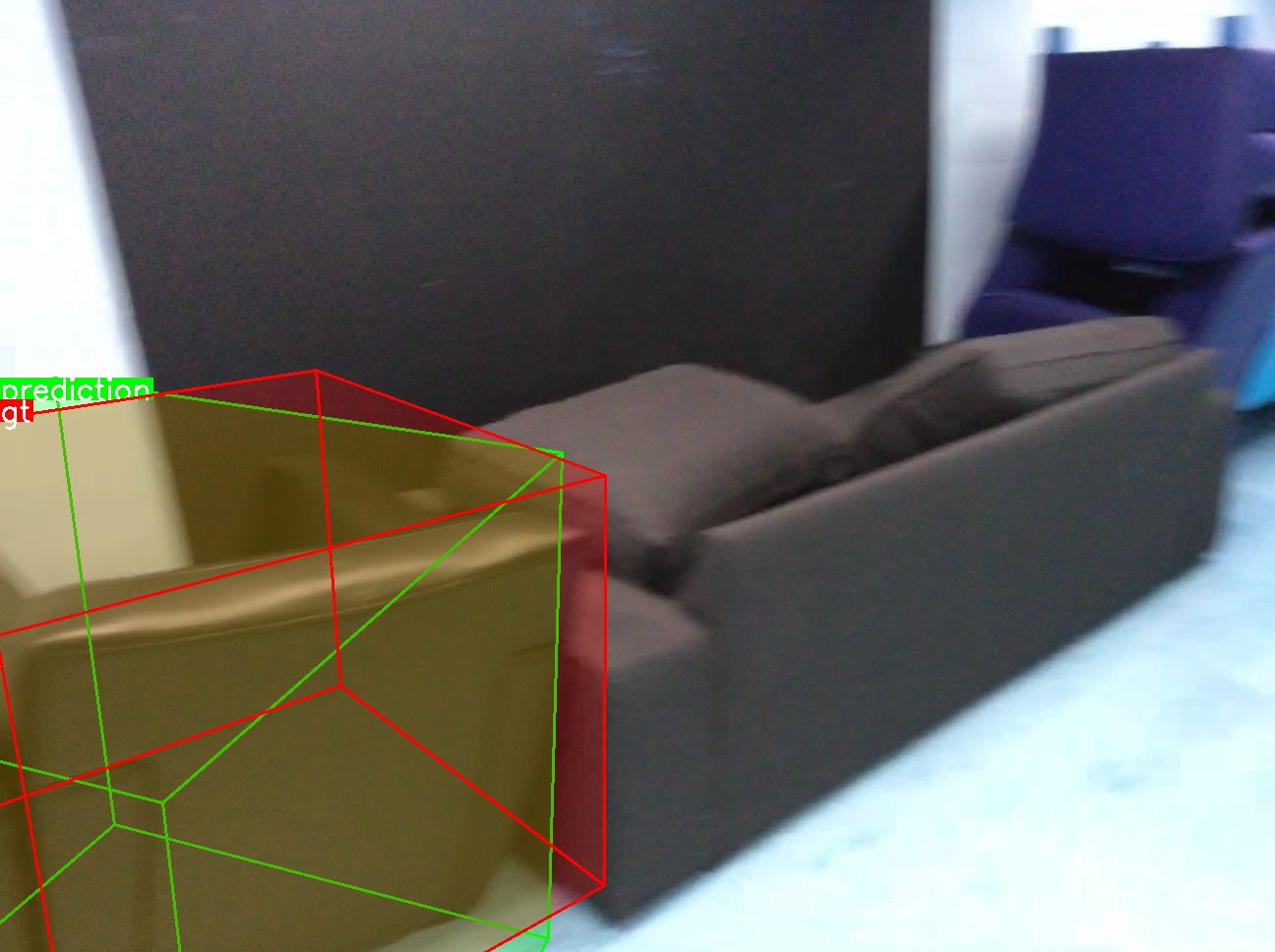} &
            \includegraphics[width=0.19\linewidth]{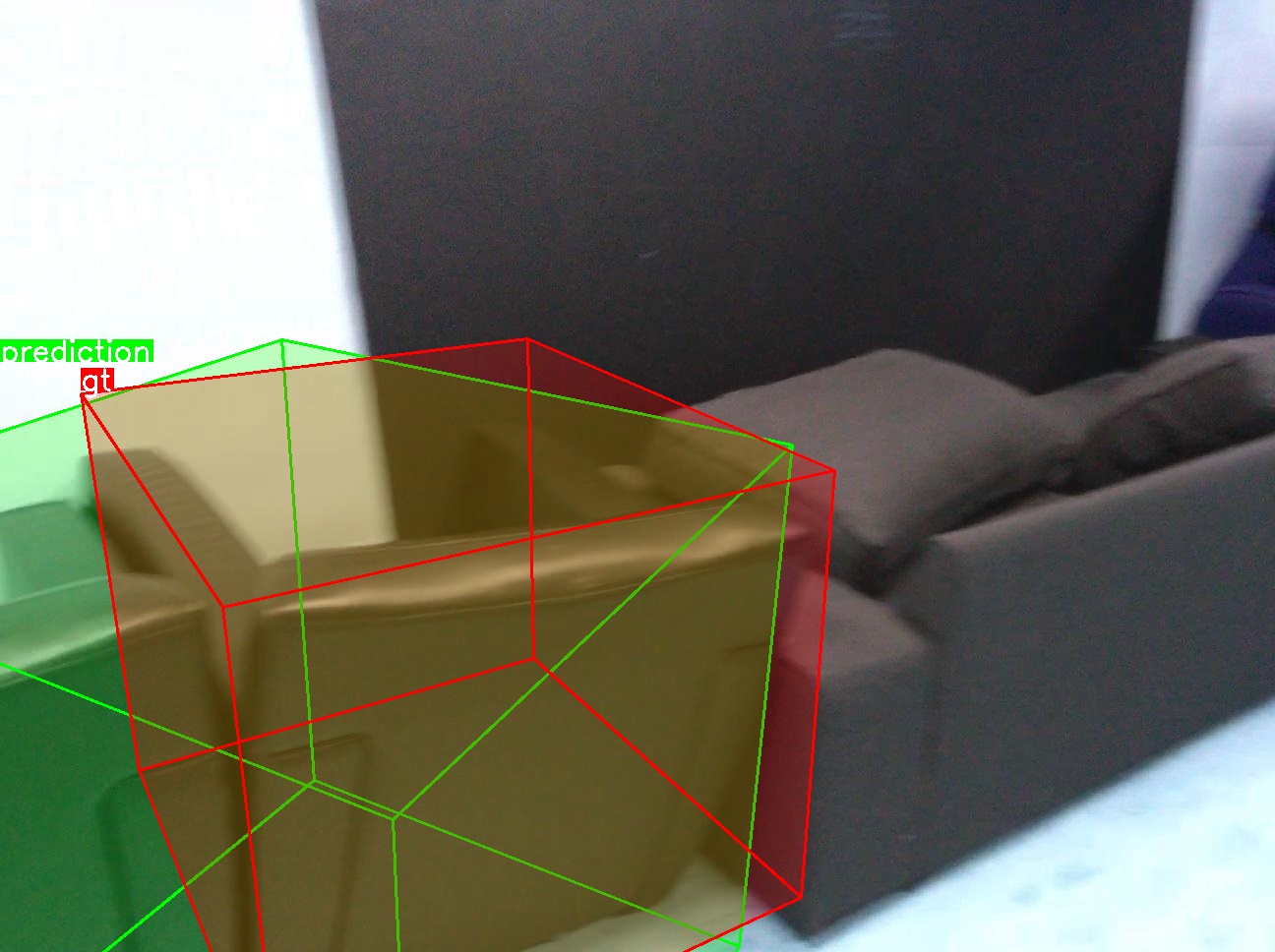} &
            \includegraphics[width=0.19\linewidth]{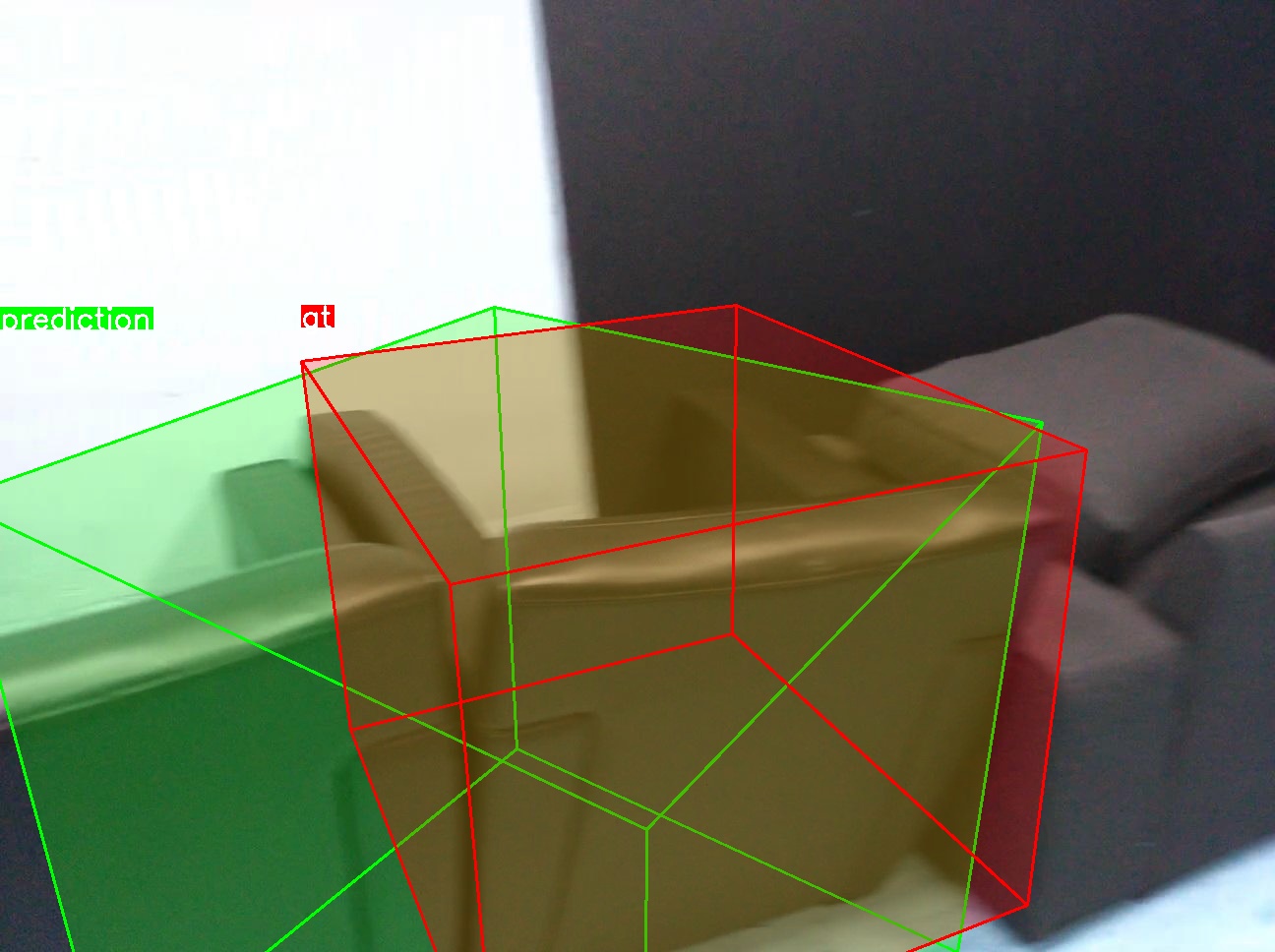} &
            \includegraphics[width=0.19\linewidth]{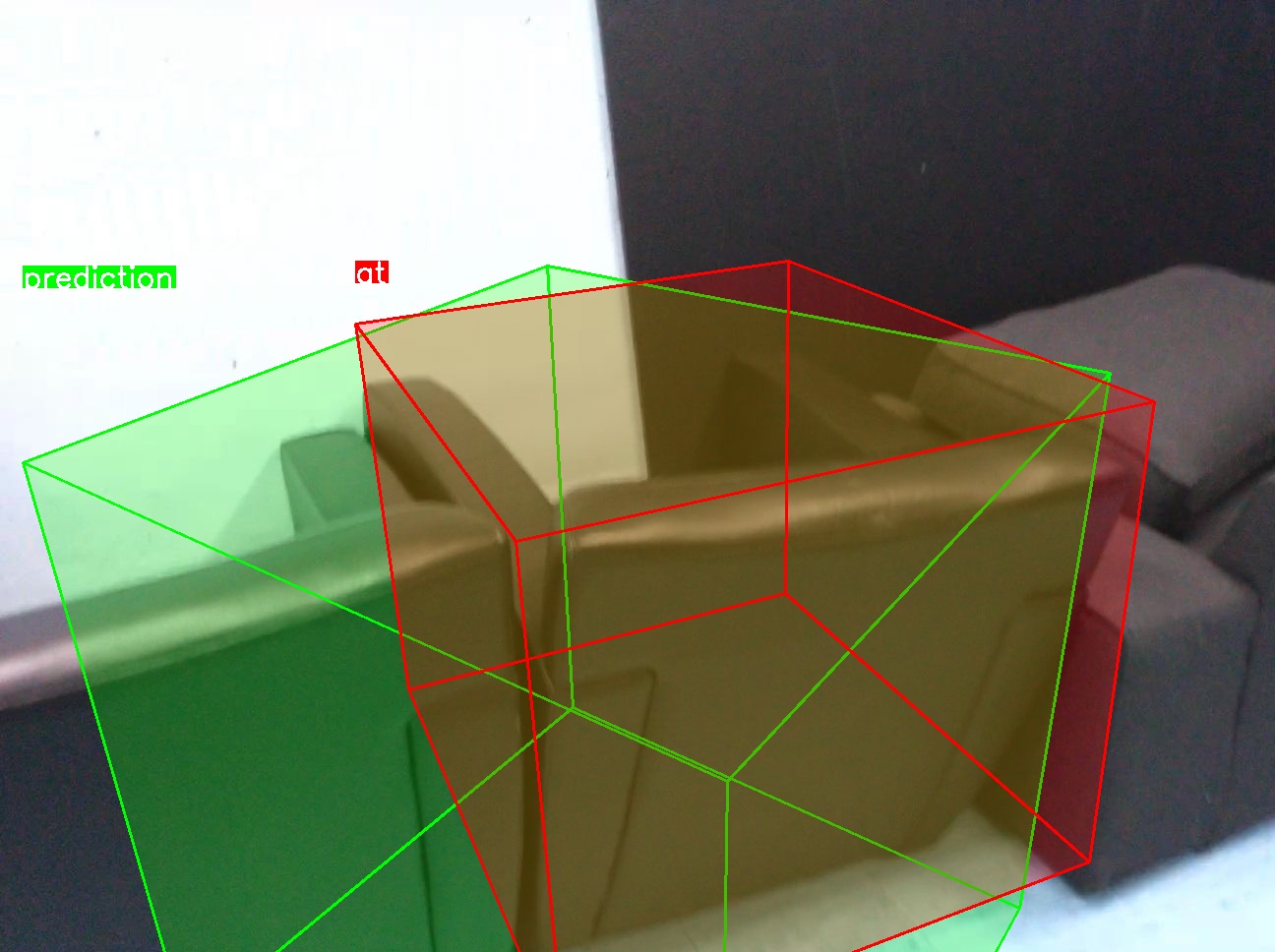} \\
            
            \includegraphics[width=0.19\linewidth]{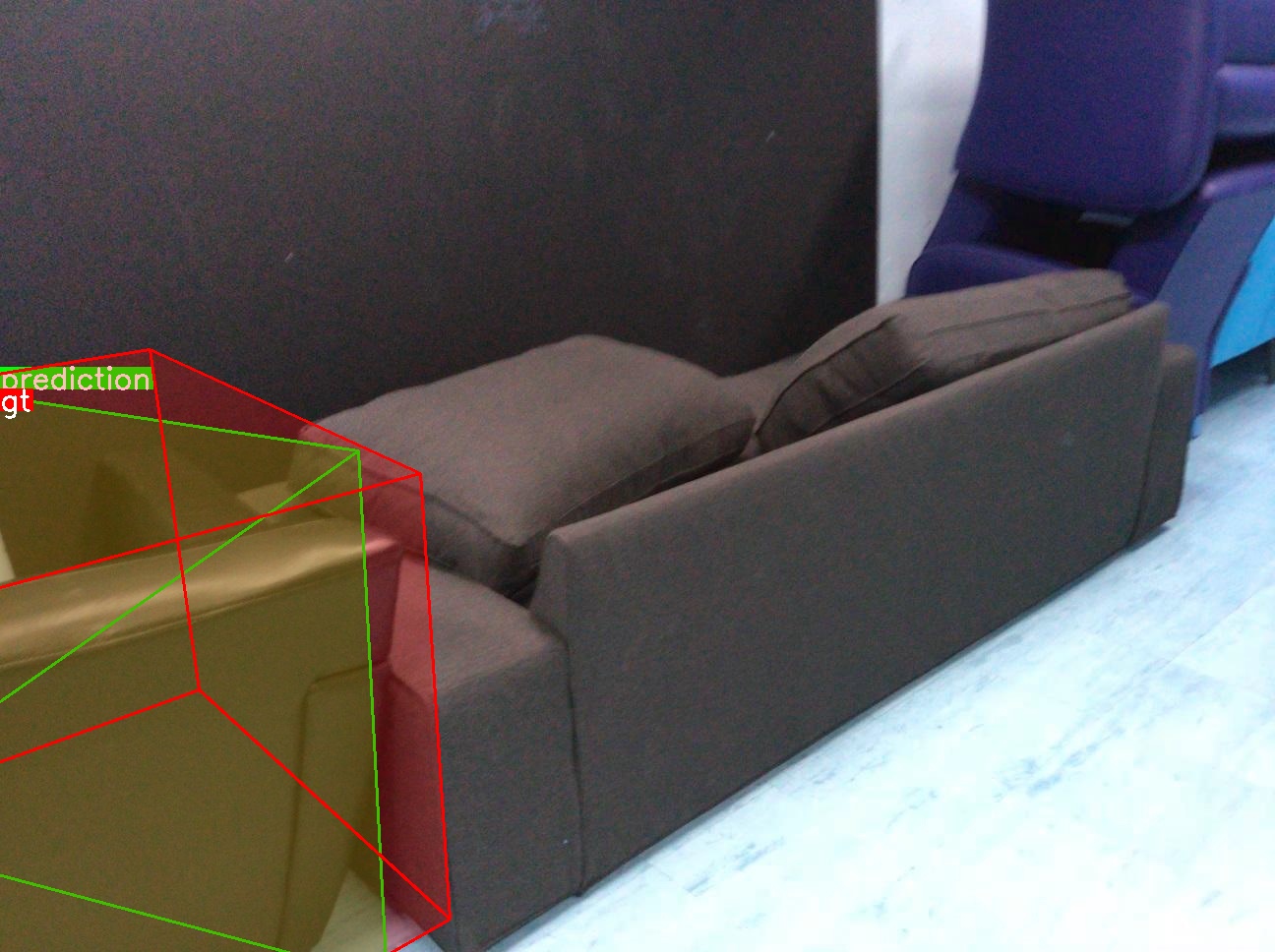} &
            \includegraphics[width=0.19\linewidth]{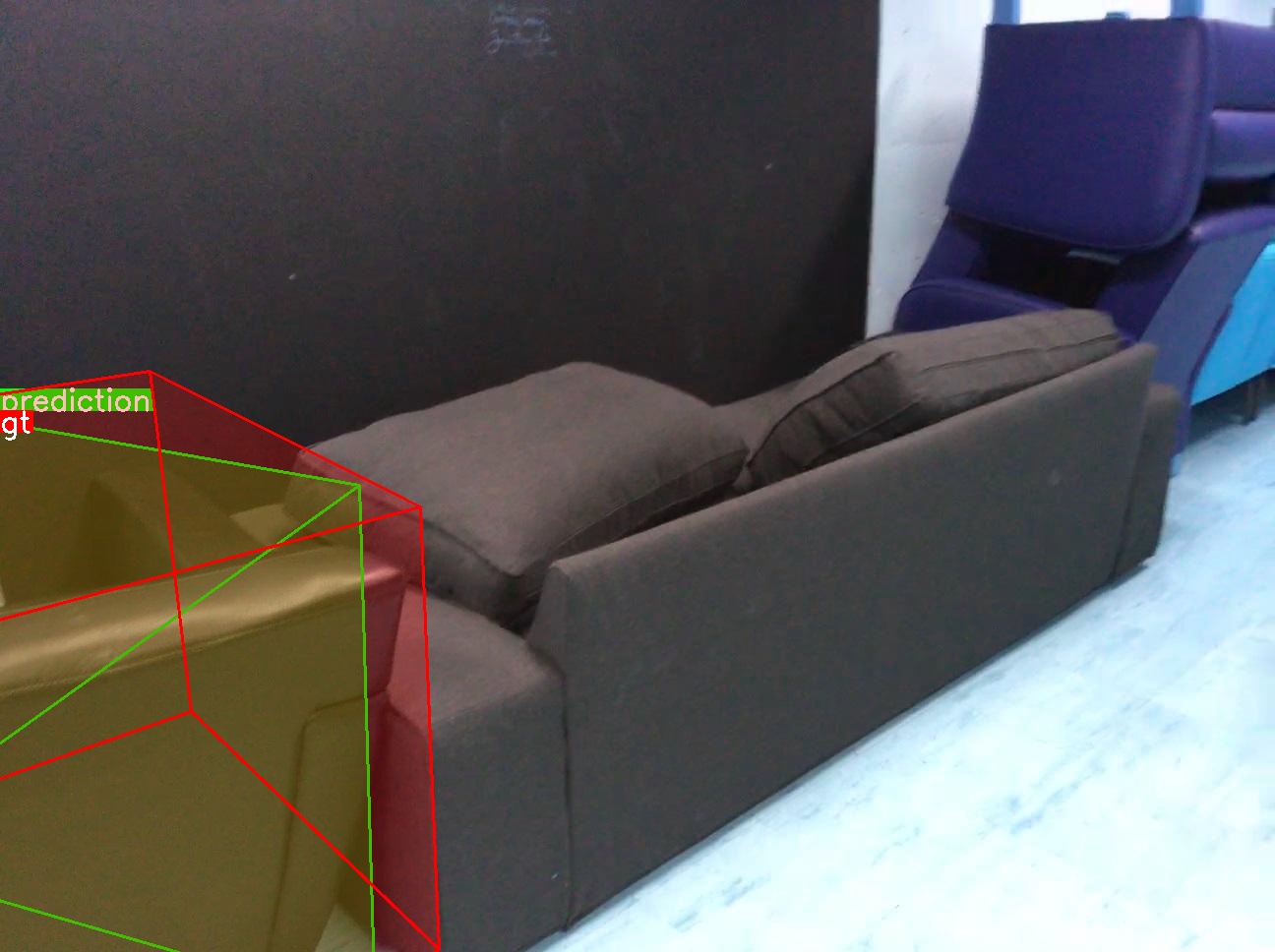} &
            \includegraphics[width=0.19\linewidth]{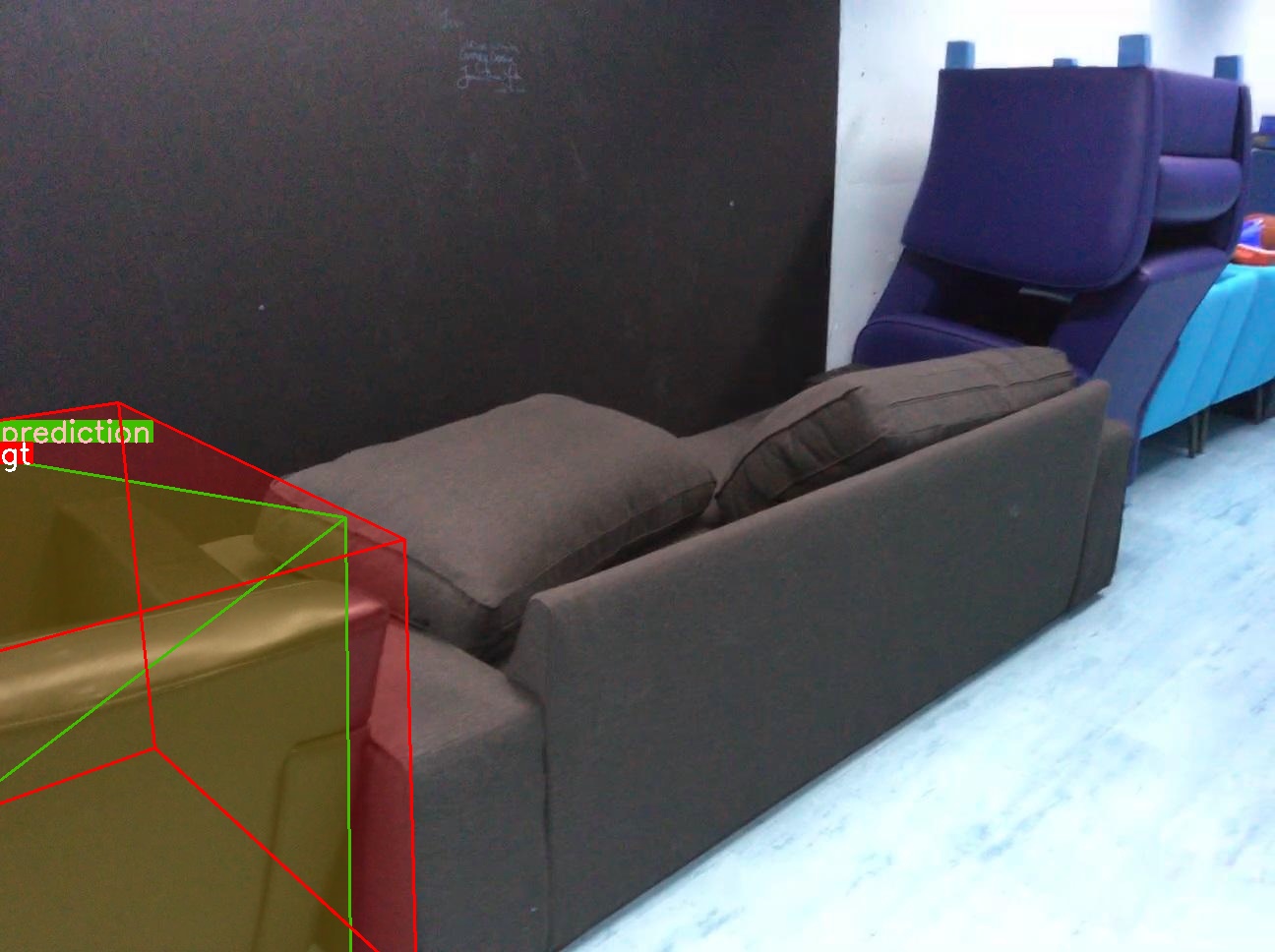} &
            \includegraphics[width=0.19\linewidth]{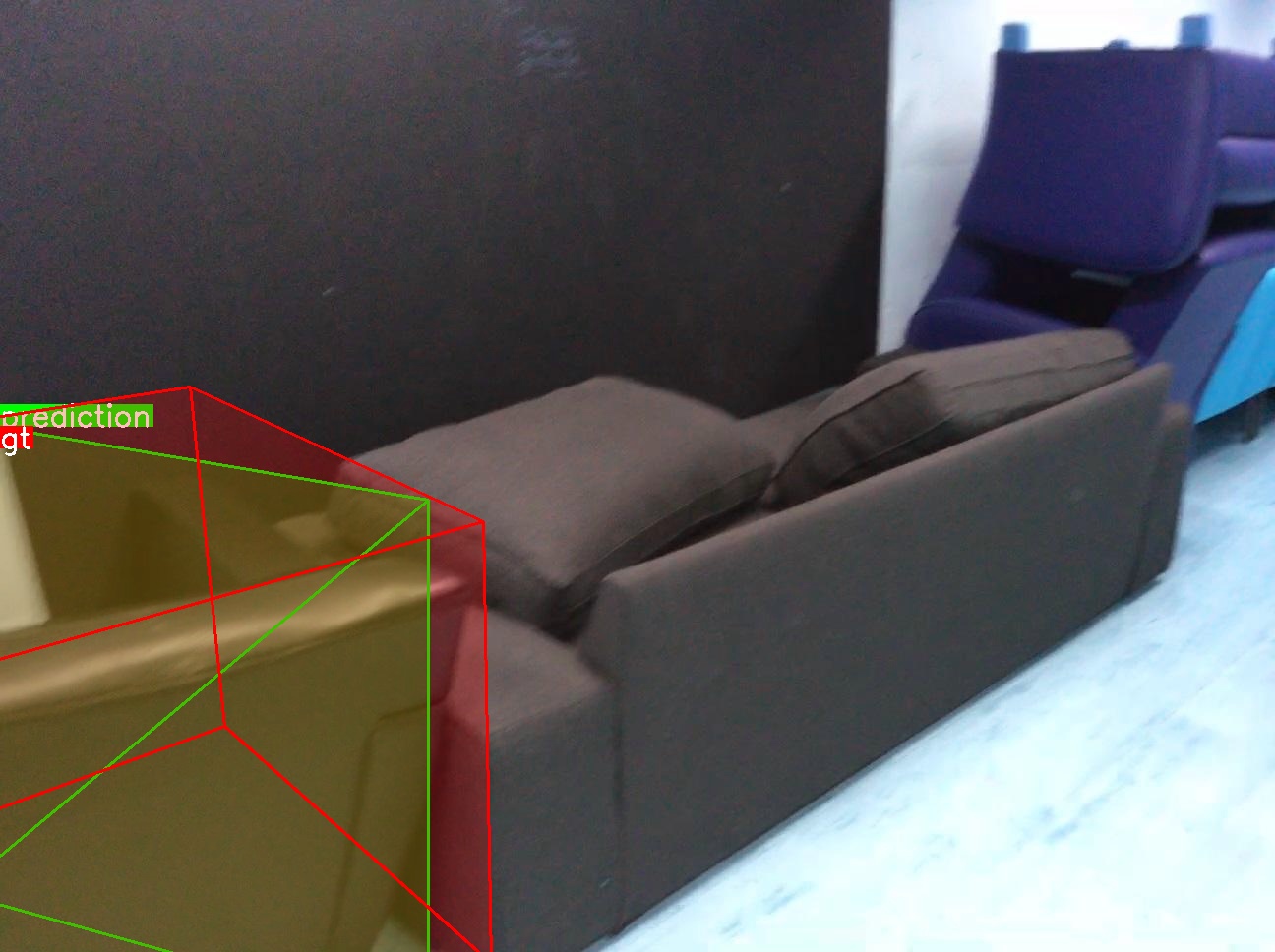} &
            \includegraphics[width=0.19\linewidth]{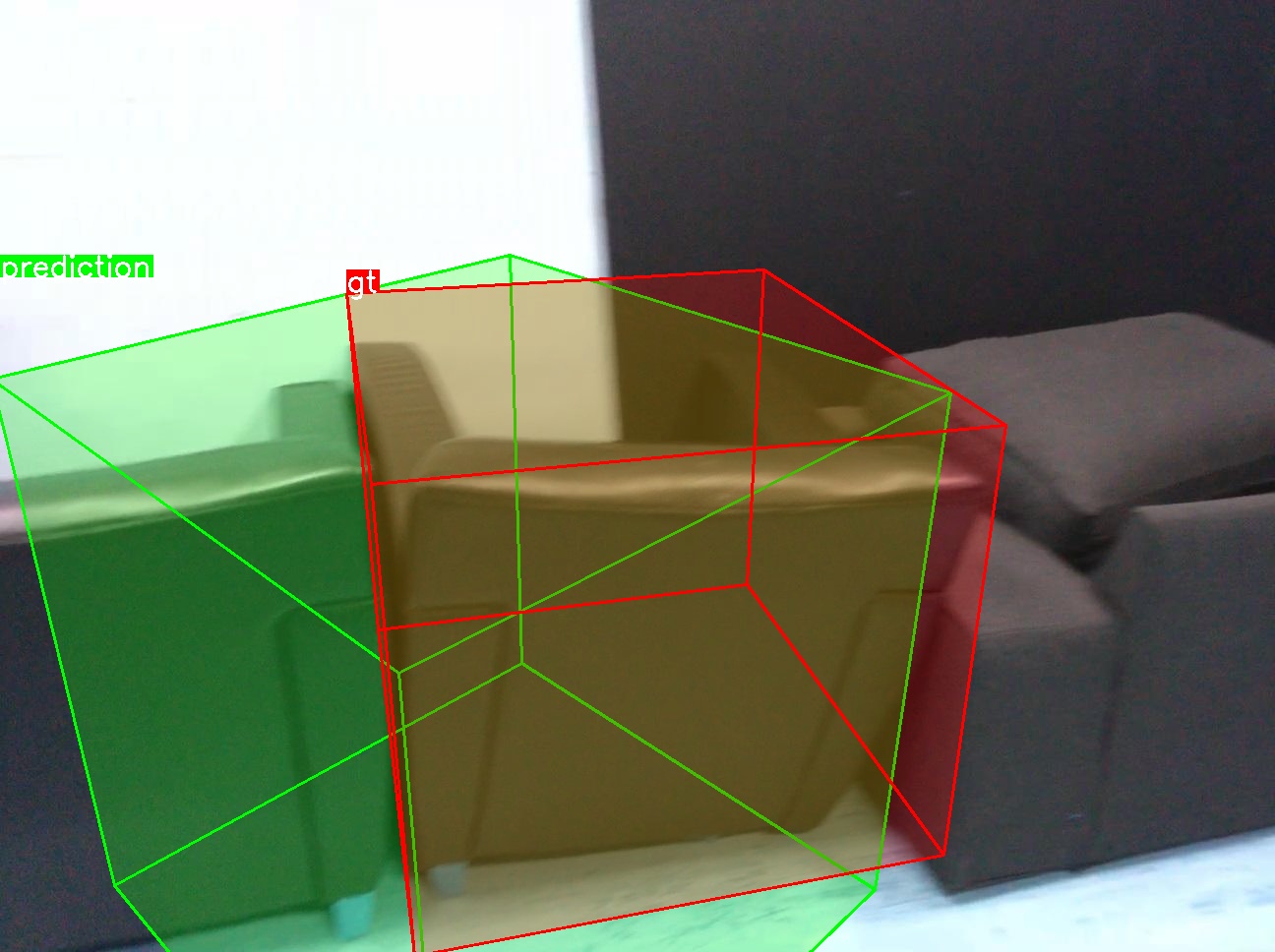} \\
        \end{tabular}
    \end{minipage}
    
    \vspace{1em} 

    \begin{minipage}{0.9\linewidth}
        \centering
        \textbf{Text:} \textit{``A black top located in center of desk. the lid is open and blue screen is visible.''}\\
        \vspace{2pt}
        
        \begin{tabular}{ccccc}
            \includegraphics[width=0.19\linewidth]{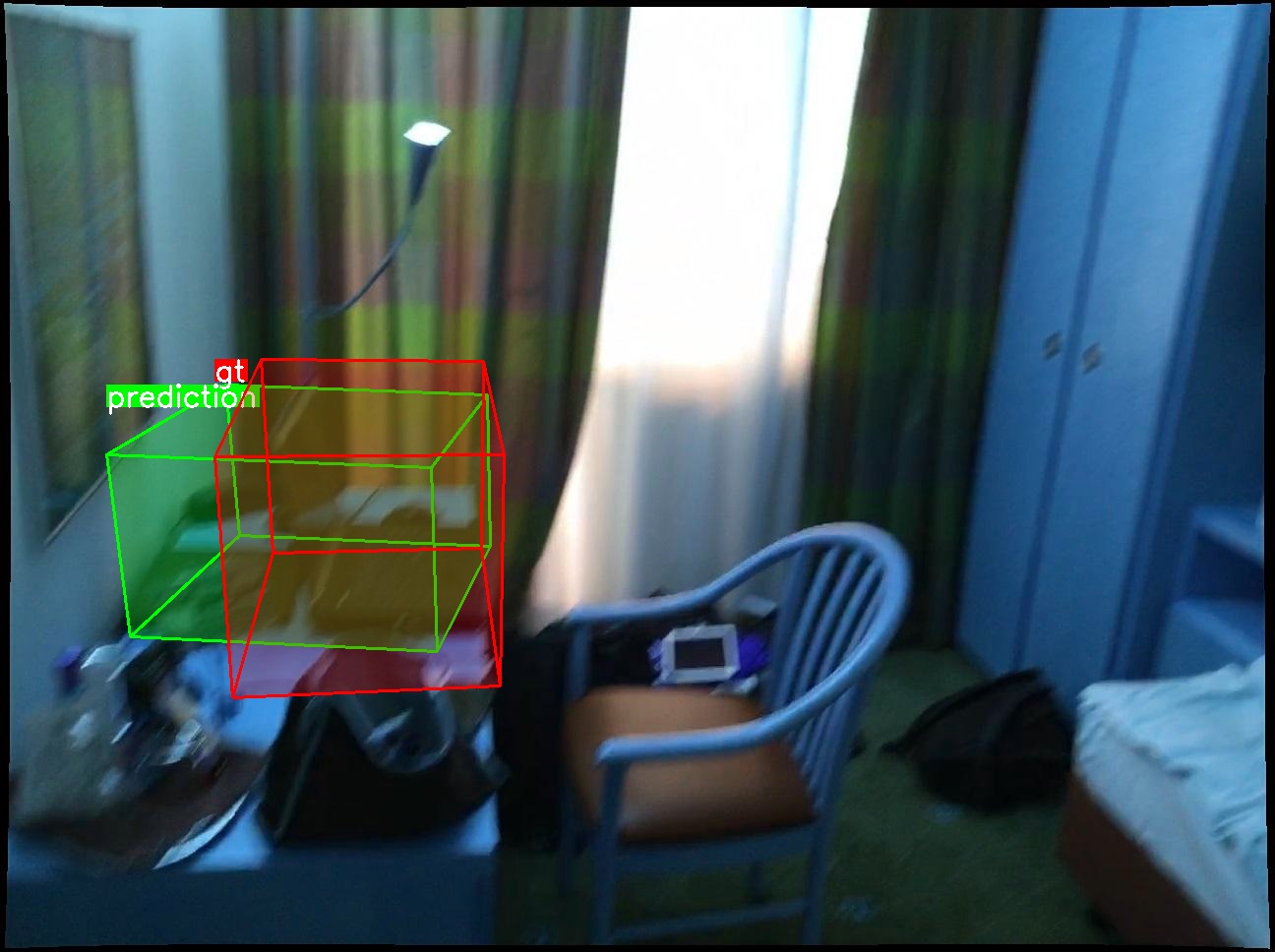} &
            \includegraphics[width=0.19\linewidth]{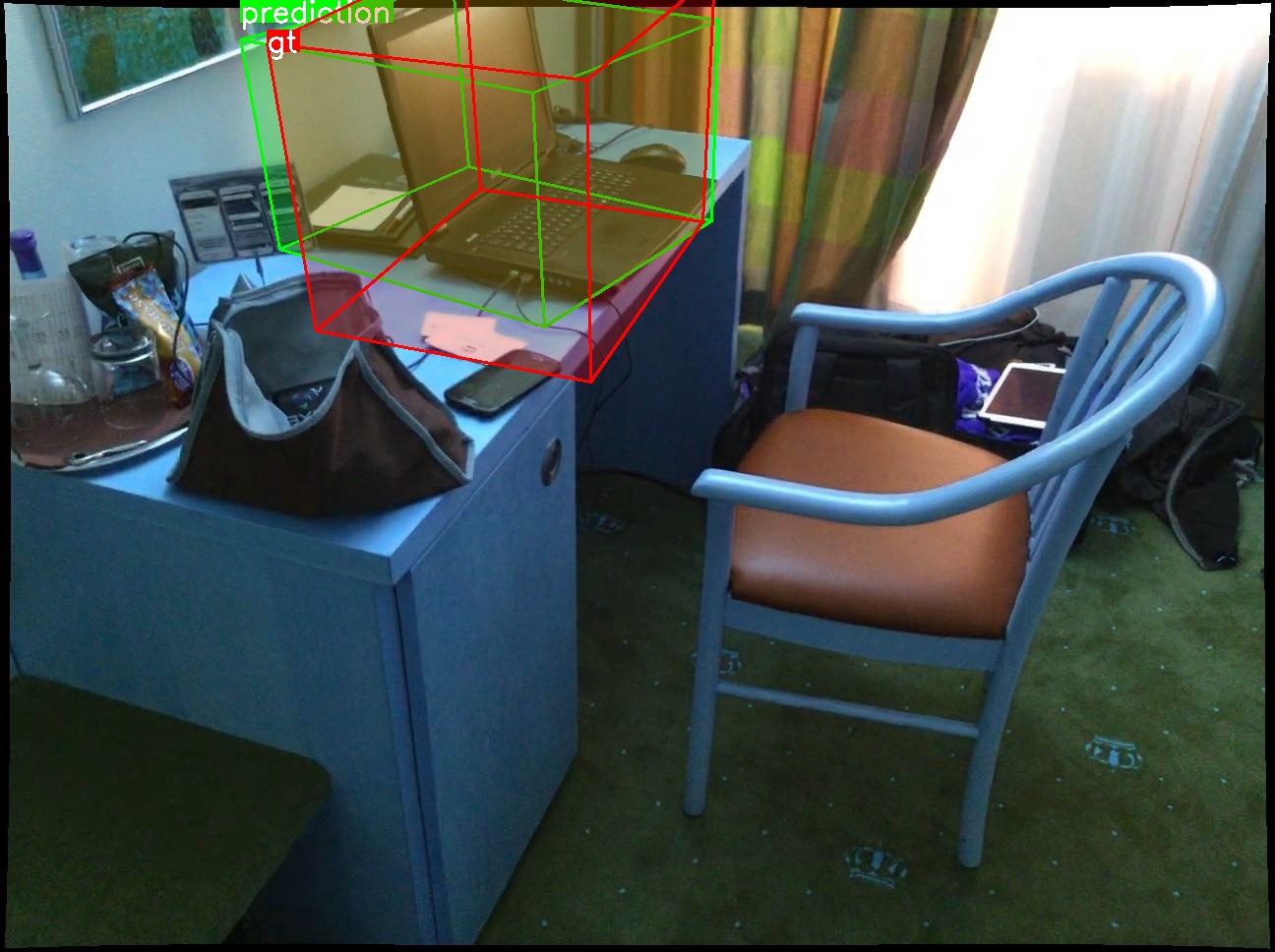} &
            \includegraphics[width=0.19\linewidth]{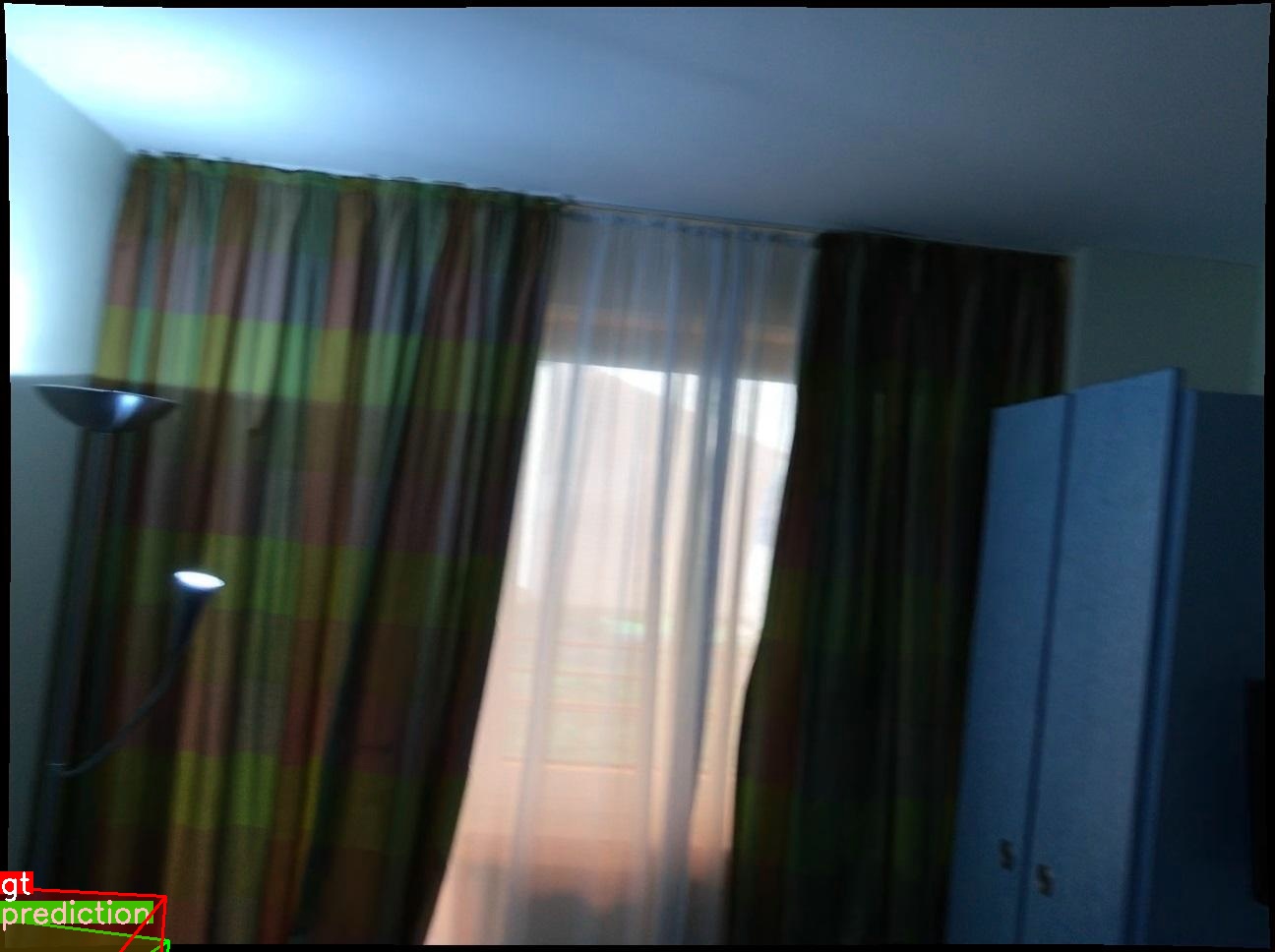} &
            \includegraphics[width=0.19\linewidth]{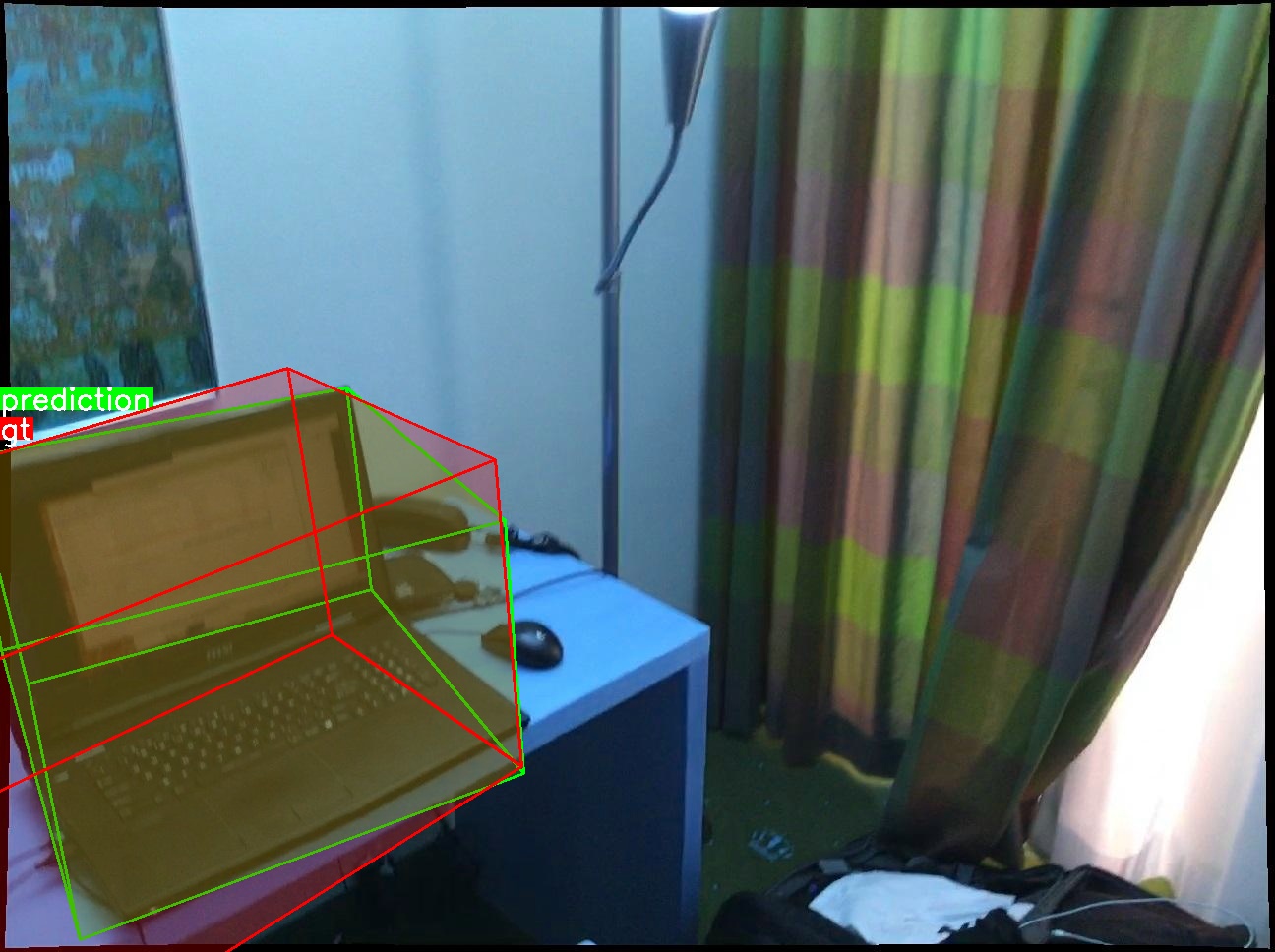} &
            \includegraphics[width=0.19\linewidth]{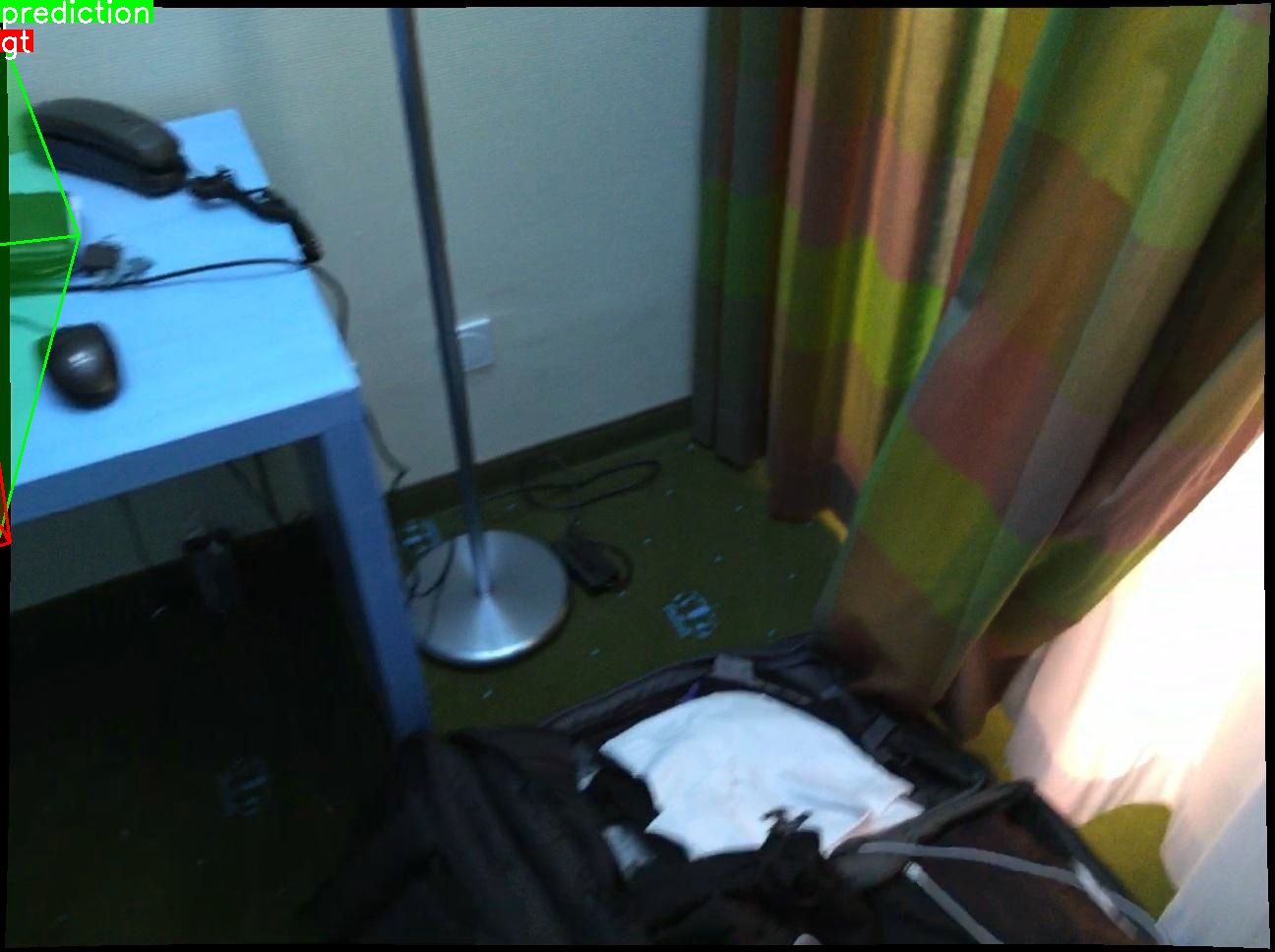} \\

            \includegraphics[width=0.19\linewidth]{Images/visualization_results/grounding/scene0046_00_idx6/frame_00016.jpg} &
            \includegraphics[width=0.19\linewidth]{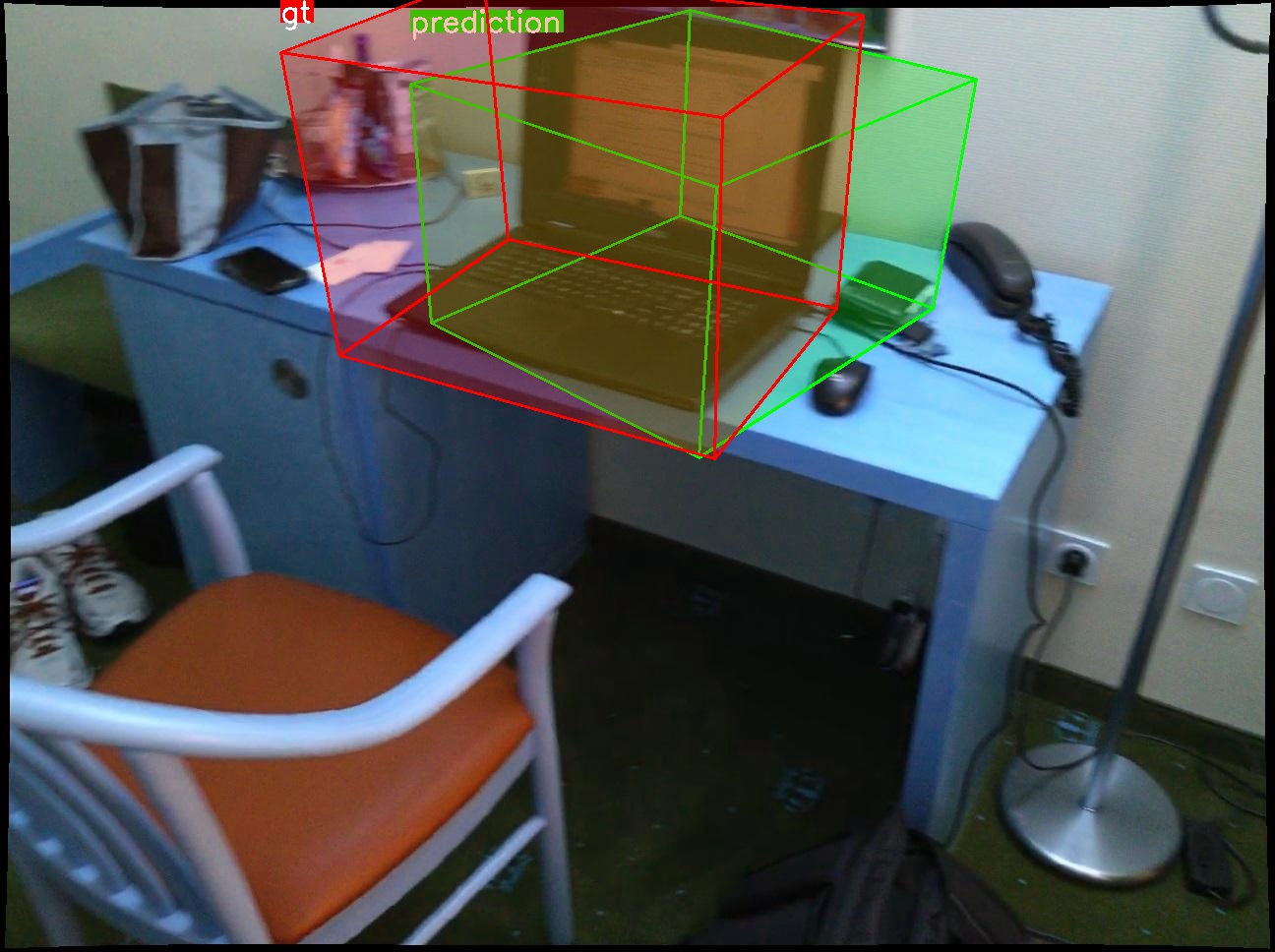} &
            \includegraphics[width=0.19\linewidth]{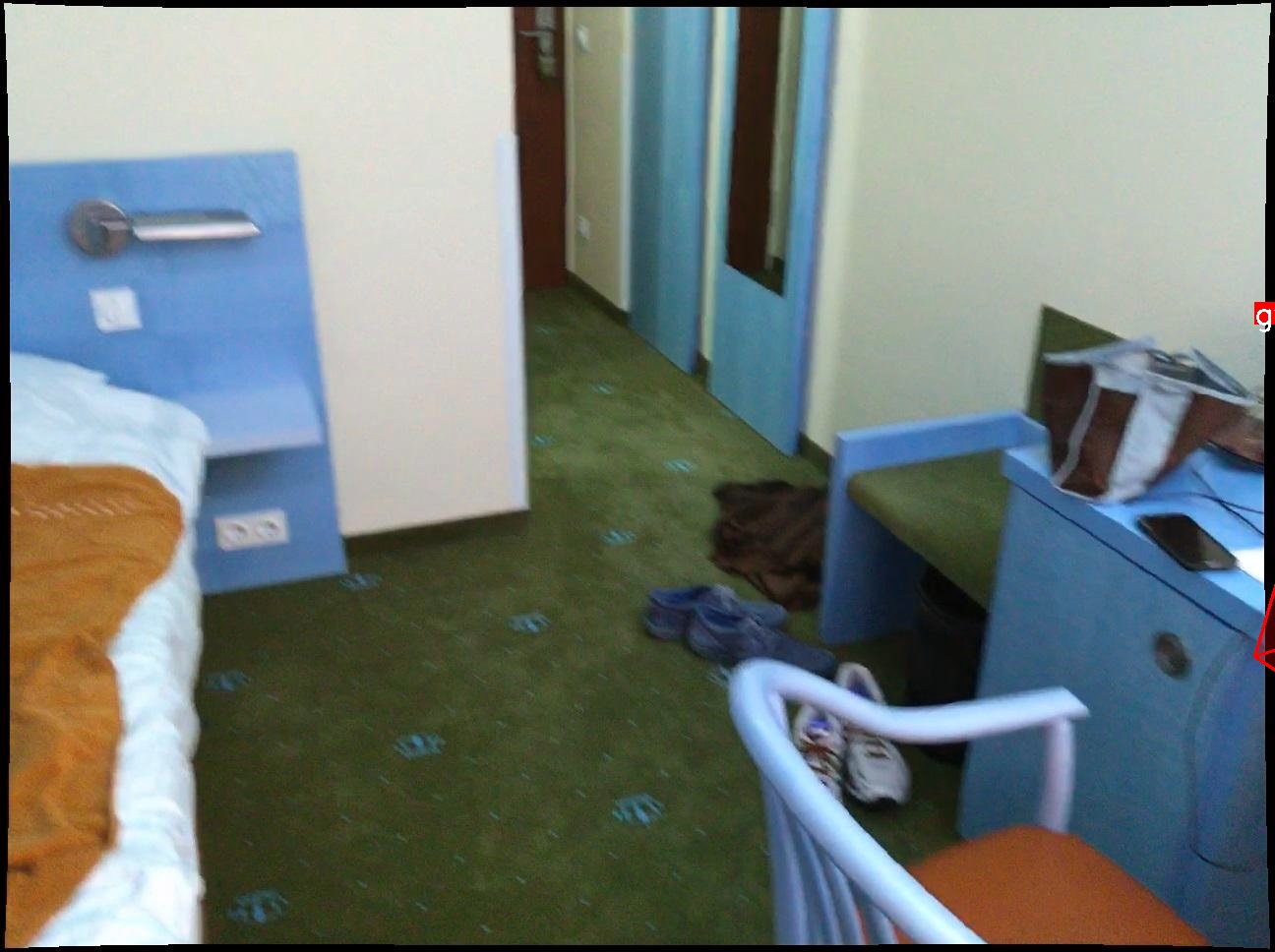} &
            \includegraphics[width=0.19\linewidth]{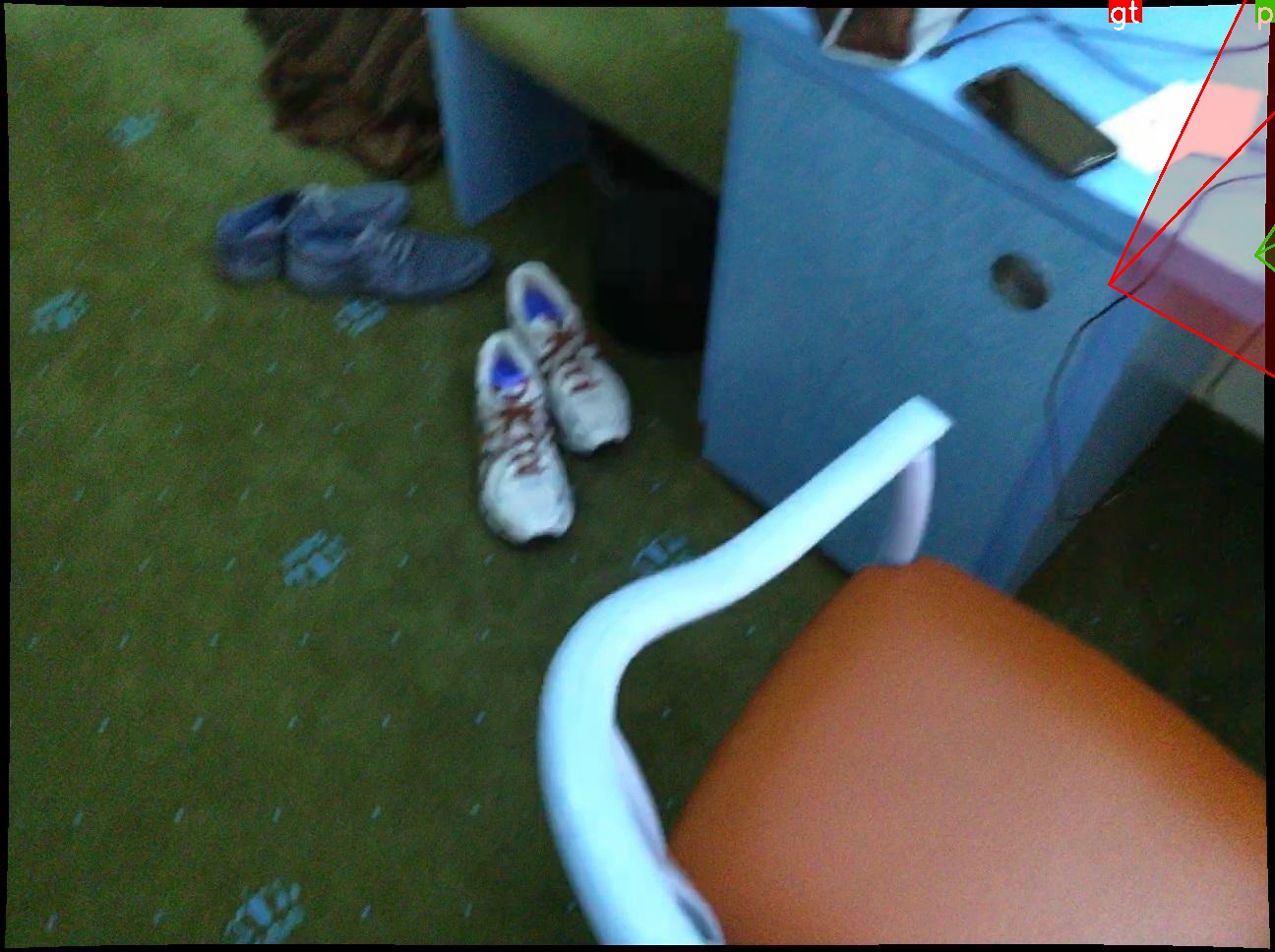} &
            \includegraphics[width=0.19\linewidth]{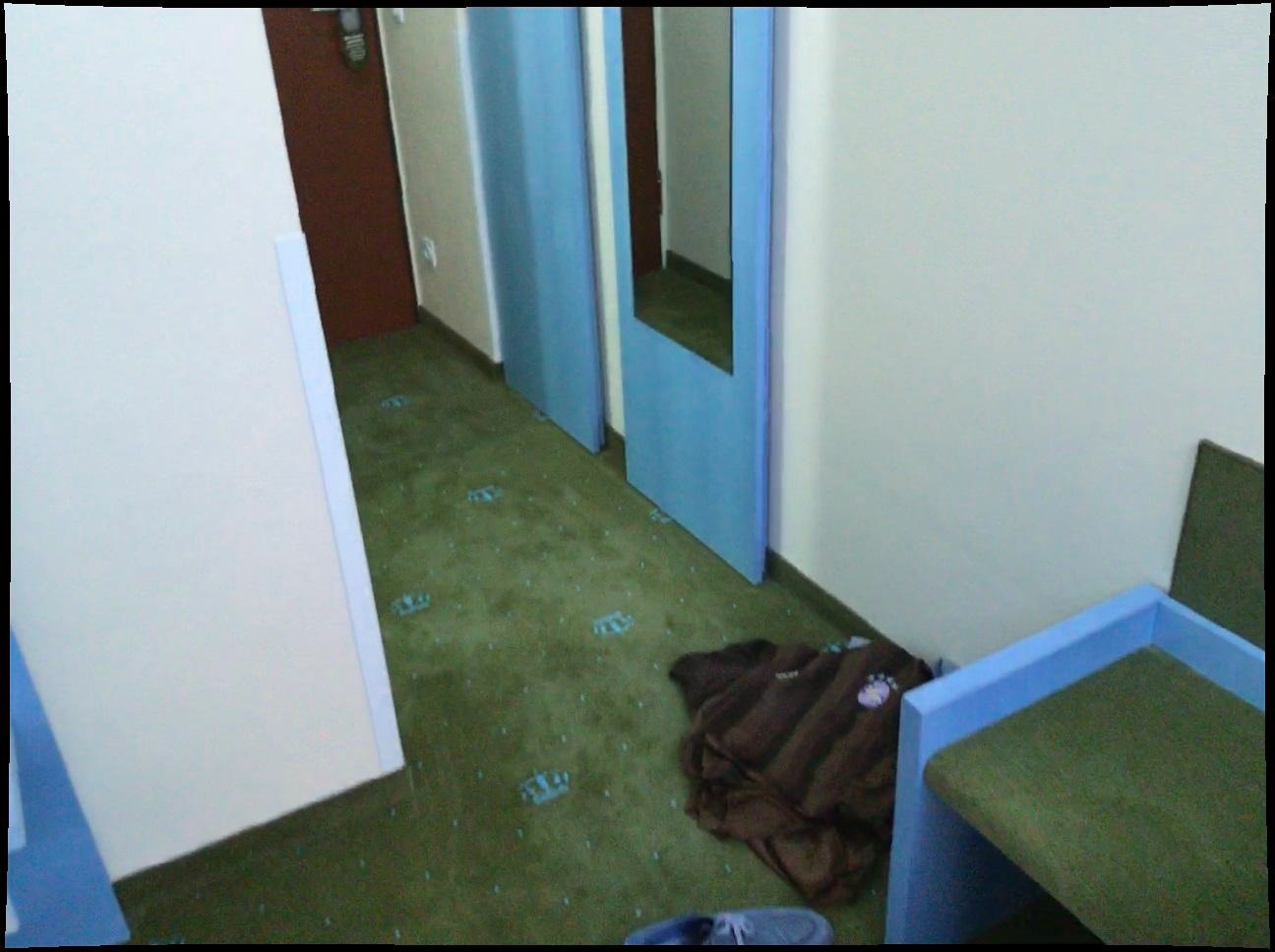} \\
        \end{tabular}
    \end{minipage}
    
    \vspace{1em} 

    \begin{minipage}{0.9\linewidth}
        \centering
        \textbf{Text:} \textit{``The brown desk is right next to the window by the chair. the desk is also right in front of the door.''}\\
        \vspace{2pt}
        
        \begin{tabular}{ccccc}
            \includegraphics[width=0.19\linewidth]{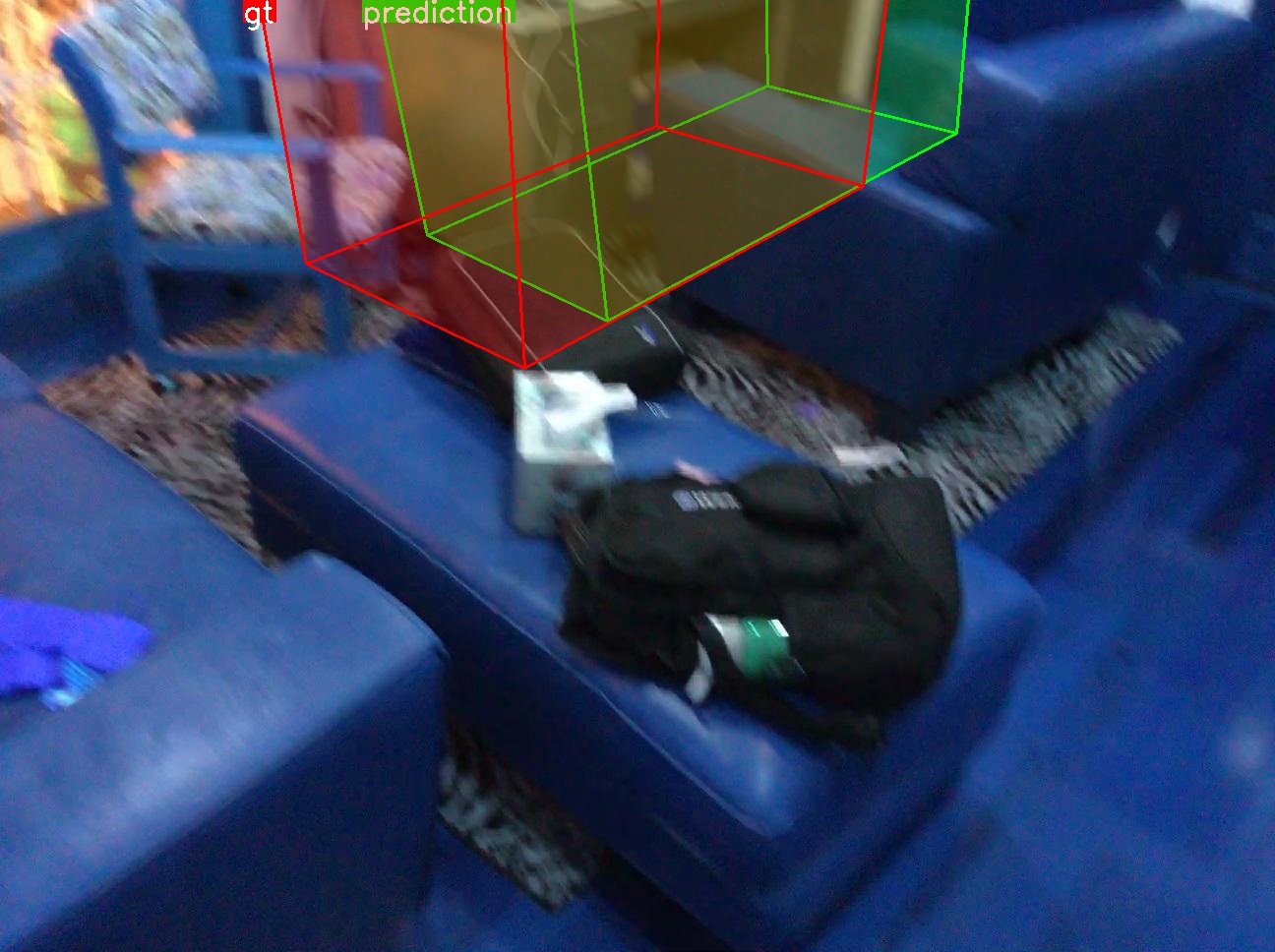} &
            \includegraphics[width=0.19\linewidth]{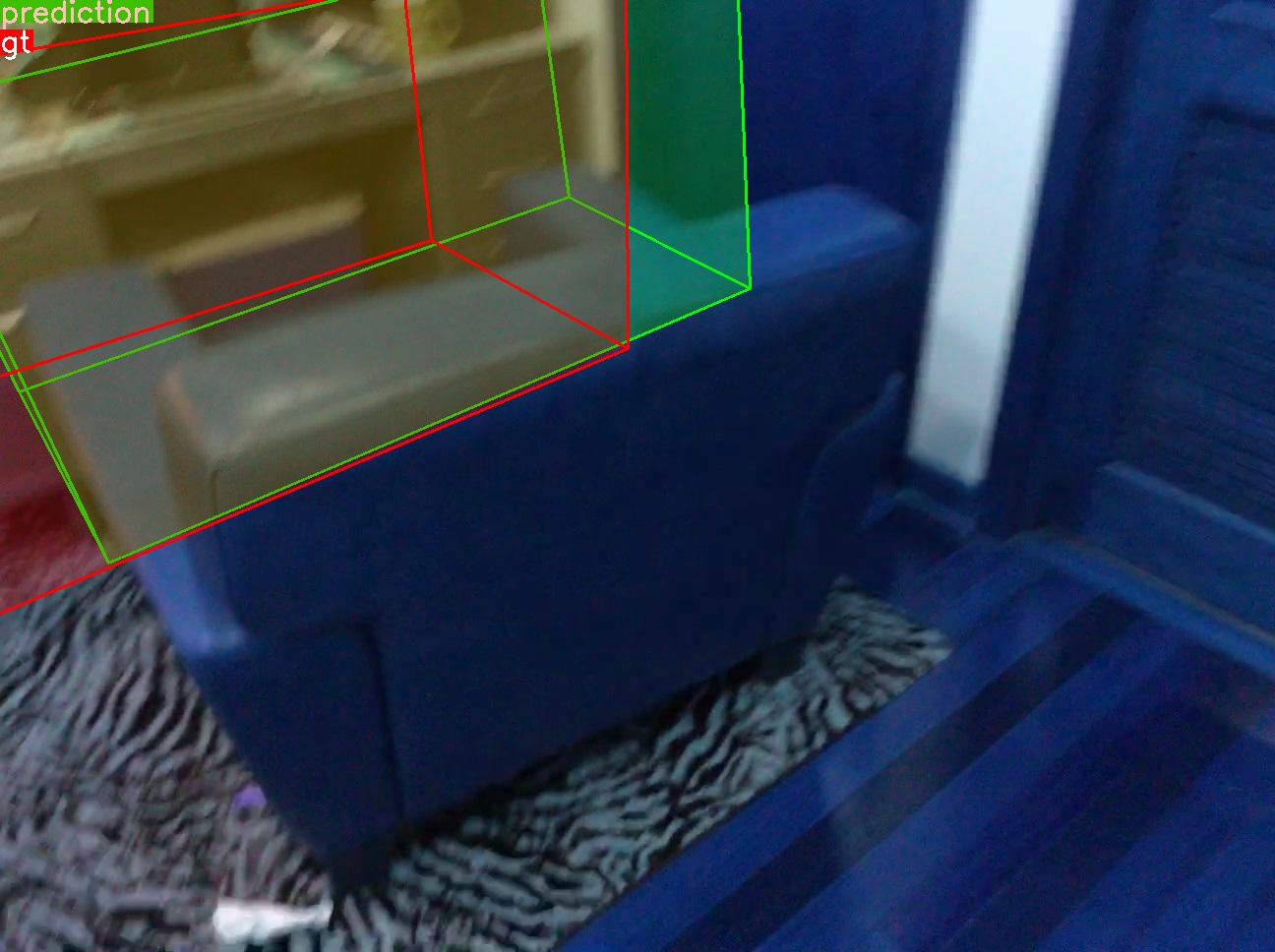} &
            \includegraphics[width=0.19\linewidth]{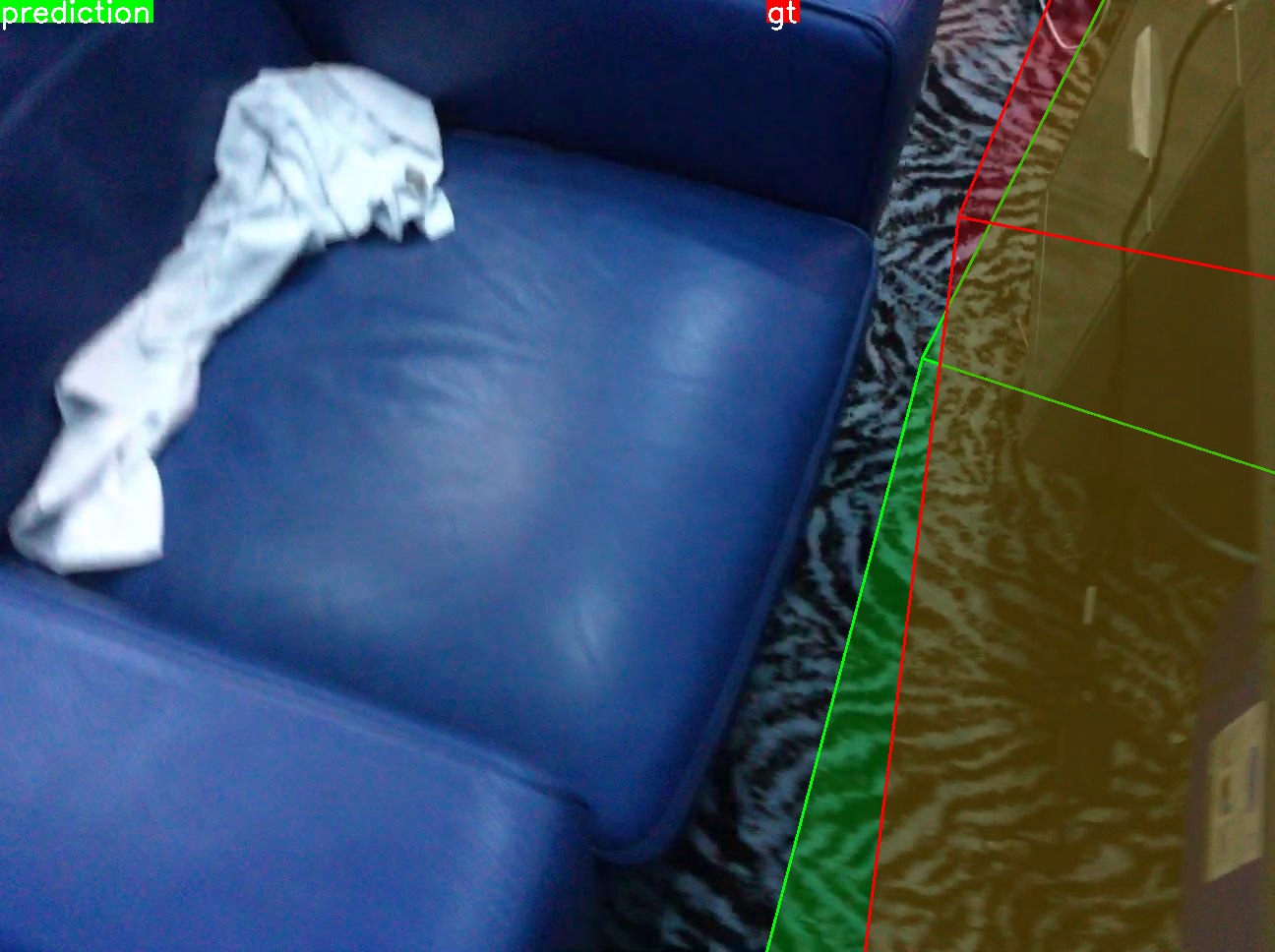} &
            \includegraphics[width=0.19\linewidth]{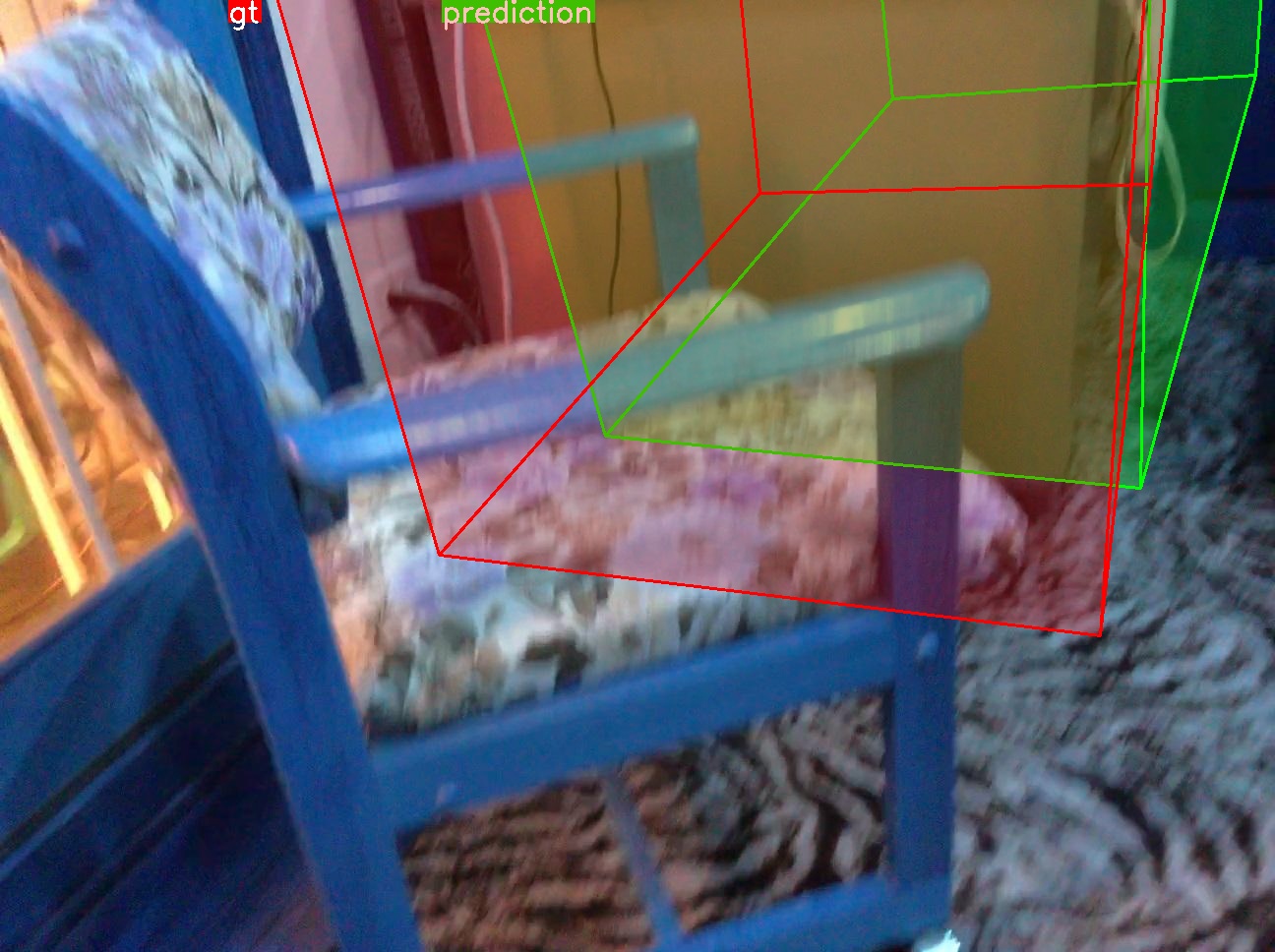} &
            \includegraphics[width=0.19\linewidth]{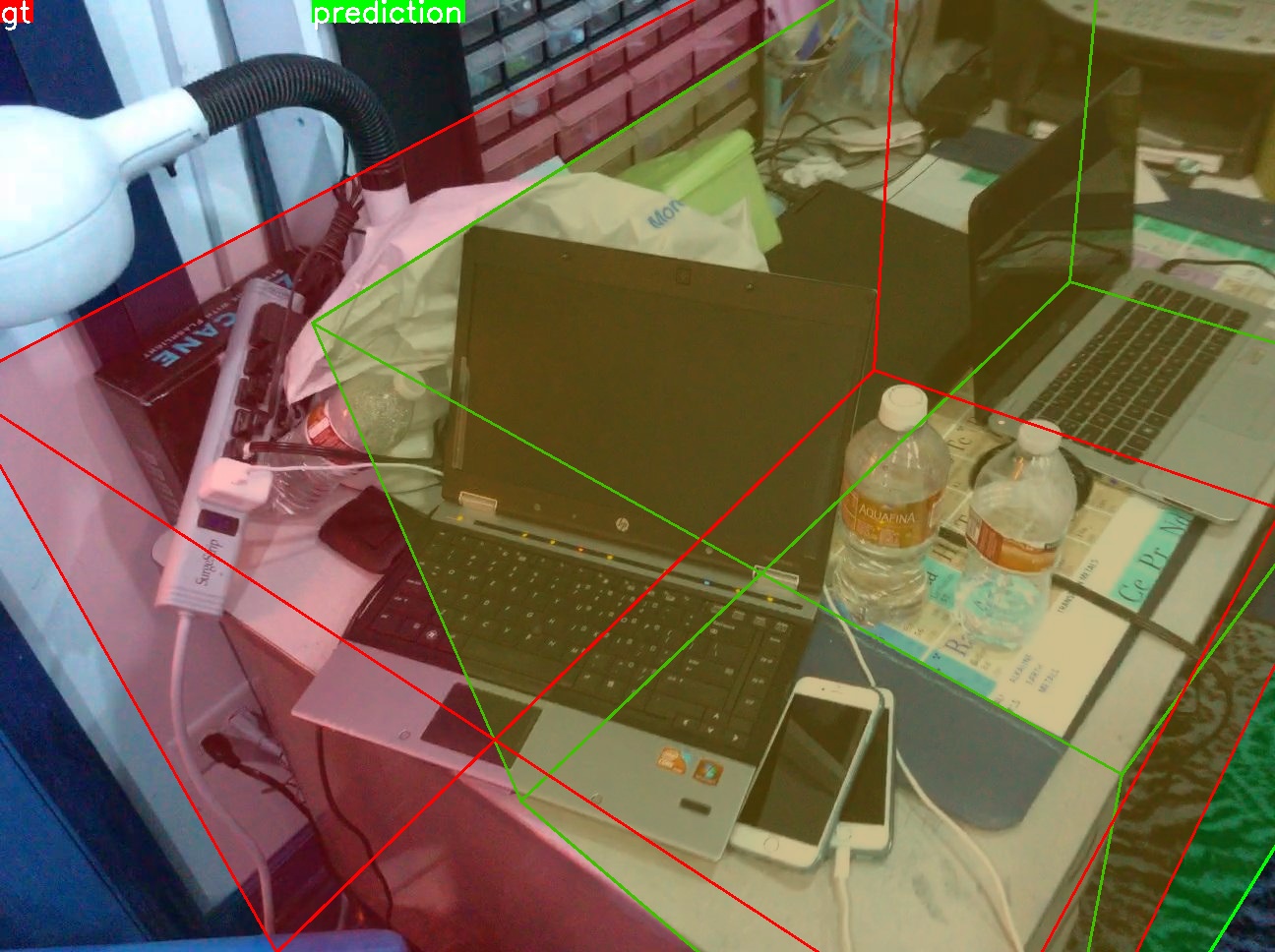} \\
            
            \includegraphics[width=0.19\linewidth]{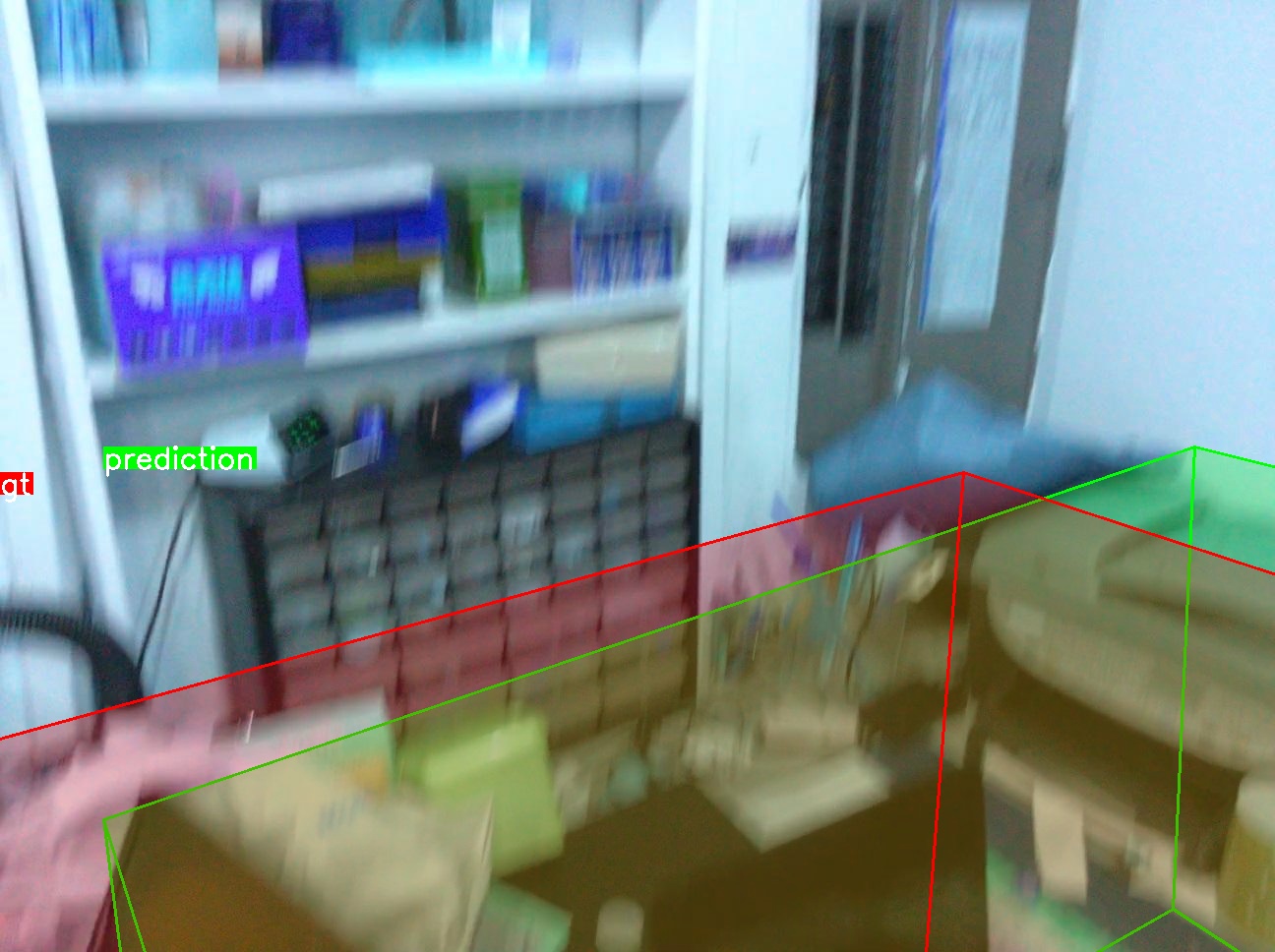} &
            \includegraphics[width=0.19\linewidth]{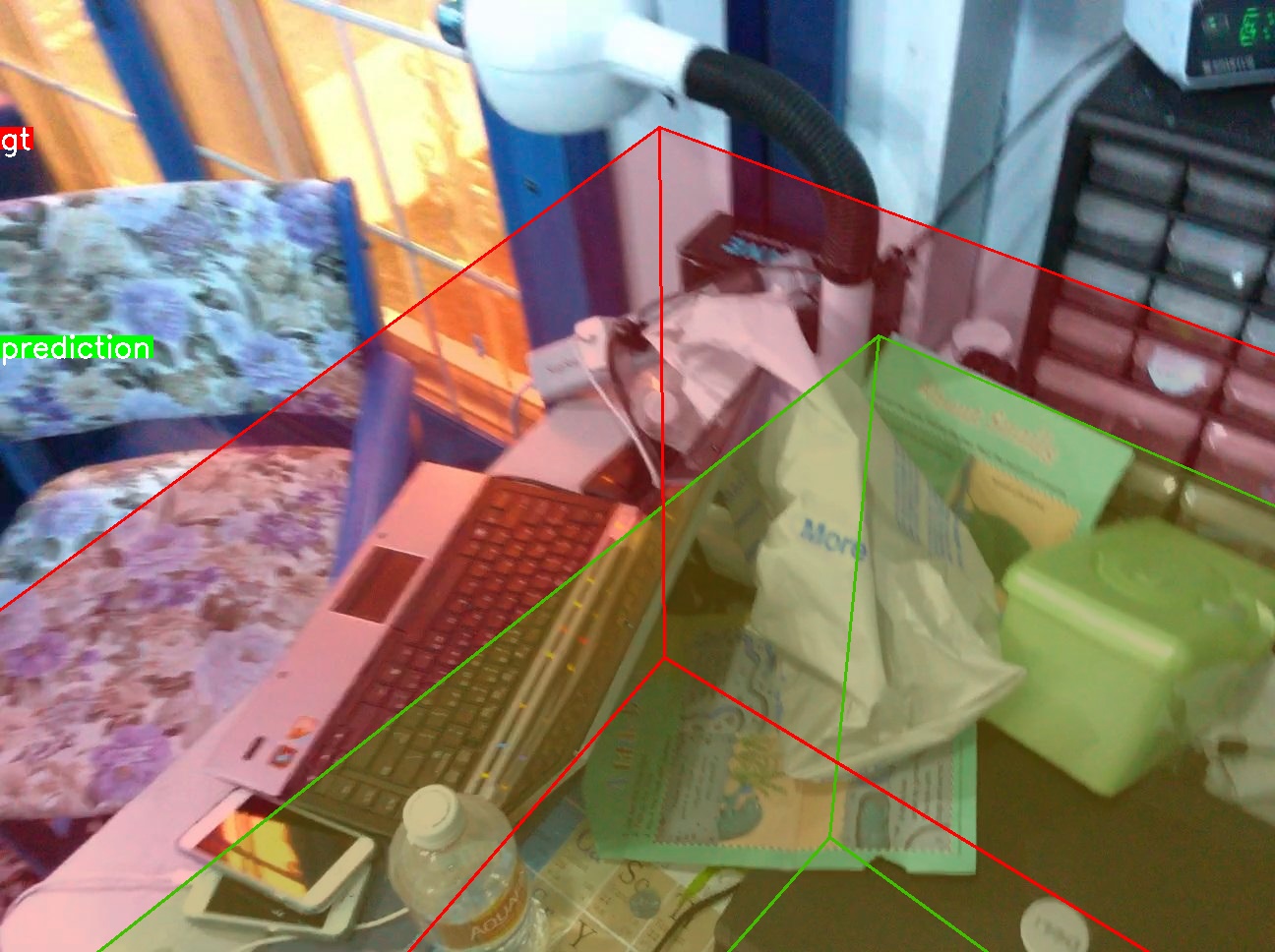} &
            \includegraphics[width=0.19\linewidth]{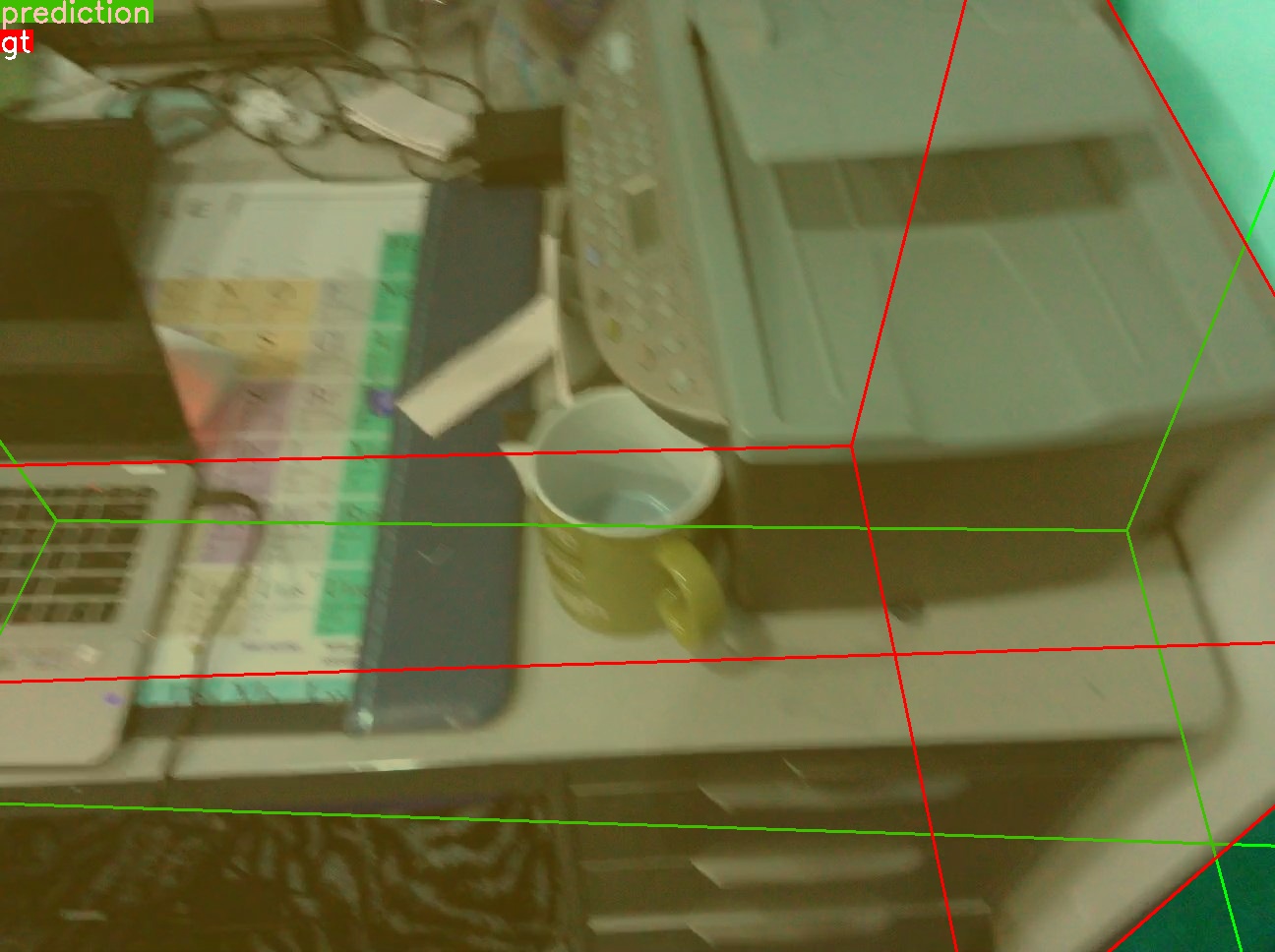} &
            \includegraphics[width=0.19\linewidth]{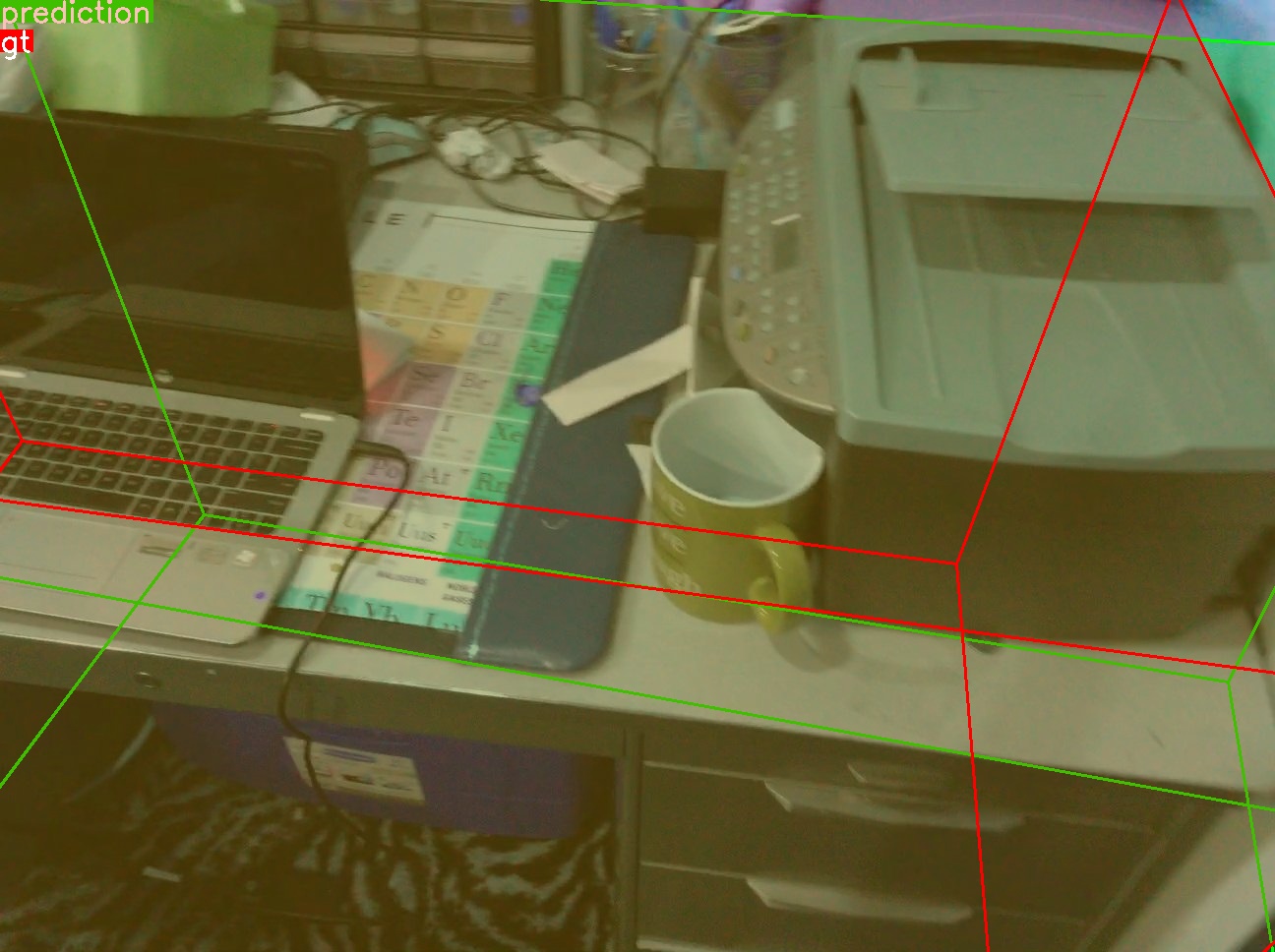} &
            \includegraphics[width=0.19\linewidth]{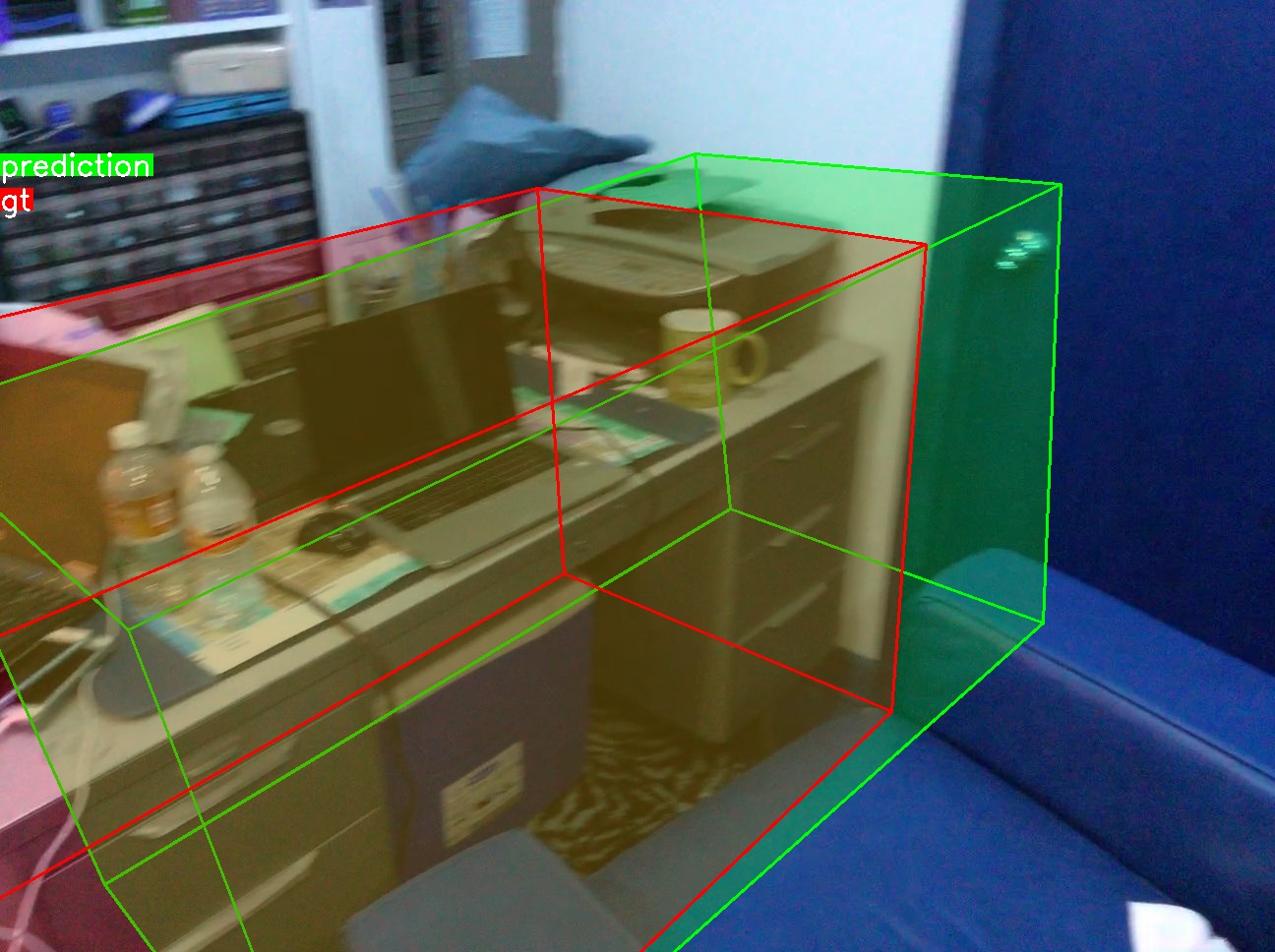} \\
        \end{tabular}
    \end{minipage}

    \vspace{1em} 

    \begin{minipage}{0.9\linewidth}
        \centering
        \textbf{Text:} \textit{``This is a black table. it is to the right of the kitchen cabinet.''}\\
        \vspace{2pt}
        
        \begin{tabular}{ccccc}
            \includegraphics[width=0.19\linewidth]{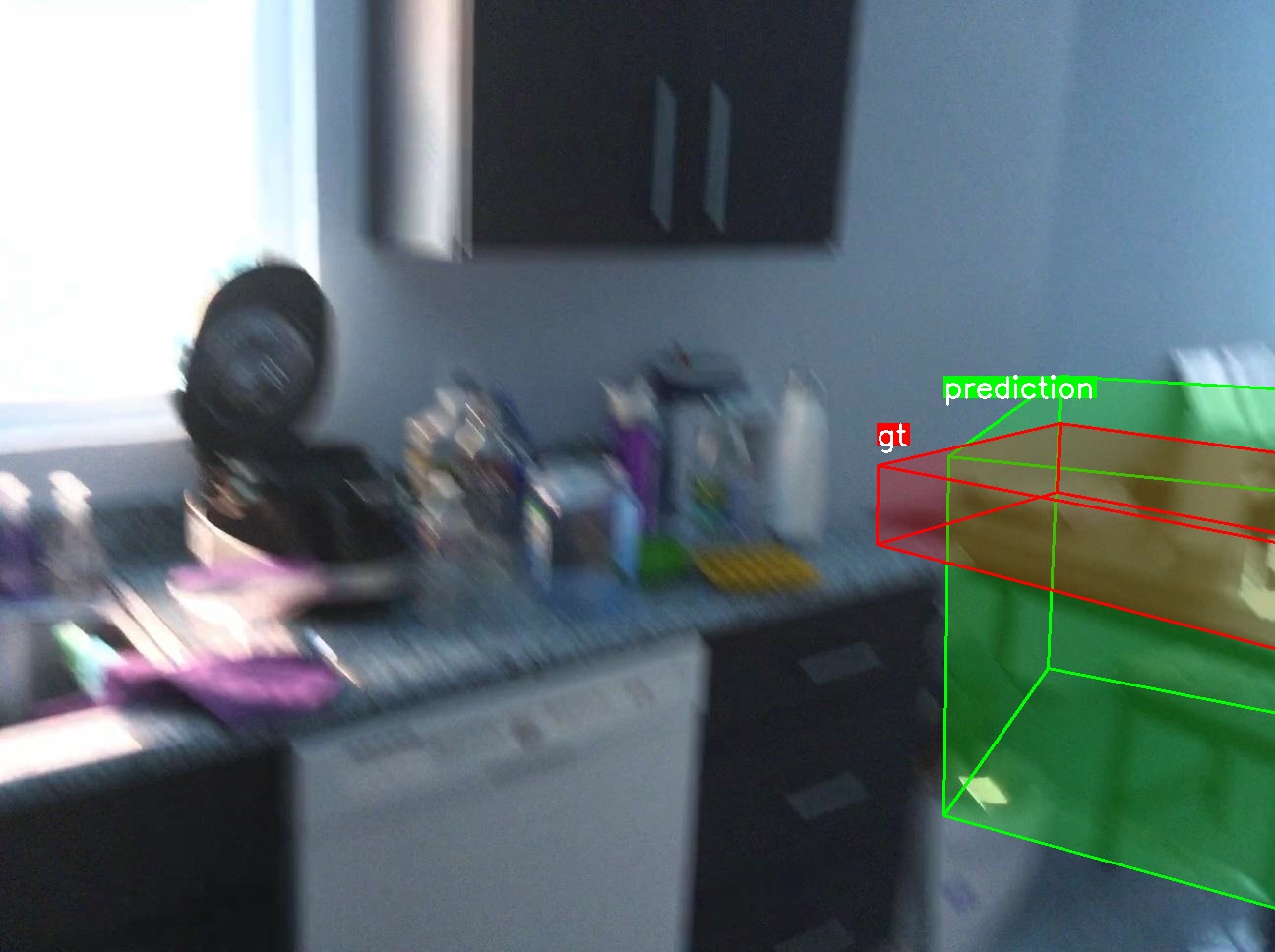} &
            \includegraphics[width=0.19\linewidth]{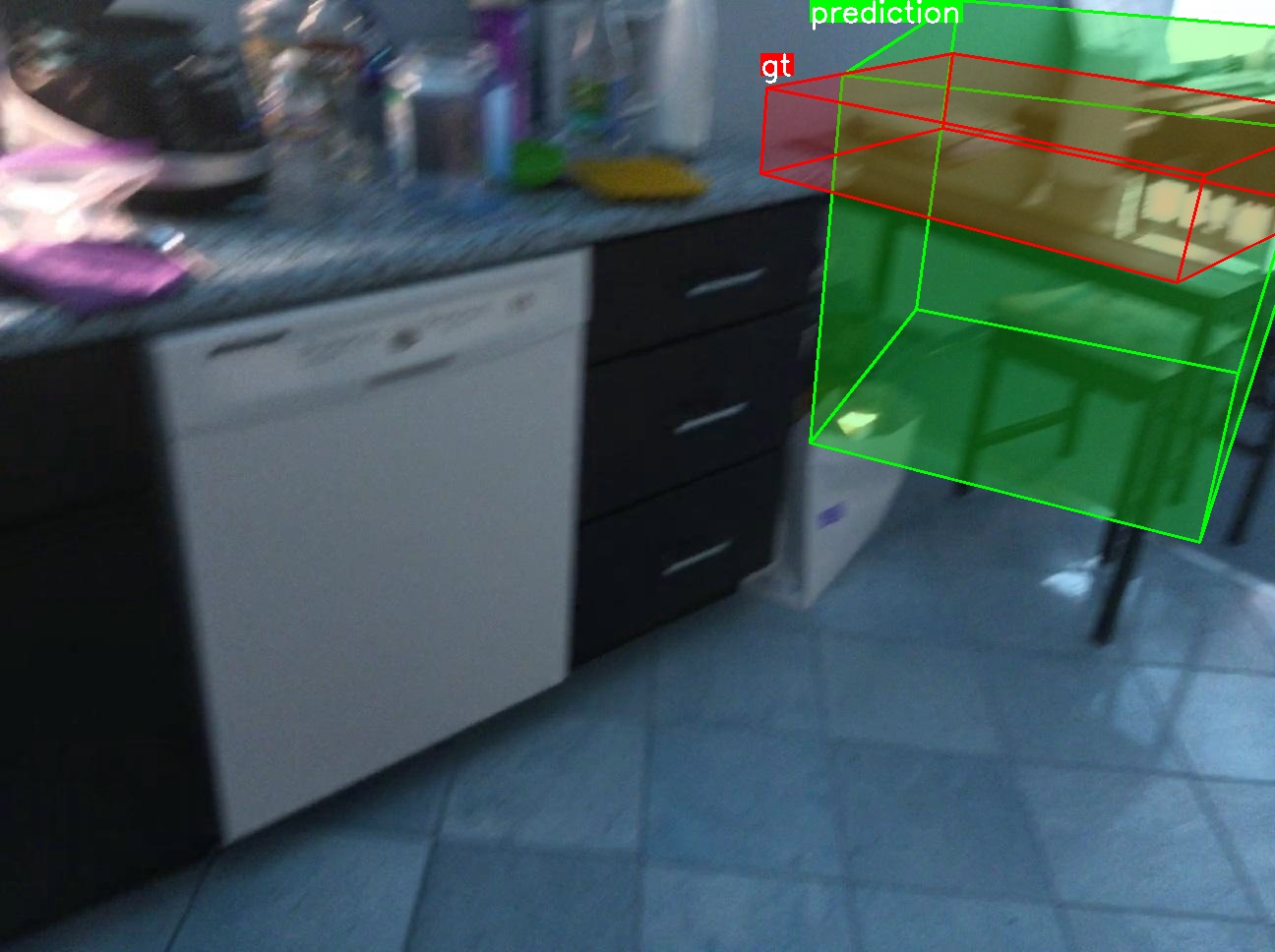} &
            \includegraphics[width=0.19\linewidth]{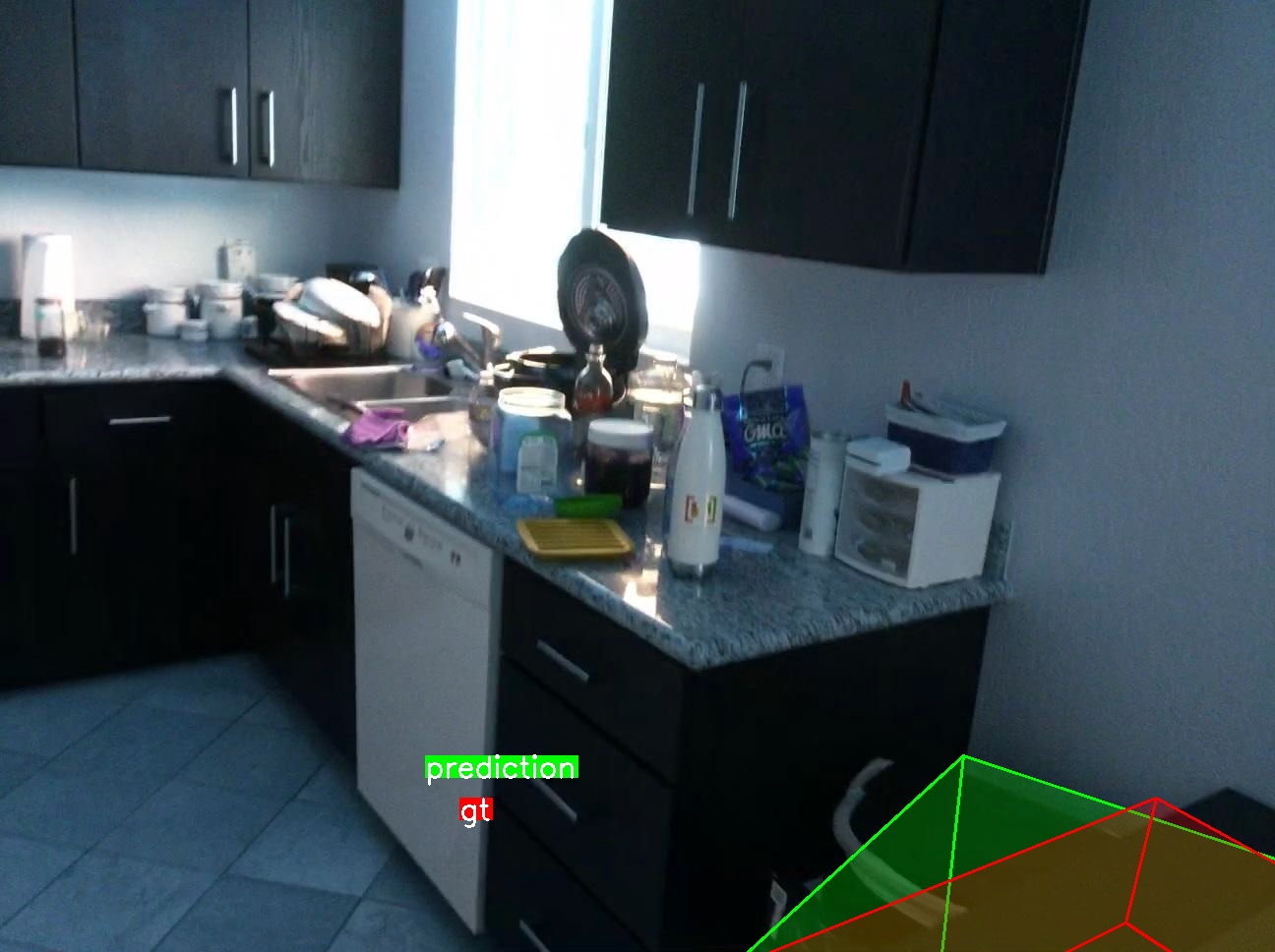} &
            \includegraphics[width=0.19\linewidth]{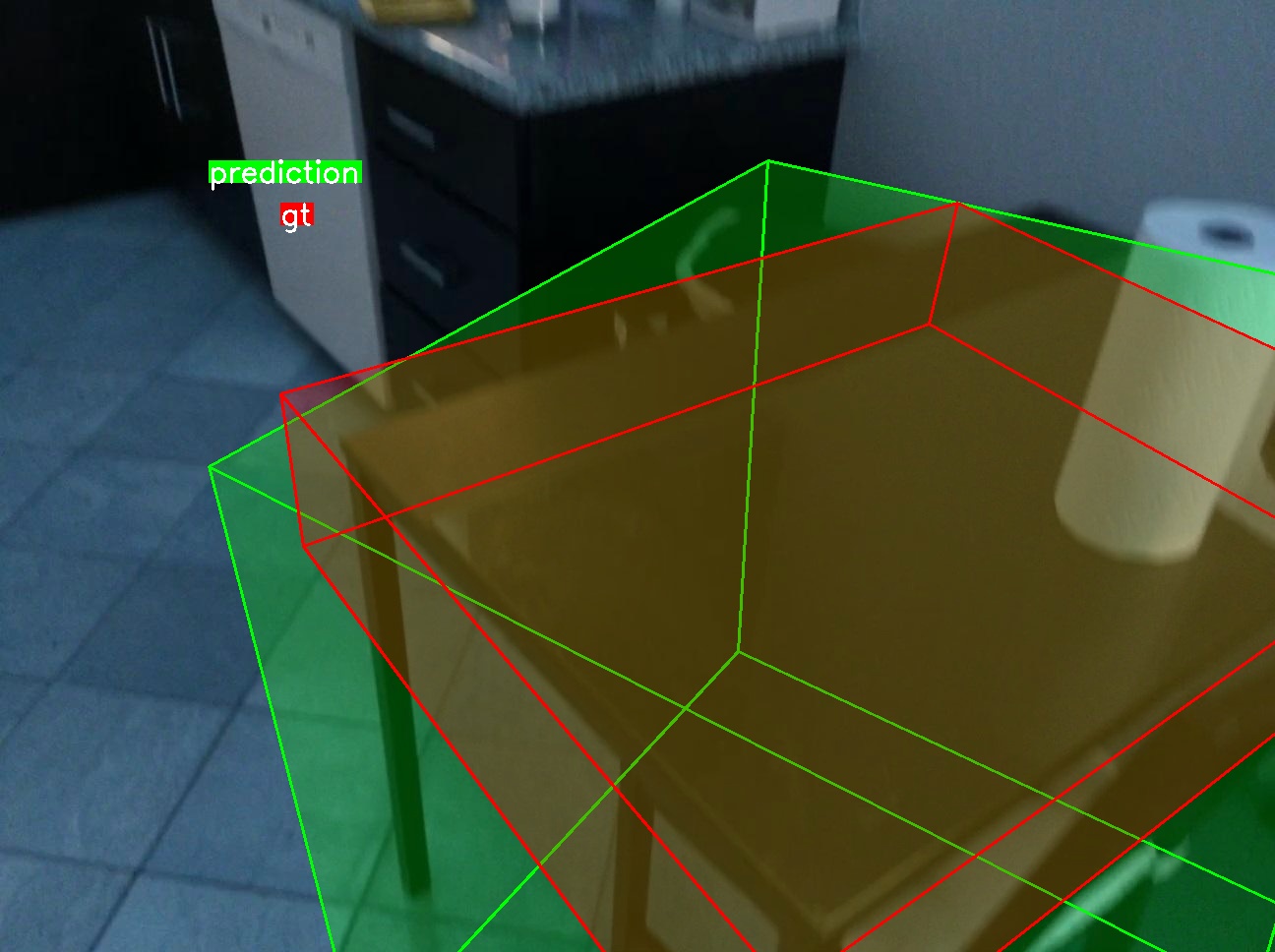} &
            \includegraphics[width=0.19\linewidth]{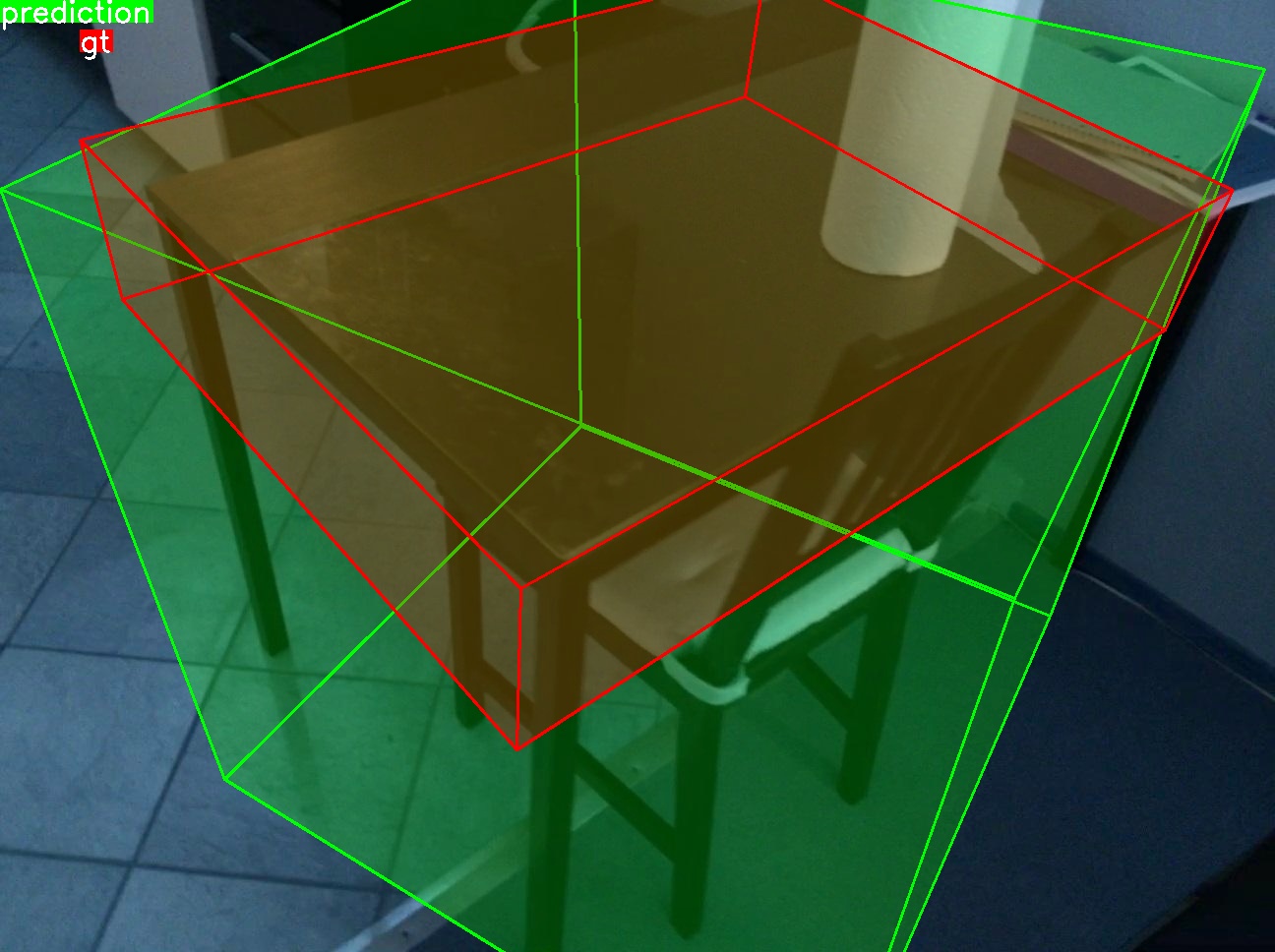} \\
            
            \includegraphics[width=0.19\linewidth]{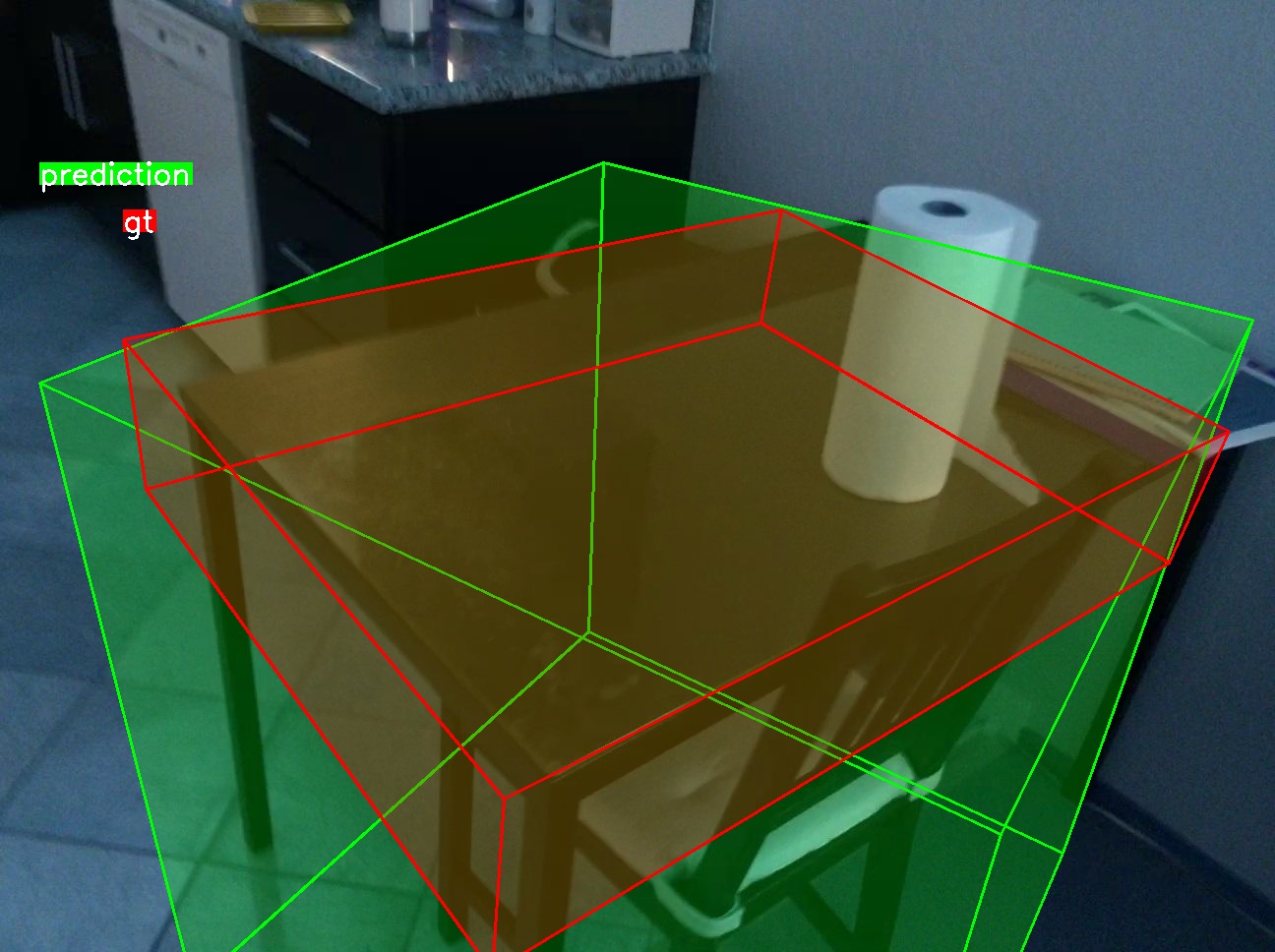} &
            \includegraphics[width=0.19\linewidth]{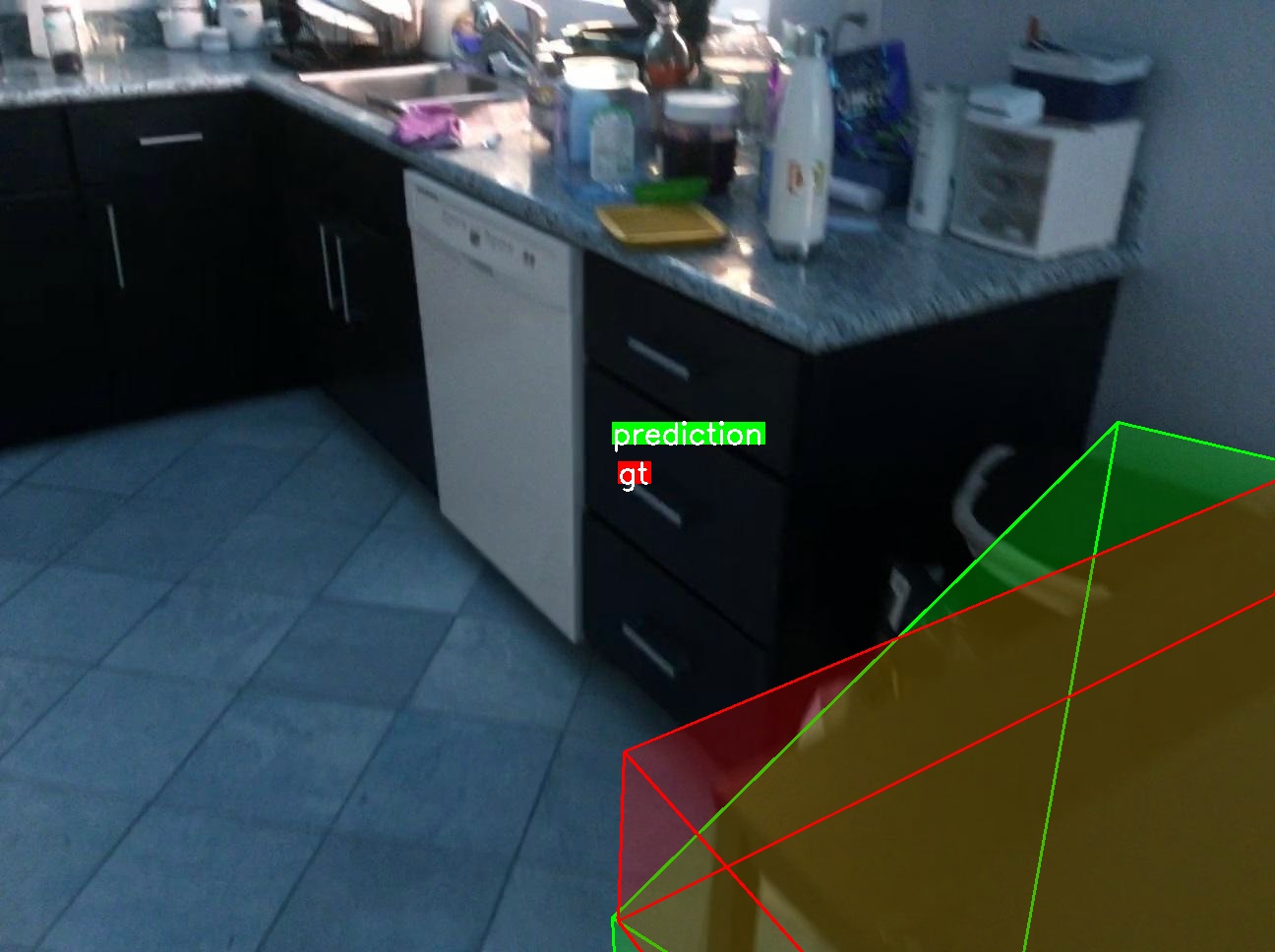} &
            \includegraphics[width=0.19\linewidth]{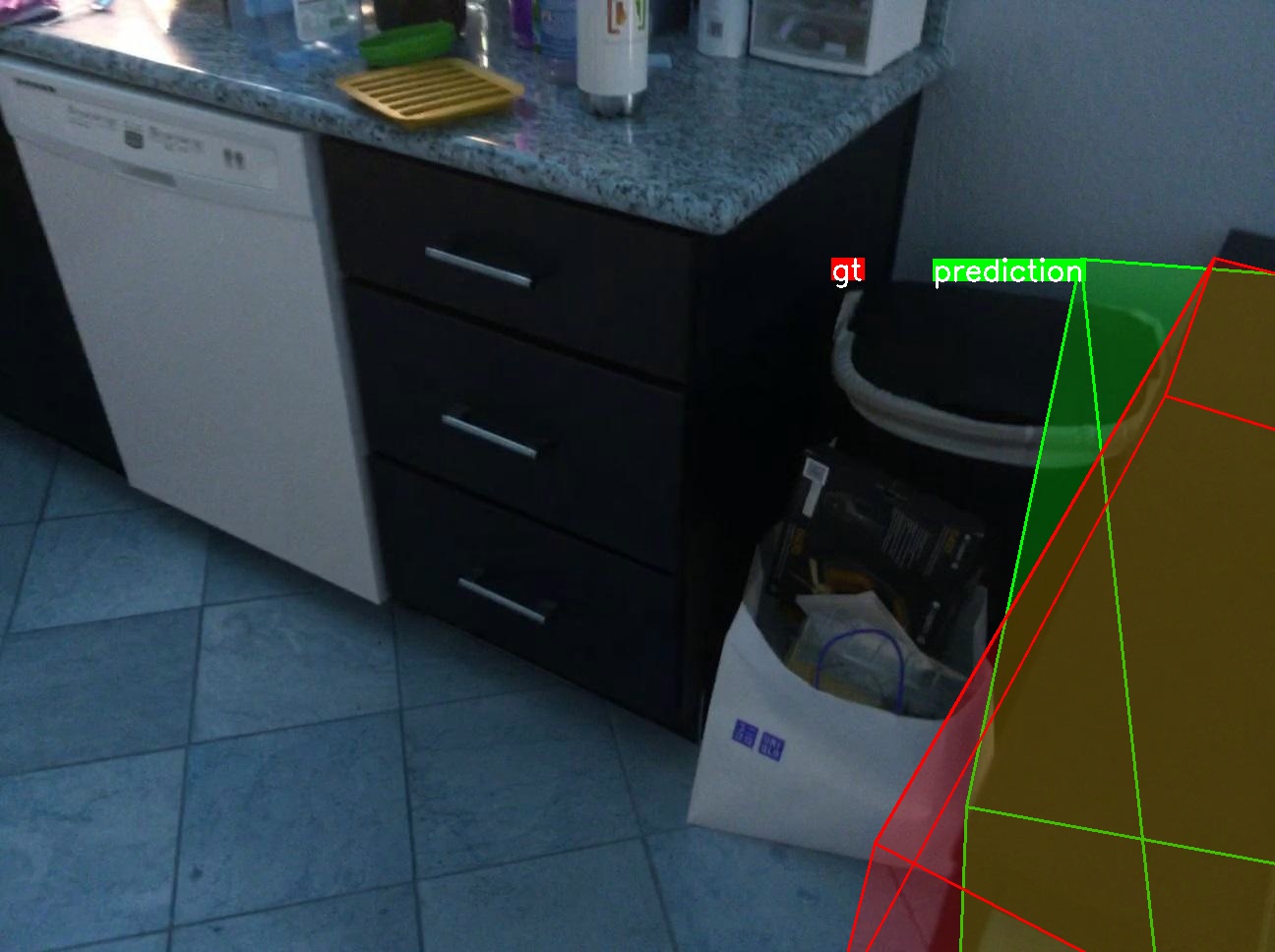} &
            \includegraphics[width=0.19\linewidth]{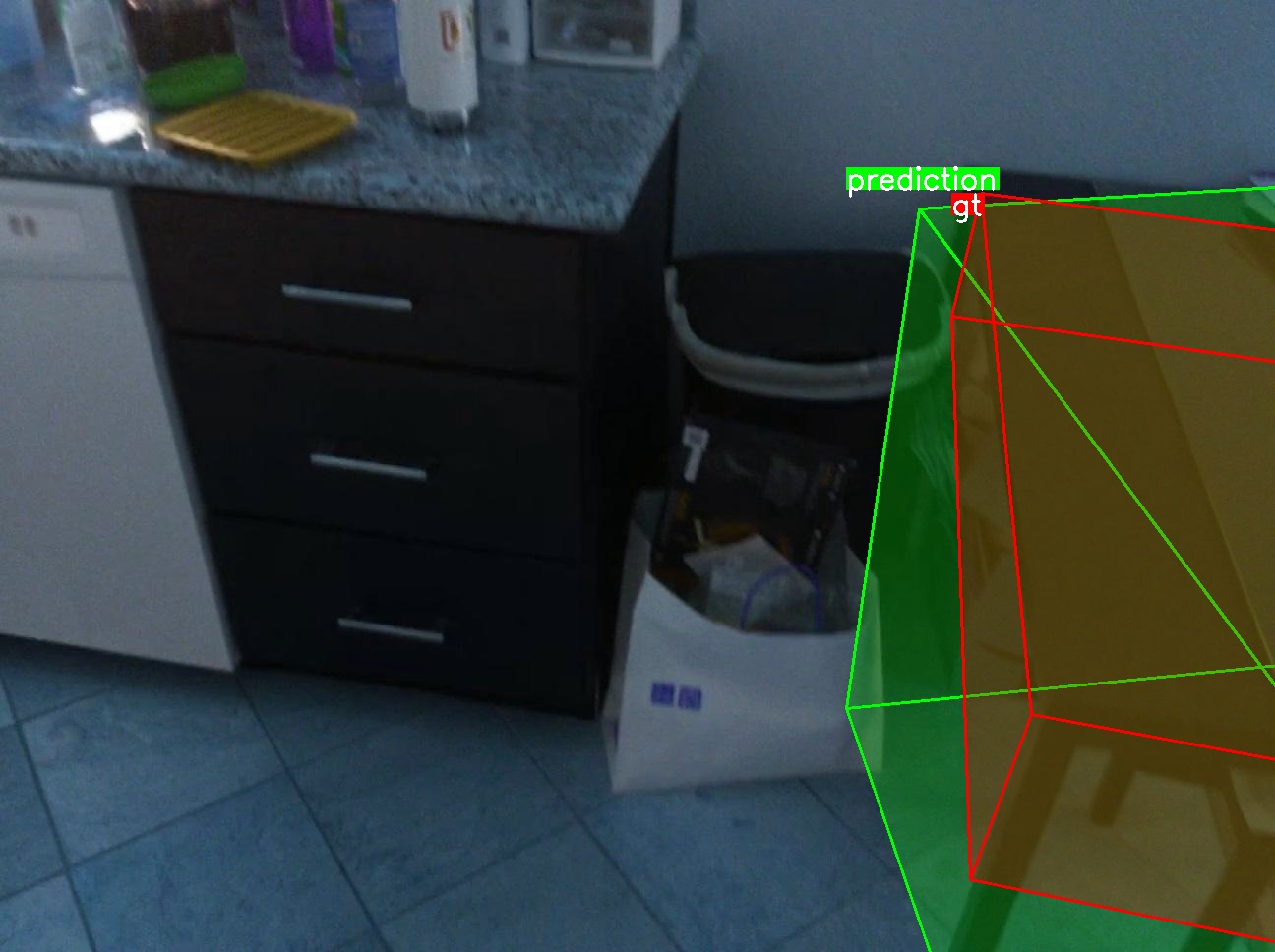} &
            \includegraphics[width=0.19\linewidth]{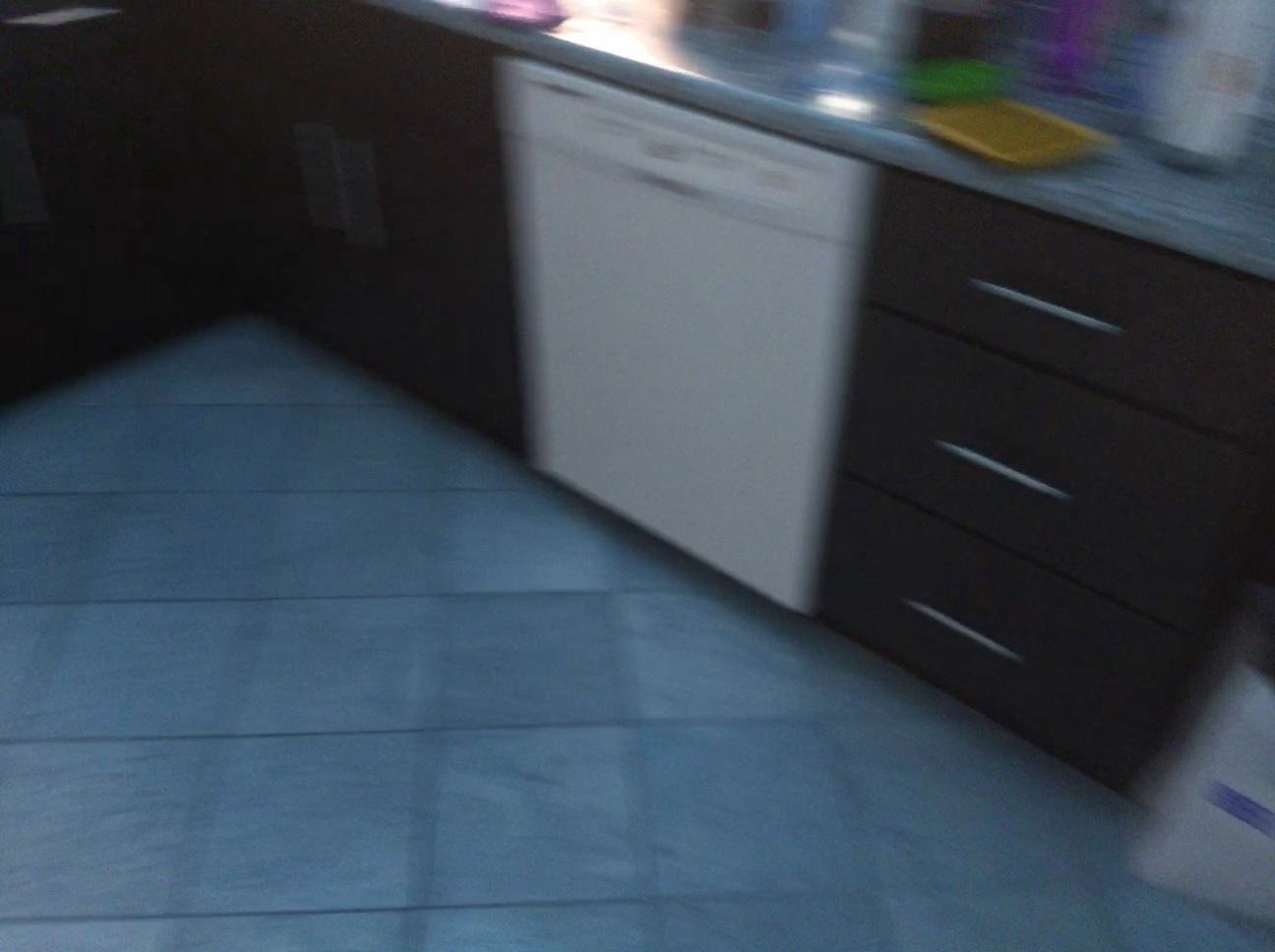} \\
        \end{tabular}
    \end{minipage}

    \caption{\textbf{Qualitative results of \textbf{3D Visual Grounding} using our \textbf{\model}}. Each block shows the grounding text followed by 10 video frames visualizing the predicted 3D bounding boxes (green) and ground truth (red).}
    \label{fig:vis_3d_grounding}
\end{figure*}

In \cref{fig:vis_3d_grounding}, we provide qualitative results of \vigrd. The evaluation is conducted under a 32-frame setting, and we choose some key frames mannually for visualization. 

\subsubsection{3D Spatial Reasoning}

\begin{figure}[t!]
\centering
\label{fig:spatial_visualization_a}

\begin{tcolorbox}[
    colback=lightgraybg,
    colframe=black!60,
    arc=4mm,
    boxrule=0.8pt,
    left=6pt,
    right=6pt,
    top=6pt,
    bottom=6pt,
    width=\textwidth
]

\textbf{Video}
\vspace{4pt}

\begin{center}
    \setlength{\tabcolsep}{1.5pt}
    \renewcommand{\arraystretch}{1.0}
    
    \resizebox{0.65\linewidth}{!}{%
    \begin{tabular}{cccccc}
        \includegraphics[width=0.155\linewidth]{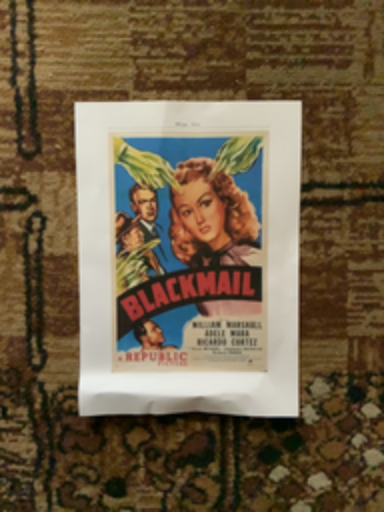} &
        \includegraphics[width=0.155\linewidth]{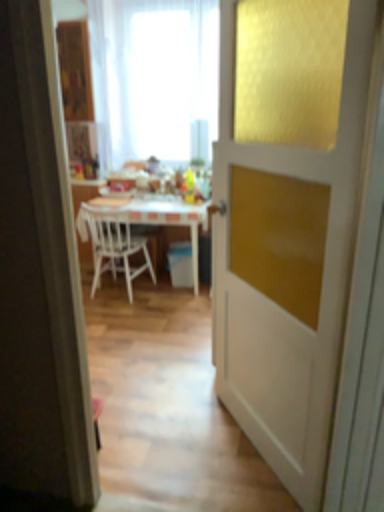} &
        \includegraphics[width=0.155\linewidth]{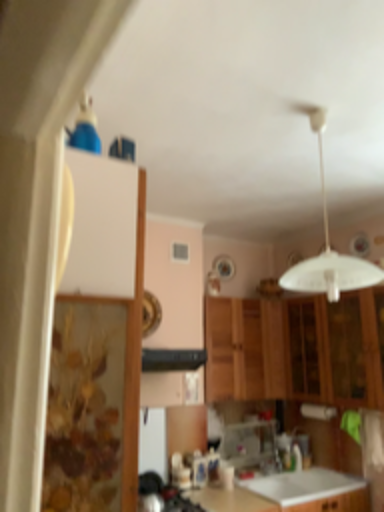} &
        \includegraphics[width=0.155\linewidth]{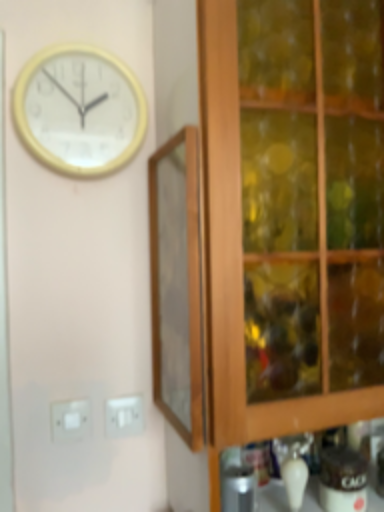} &
        \includegraphics[width=0.155\linewidth]{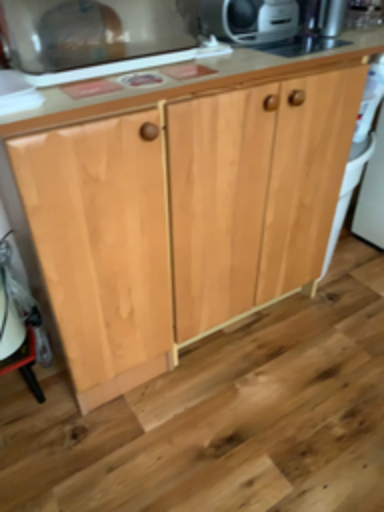} &
        \includegraphics[width=0.155\linewidth]{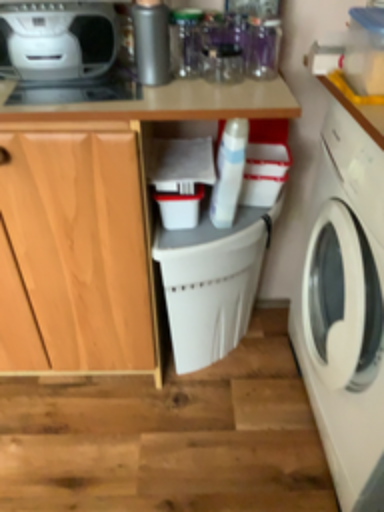} \\
        
        \includegraphics[width=0.155\linewidth]{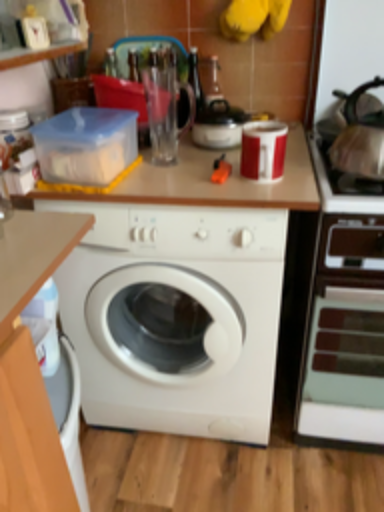} &
        \includegraphics[width=0.155\linewidth]{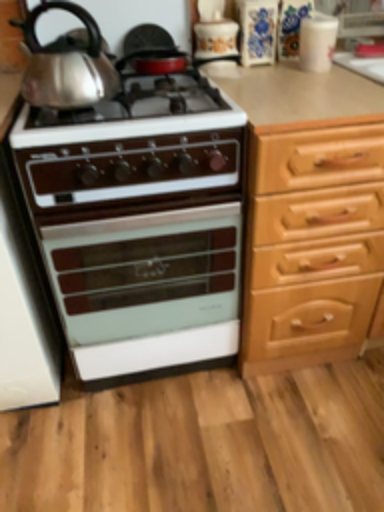} &
        \includegraphics[width=0.155\linewidth]{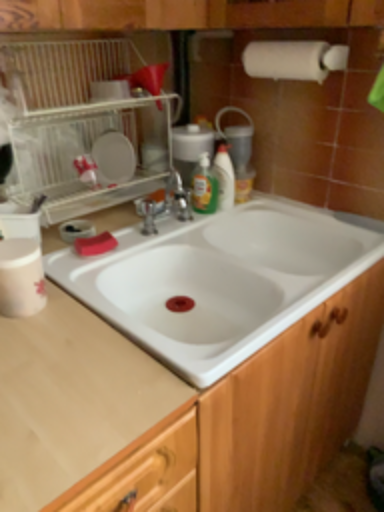} &
        \includegraphics[width=0.155\linewidth]{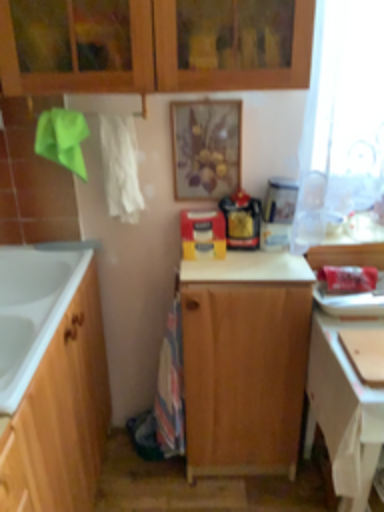} &
        \includegraphics[width=0.155\linewidth]{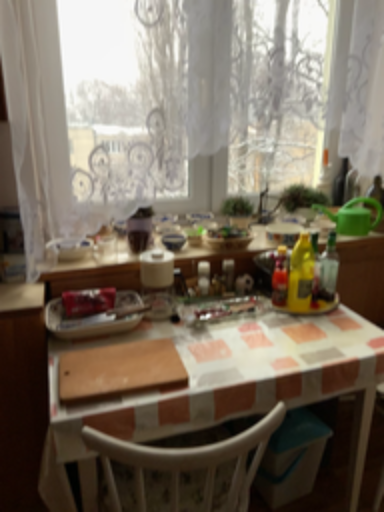} &
        \includegraphics[width=0.155\linewidth]{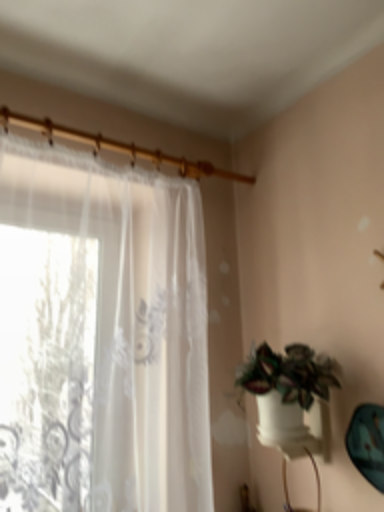} \\
    \end{tabular}
    }
\end{center}

\vspace{1pt}

\textbf{Question}

\vspace{2pt}

\textit{These are frames of a video.}
\textit{Measuring from the closest point of each object, what is the distance between the stool and the washer (in meters)?}
\vspace{4pt}

\textbf{3D-RFT-4B}

\vspace{1pt}

{\ttfamily\small
\textcolor{bluehl}{\textless think\textgreater}\\
The stool is near the table and the washer is in the lower part of the kitchen area.\\
\textcolor{bluehl}{\textless /think\textgreater}\\
\textcolor{greenhl}{\textless answer\textgreater 1.4\textless /answer\textgreater} 
}
\textcolor{teal}{\textbf{GT:} 1.3}

\end{tcolorbox}

\caption{\textbf{Qualitative example on VSI-Bench (absolute distance).}}
\label{fig:vsi_qualitative_abs_dist}
\end{figure}

\begin{figure}[t!]
\centering
\label{fig:spatial_visualization_b}

\begin{tcolorbox}[
    colback=lightgraybg,
    colframe=black!60,
    arc=4mm,
    boxrule=0.8pt,
    left=6pt,
    right=6pt,
    top=6pt,
    bottom=6pt,
    width=\textwidth
]

\textbf{Video}
\begin{center}
    \setlength{\tabcolsep}{1.5pt}
    \renewcommand{\arraystretch}{1.0}
    
    \resizebox{0.65\linewidth}{!}{%
    \begin{tabular}{cccccc}
        \includegraphics[width=0.155\linewidth]{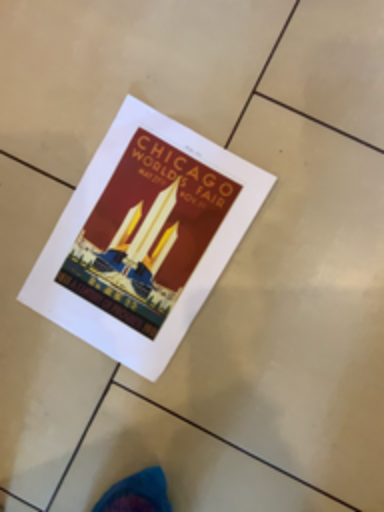} &
        \includegraphics[width=0.155\linewidth]{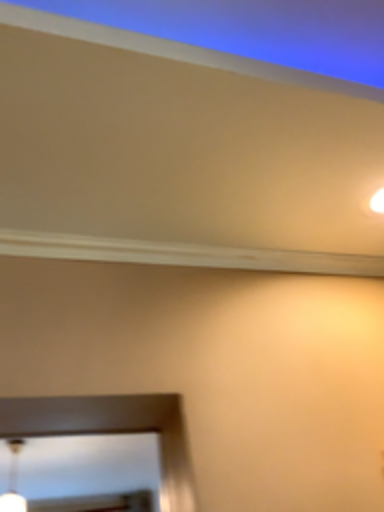} &
        \includegraphics[width=0.155\linewidth]{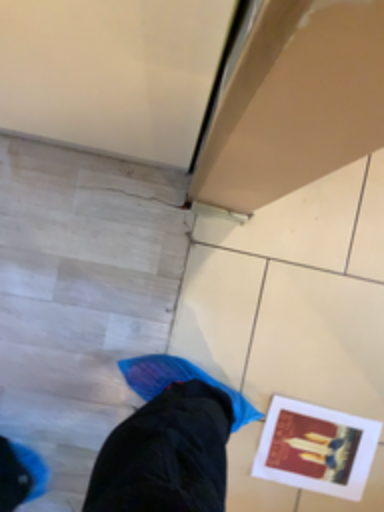} &
        \includegraphics[width=0.155\linewidth]{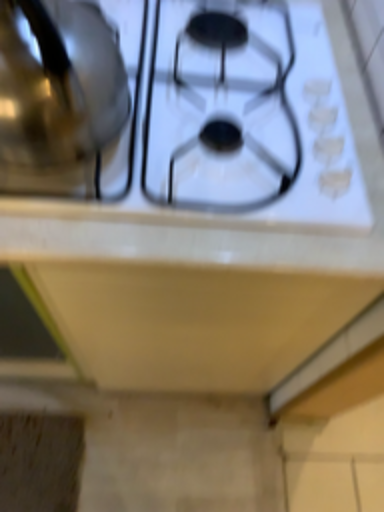} &
        \includegraphics[width=0.155\linewidth]{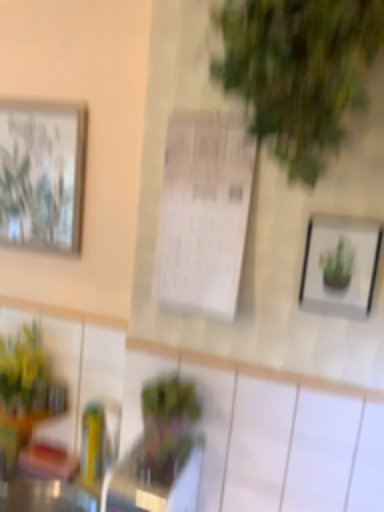} &
        \includegraphics[width=0.155\linewidth]{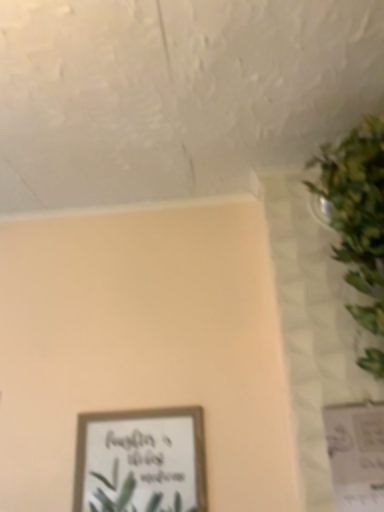} \\
        
        \includegraphics[width=0.155\linewidth]{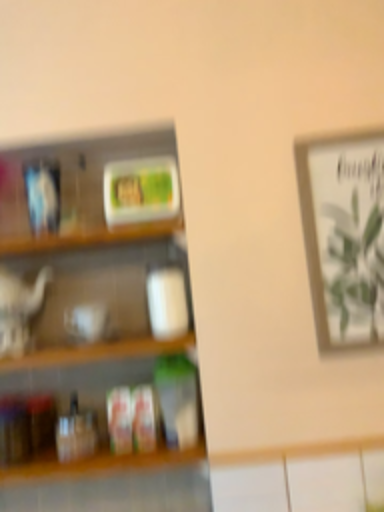} &
        \includegraphics[width=0.155\linewidth]{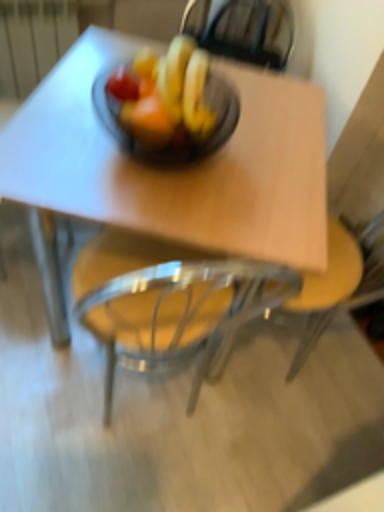} &
        \includegraphics[width=0.155\linewidth]{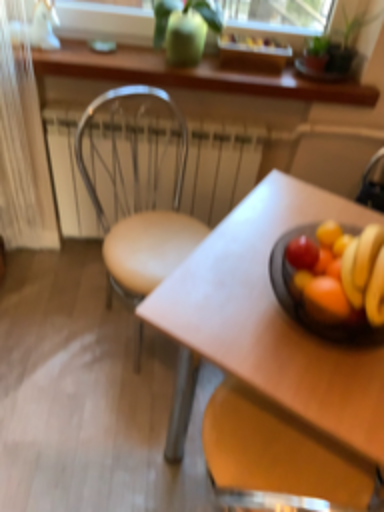} &
        \includegraphics[width=0.155\linewidth]{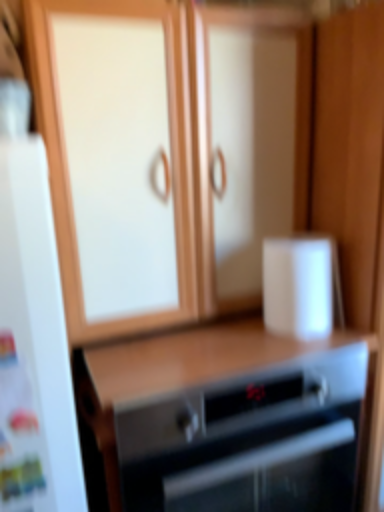} &
        \includegraphics[width=0.155\linewidth]{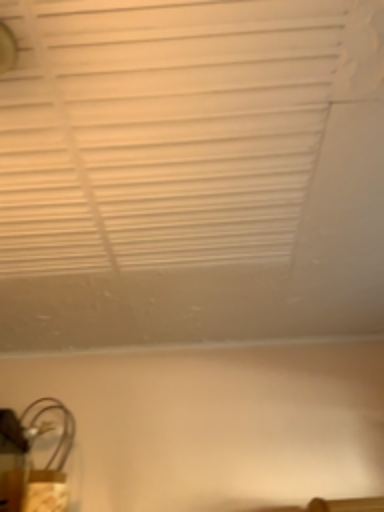} &
        \includegraphics[width=0.155\linewidth]{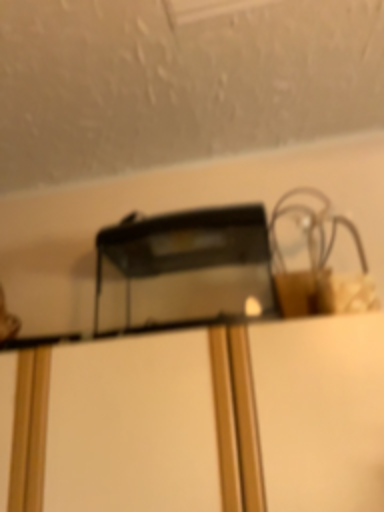} \\
    \end{tabular}
    }
\end{center}


\textbf{Question}

\vspace{2pt}

\textit{These are frames of a video.}
\textit{If I am standing by the stove and facing the refrigerator, is the table to my front-left, front-right, back-left, or back-right?}
\textit{The directions refer to the quadrants of a Cartesian plane (if I am standing at the origin and facing along the positive y-axis).}

\textit{Options:}
\textit{A. front-right}
\textit{B. front-left}
\textit{C. back-left}
\textit{D. back-right}

\vspace{4pt}

\textbf{3D-RFT-4B}

\vspace{2pt}

{\ttfamily\small
\textcolor{bluehl}{\textless think\textgreater}\\
I am standing by the stove, which is on the left side of the kitchen. The refrigerator is on the right side of the kitchen. The table is located in the middle of the kitchen, to the right of the refrigerator.\\
\textcolor{bluehl}{\textless /think\textgreater}\\
\textcolor{greenhl}{\textless answer\textgreater A\textless /answer\textgreater}
} \textcolor{teal}{\textbf{GT:} A}

\end{tcolorbox}

\caption{\textbf{Qualitative example on VSI-Bench (relative direction).}}
\label{fig:vsi_qualitative_rel_dir}
\end{figure}

\begin{figure}[t!]
\label{fig:spatial_visualization_c}
\centering

\begin{tcolorbox}[
    colback=lightgraybg,
    colframe=black!60,
    arc=4mm,
    boxrule=0.8pt,
    left=6pt,
    right=6pt,
    top=6pt,
    bottom=6pt,
    width=\textwidth
]

\textbf{Video}
\begin{center}
    \setlength{\tabcolsep}{1.5pt}
    \renewcommand{\arraystretch}{1.0}
    
    \resizebox{0.86\linewidth}{!}{%
    \begin{tabular}{cccccc}
        \includegraphics[width=0.155\linewidth]{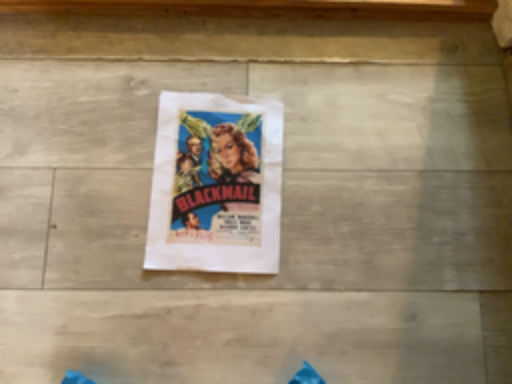} &
        \includegraphics[width=0.155\linewidth]{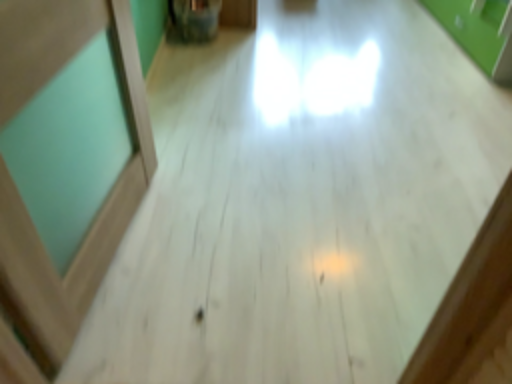} &
        \includegraphics[width=0.155\linewidth]{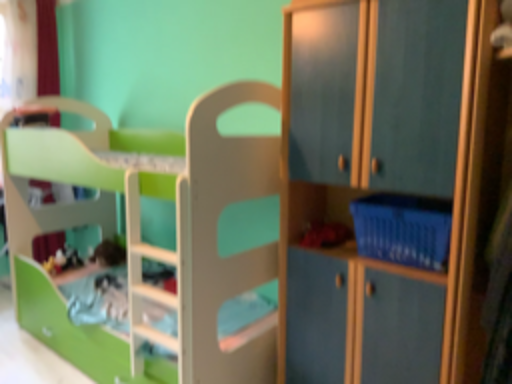} &
        \includegraphics[width=0.155\linewidth]{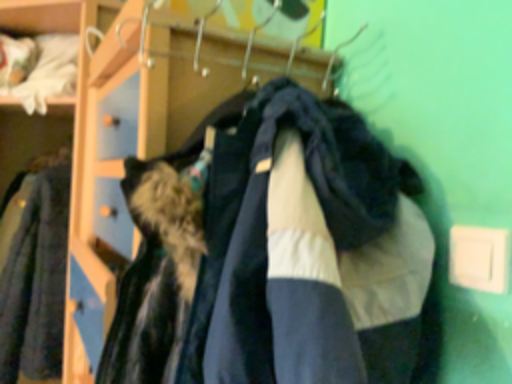} &
        \includegraphics[width=0.155\linewidth]{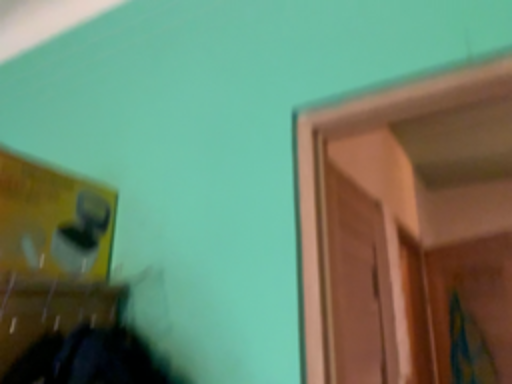} &
        \includegraphics[width=0.155\linewidth]{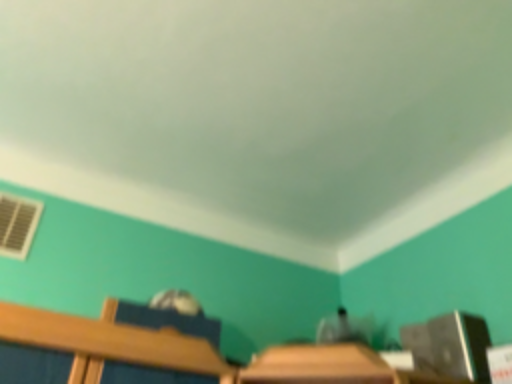} \\
        
        \includegraphics[width=0.155\linewidth]{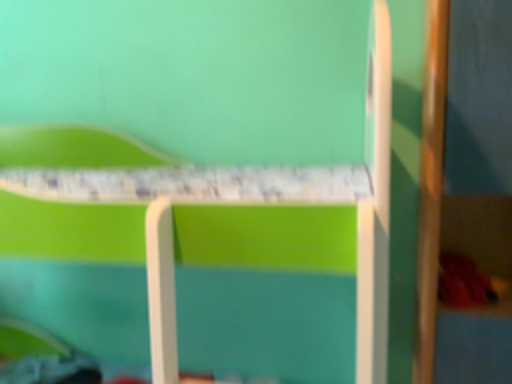} &
        \includegraphics[width=0.155\linewidth]{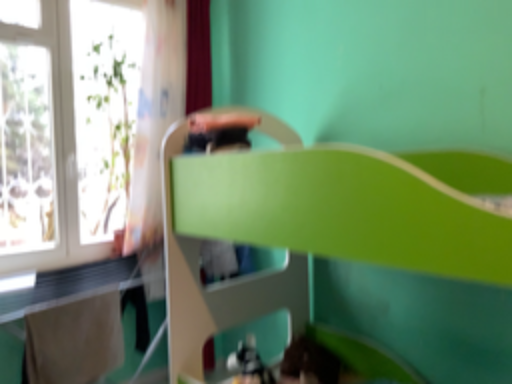} &
        \includegraphics[width=0.155\linewidth]{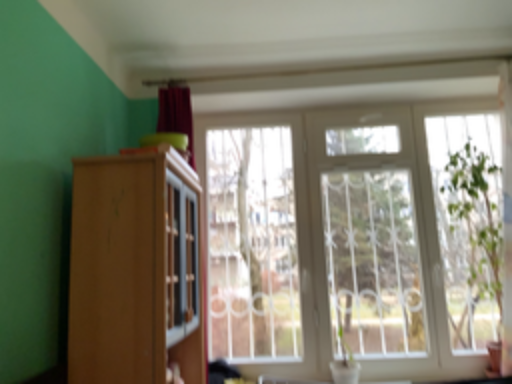} &
        \includegraphics[width=0.155\linewidth]{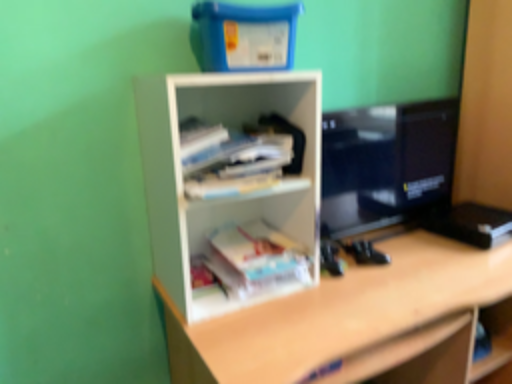} &
        \includegraphics[width=0.155\linewidth]{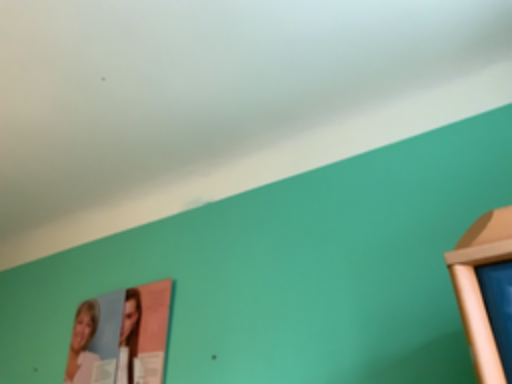} &
        \includegraphics[width=0.155\linewidth]{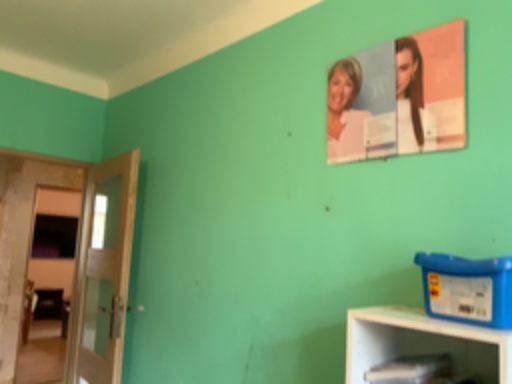} \\
    \end{tabular}
    }
\end{center}


\textbf{Question}

\vspace{2pt}

\textit{These are frames of a video.}
\textit{You are a robot beginning at the door and facing the desk. You want to navigate to the window. You will perform the following actions (Note: for each [please fill in], choose either ``turn back,'' ``turn left,'' or ``turn right.''): 1. Go forward until the desk 2. [please fill in] 3. Go forward until the bed 4. [please fill in] 5. Go forward until the window. You have reached the final destination.}
\textit{Options:}
\textit{A. Turn Right, Turn Right}
\textit{B. Turn Right, Turn Left}
\textit{C. Turn Back, Turn Right}
\textit{D. Turn Left, Turn Left}

\vspace{4pt}

\textbf{3D-RFT-4B}

\vspace{4pt}

{\ttfamily\small
\textcolor{bluehl}{\textless think\textgreater}\\
The robot starts at the door facing the desk. To reach the window, it must first go forward to the desk, then turn right to face the bed, and finally turn right again to reach the window.\\
\textcolor{bluehl}{\textless /think\textgreater}\\
\textcolor{greenhl}{\textless answer\textgreater B\textless /answer\textgreater}
} \textcolor{teal}{\textbf{GT:} B}

\end{tcolorbox}

\caption{\textbf{Qualitative example on VSI-Bench (route plan).}}
\label{fig:vsi_qualitative_route_plan}
\end{figure}

We present qualitative results for \spr in \cref{fig:spatial_visualization_a,fig:spatial_visualization_b,fig:spatial_visualization_c}. In these cases, we sample some frames from the raw video. These examples demonstrate that \model is capable of generating concise, scene-grounded \ac{cot} reasoning to facilitate accurate answer prediction.

\subsection{Training Dynamics of 3D Visual Grounding}
\begin{figure*}[t]
    \centering
    \setlength{\tabcolsep}{1pt}
    
    \begin{subfigure}[b]{0.32\linewidth}
        \centering
        \includegraphics[width=\linewidth]{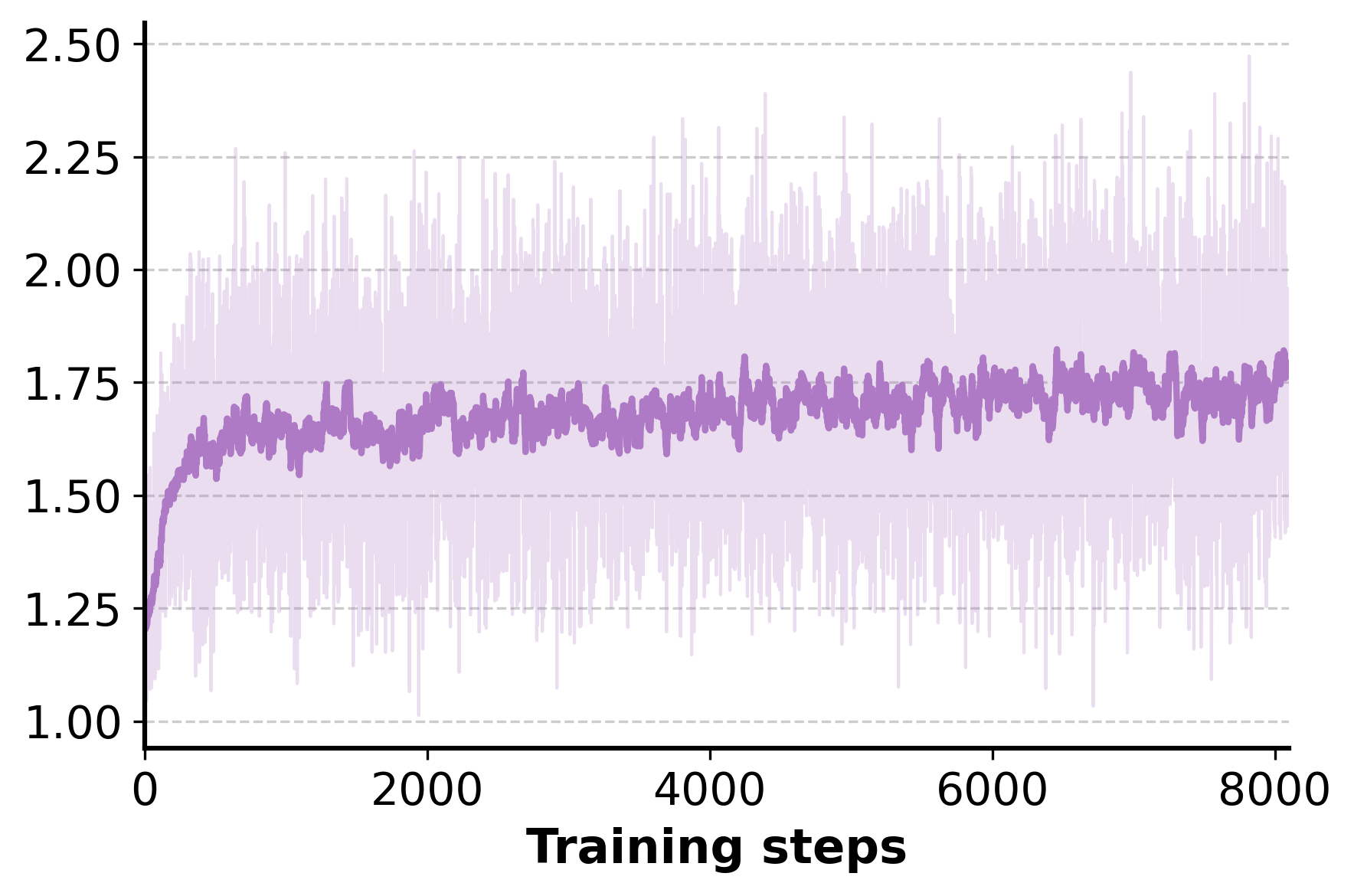}
        \caption{Total Reward}
        \label{fig:grounding_reward}
    \end{subfigure}
    \hfill
    \begin{subfigure}[b]{0.32\linewidth}
        \centering
        \includegraphics[width=\linewidth]{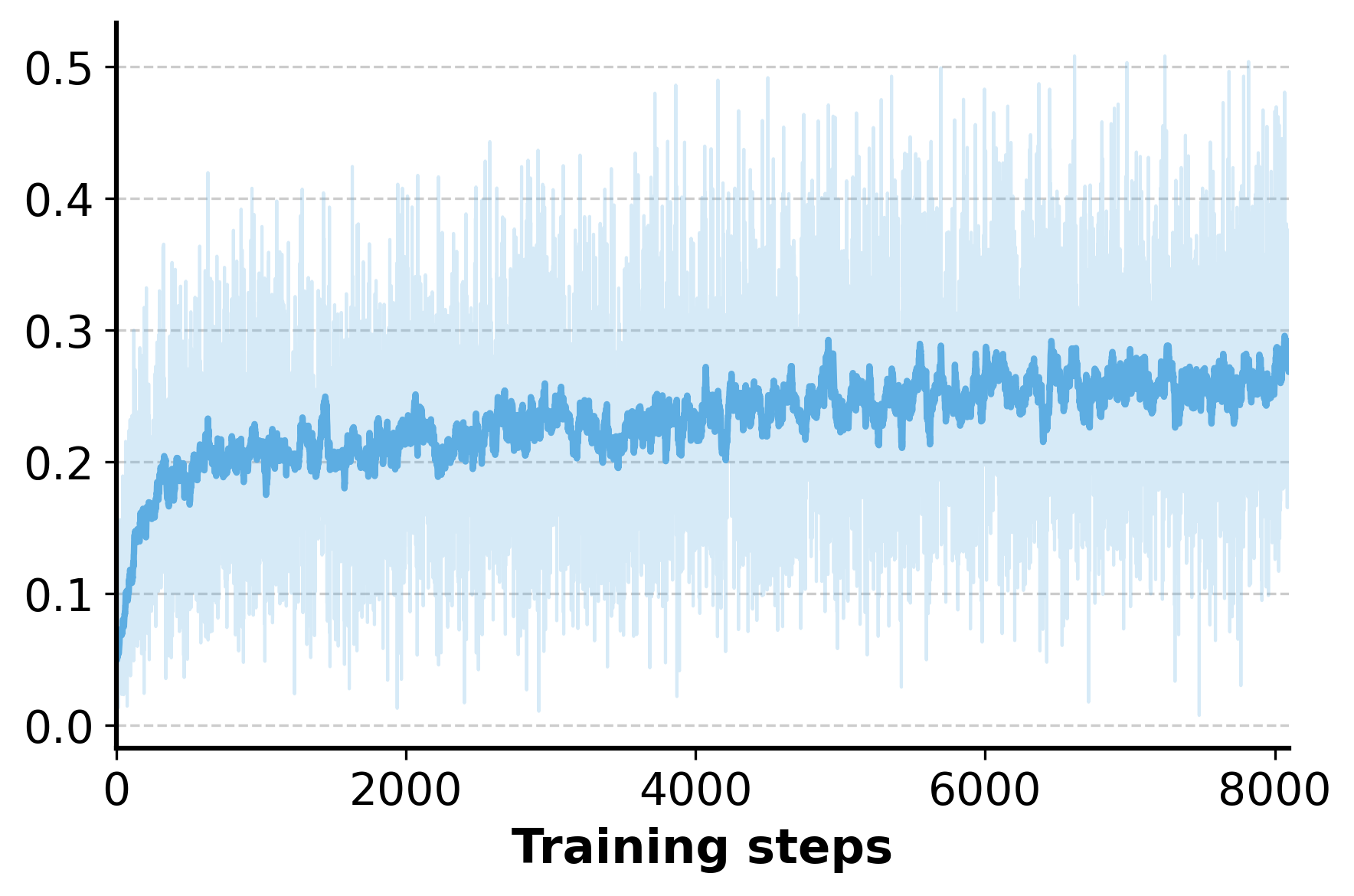}
        \caption{3D IoU Reward}
        \label{fig:grounding_iou}
    \end{subfigure}
    \hfill
    \begin{subfigure}[b]{0.32\linewidth}
        \centering
        \includegraphics[width=\linewidth]{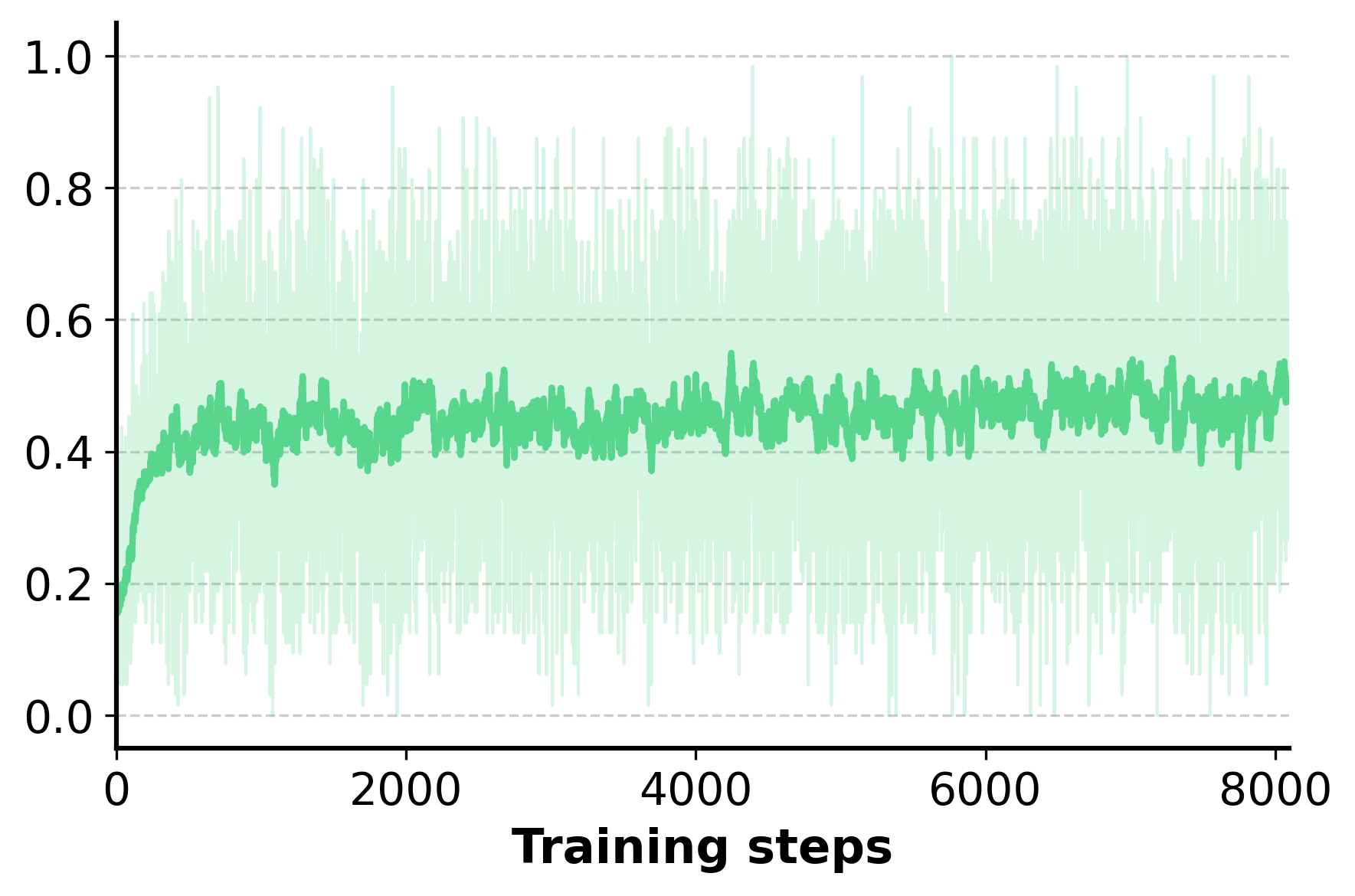}
        \caption{Frame Reward}
        \label{fig:grounding_frame}
    \end{subfigure}
    
    \caption{\textbf{Training Dynamics of 3D Visual Grounding.} We visualize the training curves for the total reward (left), the 3D IoU reward (middle), and the Frame Alignment reward (right). The curves show stable convergence and correlation with the optimization objectives.}
    \label{fig:grounding_dynamics}
\end{figure*}

We illustrate the training dynamics of \vigrd in \cref{fig:grounding_dynamics}. The curves demonstrate a consistent upward trend for both the frame reward and the 3D IoU reward, indicating a stable and effective optimization process.






\section{Discussion}

\subsection{Impact of TA Data for SFT}
\label{appendix:sec:ood}


\begin{wrapfigure}{r}{0.4\linewidth}
  \centering
  \vspace{-5em} 
  \includegraphics[width=0.8\linewidth]{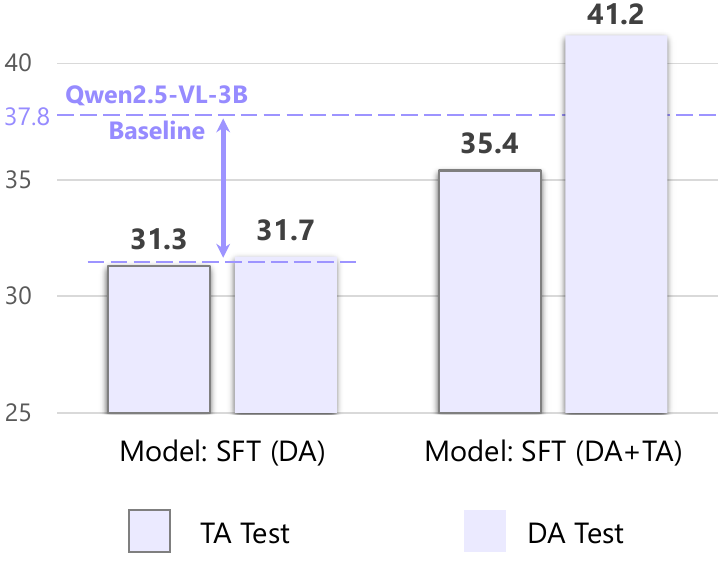}
  \caption{\textbf{Evaluation results on MindCube-Tiny}.}
  \label{appendix:fig:mindcube}
  \vspace{-1em} 
\end{wrapfigure}

We compare the \ac{ood} evaluation results of SFT (DA) and SFT (DA+TA) on MindCube-Tiny \citep{yin2025spatial}. The results in \cref{appendix:fig:mindcube} show that SFT (DA) significantly lags behind SFT (DA+TA) under both TA and DA test settings. SFT (DA) also shows notable performance drops compared to the vanilla Qwen2.5-VL-3B baseline. This indicates the potential overfitting issues in training 3D \ac{mllm} with solely DA data, appealing for mixing more TA data (\ie, \ac{cot} data) for comprehensive enhancement of \ac{mllm}'s capabilities.

\subsection{Remarks}
Unlike domains where \ac{rlvr} thrives (\eg, math and coding), 3D scene understanding presents distinct challenges. First, the scarcity of high-quality labels and the inherent sparsity of learning signals constitute major obstacles. The cost and difficulty of 3D \ac{cot} data collection further hinder the efficacy of \ac{rlvr} for 3D scene understanding. Our findings suggest that \ac{rlvr} outcomes are sensitive to \ac{cot} data quality, highlighting the refinement of \ac{cot} data as a critical future direction. Second, while math and coding tasks exhibit a straightforward reasoning loop within a unified textual modality, 3D scene understanding requires robust 3D-aware perception from video inputs before reasoning can occur. This perceptual stage represents a significant bottleneck, which could undermine the correctness and coherence of the model's thoughts. Therefore, future research should prioritize process reward designs to guarantee the soundness of reasoning within 3D scenes.


\section{Limitations}

In this work, we focus on demonstrating the efficacy of the \ac{rft} paradigm by conducting task-specific fine-tuning during the \ac{rl} stage; we do not currently explore unified, multi-task fine-tuning. Consequently, investigating how to balance diverse task rewards to enable efficient mixed training remains a critical direction for future research. Furthermore, we do not utilize \ac{cot} data for \percep tasks due to the challenges associated with collecting high-quality reasoning annotations in this domain. Acquiring such data and analyzing the impact of \ac{cot} on perception capabilities represents another promising avenue for advancing \vsu models.


\end{document}